\documentclass[DIV=15, bibliography=totoc]{scrartcl}  

\UseRawInputEncoding

\usepackage[a4paper,margin=2.25cm,bottom=2.5cm]{geometry}
\usepackage{amsmath,amssymb}

\usepackage{xcolor} 
\definecolor{lightblue}{rgb}{0,0.5,1.0}
\definecolor{linkblue}{rgb}{0,0.1,0.6}
\definecolor{citegreen}{rgb}{0,0.4,0.0}
\definecolor{linkred}{rgb}{0.8,0,0.005}
\definecolor{mailviolet}{rgb}{0.3,0,0.35}
\definecolor{tumblue}{rgb}{0,0.396,0.741}
\definecolor{darkgreen}{rgb}{0,0.4,0} 
\definecolor{darkbrown}{rgb}{0.5, 0.396, 0.09}

\usepackage[hyperindex,colorlinks=true,linkcolor=linkblue,citecolor=citegreen,urlcolor=mailviolet,filecolor=linkred]{hyperref}

\usepackage{caption, subcaption}
\usepackage{fancyhdr}

\usepackage{siunitx}
\usepackage[square,comma,nonamebreak,compress,numbers]{natbib}

\usepackage{soulpos}
\usepackage{ulem}
\usepackage{authblk} 
\usepackage{float}

\usepackage[inkscapearea=page]{svg}
\usepackage{amsmath}
\usepackage[noabbrev]{cleveref}
\usepackage{upgreek}
\usepackage{adjustbox}
\usepackage{color}
\usepackage{enumitem}
\usepackage{todonotes}

\usepackage{pgfplots}
\usepackage{pgfplotstable}
\usepackage{filecontents}
\usepackage{tikz}
\usepackage{tikzscale}
\usepackage{tikz-3dplot}

\usetikzlibrary{spy}
\usetikzlibrary{intersections}
\usetikzlibrary{arrows,shapes}
\usetikzlibrary{spy}
\usetikzlibrary{backgrounds}  
\usetikzlibrary{decorations}
\usetikzlibrary{decorations.markings}
\usetikzlibrary{positioning}
\usetikzlibrary{patterns}
\usetikzlibrary{calc} 
\usetikzlibrary{quotes} 
\usetikzlibrary{external} 
\usetikzlibrary{matrix}

\pgfplotscreateplotcyclelist{customCycleList}
{%
  {blue, mark=o},
  {red, mark=square},
  {darkgreen, mark=diamond},
  {cyan, mark=diamond},
  {darkbrown, mark=triangle*},
  {black, mark=pentagon},
}

\pgfplotsset{every axis/.append style= {
    cycle list name=customCycleList,
}}

\title{Deep Learning in Deterministic Computational Mechanics}

\author[1]{Leon Herrmann\thanks{\href{mailto:leon.herrmann@tum.de}{\texttt{leon.herrmann@tum.de}},
    Corresponding author}}
\author[1]{Stefan Kollmannsberger}

\affil[1]{Chair of Computational Modeling and Simulation, Technical University of Munich, School of Engineering and Design, Arcisstraße 21, Munich, 80\,333, Germany}

\newcommand{\publicationDate}{\today}

\date{}
\fancypagestyle{plain}{%
\fancyhf{}%
\fancyfoot[L]{
\vspace{-0.0cm} 
\footnotesize{Preprint submitted to Computational Mechanics.  \hfill
\publicationDate
\\
}} }

\crefname{paragraph}{paragraph}{paragraphs}
\Crefname{paragraph}{Paragraph}{Paragraphs}

\begin{document}     

\normalem \maketitle  
\normalfont\fontsize{11}{13}\selectfont

\vspace{-1.5cm} \hrule 

\section*{Abstract}
The rapid growth of deep learning research, including within the field of computational mechanics, has resulted in an extensive and diverse body of literature. To help researchers identify key concepts and promising methodologies within this field, we provide an overview of deep learning in deterministic computational mechanics. Five main categories are identified and explored: simulation substitution, simulation enhancement, discretizations as neural networks, generative approaches, and deep reinforcement learning. This review focuses on deep learning methods rather than applications for computational mechanics, thereby enabling researchers to explore this field more effectively. As such, the review is not necessarily aimed at researchers with extensive knowledge of deep learning --- instead, the primary audience is researchers at the verge of entering this field or those who attempt to gain an overview of deep learning in computational mechanics. The discussed concepts are, therefore, explained as simple as possible.

\vspace{0.25cm}
\noindent\textit{Keywords:} 
Deep learning; Computational mechanics, Neural networks; Surrogate model; Physics-informed; Generative 
\vspace{0.25cm}

\renewcommand{\contentsname}{Contents}
\tableofcontents

\newpage 
\section{Introduction}\label{sec:introduction}
\subsection{Motivation}\label{ssec:motivation}
In recent years, access to enormous quantities of data combined with rapid advances in machine learning has yielded outstanding results in computer vision, recommendation systems, medical diagnosis, and financial forecasting~\cite{abu-mostafa_learning_2012}. Nonetheless, the impact of learning algorithms reaches far beyond and has already found its way into many scientific disciplines~\cite{adie_deep_2018}. \\

\begin{figure} [htb]
	\centering
	\begin{subfigure}[t]{0.47\textwidth}
		\includegraphics[width=\textwidth]{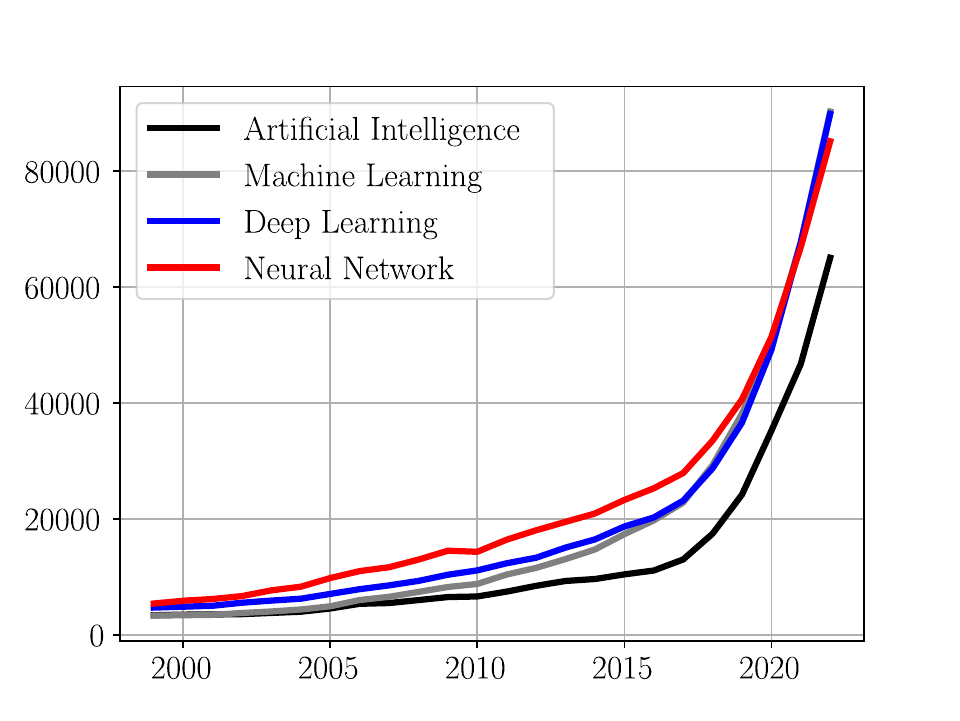}
		\caption{Publications in all fields.}
		\label{fig:AIpublications1}
	\end{subfigure}
	\hfill
	\begin{subfigure}[t]{0.47\textwidth}
		\includegraphics[width=\textwidth]{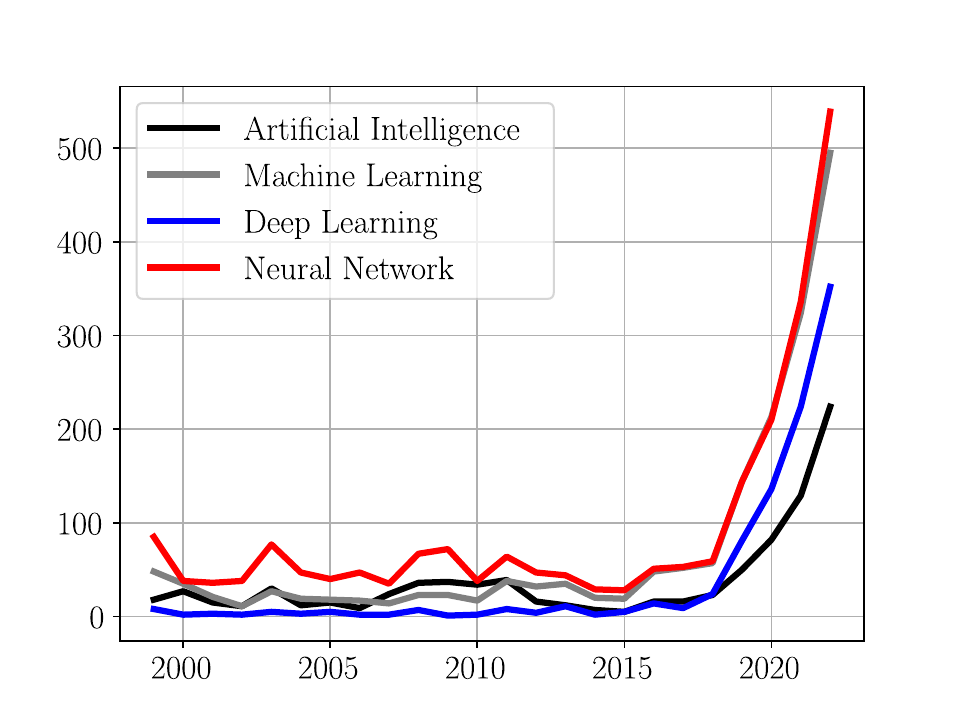}
		\caption{Publications within Computational Mechanics.}
		\label{fig:AIpublications2}
	\end{subfigure}
	\caption{Number of publications concerning artificial intelligence and some of its subtopics since 1999. Illustration inspired by~\cite{yagawa_computational_2023}.}
	\label{fig:AIpublications}
\end{figure}

The rapid interest in machine learning in general and within computational mechanics is well documented in the scientific literature. By considering the number of publications treating ``Artificial Intelligence'', ``Machine Learning'', ``Deep Learning'', and ``Neural Networks'', the interest can be quantified. \Cref{fig:AIpublications1} shows the trend in all journals of Elsevier and Springer since 1999, while~\Cref{fig:AIpublications2} depicts the trend within the computational mechanics community by considering representative journals\footnote{The considered journals are \textit{Computer Methods in Applied Mechanics and Engineering, Computers \& Mathematics with Applications, Computers \& Structures, Computational Mechanics, Engineering with Computers, Journal of Computational Physics}.} at Elsevier and Springer. The trends before 2017 differ slightly, with a steady growth in general but only limited interest within computational mechanics. However, around 2017, both curves show a shift in trend, namely a vast increase in publications highlighting the interest and potential prospects of artificial intelligence and its subtopics for a variety of applications.

\subsection{Taxonomy of Deep Learning Techniques in Computational Mechanics}\label{ssec:categories}

Due to the rapid growth~\cite{zhang_ai_2022} in deep learning research, as also seen in~\Cref{fig:AIpublications1}, we provide an overview of the various deep learning methodologies in deterministic computational mechanics. Numerous review articles on deep learning for specific applications have already emerged (see~\cite{woldseth_use_2022, shin_topology_2023} for topology optimization, \cite{adler_deep_2021} for full waveform inversion, \cite{garnier_review_2019,duraisamy_turbulence_2019, brunton_machine_2020,cai_physics-informed_2021,calzolari_deep_2021} for fluid mechanics, \cite{bock_review_2019} for continuum mechanics, \cite{bishara_state---art_2023} for material mechanics, \cite{rosenkranz_comparative_2023} for constitutive modeling, \cite{regenwetter_deep_2022} for generative design, \cite{moosavi_role_2020} for material design, and \cite{faller_neural_1996} for aeronautics)\footnote{For introductory textbooks, see~\cite{thuerey_physics-based_2021,kollmannsberger_deep_2021,brunton_data-driven_2022,karpatne_knowledge_2022,yagawa_computational_2023}.}. The aim of this work is, however, to focus on the general methods rather than applications, where similar methods are often applied to different problems. This has the potential to bridge gaps between scientific communities by highlighting similarities between methods and thereby establishing clarity on the state-of-the-art. We propose the following taxonomy in order to discuss the deep learning methods in a structured manner:
\begin{itemize}
	\item \textbf{simulation substitution} (\Cref{sec:simulationsubstitution})
	\begin{itemize}
	    \item \textbf{data-driven modeling} (\Cref{ssec:datadrivensurrogatemodeling})
	    \item \textbf{physics-informed learning} (\Cref{ssec:pinns})
	\end{itemize}
	\item \textbf{simulation enhancement} (\Cref{sec:simulationenhancement})
	\item \textbf{discretizations as neural networks} (\Cref{sec:discretizationsasNNs})
	\item \textbf{generative approaches} (\Cref{sec:generativeapproaches})
	\item \textbf{deep reinforcement learning} (\Cref{sec:deepreinforcement})
\end{itemize}
\textbf{Simulation substitution} replaces the entire simulation with a surrogate model, which in this work are neural networks (NNs). The model can be trained with supervised learning, which purely relies on labeled data and therefore is referred to as \textbf{data-driven modeling}. The generalization errors of these models can be reduced by \textbf{physics-informed learning}. Here, physics constraints are imposed on the learnable space such that only physically admissible solutions are learned. \\

\textbf{Simulation enhancement} instead only replaces components of the simulation chain, while the remaining parts are still handled by classical methods. Approaches within this category are strongly linked to their respective applications and will, therefore, be presented in the context of their specific use cases. Both data-driven and physics-informed approaches will be discussed. \\

Treating \textbf{discretizations as neural networks} is achieved by constructing a discretization from the basic building blocks of NNs, i.e., linear transformations and non-linear activation functions. Thereby, techniques within deep learning frameworks -- such as automatic differentiation, gradient-based optimization, and efficient GPU-based parallelization -- can be leveraged to improve classical simulation techniques. \\

\textbf{Generative approaches} deal with creating new content based on a data set. The goal is not, however, to recreate the data, but to generate statistically similar data. This is useful in diversifying the design space or enhancing a data set to train surrogate models. \\

Finally, in \textbf{deep reinforcement learning}, an agent learns how to interact with an environment in order to maximize rewards provided by the environment. In the case of deep reinforcement learning, the agent is modeled with NNs. In the context of computational mechanics, the environment is modeled by the governing physical equations. Reinforcement learning provides an alternative to gradient-based optimization, which is useful when gradient information is not available. 

\subsection{Deep Learning}
Before continuing with the topics specific to computational mechanics, NNs\footnote{see~\cite{goodfellow_deep_2016} for an in-depth treatment and PyTorch~\cite{paszke_pytorch_2019} or TensorFlow~\cite{martin_abadi_tensorflow_2015} for deep learning libraries} and the notation used throughout this work are briefly introduced. In essence, NNs are function approximators that are capable of approximating any continuous function~\cite{hornik_multilayer_1989}. The NN parametrized by the parameters $\boldsymbol{\theta}$ learns a function $\hat{y}=f_{NN}(x;\boldsymbol{\theta})$, which approximates the relation $y=f(x)$. The NN is constructed with nested linear transformations in combination with non-linear activation functions $\sigma$. The quality of prediction is determined by a cost function $C(\hat{y})$, which is to be minimized. Its gradients $\nabla_{\boldsymbol{\theta}} C$ with respect to the parameters $\boldsymbol{\theta}$ are used within a gradient-based optimization~\cite{goodfellow_deep_2016, kingma_adam_2017, nocedal_numerical_2006} to update the parameters $\boldsymbol{\theta}$ and thereby improve the prediction $\hat{y}$. Supervised learning relies on labeled data $x^{\mathcal{M}}, y^{\mathcal{M}}$ to establish a cost function, while unsupervised learning does not rely on labeled data. The parameters defining the user-defined training algorithm and NN architecture are referred to as hyperparameters.

\newtheorem{remark}{Notational Remark}
\begin{remark}
\normalfont
Although $x$ and $y$ may denote vector-valued quantities, we do not use bold-faced notation for them. Instead, this is reserved for 
all $N$ degrees of freedom within a problem, i.e., $\boldsymbol{x} = \{x_i\}_{i=1}^{N}$, $\boldsymbol{y} = \{y_i\}_{i=1}^{N}$. This can, for instance, be in the form of a domain $\Omega$ sampled with $N$ grid points or systems composed of $N$ degrees of freedom.
Note however, that matrices will still be denoted with capital letters in bold face.
\end{remark}

\begin{remark}
\normalfont A multitude of NN architectures will be discussed throughout this work, for which we introduce abbreviations and subscripts. Most prominent are fully-connected NNs $F_{FNN}$ (FC-NNs)~\cite{rosenblatt_perceptron_1958,goodfellow_deep_2016}, convolutional NNs $f_{CNN}$ (CNNs)~\cite{lecun_handwritten_1989, lecun_backpropagation_1989,lecun_gradient-based_1998}, recurrent NNs $f_{RNN}$ (RNNs)~\cite{rumelhart_learning_1986, hochreiter_long_1997, cho_learning_2014}, and graph NNs $f_{GNN}$ (GNNs)~\cite{kipf_semi-supervised_2017, monti_dual-primal_2018,battaglia_relational_2018}\footnote{Another architecture worth mentioning, as it has recently been applied for regression~\cite{henkes_spiking_2022,tandale_spiking_2023} are spiking NNs~\cite{gerstner_spiking_2002} specialized to run on neuromorphic hardware and thereby reduce memory and energy consumption. These are however not treated in this work.}. If the network architecture is independent of the method, the network is denoted as $f_{NN}$.
\end{remark}

\section{Simulation Substitution}\label{sec:simulationsubstitution}
In the field of computational mechanics, numerical procedures are developed to solve or find partial differential equations (PDEs). A generic PDE can be written as
\begin{equation}
\mathcal{N}[u;\lambda]=0, \qquad \text{on }\Omega\times\mathcal{T},\label{eq:pde}
\end{equation}
where a non-linear operator $\mathcal{N}$ acts on a solution $u(x,t)$ of a PDE as well as the coefficients $\lambda(x,t)$ of the PDE in the spatio-temporal domain $\Omega\times \mathcal{T}$. In the forward problem, the solution $u(x,t)$ is to be computed, while the inverse problem considers either the non-linear operator $\mathcal{N}$ or coefficients $\lambda(x,t)$ as unknowns. \\

A further distinction is made between methods treating the temporal dimension $t$ as a continuum, as in space-time approaches~\cite{hughes_space-time_1988} (\Cref{sssec:spacetime1,sssec:spacetime2})\footnote{Static problems without time-dependence can only be treated by the space-time approaches.}, or in discrete sequential time steps, as in time-stepping procedures (\Cref{sssec:timestepping1,sssec:timestepping2}). For simplicity, but without loss of generality, time-stepping procedures will be presented on PDEs with a first order derivative with respect to time:
\begin{equation}
\frac{\partial u}{\partial t} = \mathcal{N}[u;\lambda], \qquad \text{on }\Omega\times\mathcal{T}.\label{eq:timepde}
\end{equation}
Another task in computational mechanics is the forward modeling and identification of systems of ordinary differential equations (ODEs). For this, we will consider systems of the following form:
\begin{equation}
    \frac{d\boldsymbol{x}(t)}{dt} = \boldsymbol{f}(\boldsymbol{x}(t)).\label{eq:timesystem}
\end{equation}
Here, $\boldsymbol{x}(t)$ are the time-dependent degrees of freedom and $\boldsymbol{f}$ is the right-hand side defining the system of equations.\footnote{Note that a spatial discretization of the PDE~\Cref{eq:timepde} can also be written as a system of ODEs.} Both the forward problem of computing $\boldsymbol{x}(t)$ and the inverse problem of identifying $\boldsymbol{f}$ will be discussed in the following.

\subsection{Data-Driven Modeling}\label{ssec:datadrivensurrogatemodeling}
Data-driven modeling relies entirely on labeled data $x^{\mathcal{M}}, y^{\mathcal{M}}$. The NN learns the mapping between $x^{\mathcal{M}}$ and $y^{\mathcal{M}}$ with $\hat{y}_i=f_{NN}(x_i;\boldsymbol{\theta})$. Thereby an interpolation to yet unseen datapoints is established. A data-driven loss $\mathcal{L}_{\mathcal{D}}$, such as the mean squared error, for example, can be used as cost function $C$.
\begin{equation}
    C = \mathcal{L}_{\mathcal{D}}=\frac{1}{2N_{\mathcal{D}}} \sum_{i=1}^{N_{\mathcal{D}}} ||\hat{y}_i - y_i^{\mathcal{M}} ||^2_2 \label{eq:genericcost}
\end{equation}

\subsubsection{Space-Time Approaches}\label{sssec:spacetime1}
To declutter the notation, but without loss of generality, the temporal dimension $t$ is dropped in this section, as it is possible to treat it like any other spatial dimension $x$ in the scope of these methods. The goal of the upcoming methods is to either learn a forward operator $\hat{u}=F[\lambda; x]$, an inverse operator for the coefficients $\hat{\lambda} = I[u; x]$, or an inverse operator for the non-linear operator $\hat{\mathcal{N}} = O[u; \lambda; x]$.\footnote{Note that $u$ might only be partially known on the domain $\Omega$ for inverse problems.} The methods will be explained using the forward operator, but they apply analogously to the inverse operators. Only the inputs and outputs differ. \\

The solution prediction $\hat{u}_i$ at coordinate $x_i$ or $\boldsymbol{\hat{u}}_i$ on the entire domain $\Omega$ is made based on a set of inverse coefficients $\boldsymbol{\lambda}_i$. The cost function $C$ is formulated analogously to~\Cref{eq:genericcost}:
\begin{equation}
    C=\mathcal{L}_{\mathcal{D}}=\frac{1}{2N_{\mathcal{D}}} \sum_{i=1}^{N_{\mathcal{D}}} || \hat{u}_i - u_i^{\mathcal{M}} ||_2^2 \qquad \textrm{or} \qquad C=\mathcal{L}_{\mathcal{D}}=\frac{1}{2N_{\mathcal{D}}} \sum_{i=1}^{N_{\mathcal{D}}} || \boldsymbol{\hat{u}}_i - \boldsymbol{u}_i^{\mathcal{M}} ||_2^2. \label{eq:datadriven}
\end{equation}

\begin{figure}[htb]
	\centering
	\begin{tikzpicture}
	\foreach {\x} in {0,0.75,...,1.5} {
		\foreach {\y} in {0,0.75} {
			\draw[line width=0.3mm] (\x,\y) rectangle (\x+0.75,\y+0.75);
			\draw[line width=0.3mm, black, fill=white] (\x,\y) circle (0.1cm);
			\draw[line width=0.3mm, black, fill=white] (\x+0.75,\y) circle (0.1cm);
			\draw[line width=0.3mm, black, fill=white] (\x,\y+0.75) circle (0.1cm);
			\draw[line width=0.3mm, black, fill=white] (\x+0.75,\y+0.75) circle (0.1cm);
		}
	}
	\foreach {\x} in {0,0.75,...,2.25} {
		\foreach {\y} in {0,0.75,1.5} {
			\draw[line width=0.3mm] (\x+3.5,\y-0.375) rectangle (\x+0.75+3.5,\y+0.75-0.375);
		}
	}     
	\node at (1.125,2.3) {Finite Element Mesh};
	\node at (1.5+3.5,2.3) {Nodes as Pixels};
	\end{tikzpicture}
	\caption{Representation of nodes of a Cartesian grid as pixels in an image. Adapted from~\cite{zhang_label-free_2023}.}\label{fig:pixels}
\end{figure}
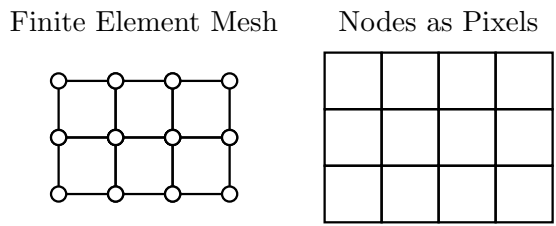

\paragraph{Fully-Connected Neural Networks}\mbox{}\\
The simplest procedure is to approximate the operator $F$ with a FC-NN $F_{FNN}$.
\begin{equation}
\hat{u}(x) = F_{FNN}(\boldsymbol{\lambda}; x; \boldsymbol{\theta}) 
\end{equation}
Example applications are flow classification~\cite{alsalman_training_2018,colvert_classifying_2018}, fluid flow in turbomachinery~\cite{pierret_turbomachinery_1999}, dynamic beam displacements from previous measurements~\cite{vurtur_badarinath_machine_2021}, wall velocity predictions in turbulence~\cite{lee_application_1997}, heat transfer~\cite{jambunathan_evaluating_1996}, prediction of source terms in turbulence models~\cite{tracey_machine_2015}, full waveform inversion~\cite{ramuhalli_electromagnetic_2002,araya-polo_deep-learning_2018,kim_geophysical_2018}, and topology optimization based on moving morphable bars~\cite{hoang_data-driven_2022}. The approach is however limited to simple problems, as an abundance of data is required. Therefore, several improvements have been proposed.
\paragraph{Image-To-Image Mapping}\mbox{}\\
One downside of the application of fully-connected NNs to problems in computational mechanics is that they often need to learn spatial relationships with respect to $x$ from scratch. CNNs inherently account for these spatial relationships due to their kernel-based structure. Therefore, image-to-image mappings using CNNs have been proposed, where an image, i.e., a uniform grid (see~\Cref{fig:pixels}) of the coefficients $\boldsymbol{\lambda}$, is used as input.
\begin{equation}
\boldsymbol{\hat{u}} = F_{CNN}(\boldsymbol{\lambda};\boldsymbol{\theta})
\end{equation}
This results in a prediction of the solution $\boldsymbol{\hat{u}}$ throughout the entire image, i.e., the domain.  

Applications include pressure and velocity predictions around airfoils~\cite{thuerey_deep_2020, chen_numerical_2021,chen_deep_2021, chen_towards_2023}, stress predictions from geometries and boundary conditions~\cite{khadilkar_deep_2019,nie_stress_2020}, steady flow predictions~\cite{guo_convolutional_2016}, detection of manufacturing features~\cite{zhang_featurenet_2018,williams_design_2019}, full waveform inversion~\cite{wu_inversionet_2018,wang_velocity_2018,yang_deep-learning_2019,zheng_applications_2019,araya-polo_deep_2019,das_convolutional_2019, wang_velocity_2020,li_deep-learning_2020,wu_seismic_2020, park_automatic_2020,ye_automatic_2022,rao_quantitative_2023}, and topology optimization~\cite{lin_investigation_2018,yu_deep_2019,abueidda_topology_2020,nakamura_deep_2020,zhang_deep_2020,zheng_generating_2021,zheng_accurate_2021,wang_deep_2022,yan_deep_2022,seo_topology_2023}. An important choice in the design of the learning algorithm is the encoding of the input data. In the case of geometries and boundary conditions, binary representations are the most straightforward approach. These are however challenging for CNNs, as discussed in~\cite{guo_convolutional_2016}. Signed distance functions~\cite{guo_convolutional_2016} or simulations on coarse grids provide superior alternatives. For inverse problems, an initial forward simulation of an initial guess of the inverse field can be used to encode the desired boundary conditions~\cite{zhang_deep_2020,wang_deep_2022,yan_deep_2022,seo_topology_2023}. Another possibility for CNNs is a decomposition of the domain. The mapping can be performed on the full domain~\cite{sosnovik_neural_2019}, smaller subdomains~\cite{joo_unit_2021}, or even individual pixels~\cite{kallioras_accelerated_2020}. In the latter two cases, interfaces require special treatment.

\paragraph{Model Order Reduction Encoding}\label{par:mor}\mbox{}\\
The disadvantage of CNN mappings is being constrained to uniform grids on rectangular domains. This can be circumvented by using GNNs, such as in~\cite{sanchez-gonzalez_learning_2020, pfaff_learning_2021,perera_graph_2022}, or point cloud-based NNs~\cite{qi_pointnet_2017,groueix_atlasnet_2018}, such as in~\cite{cunningham_investigation_2019}. To further aid the learning, the NN can be applied to a lower-dimensional space that is able to capture the data. For complex problems, mappings $e$ to low-dimensional spaces (also referred to as latent space or latent vector) $\boldsymbol{h}$ can be identified with model order reduction techniques. Thus, in the case of simulation substitution, a low-dimensional encoding $\boldsymbol{h}^\lambda=e(\boldsymbol{\lambda})$ of $\boldsymbol{\lambda}$ is identified. This is provided as input to a NN to predict the solution field $\boldsymbol{h}^u$ in a reduced latent space. The full solution field $\boldsymbol{u}$ is obtained in a decoding $d=e^{-1}$ step. The prediction is given as
\begin{equation}
\boldsymbol{\hat{u}} = d(\boldsymbol{\hat{h}}^u) = d\bigl(F_{NN}(\boldsymbol{h}^\lambda; \boldsymbol{\theta})\bigr) = d\Bigl(F_{NN}\bigl(e(\boldsymbol{\lambda});\boldsymbol{\theta}\bigr)\Bigr).
\end{equation} 

The dimensional reduction can, e.g., be performed with principal components analysis~\cite{jolliffe_principal_2002,heimann_statistical_2009}, as proposed in~\cite{bhattacharya_model_2021}, proper orthogonal decomposition~\cite{berkooz_proper_1993}, or reduced manifold learning~\cite{munoz_manifold_2023}. These techniques have been applied to learning aortic wall stresses~\cite{liang_deep_2018}, arterial wall stresses~\cite{madani_bridging_2019}, flow velocities in viscoplastic flow~\cite{muravleva_application_2018}, and the inverse problem of identifying unpressurized geometries from pressurized geometries~\cite{liang_machine_2018}. Currently, the most impressive results in data-driven surrogate modeling are achieved with model order reduction encodings combined with NNs~\cite{derouiche_data-driven_2021,hernandez_thermodynamics-informed_2023}, which can be combined with most other methodologies presented in this work.\\

Another dimensionality reduction technique are autoencoders~\cite{hinton_reducing_2006}, where $e$ and $d$ are modeled by NNs\footnote{Note that the autoencoder is modified, as it does not perform an identity mapping. Nonetheless, the idea of mapping to a reduced latent state is exploited.}. These are treated in detail in~\Cref{ssec:autoencoders} and enable non-linear encodings. An early investigation is presented in~\cite{milano_neural_2002}, where proper orthogonal decomposition is related to NNs. Application areas are the prediction of designs of acoustic scatterers from the reduced latent space~\cite{nair_grids-net_2023}, or mappings from dynamic responses of bridges to damage~\cite{fernandez-navamuel_supervised_2022}. Furthermore, it has to be stated that many of the image-to-image mapping techniques rely on NN architectures inspired by autoencoders, such as U-nets~\cite{ronneberger_u-net_2015,zhou_unet_2018}. 


\paragraph{Neural Operators}\mbox{}\\
The most recent trend in surrogate modeling with NNs are neural operators~\cite{lu_comprehensive_2022}, which map between function spaces instead of functions. Neural operators rely on the extension of the universal approximation theorem~\cite{hornik_multilayer_1989} to non-linear operators~\cite{chen_universal_1995}. The two most prominent neural operators are DeepONets\footnote{Originally proposed in~\cite{chen_universal_1995} with shallow NNs.}~\cite{lu_learning_2021} and Fourier neural operators~\cite{li_fourier_2021}.

\begin{figure}[htb]
		\centering
		\begin{tikzpicture}
		\node (B0) [draw, thick] at (0,0) {\begin{tabular}{c} $\boldsymbol{\lambda}$ \end{tabular}};
		\node (B1) [draw, thick] at (2.5,0) {\begin{tabular}{c} BranchNet$(\boldsymbol{\theta})$ \end{tabular}};
		\node (B2) [draw, thick] at (5,0) {\begin{tabular}{c} $\boldsymbol{\hat{b}}$ \end{tabular}};
		
		\node (T0) [draw, thick] at (0,-2) {\begin{tabular}{c} $x$ \end{tabular}};
		\node (T1) [draw, thick] at (2.5,-2) {\begin{tabular}{c} TrunkNet$(\boldsymbol{\theta})$ \end{tabular}};
		\node (T2) [draw, thick] at (5,-2) {\begin{tabular}{c} $\boldsymbol{\hat{t}}$ \end{tabular}};
		
		\node (G0) [draw, thick] at (8.5,-1) {\begin{tabular}{c} $\hat{u}(\boldsymbol{\lambda};x)=\sum_{i=1}^{N_P} \hat{b}_i \hat{t}_i$ \end{tabular}};
		
		\draw [line width=0.4mm,,->] (B0.east) -- (B1.west);
		\draw [line width=0.4mm,,->] (B1.east) -- (B2.west);
		\draw [line width=0.4mm,,->] (T0.east) -- (T1.west);
		\draw [line width=0.4mm,,->] (T1.east) -- (T2.west);
		\draw [line width=0.4mm,,->] (B2.south) -- (G0.west);
		\draw [line width=0.4mm,,->] (T2.north) -- (G0.west);
		\end{tikzpicture}
		\caption{DeepONet, operator learning via prediction of the basis functions $\boldsymbol{\hat{t}}$ and the corresponding coefficients $\boldsymbol{\hat{b}}$~\cite{lu_learning_2021}.}\label{fig:deepONet}
\end{figure}
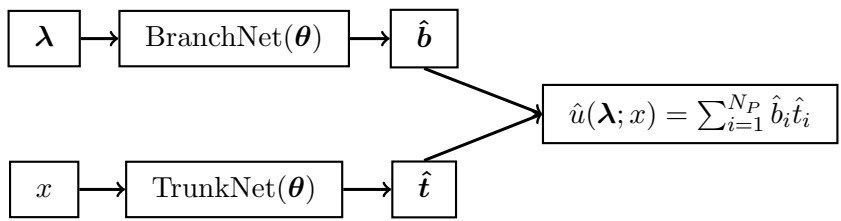

\subparagraph{DeepONet}\mbox{}\\
In DeepONets~\cite{lu_learning_2021}, illustrated in~\Cref{fig:deepONet}, the task of predicting the operator $\hat{u}(\boldsymbol{\lambda}; x)$ is split up into two sub-tasks:
\begin{itemize}
    \item the prediction of $N_P$ basis functions $\boldsymbol{\hat{t}}(x)$ (TrunkNet),
    \item the prediction of the corresponding $N_P$ problem-specific coefficients $\boldsymbol{\hat{b}}(\boldsymbol{\lambda})$ (BranchNet).
\end{itemize}
The basis is predicted by the TrunkNet with parameters $\boldsymbol{\theta}^T$ via an evaluation at coordinates $x$. The coefficients are estimated from the coefficients $\boldsymbol{\lambda}$ using the BranchNet parametrized by $\boldsymbol{\theta}^B$ and, thus, specific to the problem being solved. Taking the dot product over the evaluated basis and the coefficients yields the solution prediction $\hat{u}(\boldsymbol{\lambda}; x)$.
\begin{align}
\boldsymbol{\hat{t}}(x) &= F^T_{FNN}(x;\boldsymbol{\theta}^T) \\
\boldsymbol{\hat{b}}(\boldsymbol{\lambda}) &= F^B_{FNN}(\boldsymbol{\lambda}; \boldsymbol{\theta}^B) \\
\hat{u}(x) &= \boldsymbol{\hat{b}}(\boldsymbol{\lambda}) \cdot \boldsymbol{\hat{t}}(x)
\end{align}

Applications can be found in~\cite{lin_seamless_2021,mao_deepmmnet_2021,di_leoni_deeponet_2021,cai_deepmmnet_2021,lin_operator_2021,yin_simulating_2022,osorio_forecasting_2022,goswami_neural_2022,koric_deep_2023,clark_di_leoni_neural_2023,koric_data-driven_2023,liu_operator_2023,ahmed_multifidelity_2023}. DeepONets have also been extended with physics-informed loss functions~\cite{wang_learning_2021,goswami_physics-informed_2022,goswami_physics-informed_2022-1}.





\subparagraph{Fourier Neural Operators}\mbox{}\\
Fourier neural operators~\cite{li_fourier_2021} predict the solution $\boldsymbol{\hat{u}}$ on a uniform grid $\boldsymbol{x}$ from the spatially varying coefficients $\boldsymbol{\lambda}=\lambda(\boldsymbol{x})$. As the aim is to learn a mapping between functions, sampled on the entire domain, non-local mappings can be performed at each layer~\cite{kovachki_universal_2021}. For example, mappings such as integral kernels~\cite{li_neural_2020,li_multipole_2020}, Laplace transformations~\cite{cao_lno_2023}, and Fourier transforms~\cite{li_fourier_2021} can be employed. These transformations enhance the non-local expressivity of the NN~\cite{kovachki_universal_2021}, where Fourier transforms are particularly favorable due to the computational efficiency achievable through fast Fourier transforms. \\

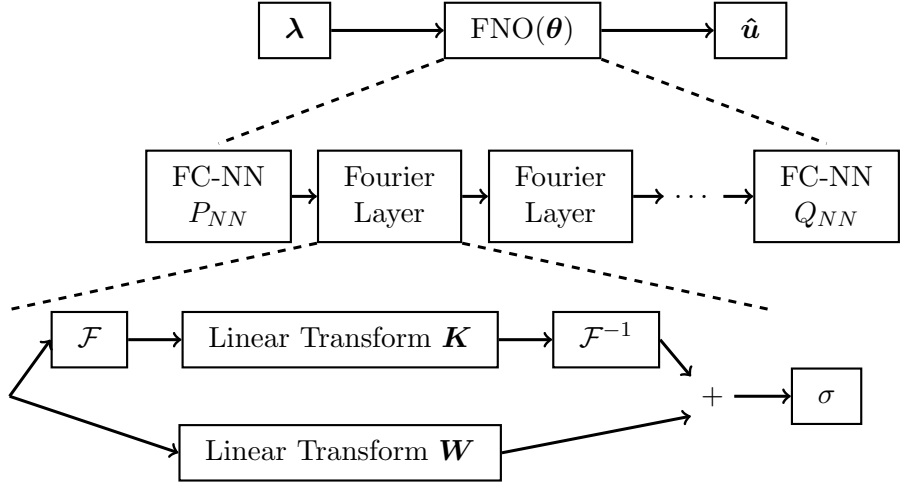
\begin{figure}[htb]
		\centering
		\begin{tikzpicture}
		\node (I0) [draw, thick] at (0,0) {\begin{tabular}{c} $\boldsymbol{\lambda}$ \end{tabular}};
		\node (I1) [draw, thick] at (3,0) {\begin{tabular}{c} FNO$(\boldsymbol{\theta})$ \end{tabular}};
		\node (I2) [draw, thick] at (6,0) {\begin{tabular}{c} $\boldsymbol{\hat{u}}$ \end{tabular}};
		
		\draw [line width=0.4mm,,->] (I0.east) -- (I1.west);
		\draw [line width=0.4mm,,->] (I1.east) -- (I2.west);
		
		\draw [line width=0.4mm,dashed] (I1.south east) -- (7,-1.5);
		\draw [line width=0.4mm,dashed] (I1.south west) -- (-1,-1.5);
		
		\node (F0) [draw, thick] at (-1,-2.2) {\begin{tabular}{c} FC-NN\\$P_{NN}$ \end{tabular}};
		\node (F1) [draw, thick] at (1.25,-2.2) {\begin{tabular}{c} Fourier\\Layer \end{tabular}};
		\node (F2) [draw, thick] at (3.5,-2.2) {\begin{tabular}{c} Fourier\\Layer \end{tabular}};
		\node (F3) at (5.25,-2.2) {\dots};
		\node (F4) [draw, thick] at (7,-2.2) {\begin{tabular}{c} FC-NN\\$Q_{NN}$ \end{tabular}};
		
		\draw [line width=0.4mm,,->] (F0.east) -- (F1.west);
		\draw [line width=0.4mm,,->] (F1.east) -- (F2.west);
		\draw [line width=0.4mm,,->] (F2.east) -- (F3.west);
		\draw [line width=0.4mm,,->] (F3.east) -- (F4.west);
		
		\draw [line width=0.4mm,dashed] (F1.south west) -- (1.25-5,-3.7);
		\draw [line width=0.4mm,dashed] (F1.south east) -- (1.25+5,-3.7);
		
		\node (f0) [draw, thick] at (-2.7,-4.1) {\begin{tabular}{c} $\mathcal{F}$ \end{tabular}};
		\node (f1) [draw, thick] at (0.6,-4.1) {\begin{tabular}{c} Linear Transform $\boldsymbol{K}$ \end{tabular}};
		\node (f1W) [draw, thick] at (0.6,-5.6) {\begin{tabular}{c} Linear Transform $\boldsymbol{W}$ \end{tabular}};
		
		\node (f2) [draw, thick] at (4.1,-4.1) {\begin{tabular}{c} $\mathcal{F}^{-1}$ \end{tabular}};
		\node (f3) at (5.5, -4.85)
		{$+$};
		\node (f4) [draw, thick] at (7, -4.85)
		{\begin{tabular}{c} $\sigma$ \end{tabular}};
		
		\draw [line width=0.4mm,->] (-3.75, -4.85) -- (f0.west);
		\draw [line width=0.4mm,->] (-3.75, -4.85) -- (f1W.west);
		\draw [line width=0.4mm,->] (f0.east) -- (f1.west);
		\draw [line width=0.4mm,->] (f1.east) -- (f2.west);
		\draw [line width=0.4mm,->] (f2.east) -- (f3.north west);
		\draw [line width=0.4mm,->] (f1W.east) -- (f3.south west);
		\draw [line width=0.4mm,->] (f3.east) -- (f4.west);
		
		\end{tikzpicture}
		\caption{Fourier neural operator, operator learning in the Fourier space~\cite{li_fourier_2021}.}\label{fig:FNO}
\end{figure}

The Fourier neural operator, as illustrated in~\Cref{fig:FNO}, consists of Fourier layers, where linear transformations $\boldsymbol{K}$ are performed after Fourier transforms $\mathcal{F}$ along the spatial dimensions $x$. Subsequently, an inverse Fourier transform $\mathcal{F}^{-1}$ is applied, which is added to the output of a linear transformation $\boldsymbol{W}$ performed outside the Fourier space. Thus, the Fourier transform can be skipped by the NN. The final step is an activation function $\sigma$. The manipulations within a Fourier layer to predict the next activation on the uniform grid $\boldsymbol{a}^{(j+1)}(\boldsymbol{x})$ can be written as
\begin{equation}
\boldsymbol{a}^{(j+1)}(\boldsymbol{x})=\sigma\bigg(\boldsymbol{W}\boldsymbol{a}^{(j)}(\boldsymbol{x})+\boldsymbol{b} +(\mathcal{F}^{-1}\Bigl[\boldsymbol{K}\mathcal{F}\bigl[\boldsymbol{a}^{(j)}(\boldsymbol{x})\bigr]\Bigr] \bigg),
\end{equation}
where $\boldsymbol{b}$ is the bias. Both the linear transformations $\boldsymbol{K}, \boldsymbol{W}$ and the bias $\boldsymbol{b}$ are learnable and thereby part of the parameters $\boldsymbol{\theta}$. Multiple Fourier layers can be employed, typically used in combination with an encoding network $P_{NN}$ and a decoding network $Q_{NN}$. 

Applications can be found in~\cite{zhu_fast_2021,song_high-frequency_2022,wei_small-data-driven_2022,rashid_learning_2022,zhang_fourier_2022,yan_robust_2022,wen_u-fnoenhanced_2022,peng_attention-enhanced_2022,you_learning_2022,kuang_fast_2023,costa_rocha_deep_2023}. An extension relying on the attention mechanisms of transformers~\cite{vaswani_attention_2017} is presented in~\cite{cao_choose_2021}. Analogously to DeepONets, Fourier neural operators have been combined with physics-informed loss functions~\cite{li_physics-informed_2023}. \\

\paragraph{Neural Network Approximation Power}\mbox{}\\
Despite the advancements in NN architectures\footnote{including architectures specifically designed to solve PDEs}, NN surrogates struggle to learn solutions of general PDEs. Typically, successes have only been achieved for parametrized PDEs with relatively small parameter spaces -- or in cases where accuracy, reliability, or generalization were disregarded. It has, however, been shown -- both for simple architectures such as FC-NNs~\cite{marcati_exponential_2023-1,gonon_deep_2023} as well as for advanced architectures such as DeepONets~\cite{marcati_exponential_2023} -- that NNs possess an excellent theoretical approximation power which can capture solutions of various PDEs. Currently, there are two obstacles that impede the identification of sufficiently good optima with these desirable NN parameter spaces~\cite{marcati_exponential_2023-1}:
\begin{itemize}
    \item training data: generalization error,
    \item training algorithm: optimization error.
\end{itemize}
A lack of sufficient training data leads to poor generalization. This might be alleviated through faster data generation using, e.g., faster and specialized classical methods~\cite{alvarez-aramberri_generation_2023}, or improved sampling strategies, i.e., finding the minimum number of required datapoints distributed in a specific manner to train the surrogate. Additionally, current training algorithms only converge to local optima. Research into improved optimization algorithms, such as current trends in computing better initial weights~\cite{bolager_sampling_2023} and thereby better local optima, attempts to reduce the optimization error. At the same time, training times are reduced drastically increasing the competitiveness.

\paragraph{Active Learning \& Transfer Learning}\mbox{}\\
Finally, an important machine learning technique independent of the NN architecture is active learning~\cite{cohn_active_1994}. Instead of precomputing a labeled data set, data is only provided when the prediction quality of the NN is insufficient. Furthermore, the data is not chosen arbitrarily, but only in the vicinity of the failed prediction. In computational mechanics, the prediction of the NN can be assessed with an error indicator. For an insufficient result, the results of a classical simulation are used to retrain the NN. Over time, the NN estimates improve in the respective domain of application. Due to the error indicator and the classical simulations, the predictions are reliable. Examples for active learning in computational mechanics can be found in~\cite{liu_knowledge_2021, haasdonk_new_2023, kalina_fetextrmann_2023}. \\

Another technique, transfer learning~\cite{pan_survey_2010,yosinski_how_2014}, aims at accelerating the NN training. Here, the NN is first trained on a similar task. Subsequently, it is applied to the task of interest -- where it converges faster than an untrained NN. Applications in computational mechanics can be found in~\cite{park_automatic_2020,kollmannsberger_transfer_2023}.

\subsubsection{Time-Stepping Procedures}\label{sssec:timestepping1}
For the time-stepping procedures, we will consider~\Cref{eq:timepde,eq:timesystem} in the following.
\paragraph{Recurrent Neural Networks}\mbox{}\\
The simplest approach to modeling time series data is by using FC-NNs to predict the next time step $t_{i+1}$ from the current time step $t_i$:
\begin{equation}
    \hat{u}(x,t_{i+1}) = F_{FNN}\bigl(x,t_i;u(x,t_i);\boldsymbol{\theta}\bigr).
\end{equation}
However, this approach lacks the ability to capture the temporal dependencies between different time steps, as each input is treated independently and without considering more than just the previous time step. Incorporating the sequential nature of the data can be achieved directly with RNNs. RNNs maintain a hidden state which captures information from the previous time steps, to be used for the next time step prediction. By unrolling the RNN, the entire time-history can be predicted.
\begin{equation}
    \{\hat{u}(x,t_2),\hat{u}(x,t_3),\dots,\hat{u}(x,t_N)\} = F_{RNN}(x;u(x,t_1);\boldsymbol{\theta})
\end{equation}
Shortcomings of RNNs, such as their tendency to struggle with learning long-term dependencies due to the problem of vanishing or exploding gradients, have been addressed by more sophisticated architectures such as long short-time memory networks (LSTM)~\cite{hochreiter_long_1997}, gated recurrent unit networks (GRU)~\cite{ballakur_empirical_2020}, and transformers~\cite{vaswani_attention_2017}. The concept of recurrent units has also been combined with other architectures, as demonstrated for CNNs~\cite{geneva_modeling_2020} and GNNs~\cite{chang_compositional_2017,mrowca_flexible_2018,sanchez-gonzalez_graph_2018,li_propagation_2019,sanchez-gonzalez_learning_2020,pfaff_learning_2021,lino_simulating_2021}. \\

Further applications of RNNs are full waveform inversion~\cite{alfarraj_petrophysical-property_2018,adler_deep_2019,fabien-ouellet_seismic_2020}, high-dimensional chaotic systems~\cite{vlachas_data-driven_2018}, fluid flow~\cite{hou_machine_2019, yagawa_computational_2023}, fracture propagation~\cite{perera_graph_2022}, sensor signals in non-linear dynamic systems~\cite{heindel_virtual_2021,heindel_data-driven_2022}, and settlement field predictions induced by tunneling~\cite{freitag_recurrent_2018}, which was extended to damage prediction in affected structures~\cite{cao_artificial_2020,cao_real-time_2022}. RNNs are often combined with reduced order model encodings~\cite{gruber_comparison_2022}, where the dynamics are predicted on the reduced latent space, as demonstrated in~\cite{gonzalez_deep_2018,holden_subspace_2019,fresca_deep_2020,fresca_comprehensive_2021,fresca_pod-dl-rom_2022,ren_phycrnet_2022,hu_accelerating_2022}. Further variations employ classical time-stepping schemes on the reduced latent space obtained by autoencoders~\cite{lee_model_2020,shen_high-order_2021}.

\paragraph{Dynamic Mode Decomposition}\mbox{}\\
Another approach that was formulated for system dynamics, i.e., \Cref{eq:timesystem} is dynamic mode decomposition (DMD)~\cite{schmid_dynamic_2010,tu_dynamic_2013}. The aim of DMD
is to identify a linear operator $\boldsymbol{A}$ that relates two successive snapshot matrices with $n$ time steps $\boldsymbol{X}=[\boldsymbol{x}(t_1),\boldsymbol{x}(t_2),\dots,\boldsymbol{x}(t_n)]^T, \boldsymbol{X}'=[\boldsymbol{x}(t_2),\boldsymbol{x}(t_3),\dots,\boldsymbol{x}(t_{n+1})]^T$:
\begin{equation}
\boldsymbol{X}'\approx \boldsymbol{A} \boldsymbol{X}.
\end{equation}
To solve this, the problem is reframed as a regression task. The operator $\boldsymbol{A}$ is approximated by minimizing the Frobenius norm of the difference between $\boldsymbol{X}'$ and $\boldsymbol{A}\boldsymbol{X}$. This minimization can be performed using the Moore-Penrose pseudoinverse $\boldsymbol{X}^\dagger$ (see, e.g.,~\cite{brunton_data_2017}):
\begin{equation}
\boldsymbol{A}=\underset{{\boldsymbol{A}}}{\text{arg min} }||\boldsymbol{X}'-\boldsymbol{A}\boldsymbol{X}||_F=\boldsymbol{X}'\boldsymbol{X}^\dagger. \label{eq:dmdregression}
\end{equation}
Once the operator is identified, it can be used to propagate the dynamics forward in time, approximating the next state $\boldsymbol{x}(t_{i+1})$ using the current state $\boldsymbol{x}(t_i)$:
\begin{equation}
\boldsymbol{x}(t_{i+1}) \approx \boldsymbol{A} \boldsymbol{x}(t_i).\label{eq:dmdprediction}
\end{equation}
This framework, is however, only valid for linear dynamics. DMD can be extended to handle non-linear systems through the application of Koopman operator theory~\cite{koopman_hamiltonian_1931}. According to Koopman operator theory, it is possible to represent a non-linear system as a linear one by using an infinite-dimensional Koopman operator $\mathcal{K}$ that acts on a transformed state $e(\boldsymbol{x}(t_i))$:
\begin{equation}
g(\boldsymbol{x}(t_{i+1})) = \mathcal{K} [e(\boldsymbol{x}(t_{i}))].
\end{equation}
In theory, the Koopman operator $\mathcal{K}$ is an infinite-dimensional linear transformation. In practice, however, finite-dimensional approximations are employed. This approach is, for example utilized in the extended DMD~\cite{williams_datadriven_2015}, where the regression from~\Cref{eq:dmdregression} is performed on a higher-dimensional state $\boldsymbol{h}(t_{i}) = e(\boldsymbol{x}(t_{i}))$ relying on a dictionary of orthonormal basis functions $\boldsymbol{\psi}(\boldsymbol{x})$. Alternatively, the dictionary can be learned using NNs, i.e., $\boldsymbol{\hat{\psi}}(\boldsymbol{x})=\psi_{NN}(\boldsymbol{x};\boldsymbol{\theta})$, as demonstrated in~\cite{li_extended_2017,yeung_learning_2019}. The NN is trained by minimizing the mismatch between predicted state $\boldsymbol{\psi}(\boldsymbol{\hat{x}}(t_{i+1}))=\boldsymbol{A} \boldsymbol{\hat{\psi}}(\boldsymbol{x}(t_i))$ (\Cref{eq:dmdprediction}) and the true state in the dictionary space. Orthogonality is not required and therefore not enforced.
\begin{equation}
	C = \frac{1}{2N}\sum_{i=1}^N ||\boldsymbol{\hat{\psi}}(\boldsymbol{x}(t_{i+1})) - \boldsymbol{A} \boldsymbol{\hat{\psi}}(\boldsymbol{x}(t_i))||_2^2
\end{equation}
When the dictionary is learned, the state predictions can be reconstructed using the Koopman mode decomposition, as explained in detail in~\cite{li_extended_2017}. \\

Alternatively, the mapping to the augmented state can be performed with autoencoders, which at the same time allows for a direct map back to the original space~\cite{takeishi_learning_2017,morton_deep_2018,lusch_deep_2018,otto_linearly_2019}. Thus, an encoder learns a reduced latent space $\boldsymbol{\hat{h}}(\boldsymbol{x})=e_{NN}(\boldsymbol{x};\boldsymbol{\theta}^e)$ and a decoder learns the inverse mapping $\boldsymbol{\hat{x}}(\boldsymbol{h})=d_{NN}(\boldsymbol{h};\boldsymbol{\theta}^d)$. The networks are trained using three losses: the autoencoder reconstruction loss $\mathcal{L}_{\mathcal{A}}$, the linear dynamics loss $\mathcal{L}_{\mathcal{R}}$, and the future state prediction loss $\mathcal{L}_{\mathcal{F}}$.
\begin{align}
\mathcal{L}_{\mathcal{A}} &= \frac{1}{2 (n+1)} \sum_{i=1}^{n+1} ||\boldsymbol{x}(t_i) - d_{NN}(e_{NN}(\boldsymbol{x}(t_i);\boldsymbol{\theta}^e);\boldsymbol{\theta}^d)||_2^2 \\
\mathcal{L}_{\mathcal{R}} &= \frac{1}{2n}\sum_{i=1}^n ||e_{NN}(\boldsymbol{x}(t_{i+1});\boldsymbol{\theta}^e) - \boldsymbol{A} e_{NN}(\boldsymbol{x}(t_i);\boldsymbol{\theta}^e)||_2^2 \\
\mathcal{L}_{\mathcal{F}} &= \frac{1}{2n}\sum_{i=1}^n||\boldsymbol{x}(t_{i+1}) - d_{NN}(\boldsymbol{A} e_{NN}(\boldsymbol{x}(t_i);\boldsymbol{\theta}^e);\boldsymbol{\theta}^d)||_2^2\\
C &= \kappa_{\mathcal{A}} \mathcal{L}_{\mathcal{A}} + \kappa_{\mathcal{R}} \mathcal{L}_{\mathcal{R}} + \kappa_{\mathcal{F}} \mathcal{L}_{\mathcal{F}}\label{eq:weigtedcost}
\end{align}
The cost function $C$ is composed of a weighted sum of the loss terms $\mathcal{L}_{\mathcal{A}},\mathcal{L}_{\mathcal{R}},\mathcal{L}_{\mathcal{F}}$ and weighting terms $\kappa_{\mathcal{A}},\kappa_{\mathcal{R}},\kappa_{\mathcal{F}}$. 
Furthermore, \cite{lusch_deep_2018} allows $\boldsymbol{A}$ to vary depending on the state. This is achieved by predicting the eigenvalues of $\boldsymbol{A}$ with an auxiliary network and constructing the matrix from these.


\subsection{Physics-Informed Learning}\label{ssec:pinns}
In supervised learning, as discussed in~\Cref{ssec:datadrivensurrogatemodeling}, the quality of prediction strongly depends on the amount of training data. Acquiring data in computational mechanics may be expensive. To reduce the amount of required data, constraints enforcing the physics have been proposed. Two main approaches exist. The physics can be enforced by modifying the cost function through a penalty term punishing unphysical predictions, thus acting as a regularizer. Possible modifications are discussed in the upcoming section. Alternatively, the physics can be enforced by construction, i.e., by reducing the learnable space to a physically meaningful space. This approach is highly specific to its application and will therefore mainly be explored in~\Cref{sec:simulationenhancement}. A brief coverage is provided in~\Cref{sssec:enorcementphysics}.

\subsubsection{Space-Time Approaches}\label{sssec:spacetime2}
Once again and without loss of generality, the temporal dimension $t$ is dropped to declutter the notation. However, in contrast to~\Cref{sssec:spacetime1}, the following methods are not equally applicable to forward and inverse problems. Thus, the prediction of the solution $\hat{u}$, the PDE coefficients $\hat{\lambda}$, and the non-linear operator $\mathcal{N}$ are treated separately.

\paragraph{Differential Equation Solving With Neural Networks}\label{par:diffEquationSolving}\mbox{}\\
The concept of solving PDEs\footnote{Typically, a single solution to a PDE is obtained. If the PDE is parametrized, multiple solutions can be obtained.} was first proposed in the 1990s~\cite{psichogios_hybrid_1992, dissanayake_neural-network-based_1994, lagaris_artificial_1998}, but was recently popularized by the so-called physics-informed neural networks (PINNs)~\cite{raissi_deep_2018} (see~\cite{karniadakis_physics-informed_2021, cuomo_scientific_2022, hao_physics-informed_2022} for recent review articles and SciANN~\cite{haghighat_sciann_2021}, SimNet~\cite{hennigh_nvidia_2021}, DeepXDE~\cite{lu_deepxde_2021} for libraries). \\

To illustrate the idea and variations of PINNs, we will consider the differential equation of a static elastic bar
\begin{equation}
\frac{d}{dx}\left(EA\frac{du}{dx}\right)+p=0, \qquad x\in\Omega. \label{eq:barequation}
\end{equation}
Here, the non-linear operator $\mathcal{N}$ is given by the left-hand side of the equation, the solution $u(x)$ is the axial displacement, and the spatially varying coefficients $\lambda(x)$ are given by the cross-sectional properties $EA(x)$ and the distributed load $p(x)$. Additionally, boundary conditions are specified, which can be in terms of Dirichlet (on $\Gamma_D$) or Neumann boundary conditions (on $\Gamma_N$):
\begin{align}
u(x) &= g(x), \qquad x\in \Gamma_D, \\
EA(x)\frac{du(x)}{dx} &= f(x), \qquad x\in \Gamma_N.
\end{align}

\subparagraph{Physics-Informed Neural Networks}\mbox{}\\
PINNs~\cite{raissi_deep_2018} approximate either the solution $u(x)$, the coefficients $\lambda(x)$, or both with FC-NNs.
\begin{align}
\hat{u}(x) &= F_{FNN}(x; \boldsymbol{\theta}^u) \\
\hat{\lambda}(x) &= I_{FNN}(x; \boldsymbol{\theta}^\lambda)
\end{align}
Instead of training the network with labeled data as in~\Cref{eq:datadriven}, the residual of the PDE is considered. The residual is evaluated at a set of $N_{\mathcal{N}}$ points, called collocation points. Taking the mean squared error over the residual evaluations yields the PDE loss
\begin{equation}
\mathcal{L}_{\mathcal{N}}=\frac{1}{2N_{\mathcal{N}}} \sum_i^{N_{\mathcal{N}}} ||\mathcal{N}[u(x_i); \lambda(x_i)]||^2_2 = \frac{1}{2N_{\mathcal{N}}} \sum_i^{N_{\mathcal{N}}}  \left( \frac{d}{dx}\left(EA(x_i)\frac{du(x_i)}{dx}\right)+p(x_i) \right)^2. \label{eq:lossPINN}
\end{equation}
The gradients of the possible predictions, i.e., $u, EA$, and $p$ with respect to $x$, are obtained with automatic differentiation~\cite{baydin_automatic_2018} through the NN approximation. 
Similarly, the boundary conditions are enforced at the $N_\mathcal{B_D}+N_\mathcal{B_N}$ boundary points.
\begin{equation}
\mathcal{L}_{\mathcal{B}}= \frac{1}{2N_{\mathcal{N}_D}} \sum_i^{N_{\mathcal{B}_D}} (u(x_i) - g)^2 + \frac{1}{2N_{\mathcal{B}_N}} \sum_i^{N_{\mathcal{B}_N}} \left( EA(x_i) \frac{du(x_i)}{dx} - f \right)^2 \label{eq:lossBC}
\end{equation}
The cost function is composed of the PDE loss $\mathcal{L}_\mathcal{N}$, boundary loss $\mathcal{L}_\mathcal{B}$, and possibly a data-driven loss $\mathcal{L}_\mathcal{D}$
\begin{equation}
C= \mathcal{L}_{\mathcal{N}} + \mathcal{L}_{\mathcal{B}} + \mathcal{L}_{\mathcal{D}}. \label{eq:costPINN}
\end{equation}
Both the deep least-squares method~\cite{cai_deep_2020} and the deep Galerkin method~\cite{sirignano_dgm_2018} are closely related. Instead of focusing on the residuals at individual collocation points as in PINNs, these methods consider the $L^2$-norm of the residuals integrated over the domain $\Omega$. 

\subparagraph{Variational Physics-Informed Neural Networks}\mbox{}\\
Computing high-order derivatives for the non-linear operator $\mathcal{N}$ is expensive. Therefore, variational PINNs~\cite{kharazmi_variational_2019, kharazmi_hp-vpinns_2021} consider the weak form of the PDE, which lowers the order of differentiation. In the case of the bar equation, the weak PDE loss is given by
\begin{equation}
\mathcal{L}_\mathcal{V} = \int_\Omega \frac{dw(x)}{dx} EA(x) \frac{du(x)}{dx}d\Omega - \int_{\Gamma_N} w(x) EA(x) \frac{du(x)}{dx} d\Gamma_N - \int_\Omega w(x) p(x) d\Omega=0, \forall w(x). \label{eq:lossVPINN}
\end{equation}
In~\cite{kharazmi_variational_2019}, trigonometric and polynomial test functions $w(x)$ are used. The cost function is obtained by replacing the PDE loss $\mathcal{L}_\mathcal{N}$ with the weak PDE loss $\mathcal{L}_\mathcal{V}$ in~\Cref{eq:costPINN}. Note that the Neumann boundary conditions are now not included in the boundary loss $\mathcal{L}_\mathcal{B}$, as they are already incorporated in the weak form in~\Cref{eq:lossVPINN}. The integrals are evaluated through numerical integration methods, such as Gaussian quadrature, Monte Carlo integration methods~\cite{morokoff_quasi-monte_1995,pharr_14_2004}, or sparse grid quadratures~\cite{novak_high_1996}. Severe inaccuracies can be introduced through the numerical integration of the NN output -- for which remedies have been proposed in~\cite{rivera_quadrature_2022}.

\subparagraph{Weak Adversarial Networks}\mbox{}\\
Instead of specifying the test functions $w(x)$, weak adversarial networks~\cite{zang_weak_2020} employ a second NN as test function
\begin{equation}
\hat{w}(x) = W_{FNN}(x;\boldsymbol{\theta}^w).
\end{equation} 
The test function is learned through a minimax optimization
\begin{equation}
\min_{\boldsymbol{\theta}^u} \max_{\boldsymbol{\theta}^w} C,
\end{equation}
where the test function $w(x)$ continually challenges the solution $u(x)$.

\subparagraph{Deep Energy Method \& Deep Ritz Method}\mbox{}\\
By minimizing the potential energy $\Pi=\Pi_i+\Pi_e$ instead, the need for test functions is overcome by the deep energy method~\cite{nguyen-thanh_deep_2019} and the deep Ritz method~\cite{e_deep_2018}. This results in the following loss term
\begin{equation}
\mathcal{L}_\mathcal{E}=\Pi_i+\Pi_e=\frac{1}{2}\int_\Omega EA(x) \left(\frac{du(x)}{dx} \right)^2 d\Omega - \int_\Gamma u(x) EA(x) \frac{du(x)}{dx} d\Gamma - \int_\Omega u(x) p(x) d\Omega. \label{eq:lossdeepenergy}
\end{equation}

Note that the inverse problem generally cannot be solved using the minimization of the potential energy. Consider, for instance, the potential energy of the bar equation in~\Cref{eq:lossdeepenergy}, which is not well-posed in the inverse setting. Here, $EA(x)$ going towards $-\infty$ in the domain $\Omega$ and going towards $\infty$ at $\Gamma_N$ minimizes the potential energy $\mathcal{L}_{\mathcal{E}}$.

\subparagraph{Extensions}\mbox{}\\
A multitude of extensions to the PINN methodology exist. For in-depth reviews, see~\cite{karniadakis_physics-informed_2021, cuomo_scientific_2022, hao_physics-informed_2022}. \\

\textbf{Learning Multiple Solutions}\\
Currently, PINNs are mainly employed to learn a single solution. As the training effort exceeds the solving effort of classical solvers, the viability of PINNs is questionable~\cite{grossmann_can_2023}. However, PINNs can also be employed to learn multiple solutions. This is achieved by providing the parametrization of the PDE, i.e., $\lambda$ as an additional input to the network, as discussed in~\Cref{sec:simulationsubstitution}. This enables a cheap prediction stage without retraining for new solutions\footnote{Importantly, the training would be without training data and would only require a definition of the parametrized PDE. Currently, this is only possible for simple PDEs with small parameter spaces.}. One possible example for this is~\cite{kashefi_physics-informed_2022}, where different geometries are captured in terms of point clouds and processed with point cloud-based NNs~\cite{qi_pointnet_2017}. \\

\textbf{Boundary Conditions}\\
The enforcement of the boundary conditions through a penalty term $\mathcal{L}_{\mathcal{B}}$ in~\Cref{eq:lossBC} leads to an unbalanced optimization, due to the competing loss terms $\mathcal{L}_{\mathcal{N}}, \mathcal{L}_{\mathcal{B}}, \mathcal{L}_{\mathcal{D}}$ in~\Cref{eq:costPINN}\footnote{Consider, for instance, a training procedure in which the PDE loss $\mathcal{L}_{\mathcal{N}}$ is first minimal, such that the PDE is fulfilled. Without fulfilment of the boundary conditions, the solution is not unique. However, the NN struggles to modify the current boundary values without violating the PDE loss and thereby increasing the total cost function $C$. The NN is thus stuck in a bad local minimum. Similar scenarios can be formulated for a too rapid minimization of the other loss terms.}. One remedy is to modify the NN output $F_{FNN}$ by multiplication of a function, such that the Dirichlet boundary conditions are satisfied a priori, i.e., $\mathcal{L}_{\mathcal{B}}=0$, as demonstrated in~\cite{berg_unified_2018, kollmannsberger_deep_2021}. 
\begin{equation}
\hat{u}(x) = G(x) + D(x) F_{FNN}(x;\boldsymbol{\theta}^u)
\end{equation}
Here, $G(x)$ is a smooth interpolation of the boundary conditions, and $D(x)$ is a signed distance function that is zero at the boundary. For Neumann boundary conditions, \cite{henkes_physics_2022} propose to predict $u$ and its derivatives $\partial u/\partial x$ with separate networks, such that the Neumann boundary conditions can be enforced strongly by modifying the derivative network. This requires an additional constraint, ensuring that the derivative predictions match the derivative of $u$. For complex domains, $G(x)$ and $D(x)$ cannot be found analytically. Therefore, \cite{berg_unified_2018} use NNs to learn $G(x)$ and $D(x)$ in a supervised manner by prescribing either the boundary values or zero at the boundary and restricting the values within the domain to be non-zero. Similarly~\cite{lagaris_neural-network_2000} proposed using radial basis function networks for $G(x)$, where $D(x)=1$ is assumed. The radial basis function networks are determined by solving a linear system of equations constructed with the boundary conditions. On uniform grids, strong enforcement can be achieved through specialized CNN kernels~\cite{ren_phycrnet_2022} with constant padding terms for Dirichlet boundary conditions and ghost cells for Neumann boundary conditions. Constrained backward propagation~\cite{ferrari_constrained_2008} has also been proposed to guarantee the enforcement of boundary conditions~\cite{rudd_constrained_2014,rudd_constrained_2015}. \\

Another possibility is to introduce weighting terms $\kappa_{\mathcal{N}}, \kappa_{\mathcal{B}}, \kappa_{\mathcal{D}}$ for each loss term. These are either hyperparameters, or they are learned during the optimization with attention mechanisms~\cite{wang_residual_2017,zhang_occluded_2018,magiera_constraint-aware_2020}. This is achieved by performing a minimax optimization with respect to all weighting terms $\boldsymbol{\kappa}=\{\kappa_{\mathcal{N}}, \kappa_{\mathcal{B}}, \kappa_{\mathcal{D}}\}$
\begin{equation}
\min_{\boldsymbol{\theta}} \max_{\boldsymbol{\kappa}} C.
\end{equation}
Expanding on this idea, each collocation point used for the loss terms can be considered an individual equality constraint~\cite{nandwani_primal_2019, mcclenny_self-adaptive_2022}. Therefore, a weighting term $\kappa_{\mathcal{N}_i}$ is allocated for each collocation point $x_i$, as illustrated for the PDE loss $\mathcal{L}_{\mathcal{N}}$ from~\Cref{eq:lossPINN}
\begin{equation}
\mathcal{L}_{\mathcal{N}} = \frac{1}{2N_{\mathcal{N}}} \sum_i^{N_\mathcal{N}} \kappa_{\mathcal{N},i} || \mathcal{N}[u(x_i);\lambda(x_i)]||^2_2.
\end{equation}
This has the added advantage that greater emphasis is assigned on more important collocation points, i.e., points which lead to larger residuals. This approach is strongly related to the approaches relying on the augmented Lagrangian method~\cite{lu_physics-informed_2021} and
competitive PINNs~\cite{zeng_competitive_2022}, where an additional NN models the penalty weights $\kappa(x)=K_{FNN}(x; \boldsymbol{\theta}^\kappa)$. This is similar to weak adversarial networks, but instead formulated using the strong form.\\



\textbf{Ansatz}\\
Another prominent topic is the question of which ansatz to choose. The type of ansatz is, for example, determined by different NN architectures (see~\cite{moser_modeling_2023} for a comparison) or combinations with classical ansatz formulations. Instead of using FC-NNs, some authors~\cite{zhu_physics-constrained_2019, geneva_modeling_2020} employ CNNs to exploit the spatial structure of the data. Irregular geometries can be handled by embedding the structure in a rectangular domain using binary encodings~\cite{han_flownet_2020} or signed distance functions~\cite{guo_convolutional_2016,bhatnagar_prediction_2019}. Another option are coordinate transformations into rectangular grids~\cite{gao_phygeonet_2021}.
The CNN requires a full-grid discretization, meaning that the coordinates $x$ are analytically independent of the prediction $\hat{u} = F_{CNN}$. Thus, the gradients of $u$ are not obtained with automatic differentiation, but with numerical differentiation, i.e., finite differences. Alternatively, the output of the CNN can represent coefficients of an interpolation, as proposed under the name spline-PINNs~\cite{wandel_spline-pinn_2022} using Hermite splines.
This again allows for an automatic differentiation. This is similarly applied for irregular geometries in~\cite{gao_physics-informed_2022}, where GNNs are used in combination with a piecewise polynomial basis. Using a classical basis has the added advantage that Dirichlet boundary conditions can be satisfied exactly. A further variation is the approximation of the coefficients of classical bases with FC-NNs. This is shown with B-splines in~\cite{moller_physics-informed_2021} in the sense of isogeometric analysis~\cite{hughes_isogeometric_2005}. This was similarly done for piecewise polynomials in~\cite{meethal_finite_2022}. However, instead of simply minimizing the PDE residual from~\Cref{eq:lossPINN} directly, the finite element discretization~\cite{hughes_finite_2000, bathe_finite_2014} is exploited. The loss $\mathcal{L}_{\mathcal{F}}$ can thus be formulated in terms of the non-linear stiffness matrix $\boldsymbol{K}$, the force vector $\boldsymbol{F}$, and the degrees of freedom $\boldsymbol{u}^h$.
\begin{equation}
\mathcal{L}_{\mathcal{F}}= ||\boldsymbol{K}(\boldsymbol{u}^h)\boldsymbol{u}^h-\boldsymbol{F}||_2^2
\end{equation}
In the forward problem, $\boldsymbol{u}^h$ is approximated by a FC-NN, whereas for the inverse problem a FC-NN predicts $\boldsymbol{K}$. Similarly, \cite{berrone_variational_2022,badia_finite_2023} map a NN onto a finite element space by using the NN evaluations at nodal coordinates as the corresponding basis function coefficents. This also allows a straightforward strong enforcement of Dirichlet boundary conditions, as demonstrated in~\cite{zhang_label-free_2023} with CNNs. The nodes are represented as pixels (see~\Cref{fig:pixels}). \\

\begin{figure}[htb]
	\centering
	\begin{tikzpicture}
	\draw[line width=0.3mm] (0,0) rectangle (0,4);
	\node at (0,-0.4) {$\boldsymbol{u}_1$};
	\draw[line width=0.3mm, black, fill=white] (0,0) circle (0.1cm);
	\node at (0.4,0) {$u_1^2$};
	\draw[line width=0.3mm, black, fill=white] (0,4) circle (0.1cm);
	\node at (0.4,4) {$u_1^1$};
	
	\draw[line width=0.3mm] (2,0) rectangle (2,4);
	\node at (2,-0.4) {$\boldsymbol{u}_2$};
	\draw[line width=0.3mm, black, fill=white] (2,0) circle (0.1cm);
	\node at (2.4,0) {$u_2^3$};
	\draw[line width=0.3mm, black, fill=white] (2,2) circle (0.1cm);
	\node at (2.4,2) {$u_2^2$};
	\draw[line width=0.3mm, black, fill=white] (2,4) circle (0.1cm);
	\node at (2.4,4) {$u_2^1$};
	
	\draw[line width=0.3mm] (4,0) rectangle (4,4);
	\node at (4,-0.4) {$\boldsymbol{u}_3$};
	\draw[line width=0.3mm, black, fill=white] (4,0) circle (0.1cm);
	\node at (4.4,0) {$u_2^5$};
	\draw[line width=0.3mm, black, fill=white] (4,1) circle (0.1cm);
	\node at (4.4,1) {$u_2^4$};
	\draw[line width=0.3mm, black, fill=white] (4,2) circle (0.1cm);
	\node at (4.4,2) {$u_2^3$};
	\draw[line width=0.3mm, black, fill=white] (4,3) circle (0.1cm);
	\node at (4.4,3) {$u_2^2$};
	\draw[line width=0.3mm, black, fill=white] (4,4) circle (0.1cm);
	\node at (4.4,4) {$u_2^1$};
	
	\draw[line width=0.3mm, ->] (0,-0.7) -- (4,-0.7);
	\node at (2,-1) {Refinement};
	
	\node (I) [draw, circle, line width=0.3mm, minimum size=0.7cm, inner sep=0] at (7, 2) {$\boldsymbol{\lambda}$};
	\node (U11) [draw, circle, line width=0.3mm, minimum size=0.7cm, inner sep=0] at (9, 4) {$u_1^1$};
	\node (U12) [draw, circle, line width=0.3mm, minimum size=0.7cm, inner sep=0] at (9, 0) {$u_1^2$};
	
	\node (U31) [draw, circle, line width=0.3mm, minimum size=0.7cm, inner sep=0] at (13, 4) {$u_3^1$};
	\node (U32) [draw, circle, line width=0.3mm, minimum size=0.7cm, inner sep=0] at (13, 3) {$u_3^2$};
	\node (U33) [draw, circle, line width=0.3mm, minimum size=0.7cm, inner sep=0] at (13, 2) {$u_3^3$};
	\node (U34) [draw, circle, line width=0.3mm, minimum size=0.7cm, inner sep=0] at (13, 1) {$u_3^4$};
	\node (U35) [draw, circle, line width=0.3mm, minimum size=0.7cm, inner sep=0] at (13, 0) {$u_3^5$};
	
	\draw [line width=0.3mm, lightgray, ->] (I.east) -- (U31.west);
	\draw [line width=0.3mm, lightgray, ->] (I.east) -- (U32.west);
	\draw [line width=0.3mm, lightgray, ->] (I.east) -- (U33.west);
	\draw [line width=0.3mm, lightgray, ->] (I.east) -- (U34.west);
	\draw [line width=0.3mm, lightgray, ->] (I.east) -- (U35.west);
	
	\node (U21) [draw, circle, line width=0.3mm, minimum size=0.7cm, inner sep=0,fill=white] at (11, 4) {$u_2^1$};
	\node (U22) [draw, circle, line width=0.3mm, minimum size=0.7cm, inner sep=0,fill=white] at (11, 2) {$u_2^2$};
	\node (U23) [draw, circle, line width=0.3mm, minimum size=0.7cm, inner sep=0,fill=white] at (11, 0) {$u_2^3$};
	
	\draw[line width=0.3mm, lightgray, ->] (I.east) -- (U11.west);
	\draw[line width=0.3mm, lightgray, ->] (I.east) -- (U12.west);
	
	\draw[line width=0.5mm, black, ->] (U11.east) -- (U21.west);
	\draw[line width=0.5mm, black, ->] (U11.east) -- (U22.west);
	\draw[line width=0.5mm, black, ->] (U12.east) -- (U22.west);
	\draw[line width=0.5mm, black, ->] (U12.east) -- (U23.west);
	
	\draw[line width=0.5mm, black, ->] (U21.east) -- (U31.west);
	\draw[line width=0.5mm, black, ->] (U21.east) -- (U32.west);
	\draw[line width=0.5mm, black, ->] (U22.east) -- (U32.west);
	\draw[line width=0.5mm, black, ->] (U22.east) -- (U33.west);
	\draw[line width=0.5mm, black, ->] (U22.east) -- (U34.west);
	\draw[line width=0.5mm, black, ->] (U23.east) -- (U34.west);
	\draw[line width=0.5mm, black, ->] (U23.east) -- (U35.west);
	

    \draw [line width=0.3mm, lightgray,->] (U22.east) -- (U33.west);
	
	\draw [line width=0.3mm, lightgray, ->] (I.east) -- (U21.west);
	\draw [line width=0.3mm, lightgray, ->] (I.east) -- (U22.west);
	\draw [line width=0.3mm, lightgray, ->] (I.east) -- (U23.west);
	
	\draw[line width=0.3mm, ->] (7,-0.7) -- (13,-0.7);
	\node at (10,-1) {Increasing Depth};
	
	\end{tikzpicture}
	\caption{Refinement expressed with NNs in terms of NN depth. Thick black lines indicate non-learnable connections and gray lines indicate learnable connections. Each added layer is composed of a projection from the coarser level and a correction obtained through the learnable connection.}\label{fig:adaptivity}
\end{figure}
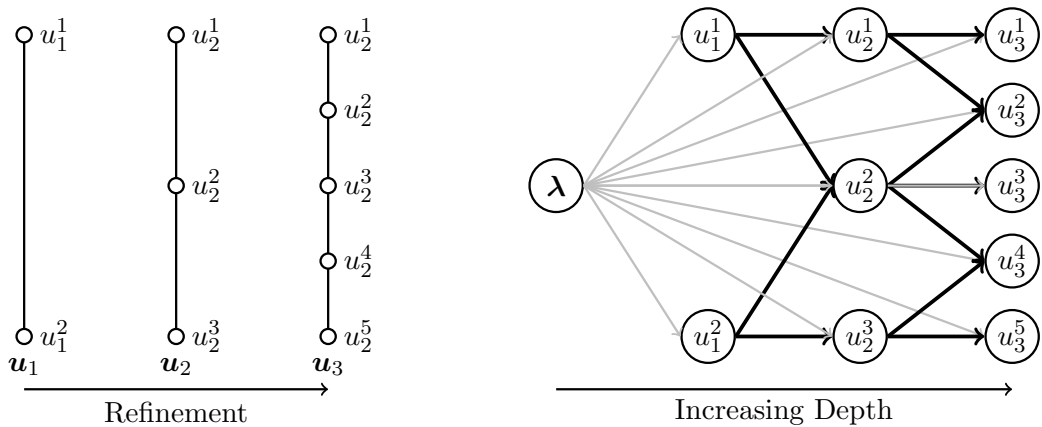

Prior information on the solution can be incorporated through a feature layer~\cite{yazdani_systems_2020}. If, for example, it is known that the solution is composed of trigonometric functions, a feature layer with trigonometric functions can be applied after the input layer. Thus, known features are given to the NN directly to aid the learning. Without known features, the task can also be modified to improve learning. Inspired by adaptivity from finite elements, refinements are progressively learned by additional layers of the NN~\cite{uriarte_finite_2022} (see~\Cref{fig:adaptivity}). 
Thus, a coarse solution $\boldsymbol{u}_1$ is learned to begin with, then refined to $\boldsymbol{u}_2$ by an additional layer, which again is refined to $\boldsymbol{u}_3$ until the deepest refinement level is reached. \\

\textbf{Domain Decomposition} \\
To improve the scalability of PINNs to more complex problems, several domain decomposition methods have been proposed. One approach are hp-variational PINNs~\cite{kharazmi_hp-vpinns_2021}, where the domain is decomposed into patches. Piecewise polynomial test functions are defined on each patch separately, while the solution is approximated by a globally acting NN. This enables a separate numerical integration of each patch, improving its accuracy. \\

In an alternative formulation, one NN can be used per subdomain. This was proposed as conservative PINNs~\cite{jagtap_conservative_2020}, where conservation laws are enforced at the interface to ensure continuity. Here, the discrepancies between both solution and flux were penalized at the interface in a least squares manner. The advantages of this approach are twofold: Firstly, parallelization is possible~\cite{shukla_parallel_2021} and, secondly, adaptivitiy can be introduced. Shallower networks can be employed for smooth solutions and deeper networks for more complex solutions. The approach was generalized for any PDE in the context of extended PINNs~\cite{karniadakis_extended_2020}. Here, the interface condition is formulated in terms of the difference in both the residual and the solution. \\

\textbf{Acceleration Methods}\mbox{}\\
Analogously to supervised learning, as discussed in~\Cref{ssec:datadrivensurrogatemodeling}, transfer learning can be applied to PINNs~\cite{chen_transfer_2021} as, e.g., demonstrated in phase-field fracture~\cite{goswami_transfer_2019} or topology optimization~\cite{he_deep_2023}. These are very suitable problems since crack and displacement fields evolve with mostly local changes in phase-field fracture. For topology optimization, only minor updates are expected between each optimization iteration~\cite{he_deep_2023}.\\

The poor performance of PINNs in their original form can also be improved with better sampling strategies. In importance sampling~\cite{nabian_efficient_2021,hanna_residual-based_2022}, the collocation point density is proportional to the value of the cost function. Alternatively, residual-based adaptive refinement~\cite{lu_deepxde_2021} adds collocation points in the vicinity of areas with a higher cost function. \\

Another essential topic for NNs is normalization of the inputs, outputs, and loss terms~\cite{kollmannsberger_physics-informed_2021,anton_identification_2021}. For time-dependent problems, it is possible to use time-dependent normalization~\cite{zong_improved_2023} to ensure that the solution is always in the same range regardless of the time step. \\
 
Furthermore, the cost function can be enhanced by including the derivative of the residual~\cite{yu_gradient-enhanced_2022} as well. The derivative should also be minimized, as both the residual and its derivative should be zero at the correct solution. However, a general problem in the cost function formulation persists. The cost function should correspond to the norm of the error, which is not necessarily the case. This means that a reduction in the cost does not necessarily yield an improvement in quality of solution. The error norm can be expressed in terms of the $H^{-1}$-norm, which, according to~\cite{taylor_deep_2023}, can efficiently be computed on rectangular domains with Fourier transforms. Thus, the $H^{-1}$-norm can directly be used as cost function and minimized. \\

Another aspect is numerical differentiation, which is advantageous for the residual of the PDE~\cite{chiu_can-pinn_2022}, as automatic differentiation may be erroneous due to spurious oscillations between collocation points. Thus, numerical differentiation enforces regularity, which was exploited in~\cite{chiu_can-pinn_2022} by coupling automatic differentiation and numerical differentiation to retain the advantages of automatic differentiation. \\

Further specialized modifications to NN architectures have been proposed. Adaptive activation functions~\cite{jagtap_adaptive_2020} have shown acceleration in convergence. Extreme learning machines~\cite{huang_extreme_2006,huang_extreme_2011} remove the need for iterations altogether. All layers are randomly initialized in extreme learning machines, and only the last layer is learnable. Without a non-linear activation function, the parameters are found with a least-squares regression. This was demonstrated for PINNs in~\cite{dong_local_2021}. Instead of only learning the last layer, the problem can be split into a non-linear and a linear regression problem, which are solved separately~\cite{dong_numerical_2022}, such that the full expressivity of NNs is retained. \\

\subparagraph{Applications To Forward Problems}\mbox{}\\
PINNs have been applied to various PDEs (see~\cite{karniadakis_physics-informed_2021, cuomo_scientific_2022, hao_physics-informed_2022} for an overview). Forward problems can, for example, be found in solid mechanics~\cite{kollmannsberger_physics-informed_2021,haghighat_physics-informed_2021,bai_introduction_2023}, fluid mechanics~\cite{kissas_machine_2020,raissi_hidden_2020,sun_surrogate_2020, jin_nsfnets_2021,cai_flow_2021,fraces_physics_2021,zhang_simulation_2022,wang_fluxnet_2023}, and thermomechanics~\cite{amini_niaki_physics-informed_2021,zhu_machine_2021}. Currently, PINNs do not outperform classical solvers such as the finite element method~\cite{markidis_old_2021, grossmann_can_2023} in terms of speed for a given accuracy of engineering relevance. In the author's experience and judgement, this is especially the case for forward problems even if the extensions mentioned above are employed. Often, the mentioned gains compared to classical forward solvers disregard the training effort and only report evaluation times. \\


Incorporating large parts of the solution in the form of measurements with the data-driven loss $\mathcal{L}_{\mathcal{D}}$ improves the performance of PINNs, which thereby can become a viable method in some cases. Yet, \cite{li_ref-nets_2022} states that data-driven methods outperform PINNs. Thus PINNs should not be regarded as a replacement for data-driven methods, but rather as a regularization technique for data-driven methods to reduce the generalization error.  \\ 



\subparagraph{Applications To Inverse Problems}\mbox{}\\
However, PINNs are in particular useful for inverse problems with full domain knowledge, i.e., the solution is available throughout the entire domain. This has, for example, been shown for the identification of material properties~\cite{chen_physics-informed_2020,zhang_physics-informed_2020,shukla_physics-informed_2020,anton_identification_2021, anton_physics-informed_2022}. 
In contrast, for inverse problems with only partial knowledge, the applicability of PINNs is limited~\cite{herrmann_use_2023}, as both forward and inverse solution have to be learned simultaneously. Most applications therefore limit themselves to simpler inversions such as size and shape optimization. Examples are published, e.g., in~\cite{rojas_parameter_2021,haghighat_physics-informed_2021,li_physics_2021,zhang_analyses_2022,depina_application_2022,xu_transfer_2023,sun_physics-informed_2023}. Exceptions that deal with the identification of entire fields can be found in full waveform inversion~\cite{rashtbehesht_physicsinformed_2022}, topology optimization~\cite{zehnder_ntopo_2021}, elasticity, and the heat equation~\cite{di_lorenzo_physics_2023}.

\paragraph{Inverse Problems}\label{par:inverseproblemspacetime}\mbox{}\\
PINNs are capable of discovering governing equations by either learning the operator $\mathcal{N}$ or the coefficients $\lambda$. The resulting operator is, however, not always interpretable, and in the case of identification of the coefficients, the underlying PDE is assumed. To discover interpretable operators, one can apply sparse regression approaches~\cite{berg_data-driven_2019}. Here, potential differential operators are assumed as an input to the non-linear operator
\begin{equation}
    \hat{\mathcal{N}}\left[x, u,\frac{\partial u}{\partial x},\frac{\partial^2 u}{\partial x^2},\dots\right]=0\label{eq:operator}.
\end{equation}
Subsequently, a NN learns the corresponding coefficients using observed solutions inserted into~\Cref{eq:operator}. The evaluation of the differential operators is achieved through automatic differentiation by first interpolating the solution with a NN. Sparsity is ensured with a $L^1$-regularization. \\

A more sophisticated and complete framework is AI-Feynman~\cite{udrescu_ai_2020}. Sequentially, dimensional analysis, polynomial regression, and brute force search algorithms are applied to identify fundamental laws in the data. If unsuccessful, a NN interpolates the data, which can thereby be queried for symmetry and separability. The identification of symmetries leads to a reduction in variables, i.e., a reduction of the input space. In the case of separability, the problem is decomposed into two subproblems. The reduced problems or subproblems are iteratively fed through the framework until an equation is identified. AI-Feynman has been successfully applied to 100 equations from the Feynman lectures~\cite{feynman_feynman_2011}.


\subsubsection{Time-Stepping Procedures}\label{sssec:timestepping2}
Again~\Cref{eq:timepde} and~\Cref{eq:timesystem} will be considered for the time-stepping procedures.

\paragraph{Physics-Informed Neural Networks}\label{par:discretetimePINN}\mbox{}\\
In the spirit of domain decomposition, parareal PINNs~\cite{meng_ppinn_2020} split up the temporal domain in subdomains $[t_i<t_{i+1}]$. A rough estimate of the solution $u$ is provided by a conjugate gradient solver on a simplified form of the PDE starting from $t_0$. PINNs are then independently applied in each subdomain to correct the estimate. Subsequently, the conjugate gradient solver is applied again, starting from $t_1$. This process is repeated until all time steps have been traversed. A closely related approach can be found in~\cite{mattey_novel_2022}, where a PINN is retrained on successive time segments. It is however ensured that previous time steps are kept fulfilled through a data-driven loss term for time segments that were already learned. \\

Another approach are the discrete-time PINNs~\cite{raissi_deep_2018}, which consider the temporal dimension in a discrete manner. The differential equation from~\Cref{eq:timepde} is discretized with the Runge-Kutta method with $q$ stages~\cite{iserles_first_2008}:
\begin{align}
u^{n+c_i} &= u^n+\Delta t \sum_{j=1}^q a_{ij} \mathcal{N}[u^{n+c_j}], \qquad i=1,\dots,q, \label{eq:rungekutta1}\\
u^{n+1} &= u^n+\Delta t \sum_{j=1}^q b_j \mathcal{N}[u^{n+c_j}], \label{eq:rungekutta2}
\end{align}
where
\begin{equation}
u^{n+c_j}(x)=u(t^n+c_j\Delta t, x), \qquad j=1,\dots,q.
\end{equation}
A NN $F_{NN}$ predicts all stages $i=1,\dots,q$ from an input $x$:
\begin{equation}
\boldsymbol{\hat{u}} = [\hat{u}^{n+c_1}(x),\dots,\hat{u}^{n+c_q}(x),\hat{u}^{n+1}(x)] = F_{NN}(x;\boldsymbol{\theta}).
\end{equation}
The cost is then constructed by rearranging~\Cref{eq:rungekutta1,eq:rungekutta2}.
\begin{align}
\hat{u}^n &= \hat{u}_i^n = \hat{u}^{n+c_i} - \Delta t \sum_{j=1}^q a_{ij} \mathcal{N}[\hat{u}^{n+c_j}],  \qquad i=1,\dots,q, \\
\hat{u}^n &= \hat{u}^n_{q+1} = \hat{u}^{n+1} - \Delta t \sum_{j=1}^q b_j \mathcal{N}[\hat{u}^{n+c_j}].
\end{align}
The $q+1$ predictions $\hat{u}_i^n, \hat{u}^n_{q+1}$ of $\hat{u}^n$ have to match the initial conditions $u^{\mathcal{M}^n}$, where the mean squared error is used as a loss function to learn all stages $\boldsymbol{\hat{u}}$. The approach has been applied to fluid mechanics~\cite{wessels_neural_2020,bai_general_2022}.


\paragraph{Inverse Problems}\mbox{}\\
As for inverse problems in the space-time approaches (\Cref{par:inverseproblemspacetime}), the non-linear operator $\mathcal{N}$ can be learned. For temporal problems, this corresponds to the right-hand side of~\Cref{eq:timepde} for PDEs and to~\Cref{eq:timesystem} for systems of ODEs. The predicted right-hand side can then be used to predict time series using a classical time-stepping scheme, as proposed in~\cite{gonzalez-garcia_identification_1998}. More sophisticated methods leaning on similar principles are presented in the following. Specifically, we will discuss PDE-Net for discovering PDEs, SINDy for discovering systems of ODEs in an interpretable sense, and an approach relying on multistep methods for systems of ODEs. The multistep approach leads to a non-interpretable, but more expressive approximation of the right-hand side.

\subparagraph{PDE-Net}\mbox{}\\
PDE-Net~\cite{long_pde-net_2018, long_pde-net_2019} is designed to learn both the system dynamics $u(x,t)$ and the underlying differential equation it follows. Given a problem of the form of~\Cref{eq:timepde}, the right-hand side can be approximated as a function of coordinates and gradients of the solution.
\begin{equation}
    \hat{\mathcal{N}}\left[x,u,\frac{\partial u}{\partial x},\frac{\partial^2 u}{\partial x^2},\dots \right] \label{eq:pdenet}
\end{equation}
The operator $\hat{\mathcal{N}}$ is approximated by NNs. The first step involves estimating spatial derivatives using learnable convolutional filters. The filters are designed to adjust their order of approximation based on the fit to the underlying measurements $u^{\mathcal{M}}$, while the type of gradient is predefined\footnote{This is enforced through constraints using moment matrices of the convolutional filters.}. Thus, the NN learns how to best approximate spatial derivatives specific to the underlying data. Subsequently, the inputs of $\hat{\mathcal{N}}$ are combined with point-wise CNNs~\cite{hua_pointwise_2018} in~\cite{long_pde-net_2018} or a symbolic network in~\cite{long_pde-net_2019}. Both yield an interpretable operator from which the analytical expression can be extracted.
In order to construct a loss function, \Cref{eq:timepde,eq:pdenet} are discretized using the forward Euler method:
\begin{equation}
u(x, t_{n+1}) = u(x, t_{n}) + \Delta t \hat{\mathcal{N}}\left[x, u, \frac{\partial u}{\partial x}, \frac{\partial^2 u}{\partial x^2}, \dots\right].
\end{equation} 
This temporal discretization is applied iteratively, and the discrepancy between the derived function and the measured data $u^{\mathcal{M}}(x, t_{n})$ serves as the loss function.

\subparagraph{SINDy}\mbox{}\\
Sparse identification of non-linear dynamic systems (SINDy)~\cite{brunton_discovering_2016} deals with the discovery of dynamic systems of the form of~\Cref{eq:timesystem}. The task is posed as a sparse regression problem. Snapshot matrices of the state $\boldsymbol{X}=[\boldsymbol{x}(t_1),\boldsymbol{x}(t_2),\dots,\boldsymbol{x}(t_n)]$ and its time derivative $\dot{\boldsymbol{X}}=[\dot{\boldsymbol{x}}(t_1),\dot{\boldsymbol{x}}(t_2),\dots,\dot{\boldsymbol{x}}(t_n)]$ are related to one another via candidate functions $\boldsymbol{\Theta}(\boldsymbol{X})$ evaluated at $\boldsymbol{X}$ using unknown coefficients $\boldsymbol{\Xi}$:
\begin{equation}
\dot{\boldsymbol{X}}=\boldsymbol{\Theta}(\boldsymbol{X})\boldsymbol{\Xi}. \label{eq:sindyloss}
\end{equation}
The coefficients $\boldsymbol{\Xi}$ are determined through sparse regression, such as sequential thresholded least squares or LASSO regression. By including partial derivatives, SINDy has been extended to the discovery of PDEs~\cite{rudy_data-driven_2017,schaeffer_learning_2017}. \\

The expressivity of SINDy can further be increased by a coordinate transformation into a representation allowing for a simpler representation of the system dynamics. This can be achieved with an autoencoder (consisting of an encoder $e_{NN}(x;\boldsymbol{\theta}^e)$ and a decoder $d_{NN}(h;\boldsymbol{\theta}^d)$, as proposed in~\cite{champion_data-driven_2019}, where the dynamics are learned on the reduced latent space $h$ using SINDy. A simultaneous optimization of the NN parameters $\boldsymbol{\theta}^e, \boldsymbol{\theta}^d$ and SINDy parameters $\boldsymbol{\Xi}$ is conducted with gradient descent. The cost is defined in terms of the autoencoder reconstruction loss $\mathcal{L}_{\mathcal{A}}$ and the residual of~\Cref{eq:sindyloss} at both the reduced latent space $\mathcal{L}_{\mathcal{R}}$ and the original space $\mathcal{L}_{\mathcal{F}}$\footnote{The encoder and decoder are derived with respect to their inputs to estimate the derivatives $\dot{\boldsymbol{x}}, \dot{\boldsymbol{h}}$ using the chain rule.}. A $L^1$-regularization for $\boldsymbol{\Xi}$ promotes sparsity. \\
\begin{align}
    \mathcal{L}_{\mathcal{A}} &= \frac{1}{2n} \sum_{i=1}^{n} ||\boldsymbol{x}(t_i) - d_{NN}\big(e_{NN}(\boldsymbol{x}(t_i);\boldsymbol{\theta}^e);\boldsymbol{\theta}^d\big)||_2^2 \\
    \mathcal{L}_{\mathcal{R}} &= \frac{1}{2n}\sum_{i=1}^n ||\underbrace{\Big(\nabla_x e_{NN}\big(\boldsymbol{x}(t_i);\boldsymbol{\theta}^e\big)\Big)\cdot \dot{\boldsymbol{x}}(t_i)}_{\dot{\boldsymbol{h}}} - \boldsymbol{\Theta}\Big(e_{NN}\big(\boldsymbol{x}(t_i);\boldsymbol{\theta}^e\big)\Big)\boldsymbol{\Xi}||_2^2 \\
    \mathcal{L}_{\mathcal{F}} &= \frac{1}{2n}\sum_{i=1}^n||\dot{\boldsymbol{x}}(t_i) - \nabla_h d_{NN}\big(\underbrace{e_{NN}(\boldsymbol{x}(t_i);\boldsymbol{\theta}^e)}_{\boldsymbol{h}};\boldsymbol{\theta}^d\big)\cdot \underbrace{\boldsymbol{\Theta}\Big(e_{NN}(\boldsymbol{x}(t_i);\boldsymbol{\theta}^e)\Big)\boldsymbol{\Xi}}_{\dot{\boldsymbol{h}}}||_2^2 \\
    C &= \kappa_{\mathcal{A}} \mathcal{L}_{\mathcal{A}} + \kappa_{\mathcal{R}} \mathcal{L}_{\mathcal{R}} + \kappa_{\mathcal{F}} \mathcal{L}_{\mathcal{F}}
\end{align}
As in~\Cref{eq:weigtedcost}, a weighted cost function with weights $\kappa_{\mathcal{A}},\kappa_{\mathcal{R}},\kappa_{\mathcal{F}}$ is employed. The reduced latent space can be exploited for forward simulations of the identified system. By solving the system with classical time-stepping schemes in the reduced latent space, the solution is obtained in the full space through the decoder, as outlined in~\cite{conti_reduced_2023}. Thus, a reduced order model of a previously unknown system is identified. The downside is, that the model is no longer interpretable in the full space.

\subparagraph{Multistep Methods}\mbox{}\\
Another approach to learning the system dynamics from~\Cref{eq:timesystem} is to approximate the right-hand side directly with a NN $\boldsymbol{\hat{f}}(\boldsymbol{x}_i)=O_{NN}(\boldsymbol{x}_i;\boldsymbol{\theta})$, $\boldsymbol{x}_i=\boldsymbol{x}(t_i)$. A residual can be formulated by considering linear multistep methods~\cite{iserles_first_2008}, a residual can be formulated. In general, these methods take the form: 
\begin{equation}
\sum_{m=0}^M [\alpha_m \boldsymbol{x}_{n-m} + \Delta t \beta_m \boldsymbol{f}(\boldsymbol{x}_{n-m})]=0,
\end{equation}
where $M, \alpha_0, \alpha_1, \beta_0, \beta_1$ are parameters specific to a multistep scheme. The scheme can be reformulated with a cost function, given as:
\begin{align}
C &= \frac{1}{N-M+1} \sum_{n=M}^N ||\boldsymbol{\hat{y}}_n||^2_2 \\
\boldsymbol{\hat{y}}_n &= \sum_{m=0}^M [\alpha_m \boldsymbol{x}_{n-m}+\Delta t \beta_m \boldsymbol{\hat{f}}(\boldsymbol{x}_{n-m})]
\end{align}
The idea of the method is strongly linked to the discrete-time PINN presented in~\Cref{par:discretetimePINN}, where a reformulation of the Runge-Kutta method yields the cost function needed to learn the forward solution.

\subsubsection{Enforcement Of Physics By Construction}\label{sssec:enorcementphysics}
Up to this point, this review only considered the case where physics are enforced indirectly through penalty terms of the PDE residual. The only exception, and the first example of enforcing physics by construction, was the strong enforcement of boundary conditions~\cite{berg_unified_2018, kollmannsberger_deep_2021,ren_phycrnet_2022} by modifying the outputs of the NN -- which led to a fulfillment of the boundary conditions independent of the NN parameters. For PDEs, this can be achieved by manipulating the output, such that the solution automatically obeys fundamental physical laws. Examples for this are, e.g., given in~\cite{kim_deep_2019}, where stream functions are predicted and subsequently differentiated to ensure conservation of mass, the incorporation of symmetries~\cite{ling_machine_2016}, or invariances~\cite{ling_reynolds_2016} by using integrity bases~\cite{smith_isotropic_1965}. Dynamical systems have been treated by learning the Lagrangian or Hamiltonian with correspondingly Lagrangian NNs~\cite{lutter_deep_2019,lutter_deep_2019-1,cranmer_lagrangian_2020} and Hamiltonian NNs~\cite{greydanus_hamiltonian_2019}. The quantities of interest are obtained through the differentiable NN and compared to labeled data. Indirectly learning the quantities of interest through the Lagrangian or Hamiltonian guarantees the conservation of energy. Enforcing the physics by construction is also referred to as physics-constrained learning, as the learnable space is constrained. More examples hereof are provided in the context of simulation enhancement in~\Cref{ssec:physicalmodeling}.

\section{Simulation Enhancement}\label{sec:simulationenhancement}
The category of simulation enhancement deals with any deep learning technique that interacts directly with and, thus, improves a component of a classical simulation. This is the most diverse category and will therefore be subdivided into the individual steps of a classical simulation pipeline:
\begin{itemize}
	\item pre-processing
	\item physical modeling
	\item numerical methods
	\item post-processing
\end{itemize}
Both data-driven and physics-informed approaches will be discussed in the following.

\subsection{Pre-processing}
The discussed pre-processing methods are trained in a supervised manner relying on the techniques presented in~\Cref{ssec:datadrivensurrogatemodeling} and on labeled data.
\subsubsection{Data Preparation}
Data preparation includes tasks, such as geometry extraction. For instance the detection of cracks from images by means of segmentation~\cite{zhang_road_2016, chen_nb-cnn_2018,jaeger_infrared_2022} can subsequently be used in simulations to assess the impact of the identified cracks. 
Also, CNNs have been used to prepare voxel data obtained from computed tomography scans, see~\cite{korshunova_image-based_2020}, where scanning artifacts are removed. Similarly NNs can be employed to enhance measurement data. This was, for example, demonstrated in~\cite{hall_barbosa_automation_1999}, where the NN acts as a denoiser for magnetic signals in the scope of non-destructive testing. Similarly, low-frequency extrapolation for full waveform inversion has been performed using NNs~\cite{ovcharenko_deep_2019, sun_extrapolated_2020, sun_deep_2022}.


\subsubsection{Initialization}
Instead of preparing the data, the simulation can be accelerated by an initialization. This can, for example, be achieved through initial guesses by NNs, providing a better starting point for classical iterative solvers ~\cite{lewis_deep_2017}\footnote{Here, the initial guess is incorporated through a regularization term.}. A tighter integration is achieved by using a pre-trained~\cite{chen_transfer_2021} NN ansatz whose parameters are subsequently tweaked by the classical solver, as demonstrated for full waveform inversion in~\cite{kollmannsberger_transfer_2023}.

\subsubsection{Meshing}\label{ssec:meshing}
Finally, many simulation techniques rely on meshes. This can be achieved indirectly with NNs, by prediction of mesh density functions~\cite{dyck_determining_1992,chedid_automatic_1996,triantafyllidis_automatic_2000,krzhizhanovskaya_meshingnet_2020,lock_meshing_2023} incorporating either expert knowledge of where small elements are needed, or relying on error estimations. Subsequently, a classical mesh generator is employed. However, NNs (specifically let-it-grow NNs~\cite{fritzke_growing_1994}) have also been proposed directly as mesh generators~\cite{alfonzetti_automatic_1996, triantafyllidis_finite-element_2002}. 






\subsection{Physical Modeling}\label{ssec:physicalmodeling}
Physical models that capture physical phenomena accurately are a core component of mechanics. Deep learning offers three main approaches for physical models. Firstly, a NN is used as the physical model directly (model substitution). Secondly, an underlying model may be assumed where a NN determines its coefficients (identification of model parameters). Lastly, the entire model can be identified by a NN (model identification). In the first approach, the NN is integrated within the simulation pipeline, while the latter two rely on incorporation of the identified models in a classical sense.\\

For illustration purposes, the approaches are mostly explained on the example of constitutive models. Here, the task is to relate the strain $\varepsilon$ to a stress $\sigma$, i.e., find a function $\sigma=f(\varepsilon)$. This can, for example, be used within a finite element framework to determine the element stiffness, as elaborated in~\cite{lefik_artificial_2003}. 

\subsubsection{Model Substitution}
In model substitution, a NN $f_{NN}$ replaces the model, yielding the prediction $\hat{\sigma}=f_{NN}(\varepsilon;\boldsymbol{\theta})$. The quality of the model is assessed with a data-driven cost function (\Cref{eq:genericcost}) using labeled data $\sigma^{\mathcal{M}},\varepsilon^{\mathcal{M}}$. The approach is applied to a variety of problems, where the key difference lies in the definition of input and output quantities. The same deep learning techniques from data-driven simulation substitution (\Cref{ssec:datadrivensurrogatemodeling}) can be employed.\\

Applications include predictions of stress from strain~\cite{lefik_artificial_2003,jang_machine_2021}, flow stresses from temperatures, strain rates and strains~\cite{lin_application_2008,li_artificial_2012}, yield functions~\cite{liu_mechanistically_2022}, crack opening responses from stresses~\cite{unger_neural_2009}, contact stiffness from penetration and contact pressure~\cite{hattori_contact_2015}, point of contact from position of neighboring nodes of finite elements~\cite{oishi_new_1970}, or control points of NURBS surfaces~\cite{oishi_surface--surface_2020}. Source terms of simplified equations or coarser discretizations have also been learned for turbulence~\cite{tracey_machine_2015,singh_machine-learning-augmented_2017,maulik_subgrid_2019} and the wave equation~\cite{fabra_finite_2022}. Here, the reference -- a high-fidelity model -- is to be captured in the best possible way by the source term. \\

Variations also predict the quantity of interest indirectly. For example, strain energy densities $\psi$ are predicted by NNs from deformation tensors $F$, and subsequently derived using automatic differentiation to obtain stresses~\cite{le_computational_2015,lu_data-driven_2019}. The approach can also be extended to incorporate uncertainty quantification~\cite{huang_learning_2020}. By extending the input space with microstructural information, an in-built homogenization is added to the constitutive model~\cite{wang_multiscale_2018, li_multiscale_2020, vlassis_geometric_2020}. Thus, the macroscale simulation considers the microstructure at the integration points in the sense of FE$^2$~\cite{frankenreiter_hybrid_2011, fish_practical_2013}, but without an additional finite element computation. Incorporation of microstructures requires a large amount of realistic training data, which can be obtained through generative approaches as discussed in~\Cref{sec:generativeapproaches}. Active learning can reduce the required number of simulations on these geometries~\cite{kalina_fetextrmann_2023}. \\

A specialized NN architecture is employed by~\cite{linka_constitutive_2021}, where a NN first estimates invariants $I$ of the deformation tensor $F$ and thereupon predicts the strain energy density, thus mimicking the classical constitutive modeling approach. Another network extension is the use of RNNs to learn history-dependent models. This was shown in~\cite{wang_multiscale_2018, li_multiscale_2020, mozaffar_deep_2019,wu_recurrent_2022} for the prediction of the stress increment from the strain-stress history, the strain energy from the strain energy history~\cite{abueidda_deep_2021}, and crack patterns based on prior cracks and crystalline orientations~\cite{hsu_using_2020,lew_deep_2021}. \\

The learned models do not, however, necessarily obey fundamental physical laws. Attempts to incorporate physics as constraints using penalty terms have been made in~\cite{liu_generic_2020,weber_constrained_2021,leng_predicting_2021}. Still, physical consistency is not guaranteed. Instead, NN architectures can be chosen such that they satisfy physical requirements
by construction. In constitutive modeling, objectivity can be enforced by using only deformation invariants as input~\cite{tac_data-driven_2022}, and polyconvexity can be enforced through the architecture, such as input-convex NNs~\cite{linden_neural_2023, klein_finite_2022, klein_polyconvex_2022, asad_mechanics-informed_2023} and neural ordinary differential equations~\cite{tac_data-driven_2022,tac_data-driven_2023}. It was demonstrated that ensuring fundamental physical aspects such as invariants combined with polyconvexivity delivers a much better behavior for unseen data, especially if the model is used in extrapolation. \\

Input-convex NNs~\cite{amos_input_2017} enforce the convexity with specialized activation functions such as log-sum-exponential, or softplus functions in combination with constraints on the NN weights to ensure that they are positive, while neural ordinary differential equations~\cite{chen_neural_2019} (discussed in~\Cref{sec:discretizationsasNNs}) approximate the strain energy density derivatives and ensure non-negative values. Alternatively, a mapping from the NN to a convex function can be defined~\cite{chen_polyconvex_2022} ensuring a convex function for any NN output. Related are also thermodynamics-based NNs~\cite{masi_thermodynamics-based_2021, masi_material_2021}, e.g., applied to complex microstructures in~\cite{masi_multiscale_2022}, which by construction obey fundamental thermodynamic laws. Training of these methods can be performed in a supervised manner, relying on strain-stress data, or unsupervised. In the unsupervised setting, the constitutive model is incorporated in a finite element solver, yielding a displacement field for a specific boundary value problem. The computed field, together with measurement data, yields a residual that is referred to as the modified constitutive relation error (mCRE)~\cite{ladeveze_updating_1994, marchand_parameter_2019, nguyen_mcre-based_2022}, which is minimized to improve the constitutive relation~\cite{benady_nn-mcre_2023, benady_modified_2023}. Instead of formulating the mismatch in terms of displacements, \cite{li_machine-learning_2019,thakolkaran_nn-euclid_2022} formulate it in terms of boundary forces. For an in-depth overview of constitutive model substitution in deep learning, see~\cite{rosenkranz_comparative_2023}.







\subsubsection{Identification Of Model Parameters}
Identification of model parameters is achieved by assuming an underlying model and training a NN to predict its parameters for a given input. In the constitutive model example, one might assume a linear elastic model expressed in terms of a constitutive tensor $c$, such that $\sigma=c\varepsilon$. The constitutive tensor can be predicted from the material distribution defined in terms of a heterogeneous elasticity modulus $\boldsymbol{E}$ defined throughout the domain 
\begin{equation}
\hat{c}=f_{NN}(\boldsymbol{E};\boldsymbol{\theta}).
\end{equation}

Typical applications are homogenization, where effective properties are predicted from the geometry and material distribution. Examples are CNN-based homogenizations on computed tomography scans~\cite{li_predicting_2019, henkes_deep_2021}, predictions of in-vivo constitutive parameters of aortic walls from its geometry~\cite{liu_estimation_2019}, predictions of elastoplastic properties~\cite{lu_extraction_2020} from instrumented indentation results relying on a multi-fidelity approach~\cite{meng_composite_2020}, prediction of stress intensity factors from the geometry in microfabricated microcantilevers~\cite{liu_machine_2020}, estimation of effective bone properties from the boundary conditions and applied stresses within a finite element, and incorporating meso-scale information by training a NN on representative volume elements~\cite{hambli_multiscale_2011}.

\subsubsection{Model Identification}
NN models as a replacement of classical approaches are not interpretable, while only identifying model parameters of known models restricts the models capacity. This gap can be bridged by the identification of models in terms of parsimonious mathematical expressions.

The typical procedure is to pose the problem in terms of candidate functions and to identify the most relevant terms. The methodology was inspired by SINDy~\cite{brunton_discovering_2016} and introduced in the framework for efficient unsupervised constitutive law identification and discovery (EUCLID)~\cite{flaschel_unsupervised_2021}. The approach is unsupervised, as the stress-strain data is only indirectly  available through the displacement field and corresponding reaction forces. The $N_I$ invariants $I_i$ of the deformation tensor $F$ are inserted into a candidate library $Q(\{I_i\}_{i=1}^{N_I})$ containing the candidate functions. Together with the corresponding weights $\boldsymbol{\theta}$, the strain density $\psi$ is determined:
\begin{equation}
\psi(\{I_i\}_{i=1}^{N_I}) = Q^T(\{I_i\}_{i=1}^{N_I}) \boldsymbol{\theta}.
\end{equation}
Through derivation of the strain density $\psi$ using automatic differentiation, the stresses $\boldsymbol{\sigma}$ are determined. The problem is then cast into the weak form with which the linear momentum balance is enforced. The weak form is then minimized with respect to $\boldsymbol{\theta}$ using a fixed-point iteration scheme (inspired by~\cite{tibshirani_regression_1996}), where a $L_p$-regularization is used to promote sparsity in $\boldsymbol{\theta}$. Despite its young age, the approach has already been applied to plasticity~\cite{flaschel_discovering_2022}, viscoelasticity~\cite{marino_automated_2023}, combinations~\cite{flaschel_automated_2023}, and has been extended to incorporate uncertainties through a Bayesian model~\cite{joshi_bayesian-euclid_2022}. Furthermore, the approach has been extended with an ensemble of input-convex NNs~\cite{thakolkaran_nn-euclid_2022}, yielding a more accurate, but less interpretable model. 

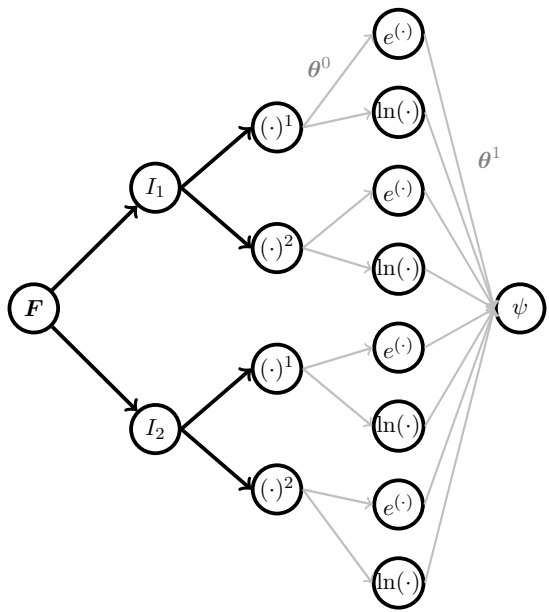
\begin{figure}[htb]
	\centering
	\begin{tikzpicture}[scale=0.8, transform shape]
	\node (F) [draw, circle, line width=0.5mm, minimum size=0.8cm,inner sep=0] at (0, 0) {$\boldsymbol{F}$};
	\node (I1) [draw, circle, line width=0.5mm, minimum size=0.8cm,inner sep=0] at (2, 2) {$I_1$};
	\node (I2) [draw, circle, line width=0.5mm, minimum size=0.8cm,inner sep=0] at (2, -2) {$I_2$};
	\node (c1) [draw, circle, line width=0.5mm, minimum size=0.8cm,inner sep=0] at (4, 3) {$(\cdot)^1$};
	\node (c2) [draw, circle, line width=0.5mm, minimum size=0.8cm,inner sep=0] at (4, 1) {$(\cdot)^2$};
	\node (c3) [draw, circle, line width=0.5mm, minimum size=0.8cm,inner sep=0] at (4, -1) {$(\cdot)^1$};
	\node (c4) [draw, circle, line width=0.5mm, minimum size=0.8cm,inner sep=0] at (4, -3) {$(\cdot)^2$};
	\node (d1) [draw, circle, line width=0.5mm, minimum size=0.8cm,inner sep=0] at (6, 7*0.65) {$e^{(\cdot)}$};
	\node (d2) [draw, circle, line width=0.5mm, minimum size=0.8cm,inner sep=0] at (6, 5*0.65) {$\ln(\cdot)$};
	\node (d3) [draw, circle, line width=0.5mm, minimum size=0.8cm,inner sep=0] at (6, 3*0.65) {$e^{(\cdot)}$};
	\node (d4) [draw, circle, line width=0.5mm, minimum size=0.8cm,inner sep=0] at (6, 0.65) {$\ln(\cdot)$};
	\node (d5) [draw, circle, line width=0.5mm, minimum size=0.8cm,inner sep=0] at (6, -0.65) {$e^{(\cdot)}$};
	\node (d6) [draw, circle, line width=0.5mm, minimum size=0.8cm,inner sep=0] at (6, -3*0.65) {$\ln(\cdot)$};
	\node (d7) [draw, circle, line width=0.5mm, minimum size=0.8cm,inner sep=0] at (6, -5*0.65) {$e^{(\cdot)}$};
	\node (d8) [draw, circle, line width=0.5mm, minimum size=0.8cm,inner sep=0] at (6, -7*0.65) {$\ln(\cdot)$};
	\node (O) [draw, circle, line width=0.5mm, minimum size=0.8cm,inner sep=0] at (8,0) {$\psi$};
	
	\draw [black, line width=0.5mm,->] (F.north east) -- (I1.south west);
	\draw [black, line width=0.5mm,->] (F.south east) -- (I2.north west);
	\draw [black, line width=0.5mm,->] (I1.east) -- (c1.west);
	\draw [black, line width=0.5mm,->] (I1.east) -- (c2.west);
	\draw [black, line width=0.5mm,->] (I2.east) -- (c3.west);
	\draw [black, line width=0.5mm,->] (I2.east) -- (c4.west);
	
	\draw [lightgray, line width=0.3mm,->] (c1.east) -- (d1.west);
	\draw [lightgray, line width=0.3mm,->] (c1.east) -- (d2.west);
	\draw [lightgray, line width=0.3mm,->] (c2.east) -- (d3.west);
	\draw [lightgray, line width=0.3mm,->] (c2.east) -- (d4.west);
	\draw [lightgray, line width=0.3mm,->] (c3.east) -- (d5.west);
	\draw [lightgray, line width=0.3mm,->] (c3.east) -- (d6.west);
	\draw [lightgray, line width=0.3mm,->] (c4.east) -- (d7.west);
	\draw [lightgray, line width=0.3mm,->] (c4.east) -- (d8.west);
	
	\foreach {\x} in {1,2,...,8} {
		\draw [lightgray, thick,->] (d\x.east) -- (O.west);
	}
	
	\node [gray] at (4.7,4) {$\boldsymbol{\theta}^{0}$};
	\node [gray] at (7.5,2.5) {$\boldsymbol{\theta}^{1}$};
	
	\end{tikzpicture}
	\caption{Automated model discovery through a sparsely connected NN with specialized activation functions acting as candidate functions. The thick black connections are not learnable, while the gray ones represent linearly weighted connections. Figure adapted and simplified from~\cite{linka_automated_2023}.}\label{fig:constitutivelawinterpretable}
\end{figure}

A similar effort was recently carried out by~\cite{linka_automated_2023, linka_new_2023}, where NNs are designed to retain interpretability. This is achieved through sparse connections in combination with specialized activation functions representing candidate functions, such that they are able to capture classical forms of constitutive terms. Through the sparse connections in the network and the specialized activation functions, the NN's weights become physical parameters, yielding an interpretable model. This is best understood by consulting~\Cref{fig:constitutivelawinterpretable}, where the strain energy density is expressed as
\begin{equation}
\hat{\psi}=\theta^1_0 e^{\theta^0_0 I_1} + \theta^1_1 \ln(\theta^0_1 I_1)
+          \theta^1_2 e^{\theta^0_2 I_1^2} + \theta^1_2 \ln(\theta^0_2 I_1^2)
+          \theta^1_3 e^{\theta^0_3 I_2} + \theta^1_4 \ln(\theta^0_4 I_2)
+          \theta^1_5 e^{\theta^0_5 I_2^2} + \theta^1_6 \ln(\theta^0_6 I_2^2).
\end{equation}
Differentiating the predicted strain energy density $\hat{\psi}$ with respect to the invariants $I_i$ yields the constitutive model, relating stress and strain.

\subsection{Numerical Methods}
This subsection describes efforts in which NNs are used to replace or enhance classical numerical schemes to solve PDEs.


\subsubsection{Algorithm Enhancement}
Classical algorithms can be enhanced by NNs, by learning corrections to commonly arising numerical errors, or by estimating tunable parameters within the algorithm. Corrections have, for example, been used for numerical quadrature~\cite{oishi_computational_2017} in the context of finite elements. Therein, NNs are used to predict adjustments to quadrature weights and positions from the nodal positions to improve the accuracy for distorted elements. Similarly, NNs have been applied as correction for strain-displacement matrices for distorted elements~\cite{jung_deep_2020}. NNs have also been employed to provide improved gradient estimates. Specifically, \cite{bar-sinai_learning_2019} modify the gradient computation to match a fine scale simulation on a coarse grid:
\begin{equation}
\frac{\partial^n u}{\partial x^n}\approx \sum_i \alpha_i^{(n)}u_i.
\end{equation}
The coefficients $\alpha_i$ are predicted by NNs from the current coarse solution. Special constraints are imposed on $\alpha_i$ to guarantee accurate derivatives. Another application are specialized strain mappings for damage mechanics embedded within individual finite elements learned by PINNs~\cite{pantidis_integrated_2023}. It has even been suggested to partially replace solvers. For example, \cite{arcones_neural_2022} replace either the fluid or structural solver by a surrogate model for fluid-structure interaction problems. \\


Learning tunable parameters was demonstrated for the estimation of the largest possible time step using a RNN acting at the latent vector of an autoencoder~\cite{han_artificial_2021}. Also, optimal test functions for finite elements were learned to improve stability~\cite{sluzalec_automatic_2023}.


\subsubsection{Multiscale Methods}\label{sssec:multiscale}
Multiscale methods have been proposed to efficiently integrate and resolve systems acting on multiple scales. One approach are the learned constitutive models from~\Cref{{ssec:physicalmodeling}} that incorporate the microstructure. This is essentially achieved through a homogenization at the mesoscale used within a macroscale simulation. \\

A related approach is element substructuring~\cite{casadei_geometric_2013, oztoprak_two-scale_2023}, where superelements mimic the behavior of a conglomerate of classic basic finite elements. In~\cite{koeppe_intelligent_2020}, the superelements are enhanced by NNs, which draw on the boundary displacements to predict the displacements and stresses within the element as well as the reaction forces at the boundary. Through assembly of the reaction forces in the global finite element system, an equilibrium is reached with a Newton-Raphson solver. Similarly, the approach in~\cite{capuano_smart_2019} learns the internal forces from the coarse degrees of freedom of the superelements. These approaches are particularly valuable, as they can seamlessly incorporate history-dependent behavior using RNNs. \\

Finally, multiscale analysis can also be performed by first solving a coarse global model with a subsequent local analysis. This is referred to as zooming methods. In~\cite{yamaguchi_zooming_2021}, a NN learns the global model and thereby predicts the boundary conditions for the local model. In a similar sense, DeepONets have been applied for the local analysis~\cite{yin_interfacing_2022}, whereas the global analysis is performed with a finite element solver. Both are conducted in an alternating fashion until convergence is reached.

\subsubsection{Optimization}\label{sssec:optimization}
Optimization is a fundamental task within computational mechanics and therefore addressed separately. It is not only used to find optimal structures, but also to solve inverse problems. Generally, the task can be formulated as minimizing a cost function $C$ with respect to parameters $\lambda$. In computational mechanics, $\lambda$ is typically fed to a forward simulation $u=F(\lambda)$, yielding a solution $u$ inserted into the cost function $C$. If the gradients $\nabla_\lambda C$ are available, gradient-based optimization is the state-of-the-art~\cite{sigmund_usefulness_2011}, where the gradients are used to update $\lambda$. In order to access the gradients, the forward simulation $F$ has to be differentiable. This requirement is, for example, utilized within the branch of deep learning called differentiable physics~\cite{thuerey_physics-based_2021}. Incorporating gradient information from the numerical solver into the NN improves learning, feedback, and generalization. An overview and introduction to differentiable physics is provided in~\cite{thuerey_physics-based_2021}, with applications in~\cite{morton_deep_2018,bar-sinai_learning_2019,chen_neural_2019,holl_learning_2020,um_solver---loop_2020,um_solver---loop_2021}\footnote{Applications of differentiable physics vary widely and are addressed throughout this work.}.\\

\begin{figure}[htb]
	\centering
	\begin{tikzpicture}
	\node (I) [draw, thick] at (-1,0) {\begin{tabular}{c} $\lambda$ \end{tabular}};
	\node (A) [draw, thick] at (1,0) {\begin{tabular}{c} $F(\lambda)$ \end{tabular}};
	\node (O) [draw, thick] at (3,0) {\begin{tabular}{c} $u$ \end{tabular}};
	\node (L) [draw, thick] at (5,-0.75) {\begin{tabular}{c} $C(u)$ \end{tabular}};
	\node (G) [draw, thick] at (2, -1.5) {\begin{tabular}{c} $\nabla_\lambda C$ \end{tabular}};
	\draw [thick,->] (I.east) -- (A.west);
	\draw [thick,->] (A.east) -- (O.west);
	\draw [thick,->] (O.east) -- (5,0) -- (L.north);
	\draw [thick,->] (L.south) -- (5,-1.5) -- (G.east);
	\draw [thick,->] (G.west) -- (-1,-1.5) -- (I.south);
	\end{tikzpicture}
	\caption{Gradient-based optimization.}\label{fig:gradientbasedoptimization}
\end{figure}
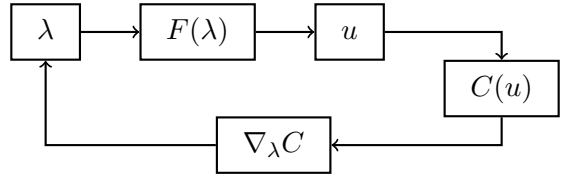

The iterative gradient-based optimization procedure is illustrated in~\Cref{fig:gradientbasedoptimization}. For an in-depth treatment of NNs in optimization, see the recent review~\cite{woldseth_use_2022}.

Inserting a learned forward operator $F$, as those discussed in~\Cref{ssec:datadrivensurrogatemodeling}, into an optimization problem provides two advantages~\cite{jensen_inversion_1999,yu_artificial_2019,chen_generative_2020,tanyu_deep_2022,zohdi_machine-learning_2023}. Firstly, a faster forward operator results in faster optimization iterations. Secondly, the gradient computation is simplified, as automatic differentiation through the forward operator $F$ is straightforward in contrast to the adjoint state method~\cite{plessix_review_2006, givoli_tutorial_2021}. Note however, that for time-stepping procedures, the computational cost might be greater for automatic differentiation, as shown in~\cite{herrmann_use_2023}. Applications include full waveform inversion~\cite{herrmann_use_2023}, topology optimization~\cite{keshavarzzadeh_robust_2021, qian_accelerating_2021, chi_universal_2021}, and control problems~\cite{lee_application_1997,pierret_turbomachinery_1999,holl_learning_2020}.\\

Similarly, an operator replacing the sensitivity computation can be learned~\cite{aulig_evolutionary_2013, aulig_topology_2014, aulig_neuro-evolutionary_2015,chi_universal_2021}. This can be achieved in a supervised manner with precomputed sensitivities to reduce the cost $C$~\cite{aulig_topology_2014,chi_universal_2021}, or by intending to maximize the improvement of the cost function after the gradient update~\cite{aulig_evolutionary_2013,aulig_neuro-evolutionary_2015}. In~\cite{aulig_evolutionary_2013,aulig_neuro-evolutionary_2015}, an evolutionary algorithm was employed for the general case that the sensitivites are not readily available. Training can adaptively be reintroduced during the optimization phase, if the cost $C$ does not decrease~\cite{chi_universal_2021}, improving the NN for the specific problem it is handling. Taking this idea to the extreme, the NN is trained on the initial gradient updates of a specific optimization. Later, solely the NN delivers the sensitivities~\cite{zhang_speeding_2021} with supervised updates every $n$ updates to improve accuracy, where $n$ is a hyperparameter. The ideas of learning a forward operator and a sensitivity operator are combined in~\cite{qian_accelerating_2021}, where it is pointed out that the sensitivity from automatic differentiation through the learned forward operator can be inaccurate, despite an accurate forward operator\footnote{Although automatic differentiation in principle has a high accuracy, oscillations between the sampled points may lead to spurious gradients with regard to the sampled points~\cite{rivera_quadrature_2022}.}. Therefore, an additional loss term is added to the cost function, enforcing the correctness of the sensitivity through labels obtained with the adjoint state method. Alternatively, the sensitivity computation can be enhanced by correcting the sensitivity computation performed on a coarse grid, as proposed in~\cite{hunter_superadjoint_2023} and related to the multiscale techniques discussed in~\Cref{sssec:multiscale}. Here, the adjoint field used for the sensitivity computation is reduced by both a proper orthogonal decomposition, and a coarser discretization. Subsequently, a NN corrects the coarse estimate through a super-resolution NN~\cite{fukami_machine-learning-based_2021}. Similarly, \cite{chi_universal_2021,senhora_machine_2022} maps the forward solution on a coarse grid to the design variable sensitivity on a fine grid. A similar application is a correction term within a fixed-point iterator, as outlined in~\cite{hsieh_learning_2019}.\\


Related to the sensitivity predictions are approaches that directly predict an updated state. The goal is to decrease the total number of iterations. In practice, a combination of predictions and classical gradient-based updates is performed~\cite{sosnovik_neural_2019, kallioras_accelerated_2020, joo_unit_2021, ye_acceleration_2021}. The main variations between the methods in the literature are the inputs and how far the forecasting is performed. In~\cite{sosnovik_neural_2019}, the update is obtained from the current state and gradient, while~\cite{kallioras_accelerated_2020} predicts the final state from the history of initial updates. The history is also considered in~\cite{joo_unit_2021}, but the prediction is performed on subpatches which are then stitched together. \\

Another option of introducing NNs to the optimization loop is to use NNs as an ansatz of $\lambda$, see e.g.~\cite{hoyer_neural_2019, xu_neural_2019, holl_learning_2020, berg_neural_2021, chen_new_2021, halle_artificial_2021, deng_topology_2020, chandrasekhar_tounn_2021, chandrasekhar_length_2021, chandrasekhar_multi-material_2021, herrmann_use_2023}. In the context of inverse problems~\cite{hoyer_neural_2019, xu_neural_2019, holl_learning_2020, berg_neural_2021, chen_new_2021, halle_artificial_2021,herrmann_use_2023}, the NN acts as regularizer on a spatially varying inverse quantity $\lambda(x)=I_{NN}(x;\boldsymbol{\theta})$, providing both smoother and sharper solutions. For topology optimization with a NN parametrization of the density function~\cite{deng_topology_2020, chandrasekhar_tounn_2021, chandrasekhar_length_2021, chandrasekhar_multi-material_2021}, no regularizing effect was observed. It was however possible to obtain a greater design diversity through different initializations of the NN. Extensions using specialized NN architectures for implicit representations~\cite{park_deepsdf_2019,michalkiewicz_implicit_2019,gropp_implicit_2020,sitzmann_implicit_2020,huang_textbackslash_2021,deng_parametric_2021} have been presented in the context of topology optimization in~\cite{zhang_tonr_2021}. Furthermore, \cite{berg_neural_2021,chandrasekhar_tounn_2021,herrmann_use_2023} showed how to conduct the gradient computation without automatic differentiation through the solver $F$. The gradient computation is split up via the chain rule:
\begin{equation}
\nabla_{\boldsymbol{\theta}}C=\nabla_{\lambda} C \cdot \nabla_{\boldsymbol{\theta}} \lambda.
\end{equation}
The first gradient $\nabla_{\lambda} C$ is computed with the adjoint state method, such that the solver can be treated as a black box. The second gradient $\nabla_{\boldsymbol{\theta}} \lambda$ is obtained through automatic differentiation. An additional advantage of the NN ansatz is that, if applied to multiple solutions with a problem specific input, the NN is trained. Thus, after sufficient inversions, the NN can be used as predictor, as presented in~\cite{biswas_prestack_2019}. The training can also be performed in combination with labeled data, yielding a semi-supervised approach, as demonstrated in~\cite{alfarraj_semi-supervised_2019, kollmannsberger_transfer_2023}.

\subsection{Post-Processing}\label{ssec:postprocessing}
Post-processing concerns the modification and interpretation of the computed solution. One motivation is to reduce the numerical error of the computed solution. This can for example be achieved with super-resolution techniques relying on specialized CNN architectures from computer vision~\cite{dong_learning_2014, dong_image_2015}. Coarse to fine mappings can be obtained in a supervised manner using matching coarse and fine simulations as labeled data, as presented for turbulent flows~\cite{fukami_super-resolution_2019,fukami_machine-learning-based_2021} and topology optimization~\cite{napier_artificial_2020, wang_deep_2021, xue_efficient_2021}. The mapping is typically performed from coarse to fine solution fields, but mappings from a posteriori errors have been proposed as well~\cite{oishi_finite_2021}. Further specialized extensions to the cost function have been suggested in the context of de-homogenization~\cite{elingaard_-homogenization_2022}. \\

The methods can analogously be applied to temporal data where the solution is refined at each time step, -- as, e.g., presented with RNNs as corrector of reduced order models~\cite{wan_data-assisted_2018}. However, coarse discretizations in dynamical models lead to an error accumulation, that increases with the number of time steps. Thus, a simple coarse-to-fine post-processing at each time step is not sufficient. To this end, \cite{um_solver---loop_2020,um_solver---loop_2021} apply a correction at each time step before the coarse solver predicts the next time step. As the correction is propagated through the solver, the sensitivities of the solver must be computed to perform the backward propagation. Therefore, a differentiable solver (i.e., differentiable physics) has to be employed. This significantly outperforms the purely supervised approach, where the entire coarse trajectory is applied without corrections in between. The number of steps performed is a hyperparameter, which increases the accuracy but comes with a higher computational effort. This concept is referred to as solver-in-the-loop. \\

Further variations perform the coarse-to-fine mapping in a patch-based manner, where the interfaces require a special treatment~\cite{sato_example-based_2018}. Another approach uses a NN to map the coarse solution to the closest fine solution stored in a database~\cite{chu_data-driven_2017}. The mapping is performed on patches of the domain. \\

Other post-processing tasks include feature extraction. After a topology optimization, NNs have been used to extract basic shapes to be used in a subsequent shape optimization~\cite{yildiz_integrated_2003, lin_artificial_2005}. Another aspect that can be ensured through post-processing is manufacturability. \\

Lastly, adaptive mesh refinement falls under the category of post-processing as well. Closely related to the meshing approaches discussed in~\Cref{ssec:meshing}, NNs have been proposed as error indicators~\cite{chen_output-based_2020,krzhizhanovskaya_meshingnet_2020} that are trained in a supervised manner. The error estimators can subsequently be employed to adapt the mesh based on the error.

\section{Discretizations As Neural Networks}\label{sec:discretizationsasNNs}
NNs are composed of linear transformations and non-linear functions, which are basic building blocks of most PDE discretizations. Thus, the motivation to utilize NNs to construct discretizations of PDEs herefore are twofold. Firstly, deep learning techniques can hereby be exploited within classical discretization frameworks. Secondly, novel NN architectures arise, which are more tailored towards many physical problems in computational mechanics but also find their use cases outside of that field.

\subsection{Finite Element Method}
\begin{figure}[htb]
	\centering
	\begin{tikzpicture}
	\node (I0) [draw, thick] at (0,0) {\begin{tabular}{c} $\alpha^e$ \end{tabular}};
	\node (I1) [draw, thick] at (3,0) {\begin{tabular}{c} $K_{ij}$ \end{tabular}};
	\node (I2) [draw, thick] at (6,0) {\begin{tabular}{c} $b_i$ \end{tabular}};
	
	\node [gray] at (1.5,0.25) {$W_{ij}^e$};
	\node [gray] at (4.5,0.25) {$u_j$};
	\draw [line width=0.4mm,->, lightgray] (I0.east) -- (I1.west);
	\draw [line width=0.4mm,->, lightgray] (I1.east) -- (I2.west);
	
	\end{tikzpicture}
	\caption{Finite element NNs, prediction of forces $b_i$ from material coefficients $\alpha^e$ via assembly of global stiffness matrix $K_{ij}$, and evaluations of equations with the displacements $u_j$~\cite{ramuhalli_finite-element_2005}.}\label{fig:feNNs}
\end{figure}
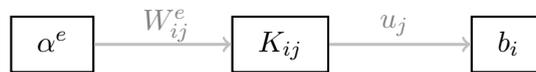

One method are finite element NNs~\cite{takeuchi_neural_1994,ramuhalli_finite-element_2005} (see~\cite{sikora_artificial_1999, xu_application_1999, guo_finite_1999, lee_neural_1990, kalkkuhl_fem-based_1999, xu_finite-element_2012} for applications), for which we consider the system of equations from a finite element discretization with the stiffness matrix $K_{ij}$, degrees of freedom $u_j$, and the body load $b_i$:
\begin{equation}
\sum_{j=1}^N K_{ij} u_j-b_i=0, i=1,2,\dots,N. \label{eq:FEMequations}
\end{equation}
Assuming constant material properties along an element and uniform elements, a pre-integration of the local stiffness matrix $k_{ij}^e=\alpha^e w_{ij}^e$ can be performed, as, e.g., shown in~\cite{yang_efficient_2012}. The goal is to pull out the material coefficients of the integration, leading to the following assembly of the global stiffness matrix:
\begin{equation}
K_{ij}=\sum_{e=1}^M \alpha^e W_{ij}^e \text{ with } W_{ij}^e=\begin{cases}
w_{ij}^e \text{ if } i, j\in e\\
0 \text{ else}´			
\end{cases}.
\end{equation}
Inserting the assembly into the system of equations from~\cref{eq:FEMequations} yields
\begin{equation}
\sum_{j=1}^N\left(\sum_{e=1}^M \alpha^e W_{ij}^e \right) u_j-b_i=0, i=1,2,\dots,N.
\end{equation}
The nested summation has a similar structure of a FC-NN, $a_i^{(l)}=\sigma(z_i^{(l)})=\sigma(\sum_{j=1}^{N^{(l)}}a_j^{(l-1)}+b_i^{(l)})$ without activation and bias (see~\Cref{fig:feNNs}):
\begin{equation}
a_i^{(2)}=\sum_{j=1}^{N^{(2)}} W_{ij}^{(1)} a_j^{(1)}=\sum_{j=1}^{N^{(2)}} W_{ij}^{(1)}(\sum_{k=1}^{N^{(1)}} W_{jk}^{(0)} a_k^{(0)}).
\end{equation}

Thus, the stiffness matrix $K_{ij}$ is the hidden layer. In a forward problem, $W_{ij}^e$ are non-learnable weights, while $u_j$ contains a mixture of learnable weights and non-learnable weights coming from the imposed Dirichlet boundary conditions. A loss can be formulated in terms of body load mismatch, as $\frac{1}{2}\sum_{i=1}^N (\hat{b}_i - b_i)^2$. In the inverse setting, $\alpha^e$ becomes learnable -- instead of $u_j$, which is then fixed. For partial domain knowledge in the inverse case, $u_j$ becomes partially learnable.

\begin{figure}[htb]
	\centering
	\begin{subfigure}[b]{0.99\textwidth}
		\centering
		\begin{tikzpicture}
		\draw [lightgray, line width=0.4mm, dashed] (-1,0) -- (0,0);
		\draw [lightgray, line width=0.4mm, dashed] (6,0) -- (7,0);
		\draw [lightgray, line width=0.4mm] (0,0) -- (6,0);
		\draw [black, line width=0.6mm] (2,0) -- (4,0);
		\fill [black] (2,0) circle (0.1cm);
		\fill [black] (4,0) circle (0.1cm);
		\fill [lightgray] (0,0) circle (0.1cm);
		\fill [lightgray] (6,0) circle (0.1cm);
		
		\node [gray] at (0.5,-0.5) {$x$};
		\draw [lightgray, |->, line width=0.3mm] (0, -0.3) -- (1, -0.3);
		
		\node [gray] at (1,0.3) {$u^{e-1}$};
		\node at (3,0.3) {$u^{e}, \Omega^e$};
		\node [gray] at (5,0.3) {$u^{e+1}$};
		\node at (2,-0.3) {$u_1^e, x_1^e$};
		\node at (4,-0.3) {$u_2^e, x_2^e$};
		
		\end{tikzpicture}
		\caption{One-dimensional linear elements with two nodes each.}
	\end{subfigure}\\
	\begin{subfigure}[b]{0.99\textwidth}
		\centering
		\begin{tikzpicture}
		\tikzstyle{neuron}=[draw=black!80,thick,minimum size=17pt,inner sep=3pt]
		\node[neuron, circle] (x) at (-1,0) {$x$};
		\node[neuron, rounded rectangle] (x1) at (1.5,0.5) {$x-x_2^e$};
		\node[neuron, rounded rectangle] (x2) at (5,0.5) {$N_1^e(x)=\frac{x-x_2^e}{x_1^e-x_2^e}$};
		
		\node[neuron, rounded rectangle] (y1) at (1.5,-0.5) {$x-x_1^e$};
		\node[neuron, rounded rectangle] (y2) at (5,-0.5) {$N_2^e(x)=\frac{x-x_1^e}{x_2^e-x_1^e}$};
		
		\node[neuron, circle, minimum size=0.7cm] (u) at (8, 0) {$u^e$};
		\node[neuron, circle, gray, inner sep=0pt] (u-1) at (8, 2) {$u^{e-1}$};
		\node[neuron, circle, gray, inner sep=0pt] (u1) at (8, -2) {$u^{e+1}$};
		
		\node[neuron, circle] (uh) at (11,0) {$u$};
		
		\node [gray] at (0.1,0.55) {$1$};
		\node [gray] at (0.1,-0.55) {$1$};
		\node [gray] (b1) at (1.5,1.7) {$b=-x_2^e$};
		\node [gray] (b2) at (1.5,-1.7) {$b=-x_1^e$};
		\node [gray] at (2.8, 0.85) {$\frac{1}{x_1^e-x_2^e}$};
		\node [gray] at (2.8, -0.85) {$\frac{1}{x_2^e-x_1^e}$};
		
		\draw [lightgray,<->, line width=0.4mm] (6.9,1.6) -- (6.9,0.4);
		\node [gray] at (7.2, 1) {$u_1^e$};
		\draw [lightgray,<->, line width=0.4mm] (6.9,-1.6) -- (6.9,-0.4);
		\node [gray] at (7.2, -1) {$u_2^e$};
		
		\draw [line width=0.6mm,lightgray] (8.6,0.2) -- (9,0.2) -- (9,0.6) -- (9.6,0.6) -- (9.6,0.2) -- (10,0.2);
		\node [gray] at (9.3,-0.5) {\begin{tabular}{c} restrict\\to $\Omega^e$ \end{tabular}};
		
		\draw [line width=0.4mm,->, lightgray] (x.east) -- (x1.west);
		\draw [line width=0.4mm,->, lightgray] (x.east) -- (y1.west);
		\draw [line width=0.4mm,->, lightgray, dashed] (x.east) -- (0,1.75);
		\draw [line width=0.4mm,->, lightgray, dashed] (x.east) -- (0,-1.75);
		\draw [line width=0.4mm,->, lightgray] (x1.east) -- (x2.west);
		\draw [line width=0.4mm,->, lightgray] (y1.east) -- (y2.west);
		\draw [line width=0.4mm,->, lightgray] (x2.east) -- (u.west);
		\draw [line width=0.4mm,->, lightgray] (y2.east) -- (u.west);
		\draw [line width=0.4mm,->, lightgray] (u.east) -- (uh.west);
		\draw [line width=0.4mm,->, lightgray, dashed] (u-1.east) -- (uh.west);
		\draw [line width=0.4mm,->, lightgray, dashed] (u1.east) -- (uh.west);
		\draw [line width=0.4mm,->, lightgray, dashed] (6.3,2.5) -- (u-1.west);
		\draw [line width=0.4mm,->, lightgray, dashed] (6.3,1.5) -- (u-1.west);
		\draw [line width=0.4mm,->, lightgray, dashed] (6.3,-2.5) -- (u1.west);
		\draw [line width=0.4mm,->, lightgray, dashed] (6.3,-1.5) -- (u1.west);
		
		\draw [line width=0.4mm,->, lightgray] (b1.south) -- (x1.north);
		\draw [line width=0.4mm,->, lightgray] (b2.north) -- (y1.south);
		
		\end{tikzpicture}
		\caption{HiDeNN.}
	\end{subfigure}
	\caption{HiDeNN with one-dimensional linear elements~\cite{zhang_hierarchical_2021}.}\label{fig:hideNN}
\end{figure}
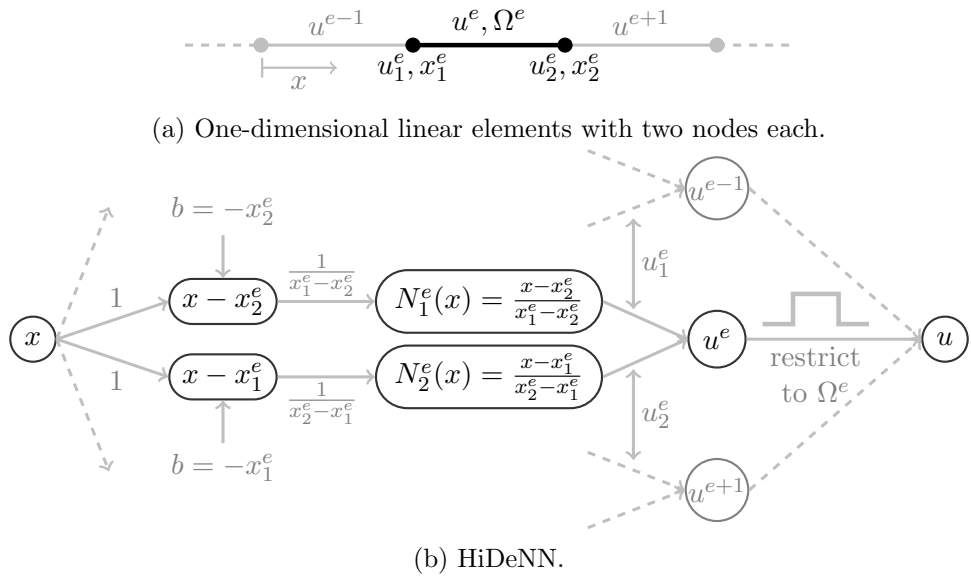

A different approach are the hierarchical deep-learning NNs (HiDeNNs)~\cite{zhang_hierarchical_2021} with extensions in~\cite{saha_hierarchical_2021,zhang_hidenn-td_2022,liu_hidenn-fem_2023,lu_convolution_2023,park_convolution_2023,li_convolution_2023}. Here, shape functions are treated as NNs constructed from basic building blocks. Consider, for example, the one-dimensional linear shape functions
\begin{align}
N_1(x)=\frac{x-x_2^e}{x_1^e-x_2^e}\\
N_2(x)=\frac{x-x_1^e}{x_2^e-x_1^e},
\end{align}
which can be represented as a NN, as shown in~\Cref{fig:hideNN}, where the weights depend on the nodal positions $x_1^e, x_2^e$. The interpolated displacement field $u^e$, which is valid in the element domain $\Omega^e$, is obtained by multiplication with the nodal displacements $u_1^e, u_2^e$, treated as shared NN weights.
\begin{equation}
u^e=N_1^e(x)u_1^e+N_2^e(x)u_2^e
\end{equation}
They are shared, as the nodal displacements $u_1^e, u_2^e$ are also used for the neighboring elements $u^{e-1}, u^{e+1}$. Finally the displacement over the entire domain $u$ is obtained by superposition of all elemental displacement fields $u^e$, which are first multiplied by a step function defined as $1$ inside the corresponding element domain $\Omega^e$ and $0$ outside. \\

A forward problem is solved with a minimization of the variational loss function, as presented in~\Cref{ssec:physicalmodeling} with the nodal values $u^e_i$ as learnable weights. According to~\cite{zhang_hierarchical_2021}, this is equivalent to iterative solution procedures in finite elements. The additional advantage is a seamless integration of $r$-refinement, i.e., the shift of nodal positions to optimal positions by making the nodal positions $x_i^e$ learnable. Special care has to be taken to avoid element inversion, which is handled by an additional term in the loss function. Inverse problems can similarly be solved by using learnable input parameters, as presented for topology optimization~\cite{li_convolution_2023}.\\

The method has been combined with reduced order modeling techniques~\cite{zhang_hidenn-td_2022}. Furthermore, the shape functions have been extended with convolutions~\cite{lu_convolution_2023,park_convolution_2023}. Specifically, a second weighting field $W(x)$ is introduced to enhance the finite element space $u^c(x)$ through convolutions:
\begin{equation}
    u^c(x)=u^e (x) * W(x).
\end{equation}
This introduces a smoothing effect over the elements and can efficiently be implemented using CNNs and, thereby, obtain a more favorable data-structure to exploit the full parallelization capabilities of GPUs~\cite{park_convolution_2023}. The enhanced space has been incorporated in the HiDeNN framework. While an independent confirmation is still missing, the authors promise a speedup of several orders of magnitude compared to traditional finite element solvers~\cite{li_convolution_2023}.




\begin{figure}[htb]
\centering
    \begin{tikzpicture}
        \draw[line width=0.3mm] (0,0) -- (4,0);
        \draw[line width=0.3mm, black, fill=white] (0,0) circle (0.1cm);
        \draw[line width=0.3mm, black, fill=white] (2,0) circle (0.1cm);
        \draw[line width=0.3mm, black, fill=white] (4,0) circle (0.1cm);

        \node at (0,-0.4) {$u_{j-1}$};
        \node at (2,-0.4) {$u_{j}$};
        \node at (4,-0.4) {$u_{j+1}$};
        \node at (0.8,0.4) {element $i$};
        \node at (3.2,0.4) {element $i+1$};
        \node at (4.4,0) {\dots};
        \node at (-0.4,0) {\dots};

        \draw[line width=0.3mm] (6,0) -- (8,0);
        \draw[line width=0.3mm, black, fill=white] (6,0) circle (0.1cm);
        \draw[line width=0.3mm, black, fill=white] (8,0) circle (0.1cm);
        \node at (6,-0.4) {$K_{i}^1$};
        \node at (8,-0.4) {$K_{i}^2$};
        \node at (7,0.4) {element $i$};
        
        \end{tikzpicture}
        \caption{Segment of one-dimensional finite element mesh with degrees of freedom (left). Local element definition with stiffness contributions (right).}\label{fig:segmentFE}
\end{figure}
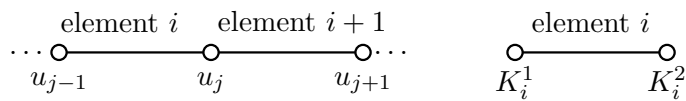

Another approach related to finite elements was presented as FEA-net~\cite{yao_fea-net_2019,yao_fea-net_2020}. Here, the matrix-vector multiplication of the global stiffness matrix $\boldsymbol{K}$ and solution vector $\boldsymbol{u}$ including the assembly of the global stiffness matrix is replaced by a convolution. In other words, the computation of the force vector $\boldsymbol{f}$ is used to compute the residual $\boldsymbol{r}$.
\begin{equation}
    \boldsymbol{r}=\boldsymbol{f} -\boldsymbol{K}\cdot \boldsymbol{u}\label{eq:feresidual}
\end{equation}
Assuming a uniform mesh with homogeneous material properties, the mesh is defined by the segment illustrated in~\Cref{fig:segmentFE}. The degree of freedom $u_j$ only interacts with the stiffness contributions $K_i^1, K_i^2, K_{i+1}^1, K_{i+1}^2$ of its neighboring elements $i$ and $i+1$. Therefore, the force component $f_j$ acting on node $j$ can be expressed by a convolution:
\begin{equation}
    f_j = [K_i^1, K_i^2+K_{i+1}^1, K_{i+1}^2] * [U_{j-1}, U_{j}, U_{j+1}]
\end{equation}
This can analogously be applied to all degrees of freedoms, with the same convolution filter $\boldsymbol{W} = [K^1, K^1 + K^2, K^2]$, assuming the same stiffness contributions for each element.
\begin{equation}
    \boldsymbol{K}\cdot \boldsymbol{u} = \boldsymbol{W} * \boldsymbol{U} \\
\end{equation}
The convolution can then be exploited in iterative schemes which minimize the residual $\boldsymbol{r}$ from~\Cref{eq:feresidual}. This saves the effort of constructing and storing the global stiffness matrix. By constructing the filter $\boldsymbol{W}$ as a function of the material properties of the adjacent elements, heterogeneities can be taken into account~\cite{yao_fea-net_2020}. If the same iterative solver is employed, FEA-Net is able to outperform classical finite elements for non-linear problems on uniform grids.

\subsection{Finite Difference Method}
Similar ideas have been proposed for finite differences~\cite{mishra_nfdtd_2005}, as employed in~\cite{herrmann_use_2023}, for example, where convolutional kernels are used as an implementation of stencils exploiting the efficient NN libraries with GPU capabilities. Here, the learnable parameters can be the finite difference stencil for inverse problems or the output for forward problems. This has, for example, been presented in the context of full waveform inversion, which is modeled as a RNN~\cite{richardson_seismic_2018, sun_theory-guided_2020}. The stencils are written as convolutional filters and repeatedly applied to the current state and the corresponding inputs. These are the wave field, the material distribution, and the source. The problem can then be regarded as a RNN. However, it is computationally expensive to perform automatic differentiation throughout the time steps for full waveform inversion, thereby obtaining the sensitivities with respect to $\gamma$ -- both regarding memory and wall clock computational time. A remedy is to combine automatic differentiation with the adjoint state method as in~\cite{chandrasekhar_tounn_2021, berg_neural_2021,herrmann_use_2023} and discussed in~\Cref{sssec:optimization}.

Taking this idea one step further, the discretized wave equation can be regarded as an analog RNN~\cite{hughes_wave_2019} where the weights are the material distribution. Here, a binary material is learned in a trainable region between source and probing location. The input $x(t)$ is encoded as a signal and emitted as source, which is measured at the probing locations $y_i(t)$ as output. By integrating the outputs, a classification of the input can be performed.

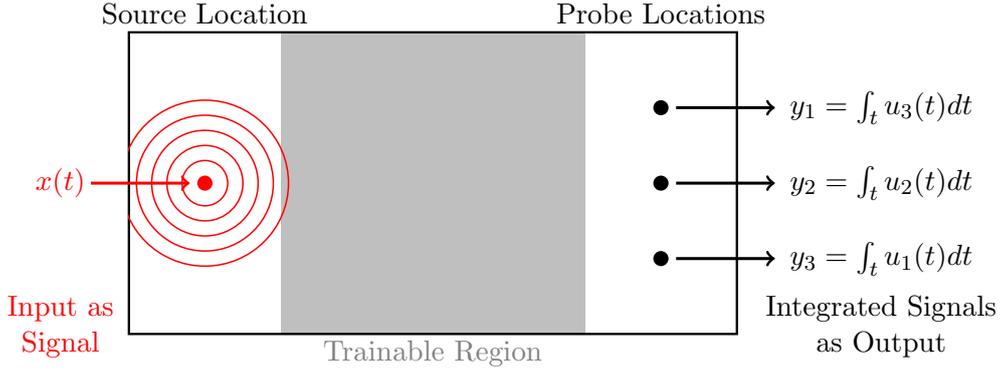
\begin{figure}[htb]
	\centering
	\begin{tikzpicture}
	\fill [lightgray, line width=0.3mm] (2,0) rectangle (6,4);
	\draw [line width=0.3mm] (0,0) rectangle (8,4);
	
	\begin{scope}
	\clip (0,0) rectangle (8,4);
	\foreach {\r} in {0.3,0.5,...,1.1} {
		\draw [line width=0.2mm, red] (1,2) circle (\r cm);
	}
	\end{scope}
	\fill [red] (1,2) circle (0.1cm);
	
	\fill [black] (7,1) circle (0.1cm);
	\fill [black] (7,2) circle (0.1cm);
	\fill [black] (7,3) circle (0.1cm);
	\draw [black, line width=0.4mm,->] (7.2,1) -- (8.5,1);
	\draw [black, line width=0.4mm,->] (7.2,2) -- (8.5,2);
	\draw [black, line width=0.4mm,->] (7.2,3) -- (8.5,3);
	\node at (9.9,1) {$y_3=\int_t u_1(t) dt$};
	\node at (9.9,2) {$y_2=\int_t u_2(t) dt$};
	\node at (9.9,3) {$y_1=\int_t u_3(t) dt$};
	
	\draw [red, line width=0.4mm,->] (-0.5,2) -- (0.8,2);
	\node [red] at (-0.9,2) {$x(t)$};
	
	\node [red] at (-0.9, 0.1) {\begin{tabular}{c}Input as\\Signal\end{tabular}};
	\node at (7, 4.25) {Probe Locations};
	\node [gray] at (4,-0.25) {Trainable Region};
	\node at (1, 4.25) {Source Location};
	\node at (9.9, 0.1) {\begin{tabular}{c}Integrated Signals\\as Output \end{tabular}};
	
	\end{tikzpicture}
	\caption{Analog RNN.}
\end{figure}

\subsection{Material Discretizations}
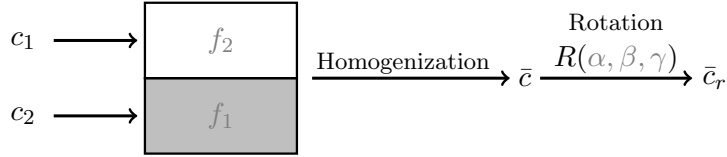
\begin{figure}[htb]
	\centering
	\begin{tikzpicture}
	\fill [lightgray, line width=0.3mm] (0,0) rectangle (2,1);
	\draw [line width=0.3mm] (0,0) rectangle (2,1);
	\draw [line width=0.3mm] (0,1) rectangle (2,2);
	
	\node [gray] at (1,0.5) {$f_1$};
	\node [gray] at (1,1.5) {$f_2$};
	
	\draw [black, line width=0.4mm,->] (2.2,1) -- (4.8,1);
	\node at (3.4,1.2) {\footnotesize Homogenization};
	\node at (5,1) {$\bar{c}$};
	\draw [black, line width=0.4mm,->] (5.2,1) -- (7.2,1);
	\node at (6.2,1.5) {\begin{tabular}{c}\footnotesize Rotation \\$R({\color{gray}\alpha,\beta,\gamma})$\end{tabular}};
	\node at (7.5,1) {$\bar{c}_r$};
	
	\node at (-1.6,1.5) {$c_1$};
	\node at (-1.6,0.5) {$c_2$};
	\draw [black, line width=0.4mm, ->] (-1.2,1.5) -- (-0.1,1.5);
	\draw [black, line width=0.4mm, ->] (-1.2,0.5) -- (-0.1,0.5);
	\end{tikzpicture}
	\caption{A single building block of the deep material network~\cite{liu_deep_2019}.}\label{fig:deepmaterialnetwork}
\end{figure}

Deep material networks~\cite{liu_deep_2019, liu_exploring_2019} construct a NN from a material distribution. An output is constructed from basic building blocks, inspired by analytical homogenization techniques. Given two materials defined in terms of their compliance tensors $c_1$, $c_2$, and volume fractions $f_1, f_2$, an analytical effective compliance tensor $\bar{c}$ is computed. The effective tensor is subsequently rotated with a rotation tensor $R$, defined in terms of the three rotation angles $\alpha, \beta, \gamma$, yielding a rotated effective tensor $\bar{c}_r$. Thus, the building block takes as input two compliance tensors $c_1,c_2$ and outputs a rotated effective compliance tensor $\bar{c}_r$, where $f_1, f_2, \alpha, \beta, \gamma$ are the learnable parameters (see~\Cref{fig:deepmaterialnetwork}). By connecting these building blocks, a large network can be created. The network is applied to homogenization tasks of RVEs~\cite{liu_deep_2019, liu_exploring_2019}, where the material of the phases is varied during evaluation.

\subsection{Neural Differential Equations}
In a more general setting, neural ordinary differential equations~\cite{chen_neural_2019} consider the forward Euler discretization of ordinary differential equations. Specifically, RNNs are viewed as Euler discretizations of continuous transformations~\cite{haber_stable_2018, ruthotto_deep_2018,lu_beyond_2020}. Consider the iterative update rule of the hidden states $y_{t+1}=y(t+\Delta t)$ of a RNN.
\begin{equation}
y_{t+1}=y_t+f(y_t;\boldsymbol{\theta})
\end{equation}
Here, $f$ is the evaluation of one recurrent unit in the RNN. In the limit of the time step size $\lim{\Delta t\rightarrow 0}$, the dynamics of the hidden units $y_t$ can be parametrized by an ordinary differential equation
\begin{equation}
\frac{dy(t)}{dt}=f(y(t),t;\boldsymbol{\theta}) \label{eq:neuralODE}
\end{equation} 
The input to the network is the initial condition $y(0)$, and the output is the solution $y(T)$ at time $T$. The output of the NN, $y(T)$, is obtained by solving~\Cref{eq:neuralODE} with a differential equation solver. The sensitivity computation for the weight update is obtained using the adjoint state method~\cite{pontriagin_mathematical_1986, givoli_tutorial_2021}, as backpropagating through each time step of the solver leads to a high memory cost. This also makes it possible to treat the solver as a black box. Similar extensions to PDEs~\cite{ruthotto_deep_2018} have been proposed by considering recurrent CNNs with residual connections, where the CNNs act as spatial gradients. \\

Similarly, \cite{yu_physics-based_2018} establish a connection between deep residual RNNs and iterative solvers. Residual connections in NNs allow information to bypass NN layers. Consider the estimation of the next state of a PDE with a classical solver $u_{t+1}=u(t+\Delta t)=F[u(t)]$. The residual $r_{t+1}=r(t+\Delta t)$ is determined in terms of the ground truth $u_{t+1}^{\mathcal{M}}$:
\begin{equation}
    r_{t+1} = u_{t+1}^{\mathcal{M}} - u_{t+1}.
\end{equation}
An iterative correction scheme is formulated with a NN. The iterations are indicated with the superindex $(k)$.
\begin{align}
    u_{t+1}^{(k+1)}&=u_{t+1}^{(k)} + f_{NN}(r_{t+1}^{(k+1)};\boldsymbol{\theta}) \\
    r_{t+1}^{(k+1)} &= u_{t+1}^{\mathcal{M}} - u_{t+1}^{(k)}
\end{align}
Note that the residual connection, i.e., $u_{t+1}^{(k)}$ as directly used in the prediction of $u_{t+1}^{(k+1)}$, allows information to pass past the recurrent unit $f_{NN}$. A related approach can be found in~\cite{ranade_discretizationnet_2021}, where an autoencoder iteratively acts on a solution until convergence. In the first iteration, a random initial solution is used as input. 


\section{Generative Approaches}\label{sec:generativeapproaches}
Generative approaches (see~\cite{regenwetter_deep_2022} for an in-depth review in the field of design and~\cite{foster_generative_2023} for a hands-on textbook) aim to model the underlying probability distribution of a data set to generate new data that resembles the training data. Three main methodologies exist: 
\begin{itemize}
    \item autoencoders,
    \item generative adversarial networks (GANs),
    \item diffusion models.
\end{itemize}
Currently, there are two prominent areas of application in computational mechanics. One area of focus is microstructure generation (\Cref{ssec:datageneration}), which aims to produce a sufficient quantity of realistic training data for surrogate models, as described in~\Cref{ssec:datadrivensurrogatemodeling}. The second key application area is generative design (\Cref{ssec:generativedesign}), which relies on algorithms to efficiently explore the design space within the constraints established by the designer. 

\subsection{Autoencoders}\label{ssec:autoencoders}
Autoencoders facilitate data generation by mapping high-dimensional training data $\{\boldsymbol{x}_i\}_{i=1}^N$ to a lower-dimensional latent space $\{\boldsymbol{h}_i\}_{i=1}^N$ which can be sampled efficiently. Specifically, an encoder $\boldsymbol{\hat{h}} = E_{NN}(\boldsymbol{x}; \boldsymbol{\theta}^e)$ transforms an input sample $\boldsymbol{x}$ to a reduced latent vector $\boldsymbol{\hat{h}}$. A corresponding decoder $\boldsymbol{\hat{x}}=D_{NN}(\boldsymbol{\hat{h}};\boldsymbol{\theta}^d)$ reconstructs the original sample $\boldsymbol{x}$ from this latent vector $\boldsymbol{\hat{h}}$. As mentioned in~\Cref{par:mor}, the encoder can serve as a tool for dimensionality reduction, whereas the decoder, within the scope of generative approaches, operates as a generator. By emulating the probability distribution of the latent space $\{\boldsymbol{\hat{h}}_i\}_{i=1}^N$, variational autoencoders~\cite{rezende_stochastic_2014, kingma_auto-encoding_2022} are able to generate new data that resembles the training data.
\subsection{Generative Adversarial Networks}\label{ssec:GANs}
GANs~\cite{goodfellow_generative_2014} emulate data distributions by setting up a two-player adversarial game between two NNs: 
\begin{itemize}
    \item the generator $G_{NN}$,
    \item the discriminator $D_{NN}$.
\end{itemize} 
The generator creates predictions $\boldsymbol{\hat{y}} = G_{NN}(\boldsymbol{\xi};\boldsymbol{\theta}_G)$ from random noise $\boldsymbol{\xi}$, while the discriminator attempts to distinguish between these generated predictions $\boldsymbol{\hat{y}}$ from real data $\boldsymbol{y}^{\mathcal{M}}$. The discriminator assigns a probability score $\hat{p}=D_{NN}(\boldsymbol{y};\boldsymbol{\theta}_D)$ which evaluates the likelihood of a datapoint $\boldsymbol{y}$ being real or generated. The quality of both the generator and the discriminator can be expressed via the following cost function:
\begin{equation}
C= \frac{1}{N_D}\sum_{i=1}^{N_D} \log \Bigl[D_{NN}(\boldsymbol{y}_i;\boldsymbol{\theta}_D)\Bigr]+\frac{1}{N_G}\sum_{i=1}^{N_G}\log\Bigl[1-D_{NN}\bigl(G_{NN}(\boldsymbol{\xi}_i;\boldsymbol{\theta}_G);\boldsymbol{\theta}_D\bigr)\Bigr]. \label{eq:ganloss}
\end{equation}
Here, $N_D$ real samples and $N_G$ generated samples are used for training. The goal for the generator is to minimize the cost function, implying that the discriminator fails to distinguish between real and generated samples. However, the discriminator strives to maximize the cost. Therefore, this is formulated as a minimax optimization problem
\begin{equation}
\min_{\boldsymbol{\theta}_G}\max_{\boldsymbol{\theta}_D} C.
\end{equation}
Convergence is ideally reached at the Nash equilibrium~\cite{nash_equilibrium_1950}, where the discriminator always outputs a probability of $1/2$, signifying its inability to distinguish between real and generated samples. \\
However, GANs can be challenging to train. Problems like mode collapse~\cite{salimans_improved_2016} can arise. Here, the generator learns only a few modes from the training data. In the extreme case, only a single sample from the training data is learned, yielding a low discriminator score, yet an undesirable outcome. To combat mode collapse, design diversity can be either promoted in the learning algorithm or the cost~\cite{salimans_improved_2016,srivastava_veegan_2017}. Another challenge lies in balancing the training of the two NNs. If the discriminator learns too quickly and manages to distinguish all generated samples, the gradient of the cost function (\Cref{eq:ganloss}) with respect to the weights becomes zero, halting further progress. A possible remedy is to use the Wasserstein distance in the cost function~\cite{arjovsky_wasserstein_2017}.\\

Additionally, GANs can be modified to include inputs that control the generated data. This can be achieved in a supervised manner with conditional GANs~\cite{mirza_conditional_2014}. The conditional GAN does not just receive random noise, but also an additional input. This supplementary input is considered by the discriminator, which assesses whether the input-output pair are real or generated. 
An unsupervised alternative are InfoGANs~\cite{chen_infogan_2016}, which disentangle the input information, i.e., the random input $\xi$, defining the generated data. This is achieved by introducing an additional parameter $c$, a latent code to the generator $G_{NN}(\xi, c;\boldsymbol{\theta}_G)$. To ensure that the parameter is used by the NN, the cost (\Cref{eq:ganloss}) is extended by a mutual information term~\cite{bridle_unsupervised_1991} $I(c, G_{NN}(x, c;\boldsymbol{\theta}_G))$ ensuring that the generated data varies meaningfully based on the input latent code $c$. \\

In comparison to variational autoencoders, GANs typically generate higher quality data. However, the advantage of autoencoders lies in their ability to construct a well-structured latent space, where proper sampling leads to smooth interpolations in the generated space. In other words, small changes in the latent space correspond to small changes in the generated space -- a characteristic not inherent to GANs. To achieve smooth interpolations, autoencoders can be combined with GANs~\cite{larsen_autoencoding_2016}, where the autoencoder acts as generator in the GAN framework, employing both an autoencoder loss and a GAN loss.

\subsection{Diffusion Models}
Diffusion models enhanced by NNs~\cite{sohl-dickstein_deep_2015, ho_denoising_2020, nichol_improved_2021} convert random noise $\boldsymbol{x}$ into a sample resembling the training data through a series of transformations. Given a data set $\{ \boldsymbol{y}^0_i\}_{i=1}^N$ that corresponds to the distribution $q(\boldsymbol{x}^0)$, a forward noising process $q(\boldsymbol{x}^t|\boldsymbol{x}^{t-1})$ is introduced. This process adds Gaussian noise to $\boldsymbol{x}^{t-1}$ at each time step $t-1$. The process is applied iteratively
\begin{equation}
q(\boldsymbol{x}^0,\boldsymbol{x}^1,\dots,\boldsymbol{x}^T)=q(\boldsymbol{x}^0)\prod_{t=1}^T q(\boldsymbol{x}^t|\boldsymbol{x}^{t-1}).
\end{equation}
After a sufficient number of iterations $T$, the resulting distribution approximates a Gaussian distribution. Consequently, a random sample from a Gaussian distribution $\boldsymbol{x}_T$ can be denoised with the reverse denoising process $q(\boldsymbol{x}^{t-1}|\boldsymbol{x}^t)$, resulting in a sample $\boldsymbol{x}^0$ that matches the original distribution $q(\boldsymbol{x}^0)$. The reverse denoising process is, however, unknown and therefore modeled as a Gaussian distribution, where the mean and covariance are learned by a NN. With the learned denoising process, data can be generated by denoising samples drawn from a Gaussian distribution. Note the similarity to autoencoders. Instead of learning a mapping to a hidden random state $\boldsymbol{h}_i$, the encoding is prescribed as the iterative application of Gaussian noise~\cite{foster_generative_2023}. \\

A related approach are normalizing flows~\cite{rezende_variational_2015} (see~\cite{kobyzev_normalizing_2021} for an introduction and extensive review). Here, a basic probability distribution is transformed through a series of invertible transformations, i.e., flows. The goal is to model distributions of interest. The individual transformations can be modeled by NNs. A normalization is required, such that each intermediate probability distribution integrates to one. 
\subsection{Applications}
\subsubsection{Data Generation}\label{ssec:datageneration}
The most straightforward application of variational autoencoders and GANs in computational mechanics is the generation of new data, based on existing examples. This has been demonstrated in~\cite{mosser_reconstruction_2017,feng_reconstruction_2019,shams_coupled_2020,xia_multi-scale_2022, henkes_three-dimensional_2022} for microstructures in~\cite{araya-polo_deep_2019} for velocity models used in full waveform inversion, and in~\cite{rawat_novel_2019} for optimized structures using GANs. Variational autoencoders have also been used to model the crossover operation in evolutionary algorithms to create new designs from parent designs~\cite{yaji_data-driven_2022}. Applications of diffusion models for microstructure generation can be found in~\cite{lee_microstructure_2023, dureth_conditional_2023,vlassis_denoising_2023}. \\

Microstructures pose a unique challenge due to their inherent three-dimensional nature, while often only two-dimensional reference images are available. This has led to the development of specialized architectures that are capable of creating three-dimensional structures from representative two-dimensional slices~\cite{feng_end--end_2020,kench_generating_2021,li_cascaded_2022}. The approach typically involves treating three-dimensional voxel data as a sequence of two-dimensional slices of pixels. Sequences of images are predicted from individual slices, ultimately forming a three-dimensional microstructure. In~\cite{zhang_3d-pmrnn_2022}, a RNN is applied to a two-dimensional reference image, yielding an additional dimension, and consequently creating a three-dimensional structure. The RNN is applied at the latent vector inside an encoder decoder architecture, such that the inputs and outputs of the RNN have a relatively small size. Similarly, \cite{zheng_rockgpt_2022, phan_size-invariant_2022} apply a transformer~\cite{vaswani_attention_2017} to the latent vector. An alternative formulation using variational autoencoder GANs is presented in~\cite{zhang_slice--voxel_2021} to reconstruct three-dimensional voxel models of porous media from two-dimensional images. \\

The generated data sets can subsequently be leveraged to train surrogate models, as demonstrated in~\cite{rawat_novel_2019, rawat_application_2019, rawat_novel_2019-1, shen_new_2019} where CNNs were used to verify the physical properties of designs, and in the study by~\cite{wessels_computational_2022} on the homogenization of microstructures with CNNs. Similarly, \cite{mosser_stochastic_2020,araya-polo_deep_2019} generate realistic material distributions, such as velocity distributions, to train an inverse operator for full waveform inversion.




\subsubsection{Generative Design \& Design Optimization}\label{ssec:generativedesign}
Within generative design, the generator can also be considered as a reparametrization of the design space that reduces the number of design variables. With autoencoders, the latent vector serves as the design parameter~\cite{guo_indirect_2018,vulimiri_integrating_2021}, which is then optimized\footnote{It is worth noting, that to ensure designs that are physically meaningful, a style transfer technique can be implemented~\cite{gatys_neural_2016}. Here, the training data is perceived as a style, and the Gram matrices' difference, characterizing the distribution of visual patterns or textures in the generated designs, is minimized.}. In the context of GANs, the optimization task is aimed at the random input $\boldsymbol{\xi}$ provided to the generator. This approach is demonstrated in various studies, such as ship hull design parameterized by NURBS surfaces~\cite{khan_shiphullgan_2023}, airfoil shapes expressed with B\'{e}zier curves~\cite{chen_inverse_2022, chen_beziergan_2021}, structural optimization~\cite{chen_mo-padgan_2021}, and full waveform inversion~\cite{richardson_generative_2018}. For optimization, variational autoencoder GANs are particularly important, as the GAN ensures high quality designs, while the autoencoder ensures well-behaving gradients. This was shown for microstructure optimization in~\cite{zhang_da-vegan_2023}.\\

An important requirement for generative design is design diversity. Achieving this involves ensuring that the entire design space is spanned by the generated data. For this, the cost function can be extended, as presented in~\cite{chen_padgan_2021}, using determinantal point processes~\cite{kulesza_determinantal_2012} or in~\cite{khan_shiphullgan_2023} with a space-filling term~\cite{bates_formulation_2003}. \\

Other strategies are specifically focused on promoting design diversity. This involves identifying novel designs via a novelty score~\cite{heyrani_nobari_creativegan_2021}. The novelty within these designs is segmented and used to modify the GAN using methods outlined in~\cite{bau_rewriting_2020}. An alternative approach proposed by~\cite{elgammal_can_2017} quantifies creativity and maximizes it. This is achieved by performing a classification in pre-determined categories by the discriminator. If the classification is unsuccessful, the design must lie outside the categories and is therefore deemed creative. Thus the generator then seeks to minimize the classification accuracy.\\

However, some applications necessitate a resemblance to prior designs due to factors such as aesthetics~\cite{oh_deep_2019} or manufacturability~\cite{greminger_generative_2020}. In~\cite{oh_deep_2019}, a pixel-wise $L^1$-distance to previous designs is included in the loss\footnote{Similarly, this loss can be used to filter out designs that are too similar.}. A complete workflow with generative design enforcing resemblance of previous designs and surrogate model training for the quantification of mechanical properties is described in~\cite{yoo_integrating_2021}. Another option is the use of style transfer techniques~\cite{gatys_neural_2016}, which in~\cite{zhang_machine-learning_2023} is incorporated into a conventional topology optimization scheme~\cite{bendsoe_topology_2003} as a constraint in the loss. These are tools with the purpose of incorporating vague constraints based on previous designs for topology optimization.\\

GANs can also be applied to inverse problems, as presented in~\cite{yang_fwigan_2023} for full waveform inversion. The generator predicts the material distribution, which is used in a differentiable simulation providing the forward solution in the form of a seismogram. The discriminator attempts to distinguish between the seismogram indirectly coming from the generator and the measured seismograms. The underlying material distribution is determined through gradient descent.

\subsubsection{Conditional Generation}
As stated earlier, GANs can take specific inputs to dictate the output's nature. The key difference to data-driven surrogate models from~\Cref{ssec:datadrivensurrogatemodeling} is that GANs provide a tool to generate multiple outputs given the same conditional input. They are thus applicable to problems with multiple solutions, such as design optimization or data generation. \\

Examples of conditional generation are rendered cars from car sketches~\cite{radhakrishnan_creative_2018}, hierarchical shape generation~\cite{chen_synthesizing_2019}, where the child shape considers its parent shape and topology optimization with predictions of optimal structures from initial fields, e.g., strain energy, of the unoptimized structure~\cite{nie_topologygan_2020,hertlein_generative_2021}. Physical properties can also be used as input. The properties are computed by a differentiable solver after generation and are incorporated in the loss. This was, e.g., presented in~\cite{heyrani_nobari_range-gan_2021} for airplane shapes, and in~\cite{wang_ih-gan_2022} for inverse homogenization. For full waveform inversion, \cite{duque_automated_2019} trains a conditional GAN with seismograms as input to predict the corresponding velocity distributions. A similar effort is made by~\cite{wang_seismic_2022} with CycleGANs~\cite{zhu_unpaired_2020} to circumvent the need for paired data. Here, one generator generates a seismogram $\hat{y}=G_y(x)$ and another a corresponding velocity distribution $\hat{x}=G_x(y)$. The predictions are judged by two separate discriminators. Additionally, a cycle-consistency loss ensures that a prediction from a prediction, i.e., $G_y(\hat{x})$ or $G_x(\hat{y})$, matches the initial input $x$ or $y$. This cycle-consistency loss ensures, that the learned transformations preserve the essential features and structures of the original seismograms or velocity distributions when they are transformed from seismogram to velocity distribution and back again. \\

Lastly, coarse-to-fine mappings as previously discussed in~\Cref{ssec:postprocessing}, can also be learned by GANs. This was, for example, demonstrated in topology optimization, where a conditional GAN refines coarse designs obtained from classical optimizations~\cite{li_non-iterative_2019,nie_topologygan_2020} or CNN predictions~\cite{yu_deep_2019}. For temporal problems, such as fluid flows, the temporal coherence between time steps poses an additional challenge. Temporal coherence can be ensured by a second discriminator, which receives three consecutive frames of either the generator or the real data and decides if they are real or generated. The method is referred to as tempoGAN~\cite{xie_tempogan_2018}.




\subsubsection{Anomaly Detection}
Finally, a last application of generative models is anomaly detection, see~\cite{pang_deep_2022} for a review. This is particularly valuable for non-destructive testing, where flawed specimens can be identified in terms of anomalies. The approach relies on generative models and attempts to reconstruct the geometry. At first, the generative model is trained on structures without flaws. During evaluation, the structures to be tested are then fed through the NN. In case of an autoencoder, as in~\cite{hawkins_outlier_2002}, it is fed through the encoder and decoder. For a GAN, as discussed, e.g., in~\cite{schlegl_unsupervised_2017,zenati_efficient_2019,schlegl_f-anogan_2019}, the input of the generator is optimized to fit the output as well as possible. The mismatch in reconstruction then provides a spatially dependent measure of where an anomaly, i.e., defect is located. \\

Another approach is to use the discriminator directly, as presented in~\cite{henkes_gan_2023}. If a flawed specimen is given to the discriminator, it will be categorized as fake, as it was not part of the undamaged structures during training. The discriminator can also be used to check if the domain of application of a surrogate model is valid. Trained on the same training data as the surrogate model, the discriminator estimates the dissimilarity between the data to be tested and the training data. For large discrepancies, the discriminator detects that the surrogate model becomes invalid.\footnote{Note however, that the discriminator does not guarantee an accurate assessment of the validity of the surrogate model.}

\section{Deep Reinforcement Learning}\label{sec:deepreinforcement}
\begin{figure}[htb]
	\centering
	\begin{tikzpicture}
	\node (A) [draw, thick] at (0,0) {\begin{tabular}{c}Agent\end{tabular}};
	\node (E) [draw, thick] at (0,-2) {\begin{tabular}{c}Environment\end{tabular}};
	\draw [lightgray, thick,->] (A.east) -- (2,0) -- (2,-2) -- (E.east);
	\draw [lightgray, thick,->] (E.west) -- (-2,-2) -- (-2,0) -- (A.west);
	\node [gray] at (3,-1) {action $a_t$};
	\node [gray] at (-3,-1) {\begin{tabular}{c}state $s_t$\\ reward $r_t$\end{tabular}};
	\end{tikzpicture}
	\caption{Reinforcement learning in which an agent interacts with an environment with actions $a_t$, states $s_t$, and rewards $r_t$. Figure adapted from~\cite{sutton_reinforcement_2018}.}\label{fig:reinforcement}
\end{figure}
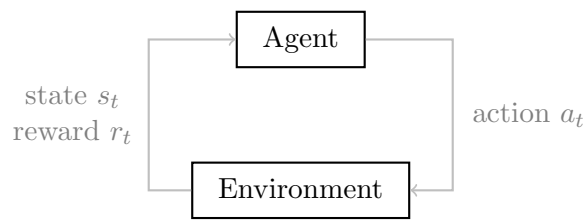

In reinforcement learning, an agent interacts with an environment through a sequence of actions $a_t$, which is illustrated in~\Cref{fig:reinforcement}. Upon executing an action $a_t$, the agent receives an updated state $s_{t+1}$ and reward $r_{t+1}$ from the environment. The agent's objective is to maximize the cumulative reward $R_{\Sigma}$. The environment can be treated as a black box. This presents an advantage in computational mechanics when differentiable physics are not feasible. Reinforcement learning has achieved impressive results such as human-level performance in games like Atari~\cite{mnih_human-level_2015}, Go~\cite{silver_mastering_2017}, and StarCraft II~\cite{vinyals_grandmaster_2019}. Further, reinforcement learning has been successfully been demonstrated in robotics~\cite{kober_reinforcement_2013}. An example hereof is learning complex maneuvers for autonomous helicopter flight~\cite{kim_autonomous_2003,abbeel_application_2006,abbeel_autonomous_2010}.

A comprehensive review of reinforcement learning exceeds the scope of this work, since it represents a major branch of machine learning. An introduction is, e.g., given in~\cite{garnier_review_2019,brunton_data-driven_2022}, and an in-depth textbook is~\cite{sutton_reinforcement_2018}. However, at the intersection of these domains lies deep reinforcement learning, which employs NNs to model the agent's actions. In~\Cref{sec:appendixdrl}, we present the main concepts of deep reinforcement learning and delve into two prominent methodologies: deep policy networks (\Cref{ssec:deeppolicynetworks}) and deep Q-learning (\Cref{ssec:deepqlearning}) in view of applications in computational mechanics. \\

\subsection{Applications}
Deep reinforcement learning is mainly used for inverse problems (see~\cite{garnier_review_2019} for a review within fluid mechanics), where the PDE solver is treated as a black box, and assumed to not be differentiable. \\

The most prominent application are control problems. One example is discovering swimming strategies for fish -- with the goal of efficiently minimizing the distance to a leader fish~\cite{novati_synchronised_2017,verma_efficient_2018}. The environment is given by the Navier Stokes equation. Another example is balancing rigid bodies with fluid jets while using as little force as possible~\cite{ma_fluid_2018}. Similarly, \cite{rabault_artificial_2019} control jets in order to reduce the drag around a cylinder. Reducing the drag around a cylinder is also achieved by controlling small rotating cylinders in the wake of the flow~\cite{fan_reinforcement_2020}. A more complex example is controlling unmanned aerial vehicles~\cite{xu_learning_2019}. The control schemes are learned by interacting with simulations and, subsequently, applied in experiments. \\

Further applications in connection with inverse problems are learning filters to perturb flows in order to match target flows~\cite{lee_flow_2018}. Also, constitutive laws can be identified. The individual arithmetic manipulations within a constitutive law can be represented as graphs. An agent constructs the graph in order to best match simulation and measurement~\cite{wang_meta-modeling_2019}, which yields an interpretable law. \\

Topology optimization has also been tackled by reinforcement learning. Specifically, the ability to predict only binary states (material or no material) is desirable -- instead of intermediate states, as in solid isotropic material with penalization~\cite{bendsoe_optimal_1989,bendsoe_topology_2004}. This has been shown with binary truss structures, modeled with graphs in order to minimize the total structural volume under stress constraints. In~\cite{hayashi_reinforcement_2020}, an agent removes trusses from existing structures, and trusses are added in~\cite{zhu_machine-specified_2021}. Similarly, \cite{sun_generative_2020} removes finite elements in solid structures to modify the topology. Instead, \cite{jang_generative_2022} pursues design diversity. Here a NN surrogate model predicts near optimal structures from reference designs. The agent then learns to generate reference designs as input, such that the corresponding optimal structures are as diverse as possible. \\

Also, high-dimensional PDEs have been solved with reinforcement learning~\cite{han_solving_2018,e_deep_2017}. This is achieved by recasting the PDE into stochastic control problems, thereby solving these with reinforcement learning. \\

Finally, adaptive mesh refinement algorithms have been learned by reinforcement learning~\cite{yang_reinforcement_2023}. An agent decides whether an element is to be refined based on the current state, i.e., the mesh and solution. The reward is subsequently defined in terms of the error reduction, which is computed with a ground truth solution. The trained agent can thus be applied to adaptive mesh refinement to previously unseen simulations.

\subsubsection{Extensions}
Each interaction with the environment requires solving the differential equation, which, due to the many interactions, makes reinforcement learning expensive. The learning can be accelerated through some basic modifications. The learning can be perfectly parallelized by using multiple environments simultaneously~\cite{rabault_accelerating_2019}, or by using multiple agents within the same environment~\cite{novati_automating_2020}. Another idea is to construct a surrogate model of the environment and thereby exploit model-based approaches~\cite{liu_physics-informed_2021,shi_physics-informed_2023,ramesh_physics-informed_2023,rodwell_physics-informed_2023}. The general procedure consists of three steps: 
\begin{itemize}
    \item model learning: learn surrogate of environment,
    \item behavior learning: learn policy or value function,
    \item environment interaction: apply learned policy and collect data.
\end{itemize}
Most approaches construct the surrogate with data-driven modeling (\Cref{ssec:datadrivensurrogatemodeling}), but physics-informed approaches have been proposed as well~\cite{liu_physics-informed_2021,ramesh_physics-informed_2023} (\Cref{ssec:physicalmodeling}).

\section{Conclusion}

In order to structure the state-of-the-art, an overview of the most prominent deep learning methods employed in computational mechanics was presented. Six main categories were identified: simulation substitution, simulation enhancement, discretizations as NNs, generative approaches, and deep reinforcement learning. \\

Despite the variety and abundance of the literature, few approaches are competitive in comparison to classical methods. With only few exceptions, current research is still in its early stages, with a focus on showcasing possibilities without focusing too much attention on accuracy and efficiency. Future research must, nevertheless, shift its focus to incorporate more in-depth investigations into the performance of the developed methods -- including thorough and meaningful comparisons to classical methods. This is in agreement with the recent review article on deep learning in topology optimization~\cite{woldseth_use_2022}, where critical and fair assessments are requested. This includes the determination of generalization capabilities, greater transparency by including, e.g., worst case performances to illustrate reliability, and computation times without disregard of the training time. \\

In line with this, and to the best of our knowledge, we provide a final overview outlining the potentials and limitations of the discussed methods.
\begin{itemize}
    \item Simulation substitution has potential for surrogate modeling of parameterized models that need to be evaluated many times. This is however, currently only realizable for small parameter spaces, due to the amount of data required. Complex problems can still be solved if they are first reduced to a low-dimensional space through model order reduction techniques. Physics-informed learning further reduces the amount of required data and improves the generalization capabilities. However, enforcing physics through penalty terms increases the computational effort, where the solutions still do not necessarily satisfy the corresponding physical laws. Instead, enforcing physics by construction, which guarantees the enforced physics, seems more favorable.
    \item Simulation enhancement is currently one of the most useful approaches. It is in particular beneficial for tasks where classical methods show difficulties. An excellent example for this is the formulation of constitutive laws, which are inherently phenomenological and thereby well-suited to be identified from data using tools such as deep learning. In addition, simulation enhancement, makes it possible to draw on insights gained from classical methods developed since the inception of computational mechanics. Furthermore, it is currently more realistic to learn smaller components of the simulation chain with NNs rather than the entire model. These components should ideally be expensive and have limited requirements regarding the accuracy. Lastly, it is also easier to assess whether a method enhanced by deep learning outperforms the classical method, as direct and fair comparisons are readily possible.
    \item An important research direction is to employ discretizations as NNs, as this offers the potential to discover NNs tailored to computational mechanics tasks, such as CNNs for computer vision and RNNs and transformers for natural language processing. Their main benefit comes from exploiting the computational benefits of tools and hardware that were created for the wider community of deep learning -- such as NN libraries and GPUs. 
    \item Generative approaches have been shown to be highly versatile in applications of computational mechanics since the accuracy of a specific instance under investigation is less of a concern here. 
    They have been used for realistic data generation to train other machine learning models, incorporate vague constraints based on data within optimization frameworks, and detect anomalies.
    \item Deep reinforcement learning has already shown encouraging results -- for example in controlling unmanned vehicles in complex physics environments. It is mainly applicable for problems where efficient differentiable physics solvers are unavailable, which is why it is popular in control problems for turbulence. In the presence of differentiable solvers, gradient-based methods are however still the state-of-the-art~\cite{sigmund_usefulness_2011} and thus preferred.
\end{itemize}


\section*{Acknowledgements}
The authors gratefully acknowledge the funding through the joint research project Geothermal-Alliance Bavaria (GAB) by the Bavarian State Ministry of Science and the Arts (StMWK) as well as the Georg Nemetschek Institut (GNI) under the project DeepMonitor. 

\section*{Declarations}
\textbf{Conflict of interest} No potential conflict of interest was reported by the authors.

\appendix
\renewcommand\thesection{\Alph{section}}

\section{Deep Reinforcement Learning}\label{sec:appendixdrl}
In reinforcement learning, the environment is commonly modeled as a Markov Decision Process (MDP). This mathematical model is defined by a set of all possible states $S$, actions $A$, and associated rewards $R$. Furthermore, the probability of getting to the next state $s_{t+1}$ from the previous $s_t$ with action $a_t$ is given by $\mathbb{P}(s_{t+1}|s_t,a_t)$. Thus, the environment is not necessarily deterministic. One key aspect of a Markov Decision Process is the Markov property, stating that future states depend solely on the current state and action, and not the history of states and actions. \\

The goal of a reinforcement learning algorithm is to determine a policy $\pi(s,a)$ which dictates the next action $a_t$ in order to maximize the cumulative reward $R_{\Sigma}$. The cumulative reward $R_{\Sigma}$ is discounted by a discount factor $\gamma^t$ in order to give more importance to immediate rewards.
\begin{equation}
    R_{\Sigma}=\sum_{t=0}^\infty \gamma^t r_t
\end{equation}\\
The quality of a policy $\pi(s,a)$ can be assessed by a state-value function $V_{\pi}(s)$, defined as the expected future reward given the current state $s$ and following the policy $\pi$. Similarly, an action-value function $Q_{\pi}(s)$ determines the expected future reward given the current state $s$ and action $a$ and then following the policy $\pi$. The expected value along a policy $\pi$ is denoted as $\mathbb{E}_\pi$.
\begin{align}
    V_{\pi}(s)=\mathbb{E}_\pi\bigl[R_{\Sigma}(t)|s\bigr] \\
    Q_{\pi}(s,a)=\mathbb{E}_\pi\bigl[R_{\Sigma}(t)|s,a\bigr] \\
\end{align}
The optimal value and quality function correspondingly follow the optimal policy:
\begin{align}
    V(s)= \max_{\pi} V_{\pi}(s), \\
    Q(s,a)=\max_{\pi}Q_{\pi}(s).
\end{align}
The approaches can be subdivided into model-based and model-free. Model-based methods incorporate a model of the environment. In the most general sense, a probabilistic environment, this entitles the probability distribution of the next state $\mathbb{P}(s_{t+1}|s_t,a_t)$ and of the next reward $\mathbb{R}(r_{t+1}|s_{t+1},s_t,a_t)$. The model of the environment can be cheaply sampled to improve the policy $\pi$ with model-free reinforcement learning techniques~\cite{sutton_dyna_1991,janner_when_2019,kaiser_model-based_2020,luo_algorithmic_2021} discussed in the sequel (\Cref{ssec:deeppolicynetworks,ssec:deepqlearning}). However, if the model is differentiable, the gradient of the reward can directly be used to update the policy~\cite{deisenroth_pilco_2011,levine_learning_2014,heess_learning_2015,clavera_model-augmented_2020,hafner_dream_2020,hafner_mastering_2022}. This is identical to the optimization through differentiable physics solvers discussed in~\Cref{sssec:optimization}. Model-free reinforcement learning techniques can be used to enhance the optimization. \\






A further distinction is made between policy-based~\cite{williams_simple_1992,sutton_policy_1999,kakade_natural_2001,silver_deterministic_2014,schulman_trust_2015} and value-based~\cite{watkins_q-learning_1992,hasselt_deep_2016,wang_dueling_2016} approaches. Policy-based methods, such as deep policy networks~\cite{brunton_data-driven_2022} (\Cref{ssec:deeppolicynetworks}), directly optimize the policy. By contrast, value-based methods, such as deep Q-learning~\cite{wang_dueling_2016} (\Cref{ssec:deepqlearning}) learn the value function from which the optimal policy is selected. Actor-critic methods, such as proximal policy optimization~\cite{schulman_proximal_2017} combine the ideas with an actor that performs a policy and a critic that judges its quality. Both can be modeled by NNs.

\subsection{Deep Policy Networks}\label{ssec:deeppolicynetworks}
In deep policy networks, the policy, i.e., the mapping of states to actions, is modeled by a NN $\hat{a}=\pi(s;\boldsymbol{\theta})$. The quality of the NN is assessed by the expected cumulative reward $R_{\Sigma}$, formulated in terms of the action-value function $Q(s,a)$.
\begin{equation}
C = R_{\Sigma} = \mathbb{E}\bigl[Q(s,a)\bigr]\\
\end{equation}
Its gradient (see~\cite{brunton_data-driven_2022,sutton_policy_1999,silver_deterministic_2014} for a derivation), given as: 
\begin{equation}
    \nabla_{\boldsymbol{\theta}} R_{\Sigma} = \mathbb{E}\bigl[Q(s,a) \nabla_{\boldsymbol{\theta}} \log\bigl(\pi(s,a;\boldsymbol{\theta})\bigr)\bigr]
\end{equation}
can be applied within a gradient ascent scheme to learn the optimal policy.


\subsection{Deep Q-Learning}\label{ssec:deepqlearning}
Deep Q-learning identifies the optimal action-value function $Q(s,a)$ from which the optimal policy is extracted. Q-Learning relies on the Bellman optimality criterion~\cite{bellman_markovian_1957,dolcetta_approximate_1984}. By separating the reward $r_0$ at the first step, the recursion formula of the optimal state-value function, i.e., the Bellman optimality criterion, can be established:
\begin{align}
V(s) &= \max_\pi \mathbb{E}_\pi \Bigl[\sum_{t=0}^\infty \gamma^t r_t| s_0=s\Bigr] \\
&= \max_\pi \mathbb{E}_\pi \Bigl[r_0+\sum_{t=1}^\infty \gamma^t r_t| s_1=s'\Bigr]\\
&= \max_\pi \mathbb{E}_\pi \bigl[r_0 + \gamma V(s')\bigr].
\end{align}
Here, $s'$ represents the next state after $s$. This can be done analogously for the action-value function.
\begin{equation}
    Q(s,a) = \max_\pi \mathbb{E}_\pi \bigl[r_0 + \gamma Q(s',a')\bigr]
\end{equation}
The recursion enables an update formula, referred to as temporal difference (TD) learning~\cite{sutton_learning_1988,bradtke_linear_1996}. Specifically, the current estimate $Q^{(m)}$ at state $s_t$ is compared to the more accurate estimate at the next state $s_{t+1}$ using the obtained reward $r_t$, referred to as the TD target estimate. The difference is the TD error, which in combination with a learning rate $\alpha$ is used to update the function $Q^{(m)}$: 
\begin{equation}
    Q^{(m+1)}(s_t, a_t) = Q^{(m)}(s_t,a_t)+\alpha \overbrace{\bigl( \underbrace{r_t+\gamma \max_a Q(s_{t+1},a)}_{\textrm{TD target estimate}} - \underbrace{Q^{(m)}(s_t,a_t)}_{\textrm{model prediction}}\bigr)}^{\textrm{TD error}}. 
\end{equation}
Here, the TD target estimate only looks one step ahead -- and is therefore referred to as TD(0). The generalization is called TD(N). In the limit $N\rightarrow \infty$, the method is equivalent to Monte Carlo learning, where all steps are performed and a true target is obtained. \\

Deep Q-learning introduces a NN for the action-value function $Q(s,a;\boldsymbol{\theta})$. Its quality is assessed with a loss composed of the mean squared error of the TD error.
\begin{equation}
    C = \mathbb{E}\Bigl[\frac{1}{2}\bigl(r_t + \gamma \max_a Q(s_{t+1},a;\boldsymbol{\theta}) - Q(s_t,a_t;\boldsymbol{\theta})\bigr)^2\Bigr]
\end{equation}

Lastly, the optimal policy $\pi(s)$ maximizing the action-value function $Q(s,a;\boldsymbol{\theta})$ is extracted:
\begin{equation}
    \pi(s)=\underset{a}{\arg\max}\textrm{ }Q(s,a;\boldsymbol{\theta})
\end{equation}

\newpage
\bibliographystyle{ieeetr}

\setlength{\bibsep}{3pt}
\setlength{\bibhang}{0.75cm}{\fontsize{9}{9}\selectfont\bibliography{library}}

\begin{thebibliography}{100}

\bibitem{abu-mostafa_learning_2012}
Y.~S. Abu-Mostafa, M.~Magdon-Ismail, and H.-T. Lin, {\em Learning {From}
  {Data}}.
\newblock S.l.: AMLBook, 2012.

\bibitem{adie_deep_2018}
J.~Adie, Y.~Juntao, X.~Zhang, and S.~See, ``Deep {Learning} for {Computational}
  {Science} and {Engineering},'' 2018.

\bibitem{yagawa_computational_2023}
G.~Yagawa and A.~Oishi, {\em Computational mechanics with deep learning: an
  introduction}.
\newblock Cham, Switzerland: Springer, 2023.

\bibitem{zhang_ai_2022}
D.~Zhang, N.~Maslej, E.~Brynjolfsson, J.~Etchemendy, T.~Lyons, J.~Manyika,
  H.~Ngo, J.~C. Niebles, M.~Sellitto, E.~Sakhaee, Y.~Shoham, J.~Clark, and
  R.~Perrault, ``The {AI} {Index} 2022 {Annual} {Report},'' May 2022.
\newblock arXiv:2205.03468 [cs].

\bibitem{woldseth_use_2022}
R.~V. Woldseth, N.~Aage, J.~A. Bærentzen, and O.~Sigmund, ``On the use of
  artificial neural networks in topology optimisation,'' {\em Structural and
  Multidisciplinary Optimization}, vol.~65, p.~294, Oct. 2022.

\bibitem{shin_topology_2023}
S.~Shin, D.~Shin, and N.~Kang, ``Topology optimization via machine learning and
  deep learning: a review,'' {\em Journal of Computational Design and
  Engineering}, vol.~10, pp.~1736--1766, July 2023.

\bibitem{adler_deep_2021}
A.~Adler, M.~Araya-Polo, and T.~Poggio, ``Deep {Learning} for {Seismic}
  {Inverse} {Problems}: {Toward} the {Acceleration} of {Geophysical} {Analysis}
  {Workflows},'' {\em IEEE Signal Processing Magazine}, vol.~38, pp.~89--119,
  Mar. 2021.

\bibitem{garnier_review_2019}
P.~Garnier, J.~Viquerat, J.~Rabault, A.~Larcher, A.~Kuhnle, and E.~Hachem, ``A
  review on {Deep} {Reinforcement} {Learning} for {Fluid} {Mechanics},'' {\em
  arXiv:1908.04127 [physics]}, Aug. 2019.
\newblock arXiv: 1908.04127.

\bibitem{duraisamy_turbulence_2019}
K.~Duraisamy, G.~Iaccarino, and H.~Xiao, ``Turbulence {Modeling} in the {Age}
  of {Data},'' {\em Annual Review of Fluid Mechanics}, vol.~51, pp.~357--377,
  Jan. 2019.

\bibitem{brunton_machine_2020}
S.~Brunton, B.~Noack, and P.~Koumoutsakos, ``Machine {Learning} for {Fluid}
  {Mechanics},'' {\em Annual Review of Fluid Mechanics}, vol.~52, pp.~477--508,
  Jan. 2020.
\newblock arXiv: 1905.11075.

\bibitem{cai_physics-informed_2021}
S.~Cai, Z.~Mao, Z.~Wang, M.~Yin, and G.~E. Karniadakis, ``Physics-informed
  neural networks ({PINNs}) for fluid mechanics: a review,'' {\em Acta
  Mechanica Sinica}, vol.~37, pp.~1727--1738, Dec. 2021.

\bibitem{calzolari_deep_2021}
G.~Calzolari and W.~Liu, ``Deep learning to replace, improve, or aid {CFD}
  analysis in built environment applications: {A} review,'' {\em Building and
  Environment}, vol.~206, p.~108315, Dec. 2021.

\bibitem{bock_review_2019}
F.~E. Bock, R.~C. Aydin, C.~J. Cyron, N.~Huber, S.~R. Kalidindi, and
  B.~Klusemann, ``A {Review} of the {Application} of {Machine} {Learning} and
  {Data} {Mining} {Approaches} in {Continuum} {Materials} {Mechanics},'' {\em
  Frontiers in Materials}, vol.~6, p.~110, May 2019.

\bibitem{bishara_state---art_2023}
D.~Bishara, Y.~Xie, W.~K. Liu, and S.~Li, ``A {State}-of-the-{Art} {Review} on
  {Machine} {Learning}-{Based} {Multiscale} {Modeling}, {Simulation},
  {Homogenization} and {Design} of {Materials},'' {\em Archives of
  Computational Methods in Engineering}, vol.~30, pp.~191--222, Jan. 2023.

\bibitem{rosenkranz_comparative_2023}
M.~Rosenkranz, K.~A. Kalina, J.~Brummund, and M.~Kästner, ``A comparative
  study on different neural network architectures to model inelasticity,'' {\em
  International Journal for Numerical Methods in Engineering}, p.~nme.7319,
  July 2023.

\bibitem{regenwetter_deep_2022}
L.~Regenwetter, A.~H. Nobari, and F.~Ahmed, ``Deep {Generative} {Models} in
  {Engineering} {Design}: {A} {Review},'' {\em arXiv:2110.10863 [cs, stat]},
  Feb. 2022.
\newblock arXiv: 2110.10863.

\bibitem{moosavi_role_2020}
S.~M. Moosavi, K.~M. Jablonka, and B.~Smit, ``The {Role} of {Machine}
  {Learning} in the {Understanding} and {Design} of {Materials},'' {\em Journal
  of the American Chemical Society}, vol.~142, pp.~20273--20287, Dec. 2020.

\bibitem{faller_neural_1996}
W.~E. Faller and S.~J. Schreck, ``Neural networks: {Applications} and
  opportunities in aeronautics,'' {\em Progress in Aerospace Sciences},
  vol.~32, pp.~433--456, Oct. 1996.

\bibitem{thuerey_physics-based_2021}
N.~Thuerey, P.~Holl, M.~Mueller, P.~Schnell, F.~Trost, and K.~Um, {\em
  Physics-based {Deep} {Learning}}.
\newblock WWW, 2021.

\bibitem{kollmannsberger_deep_2021}
S.~Kollmannsberger, D.~D'Angella, M.~Jokeit, and L.~Herrmann, {\em Deep
  {Learning} in {Computational} {Mechanics}: {An} {Introductory} {Course}},
  vol.~977 of {\em Studies in {Computational} {Intelligence}}.
\newblock Cham: Springer International Publishing, 2021.

\bibitem{brunton_data-driven_2022}
S.~L. Brunton and J.~N. Kutz, {\em Data-driven science and engineering: machine
  learning, dynamical systems, and control}.
\newblock Cambridge, United Kingdom, New York, NY: Cambridge University Press,
  2022.

\bibitem{karpatne_knowledge_2022}
A.~Karpatne, R.~Kannan, and V.~Kumar, eds., {\em Knowledge {Guided} {Machine}
  {Learning}: {Accelerating} {Discovery} using {Scientific} {Knowledge} and
  {Data}}.
\newblock New York: Chapman and Hall/CRC, Aug. 2022.

\bibitem{goodfellow_deep_2016}
I.~Goodfellow, Y.~Bengio, and A.~Courville, {\em Deep {Learning}}.
\newblock MIT Press, 2016.

\bibitem{paszke_pytorch_2019}
A.~Paszke, S.~Gross, F.~Massa, A.~Lerer, J.~Bradbury, G.~Chanan, T.~Killeen,
  Z.~Lin, N.~Gimelshein, L.~Antiga, A.~Desmaison, A.~Köpf, E.~Yang, Z.~DeVito,
  M.~Raison, A.~Tejani, S.~Chilamkurthy, B.~Steiner, L.~Fang, J.~Bai, and
  S.~Chintala, ``{PyTorch}: {An} {Imperative} {Style}, {High}-{Performance}
  {Deep} {Learning} {Library},'' {\em arXiv:1912.01703 [cs, stat]}, Dec. 2019.
\newblock arXiv: 1912.01703.

\bibitem{martin_abadi_tensorflow_2015}
{Martín Abadi}, {Ashish Agarwal}, {Paul Barham}, {Eugene Brevdo}, {Zhifeng
  Chen}, {Craig Citro}, {Greg S. Corrado}, {Andy Davis}, {Jeffrey Dean},
  {Matthieu Devin}, {Sanjay Ghemawat}, {Ian Goodfellow}, {Andrew Harp},
  {Geoffrey Irving}, {Michael Isard}, Y.~Jia, {Rafal Jozefowicz}, {Lukasz
  Kaiser}, {Manjunath Kudlur}, {Josh Levenberg}, {Dandelion Mané}, {Rajat
  Monga}, {Sherry Moore}, {Derek Murray}, {Chris Olah}, {Mike Schuster},
  {Jonathon Shlens}, {Benoit Steiner}, {Ilya Sutskever}, {Kunal Talwar}, {Paul
  Tucker}, {Vincent Vanhoucke}, {Vijay Vasudevan}, {Fernanda Viégas}, {Oriol
  Vinyals}, {Pete Warden}, {Martin Wattenberg}, {Martin Wicke}, {Yuan Yu}, and
  {Xiaoqiang Zheng}, ``{TensorFlow}: {Large}-{Scale} {Machine} {Learning} on
  {Heterogeneous} {Systems},'' 2015.

\bibitem{hornik_multilayer_1989}
K.~Hornik, M.~Stinchcombe, and H.~White, ``Multilayer feedforward networks are
  universal approximators,'' {\em Neural Networks}, vol.~2, pp.~359--366, Jan.
  1989.

\bibitem{kingma_adam_2017}
D.~P. Kingma and J.~Ba, ``Adam: {A} {Method} for {Stochastic} {Optimization},''
  {\em arXiv:1412.6980 [cs]}, Jan. 2017.
\newblock arXiv: 1412.6980.

\bibitem{nocedal_numerical_2006}
J.~Nocedal and S.~J. Wright, {\em Numerical optimization}.
\newblock Springer series in operations research, New York: Springer, 2nd
  ed~ed., 2006.
\newblock OCLC: ocm68629100.

\bibitem{rosenblatt_perceptron_1958}
F.~Rosenblatt, ``The perceptron: a probabilistic model for information storage
  and organization in the brain,'' {\em Psychological Review}, vol.~65,
  pp.~386--408, Nov. 1958.

\bibitem{lecun_handwritten_1989}
Y.~LeCun, B.~Boser, J.~Denker, D.~Henderson, R.~Howard, W.~Hubbard, and
  L.~Jackel, ``Handwritten {Digit} {Recognition} with a {Back}-{Propagation}
  {Network},'' in {\em Advances in {Neural} {Information} {Processing}
  {Systems}}, vol.~2, Morgan-Kaufmann, 1989.

\bibitem{lecun_backpropagation_1989}
Y.~LeCun, B.~Boser, J.~S. Denker, D.~Henderson, R.~E. Howard, W.~Hubbard, and
  L.~D. Jackel, ``Backpropagation {Applied} to {Handwritten} {Zip} {Code}
  {Recognition},'' {\em Neural Computation}, vol.~1, pp.~541--551, Dec. 1989.

\bibitem{lecun_gradient-based_1998}
Y.~Lecun, L.~Bottou, Y.~Bengio, and P.~Haffner, ``Gradient-based learning
  applied to document recognition,'' {\em Proceedings of the IEEE}, vol.~86,
  pp.~2278--2324, Nov. 1998.
\newblock Conference Name: Proceedings of the IEEE.

\bibitem{rumelhart_learning_1986}
D.~E. Rumelhart, G.~E. Hinton, and R.~J. Williams, ``Learning representations
  by back-propagating errors,'' {\em Nature}, vol.~323, pp.~533--536, Oct.
  1986.

\bibitem{hochreiter_long_1997}
S.~Hochreiter and J.~Schmidhuber, ``Long {Short}-{Term} {Memory},'' {\em Neural
  Computation}, vol.~9, pp.~1735--1780, Nov. 1997.

\bibitem{cho_learning_2014}
K.~Cho, B.~van Merriënboer, C.~Gulcehre, D.~Bahdanau, F.~Bougares, H.~Schwenk,
  and Y.~Bengio, ``Learning {Phrase} {Representations} using {RNN}
  {Encoder}–{Decoder} for {Statistical} {Machine} {Translation},'' in {\em
  Proceedings of the 2014 {Conference} on {Empirical} {Methods} in {Natural}
  {Language} {Processing} ({EMNLP})}, (Doha, Qatar), pp.~1724--1734,
  Association for Computational Linguistics, Oct. 2014.

\bibitem{kipf_semi-supervised_2017}
T.~N. Kipf and M.~Welling, ``Semi-{Supervised} {Classification} with {Graph}
  {Convolutional} {Networks},'' Feb. 2017.
\newblock arXiv:1609.02907 [cs, stat].

\bibitem{monti_dual-primal_2018}
F.~Monti, O.~Shchur, A.~Bojchevski, O.~Litany, S.~Günnemann, and M.~M.
  Bronstein, ``Dual-{Primal} {Graph} {Convolutional} {Networks},'' June 2018.
\newblock arXiv:1806.00770 [cs, stat].

\bibitem{battaglia_relational_2018}
P.~W. Battaglia, J.~B. Hamrick, V.~Bapst, A.~Sanchez-Gonzalez, V.~Zambaldi,
  M.~Malinowski, A.~Tacchetti, D.~Raposo, A.~Santoro, R.~Faulkner, C.~Gulcehre,
  F.~Song, A.~Ballard, J.~Gilmer, G.~Dahl, A.~Vaswani, K.~Allen, C.~Nash,
  V.~Langston, C.~Dyer, N.~Heess, D.~Wierstra, P.~Kohli, M.~Botvinick,
  O.~Vinyals, Y.~Li, and R.~Pascanu, ``Relational inductive biases, deep
  learning, and graph networks,'' Oct. 2018.
\newblock arXiv:1806.01261 [cs, stat].

\bibitem{henkes_spiking_2022}
A.~Henkes, J.~K. Eshraghian, and H.~Wessels, ``Spiking neural networks for
  nonlinear regression,'' Oct. 2022.
\newblock arXiv:2210.03515 [cs].

\bibitem{tandale_spiking_2023}
S.~B. Tandale and M.~Stoffel, ``Spiking recurrent neural networks for
  neuromorphic computing in nonlinear structural mechanics,'' {\em Computer
  Methods in Applied Mechanics and Engineering}, vol.~412, p.~116095, July
  2023.

\bibitem{gerstner_spiking_2002}
W.~Gerstner and W.~M. Kistler, {\em Spiking {Neuron} {Models}: {Single}
  {Neurons}, {Populations}, {Plasticity}}.
\newblock Cambridge University Press, 1~ed., Aug. 2002.

\bibitem{hughes_space-time_1988}
T.~J.~R. Hughes and G.~M. Hulbert, ``Space-time finite element methods for
  elastodynamics: {Formulations} and error estimates,'' {\em Computer Methods
  in Applied Mechanics and Engineering}, vol.~66, pp.~339--363, Feb. 1988.

\bibitem{alsalman_training_2018}
M.~Alsalman, B.~Colvert, and E.~Kanso, ``Training bioinspired sensors to
  classify flows,'' {\em Bioinspiration \& Biomimetics}, vol.~14, p.~016009,
  Nov. 2018.

\bibitem{colvert_classifying_2018}
B.~Colvert, M.~Alsalman, and E.~Kanso, ``Classifying vortex wakes using neural
  networks,'' {\em Bioinspiration \& Biomimetics}, vol.~13, p.~025003, Feb.
  2018.

\bibitem{pierret_turbomachinery_1999}
S.~Pierret and R.~A. Van Den~Braembussche, ``Turbomachinery {Blade} {Design}
  {Using} a {Navier}–{Stokes} {Solver} and {Artificial} {Neural} {Network},''
  {\em Journal of Turbomachinery}, vol.~121, pp.~326--332, Apr. 1999.

\bibitem{vurtur_badarinath_machine_2021}
P.~Vurtur~Badarinath, M.~Chierichetti, and F.~Davoudi~Kakhki, ``A {Machine}
  {Learning} {Approach} as a {Surrogate} for a {Finite} {Element} {Analysis}:
  {Status} of {Research} and {Application} to {One} {Dimensional} {Systems},''
  {\em Sensors}, vol.~21, p.~1654, Jan. 2021.

\bibitem{lee_application_1997}
C.~Lee, J.~Kim, D.~Babcock, and R.~Goodman, ``Application of neural networks to
  turbulence control for drag reduction,'' {\em Physics of Fluids}, vol.~9,
  pp.~1740--1747, June 1997.

\bibitem{jambunathan_evaluating_1996}
K.~Jambunathan, S.~L. Hartle, S.~Ashforth-Frost, and V.~N. Fontama,
  ``Evaluating convective heat transfer coefficients using neural networks,''
  {\em International Journal of Heat and Mass Transfer}, vol.~39,
  pp.~2329--2332, July 1996.

\bibitem{tracey_machine_2015}
B.~D. Tracey, K.~Duraisamy, and J.~J. Alonso, ``A {Machine} {Learning}
  {Strategy} to {Assist} {Turbulence} {Model} {Development},'' in {\em 53rd
  {AIAA} {Aerospace} {Sciences} {Meeting}}, (Kissimmee, Florida), American
  Institute of Aeronautics and Astronautics, Jan. 2015.

\bibitem{ramuhalli_electromagnetic_2002}
P.~Ramuhalli, L.~Udpa, and S.~Udpa, ``Electromagnetic {NDE} signal inversion by
  function-approximation neural networks,'' {\em IEEE Transactions on
  Magnetics}, vol.~38, pp.~3633--3642, Nov. 2002.

\bibitem{araya-polo_deep-learning_2018}
M.~Araya-Polo, J.~Jennings, A.~Adler, and T.~Dahlke, ``Deep-learning
  tomography,'' {\em The Leading Edge}, vol.~37, pp.~58--66, Jan. 2018.

\bibitem{kim_geophysical_2018}
Y.~Kim and N.~Nakata, ``Geophysical inversion versus machine learning in
  inverse problems,'' {\em The Leading Edge}, vol.~37, pp.~894--901, Dec. 2018.

\bibitem{hoang_data-driven_2022}
V.-N. Hoang, N.-L. Nguyen, D.~Q. Tran, Q.-V. Vu, and H.~Nguyen-Xuan,
  ``Data-driven geometry-based topology optimization,'' {\em Structural and
  Multidisciplinary Optimization}, vol.~65, p.~69, Jan. 2022.

\bibitem{zhang_label-free_2023}
X.~Zhang and K.~Garikipati, ``Label-free learning of elliptic partial
  differential equation solvers with generalizability across boundary value
  problems,'' {\em Computer Methods in Applied Mechanics and Engineering},
  p.~116214, July 2023.

\bibitem{thuerey_deep_2020}
N.~Thuerey, K.~Weißenow, L.~Prantl, and X.~Hu, ``Deep {Learning} {Methods} for
  {Reynolds}-{Averaged} {Navier}–{Stokes} {Simulations} of {Airfoil}
  {Flows},'' {\em AIAA Journal}, vol.~58, pp.~25--36, Jan. 2020.

\bibitem{chen_numerical_2021}
L.-W. Chen, B.~A. Cakal, X.~Hu, and N.~Thuerey, ``Numerical investigation of
  minimum drag profiles in laminar flow using deep learning surrogates,'' {\em
  Journal of Fluid Mechanics}, vol.~919, p.~A34, July 2021.

\bibitem{chen_deep_2021}
X.~Chen, X.~Zhao, Z.~Gong, J.~Zhang, W.~Zhou, X.~Chen, and W.~Yao, ``A deep
  neural network surrogate modeling benchmark for temperature field prediction
  of heat source layout,'' {\em Science China Physics, Mechanics \& Astronomy},
  vol.~64, p.~1, Sept. 2021.

\bibitem{chen_towards_2023}
L.-W. Chen and N.~Thuerey, ``Towards high-accuracy deep learning inference of
  compressible flows over aerofoils,'' {\em Computers \& Fluids}, vol.~250,
  p.~105707, Jan. 2023.

\bibitem{khadilkar_deep_2019}
A.~Khadilkar, J.~Wang, and R.~Rai, ``Deep learning–based stress prediction
  for bottom-up {SLA} {3D} printing process,'' {\em The International Journal
  of Advanced Manufacturing Technology}, vol.~102, pp.~2555--2569, June 2019.

\bibitem{nie_stress_2020}
Z.~Nie, H.~Jiang, and L.~B. Kara, ``Stress {Field} {Prediction} in
  {Cantilevered} {Structures} {Using} {Convolutional} {Neural} {Networks},''
  {\em Journal of Computing and Information Science in Engineering}, vol.~20,
  p.~011002, Feb. 2020.

\bibitem{guo_convolutional_2016}
X.~Guo, W.~Li, and F.~Iorio, ``Convolutional {Neural} {Networks} for {Steady}
  {Flow} {Approximation},'' in {\em Proceedings of the 22nd {ACM} {SIGKDD}
  {International} {Conference} on {Knowledge} {Discovery} and {Data} {Mining}},
  (San Francisco California USA), pp.~481--490, ACM, Aug. 2016.

\bibitem{zhang_featurenet_2018}
Z.~Zhang, P.~Jaiswal, and R.~Rai, ``{FeatureNet}: {Machining} feature
  recognition based on {3D} {Convolution} {Neural} {Network},'' {\em
  Computer-Aided Design}, vol.~101, pp.~12--22, Aug. 2018.

\bibitem{williams_design_2019}
G.~Williams, N.~A. Meisel, T.~W. Simpson, and C.~McComb, ``Design {Repository}
  {Effectiveness} for {3D} {Convolutional} {Neural} {Networks}: {Application}
  to {Additive} {Manufacturing},'' {\em Journal of Mechanical Design},
  vol.~141, p.~111701, Nov. 2019.

\bibitem{wu_inversionet_2018}
Y.~Wu, Y.~Lin, and Z.~Zhou, ``Inversionet: {Accurate} and efficient
  seismic-waveform inversion with convolutional neural networks,'' in {\em
  {SEG} {Technical} {Program} {Expanded} {Abstracts} 2018}, (Anaheim,
  California), pp.~2096--2100, Society of Exploration Geophysicists, Aug. 2018.

\bibitem{wang_velocity_2018}
W.~Wang, F.~Yang, and J.~Ma, ``Velocity model building with a modified fully
  convolutional network,'' in {\em {SEG} {Technical} {Program} {Expanded}
  {Abstracts} 2018}, (Anaheim, California), pp.~2086--2090, Society of
  Exploration Geophysicists, Aug. 2018.

\bibitem{yang_deep-learning_2019}
F.~Yang and J.~Ma, ``Deep-learning inversion: {A} next-generation seismic
  velocity model building method,'' {\em GEOPHYSICS}, vol.~84, pp.~R583--R599,
  July 2019.

\bibitem{zheng_applications_2019}
Y.~Zheng, Q.~Zhang, A.~Yusifov, and Y.~Shi, ``Applications of supervised deep
  learning for seismic interpretation and inversion,'' {\em The Leading Edge},
  vol.~38, pp.~526--533, July 2019.

\bibitem{araya-polo_deep_2019}
M.~Araya-Polo, S.~Farris, and M.~Florez, ``Deep learning-driven velocity model
  building workflow,'' {\em The Leading Edge}, vol.~38, pp.~872a1--872a9, Nov.
  2019.

\bibitem{das_convolutional_2019}
V.~Das, A.~Pollack, U.~Wollner, and T.~Mukerji, ``Convolutional neural network
  for seismic impedance inversion,'' {\em GEOPHYSICS}, vol.~84, pp.~R869--R880,
  Nov. 2019.

\bibitem{wang_velocity_2020}
W.~Wang and J.~Ma, ``Velocity model building in a crosswell acquisition
  geometry with image-trained artificial neural networks,'' {\em GEOPHYSICS},
  vol.~85, pp.~U31--U46, Mar. 2020.

\bibitem{li_deep-learning_2020}
S.~Li, B.~Liu, Y.~Ren, Y.~Chen, S.~Yang, Y.~Wang, and P.~Jiang,
  ``Deep-{Learning} {Inversion} of {Seismic} {Data},'' {\em IEEE Transactions
  on Geoscience and Remote Sensing}, vol.~58, pp.~2135--2149, Mar. 2020.

\bibitem{wu_seismic_2020}
B.~Wu, D.~Meng, L.~Wang, N.~Liu, and Y.~Wang, ``Seismic {Impedance} {Inversion}
  {Using} {Fully} {Convolutional} {Residual} {Network} and {Transfer}
  {Learning},'' {\em IEEE Geoscience and Remote Sensing Letters}, vol.~17,
  pp.~2140--2144, Dec. 2020.

\bibitem{park_automatic_2020}
M.~J. Park and M.~D. Sacchi, ``Automatic velocity analysis using convolutional
  neural network and transfer learning,'' {\em GEOPHYSICS}, vol.~85,
  pp.~V33--V43, Jan. 2020.

\bibitem{ye_automatic_2022}
J.~Ye and N.~Toyama, ``Automatic defect detection for ultrasonic wave
  propagation imaging method using spatio-temporal convolution neural
  networks,'' {\em Structural Health Monitoring}, vol.~21, pp.~2750--2767, Nov.
  2022.

\bibitem{rao_quantitative_2023}
J.~Rao, F.~Yang, H.~Mo, S.~Kollmannsberger, and E.~Rank, ``Quantitative
  reconstruction of defects in multi-layered bonded composites using fully
  convolutional network-based ultrasonic inversion,'' {\em Journal of Sound and
  Vibration}, vol.~542, p.~117418, Jan. 2023.

\bibitem{lin_investigation_2018}
Q.~Lin, J.~Hong, Z.~Liu, B.~Li, and J.~Wang, ``Investigation into the topology
  optimization for conductive heat transfer based on deep learning approach,''
  {\em International Communications in Heat and Mass Transfer}, vol.~97,
  pp.~103--109, Oct. 2018.

\bibitem{yu_deep_2019}
Y.~Yu, T.~Hur, J.~Jung, and I.~G. Jang, ``Deep learning for determining a
  near-optimal topological design without any iteration,'' {\em Structural and
  Multidisciplinary Optimization}, vol.~59, pp.~787--799, Mar. 2019.

\bibitem{abueidda_topology_2020}
D.~W. Abueidda, S.~Koric, and N.~A. Sobh, ``Topology optimization of {2D}
  structures with nonlinearities using deep learning,'' {\em Computers \&
  Structures}, vol.~237, p.~106283, Sept. 2020.

\bibitem{nakamura_deep_2020}
K.~Nakamura and Y.~Suzuki, ``Deep learning-based topological optimization for
  representing a user-specified design area,'' Apr. 2020.

\bibitem{zhang_deep_2020}
Y.~Zhang, B.~Peng, X.~Zhou, C.~Xiang, and D.~Wang, ``A deep {Convolutional}
  {Neural} {Network} for topology optimization with strong generalization
  ability,'' Mar. 2020.
\newblock arXiv:1901.07761 [cs, stat].

\bibitem{zheng_generating_2021}
S.~Zheng, Z.~He, and H.~Liu, ``Generating three-dimensional structural
  topologies via a {U}-{Net} convolutional neural network,'' {\em Thin-Walled
  Structures}, vol.~159, p.~107263, Feb. 2021.

\bibitem{zheng_accurate_2021}
S.~Zheng, H.~Fan, Z.~Zhang, Z.~Tian, and K.~Jia, ``Accurate and real-time
  structural topology prediction driven by deep learning under moving morphable
  component-based framework,'' {\em Applied Mathematical Modelling}, vol.~97,
  pp.~522--535, Sept. 2021.

\bibitem{wang_deep_2022}
D.~Wang, C.~Xiang, Y.~Pan, A.~Chen, X.~Zhou, and Y.~Zhang, ``A deep
  convolutional neural network for topology optimization with perceptible
  generalization ability,'' {\em Engineering Optimization}, vol.~54,
  pp.~973--988, June 2022.

\bibitem{yan_deep_2022}
J.~Yan, Q.~Zhang, Q.~Xu, Z.~Fan, H.~Li, W.~Sun, and G.~Wang, ``Deep learning
  driven real time topology optimisation based on initial stress learning,''
  {\em Advanced Engineering Informatics}, vol.~51, p.~101472, Jan. 2022.

\bibitem{seo_topology_2023}
J.~Seo and R.~K. Kapania, ``Topology optimization with advanced {CNN} using
  mapped physics-based data,'' {\em Structural and Multidisciplinary
  Optimization}, vol.~66, p.~21, Jan. 2023.

\bibitem{sosnovik_neural_2019}
I.~Sosnovik and I.~Oseledets, ``Neural networks for topology optimization,''
  {\em Russian Journal of Numerical Analysis and Mathematical Modelling},
  vol.~34, pp.~215--223, Aug. 2019.

\bibitem{joo_unit_2021}
Y.~Joo, Y.~Yu, and I.~G. Jang, ``Unit {Module}-{Based} {Convergence}
  {Acceleration} for {Topology} {Optimization} {Using} the {Spatiotemporal}
  {Deep} {Neural} {Network},'' {\em IEEE Access}, vol.~9, pp.~149766--149779,
  2021.

\bibitem{kallioras_accelerated_2020}
N.~A. Kallioras, G.~Kazakis, and N.~D. Lagaros, ``Accelerated topology
  optimization by means of deep learning,'' {\em Structural and
  Multidisciplinary Optimization}, vol.~62, pp.~1185--1212, Sept. 2020.

\bibitem{sanchez-gonzalez_learning_2020}
A.~Sanchez-Gonzalez, J.~Godwin, T.~Pfaff, R.~Ying, J.~Leskovec, and P.~W.
  Battaglia, ``Learning to {Simulate} {Complex} {Physics} with {Graph}
  {Networks},'' {\em arXiv:2002.09405 [physics, stat]}, Sept. 2020.
\newblock arXiv: 2002.09405.

\bibitem{pfaff_learning_2021}
T.~Pfaff, M.~Fortunato, A.~Sanchez-Gonzalez, and P.~W. Battaglia, ``Learning
  {Mesh}-{Based} {Simulation} with {Graph} {Networks},'' {\em arXiv:2010.03409
  [cs]}, June 2021.
\newblock arXiv: 2010.03409.

\bibitem{perera_graph_2022}
R.~Perera, D.~Guzzetti, and V.~Agrawal, ``Graph neural networks for simulating
  crack coalescence and propagation in brittle materials,'' {\em Computer
  Methods in Applied Mechanics and Engineering}, vol.~395, p.~115021, May 2022.

\bibitem{qi_pointnet_2017}
C.~R. Qi, H.~Su, K.~Mo, and L.~J. Guibas, ``{PointNet}: {Deep} {Learning} on
  {Point} {Sets} for {3D} {Classification} and {Segmentation},'' Apr. 2017.

\bibitem{groueix_atlasnet_2018}
T.~Groueix, M.~Fisher, V.~G. Kim, B.~C. Russell, and M.~Aubry, ``{AtlasNet}:
  {A} {Papier}-{M}{\textbackslash}{\textasciicircum}ach{\textbackslash}'e
  {Approach} to {Learning} {3D} {Surface} {Generation},'' July 2018.
\newblock arXiv:1802.05384 [cs].

\bibitem{cunningham_investigation_2019}
J.~D. Cunningham, T.~W. Simpson, and C.~S. Tucker, ``An {Investigation} of
  {Surrogate} {Models} for {Efficient} {Performance}-{Based} {Decoding} of {3D}
  {Point} {Clouds},'' {\em Journal of Mechanical Design}, vol.~141, p.~121401,
  Dec. 2019.

\bibitem{jolliffe_principal_2002}
I.~T. Jolliffe, {\em Principal component analysis}.
\newblock Springer series in statistics, New York: Springer, 2nd ed~ed., 2002.

\bibitem{heimann_statistical_2009}
T.~Heimann and H.-P. Meinzer, ``Statistical shape models for {3D} medical image
  segmentation: {A} review,'' {\em Medical Image Analysis}, vol.~13,
  pp.~543--563, Aug. 2009.

\bibitem{bhattacharya_model_2021}
K.~Bhattacharya, B.~Hosseini, N.~B. Kovachki, and A.~M. Stuart, ``Model
  {Reduction} {And} {Neural} {Networks} {For} {Parametric} {PDEs},'' {\em The
  SMAI Journal of computational mathematics}, vol.~7, pp.~121--157, 2021.

\bibitem{berkooz_proper_1993}
G.~Berkooz, P.~Holmes, and J.~L. Lumley, ``The {Proper} {Orthogonal}
  {Decomposition} in the {Analysis} of {Turbulent} {Flows},'' {\em Annual
  Review of Fluid Mechanics}, vol.~25, pp.~539--575, Jan. 1993.

\bibitem{munoz_manifold_2023}
D.~Muñoz, O.~Allix, F.~Chinesta, J.~J. Ródenas, and E.~Nadal, ``Manifold
  learning for coherent design interpolation based on geometrical and
  topological descriptors,'' {\em Computer Methods in Applied Mechanics and
  Engineering}, vol.~405, p.~115859, Feb. 2023.

\bibitem{liang_deep_2018}
L.~Liang, M.~Liu, C.~Martin, and W.~Sun, ``A deep learning approach to estimate
  stress distribution: a fast and accurate surrogate of finite-element
  analysis,'' {\em Journal of The Royal Society Interface}, vol.~15,
  p.~20170844, Jan. 2018.

\bibitem{madani_bridging_2019}
A.~Madani, A.~Bakhaty, J.~Kim, Y.~Mubarak, and M.~R.~K. Mofrad, ``Bridging
  {Finite} {Element} and {Machine} {Learning} {Modeling}: {Stress} {Prediction}
  of {Arterial} {Walls} in {Atherosclerosis},'' {\em Journal of Biomechanical
  Engineering}, vol.~141, p.~084502, Aug. 2019.

\bibitem{muravleva_application_2018}
E.~Muravleva, I.~Oseledets, and D.~Koroteev, ``Application of machine learning
  to viscoplastic flow modeling,'' {\em Physics of Fluids}, vol.~30, p.~103102,
  Oct. 2018.

\bibitem{liang_machine_2018}
L.~Liang, M.~Liu, C.~Martin, and W.~Sun, ``A machine learning approach as a
  surrogate of finite element analysis-based inverse method to estimate the
  zero-pressure geometry of human thoracic aorta,'' {\em International Journal
  for Numerical Methods in Biomedical Engineering}, vol.~34, p.~e3103, Aug.
  2018.

\bibitem{derouiche_data-driven_2021}
K.~Derouiche, S.~Garois, V.~Champaney, M.~Daoud, K.~Traidi, and F.~Chinesta,
  ``Data-{Driven} {Modeling} for {Multiphysics} {Parametrized}
  {Problems}-{Application} to {Induction} {Hardening} {Process},'' {\em
  Metals}, vol.~11, p.~738, Apr. 2021.

\bibitem{hernandez_thermodynamics-informed_2023}
Q.~Hernández, A.~Badías, F.~Chinesta, and E.~Cueto, ``Thermodynamics-informed
  neural networks for physically realistic mixed reality,'' {\em Computer
  Methods in Applied Mechanics and Engineering}, vol.~407, p.~115912, Mar.
  2023.

\bibitem{hinton_reducing_2006}
G.~E. Hinton and R.~R. Salakhutdinov, ``Reducing the {Dimensionality} of {Data}
  with {Neural} {Networks},'' {\em Science}, vol.~313, pp.~504--507, July 2006.

\bibitem{milano_neural_2002}
M.~Milano and P.~Koumoutsakos, ``Neural {Network} {Modeling} for {Near} {Wall}
  {Turbulent} {Flow},'' {\em Journal of Computational Physics}, vol.~182,
  pp.~1--26, Oct. 2002.

\bibitem{nair_grids-net_2023}
S.~Nair, T.~F. Walsh, G.~Pickrell, and F.~Semperlotti, ``{GRIDS}-{Net}:
  {Inverse} shape design and identification of scatterers via geometric
  regularization and physics-embedded deep learning,'' {\em Computer Methods in
  Applied Mechanics and Engineering}, vol.~414, p.~116167, Sept. 2023.

\bibitem{fernandez-navamuel_supervised_2022}
A.~Fernandez-Navamuel, D.~Zamora-Sánchez, Ã.~J. Omella, D.~Pardo,
  D.~Garcia-Sanchez, and F.~Magalhães, ``Supervised {Deep} {Learning} with
  {Finite} {Element} simulations for damage identification in bridges,'' {\em
  Engineering Structures}, vol.~257, p.~114016, Apr. 2022.

\bibitem{ronneberger_u-net_2015}
O.~Ronneberger, P.~Fischer, and T.~Brox, ``U-{Net}: {Convolutional} {Networks}
  for {Biomedical} {Image} {Segmentation},'' in {\em Medical {Image}
  {Computing} and {Computer}-{Assisted} {Intervention} – {MICCAI} 2015}
  (N.~Navab, J.~Hornegger, W.~M. Wells, and A.~F. Frangi, eds.), Lecture
  {Notes} in {Computer} {Science}, (Cham), pp.~234--241, Springer International
  Publishing, 2015.

\bibitem{zhou_unet_2018}
Z.~Zhou, M.~M.~R. Siddiquee, N.~Tajbakhsh, and J.~Liang, ``{UNet}++: {A}
  {Nested} {U}-{Net} {Architecture} for {Medical} {Image} {Segmentation},''
  July 2018.
\newblock arXiv:1807.10165 [cs, eess, stat].

\bibitem{lu_comprehensive_2022}
L.~Lu, X.~Meng, S.~Cai, Z.~Mao, S.~Goswami, Z.~Zhang, and G.~E. Karniadakis,
  ``A comprehensive and fair comparison of two neural operators (with practical
  extensions) based on {FAIR} data,'' {\em Computer Methods in Applied
  Mechanics and Engineering}, vol.~393, p.~114778, Apr. 2022.

\bibitem{chen_universal_1995}
T.~Chen and H.~Chen, ``Universal approximation to nonlinear operators by neural
  networks with arbitrary activation functions and its application to dynamical
  systems,'' {\em IEEE Transactions on Neural Networks}, vol.~6, pp.~911--917,
  July 1995.

\bibitem{lu_learning_2021}
L.~Lu, P.~Jin, G.~Pang, Z.~Zhang, and G.~E. Karniadakis, ``Learning nonlinear
  operators via {DeepONet} based on the universal approximation theorem of
  operators,'' {\em Nature Machine Intelligence}, vol.~3, pp.~218--229, Mar.
  2021.

\bibitem{li_fourier_2021}
Z.-Y. Li, N.~B. Kovachki, K.~Azizzadenesheli, B.~Liu, K.~Bhattacharya,
  A.~Stuart, and A.~Anandkumar, ``Fourier neural operator for parametric
  partial differential equations,'' {\em ArXiv}, vol.~abs/2010.08895, 2021.

\bibitem{lin_seamless_2021}
C.~Lin, M.~Maxey, Z.~Li, and G.~E. Karniadakis, ``A seamless multiscale
  operator neural network for inferring bubble dynamics,'' {\em Journal of
  Fluid Mechanics}, vol.~929, p.~A18, Dec. 2021.

\bibitem{mao_deepmmnet_2021}
Z.~Mao, L.~Lu, O.~Marxen, T.~A. Zaki, and G.~E. Karniadakis, ``{DeepM}\&{Mnet}
  for hypersonics: {Predicting} the coupled flow and finite-rate chemistry
  behind a normal shock using neural-network approximation of operators,'' {\em
  Journal of Computational Physics}, vol.~447, p.~110698, Dec. 2021.

\bibitem{di_leoni_deeponet_2021}
P.~C. Di~Leoni, L.~Lu, C.~Meneveau, G.~Karniadakis, and T.~A. Zaki,
  ``{DeepONet} prediction of linear instability waves in high-speed boundary
  layers,'' May 2021.
\newblock arXiv:2105.08697 [physics].

\bibitem{cai_deepmmnet_2021}
S.~Cai, Z.~Wang, L.~Lu, T.~A. Zaki, and G.~E. Karniadakis, ``{DeepM}\&{Mnet}:
  {Inferring} the electroconvection multiphysics fields based on operator
  approximation by neural networks,'' {\em Journal of Computational Physics},
  vol.~436, p.~110296, July 2021.

\bibitem{lin_operator_2021}
C.~Lin, Z.~Li, L.~Lu, S.~Cai, M.~Maxey, and G.~E. Karniadakis, ``Operator
  learning for predicting multiscale bubble growth dynamics,'' {\em The Journal
  of Chemical Physics}, vol.~154, p.~104118, Mar. 2021.
\newblock arXiv:2012.12816 [physics].

\bibitem{yin_simulating_2022}
M.~Yin, E.~Ban, B.~V. Rego, E.~Zhang, C.~Cavinato, J.~D. Humphrey, and
  G.~Em~Karniadakis, ``Simulating progressive intramural damage leading to
  aortic dissection using {DeepONet}: an operator–regression neural
  network,'' {\em Journal of The Royal Society Interface}, vol.~19,
  p.~20210670, Feb. 2022.

\bibitem{osorio_forecasting_2022}
J.~D. Osorio, Z.~Wang, G.~Karniadakis, S.~Cai, C.~Chryssostomidis, M.~Panwar,
  and R.~Hovsapian, ``Forecasting solar-thermal systems performance under
  transient operation using a data-driven machine learning approach based on
  the deep operator network architecture,'' {\em Energy Conversion and
  Management}, vol.~252, p.~115063, Jan. 2022.

\bibitem{goswami_neural_2022}
S.~Goswami, D.~S. Li, B.~V. Rego, M.~Latorre, J.~D. Humphrey, and G.~E.
  Karniadakis, ``Neural operator learning of heterogeneous mechanobiological
  insults contributing to aortic aneurysms,'' {\em Journal of The Royal Society
  Interface}, vol.~19, p.~20220410, Aug. 2022.

\bibitem{koric_deep_2023}
S.~Koric, A.~Viswantah, D.~W. Abueidda, N.~A. Sobh, and K.~Khan, ``Deep
  learning operator network for plastic deformation with variable loads and
  material properties,'' {\em Engineering with Computers}, May 2023.

\bibitem{clark_di_leoni_neural_2023}
P.~Clark Di~Leoni, L.~Lu, C.~Meneveau, G.~E. Karniadakis, and T.~A. Zaki,
  ``Neural operator prediction of linear instability waves in high-speed
  boundary layers,'' {\em Journal of Computational Physics}, vol.~474,
  p.~111793, Feb. 2023.

\bibitem{koric_data-driven_2023}
S.~Koric and D.~W. Abueidda, ``Data-driven and physics-informed deep learning
  operators for solution of heat conduction equation with parametric heat
  source,'' {\em International Journal of Heat and Mass Transfer}, vol.~203,
  p.~123809, Apr. 2023.

\bibitem{liu_operator_2023}
C.~Liu, Q.~He, A.~Zhao, T.~Wu, Z.~Song, B.~Liu, and C.~Feng, ``Operator
  {Learning} for {Predicting} {Mechanical} {Response} of {Hierarchical}
  {Composites} with {Applications} of {Inverse} {Design},'' {\em International
  Journal of Applied Mechanics}, vol.~15, p.~2350028, May 2023.

\bibitem{ahmed_multifidelity_2023}
S.~E. Ahmed and P.~Stinis, ``A multifidelity deep operator network approach to
  closure for multiscale systems,'' {\em Computer Methods in Applied Mechanics
  and Engineering}, vol.~414, p.~116161, Sept. 2023.

\bibitem{wang_learning_2021}
S.~Wang, H.~Wang, and P.~Perdikaris, ``Learning the solution operator of
  parametric partial differential equations with physics-informed
  {DeepONets},'' {\em Science Advances}, vol.~7, p.~eabi8605, Oct. 2021.

\bibitem{goswami_physics-informed_2022}
S.~Goswami, M.~Yin, Y.~Yu, and G.~E. Karniadakis, ``A physics-informed
  variational {DeepONet} for predicting crack path in quasi-brittle
  materials,'' {\em Computer Methods in Applied Mechanics and Engineering},
  vol.~391, p.~114587, Mar. 2022.

\bibitem{goswami_physics-informed_2022-1}
S.~Goswami, A.~Bora, Y.~Yu, and G.~E. Karniadakis, ``Physics-{Informed} {Deep}
  {Neural} {Operator} {Networks},'' July 2022.
\newblock arXiv:2207.05748 [cs, math].

\bibitem{kovachki_universal_2021}
N.~Kovachki, S.~Lanthaler, and S.~Mishra, ``On universal approximation and
  error bounds for {Fourier} neural operators,'' {\em The Journal of Machine
  Learning Research}, vol.~22, pp.~290:13237--290:13312, Jan. 2021.

\bibitem{li_neural_2020}
Z.~Li, N.~Kovachki, K.~Azizzadenesheli, B.~Liu, K.~Bhattacharya, A.~Stuart, and
  A.~Anandkumar, ``Neural {Operator}: {Graph} {Kernel} {Network} for {Partial}
  {Differential} {Equations},'' Mar. 2020.
\newblock arXiv:2003.03485 [cs, math, stat].

\bibitem{li_multipole_2020}
Z.~Li, N.~Kovachki, K.~Azizzadenesheli, B.~Liu, K.~Bhattacharya, A.~Stuart, and
  A.~Anandkumar, ``Multipole graph neural operator for parametric partial
  differential equations,'' in {\em Proceedings of the 34th {International}
  {Conference} on {Neural} {Information} {Processing} {Systems}}, {NIPS}'20,
  (Red Hook, NY, USA), pp.~6755--6766, Curran Associates Inc., Dec. 2020.

\bibitem{cao_lno_2023}
Q.~Cao, S.~Goswami, and G.~E. Karniadakis, ``{LNO}: {Laplace} {Neural}
  {Operator} for {Solving} {Differential} {Equations},'' May 2023.
\newblock arXiv:2303.10528 [cs].

\bibitem{zhu_fast_2021}
C.~Zhu, H.~Ye, and B.~Zhan, ``Fast {Solver} of {2D} {Maxwell}'s {Equations}
  {Based} on {Fourier} {Neural} {Operator},'' in {\em 2021 {Photonics} \&
  {Electromagnetics} {Research} {Symposium} ({PIERS})}, (Hangzhou, China),
  pp.~1635--1643, IEEE, Nov. 2021.

\bibitem{song_high-frequency_2022}
C.~Song and Y.~Wang, ``High-frequency wavefield extrapolation using the
  {Fourier} neural operator,'' {\em Journal of Geophysics and Engineering},
  vol.~19, pp.~269--282, Apr. 2022.

\bibitem{wei_small-data-driven_2022}
W.~Wei and L.-Y. Fu, ``Small-data-driven fast seismic simulations for complex
  media using physics-informed {Fourier} neural operators,'' {\em GEOPHYSICS},
  vol.~87, pp.~T435--T446, Nov. 2022.

\bibitem{rashid_learning_2022}
M.~M. Rashid, T.~Pittie, S.~Chakraborty, and N.~A. Krishnan, ``Learning the
  stress-strain fields in digital composites using {Fourier} neural operator,''
  {\em iScience}, vol.~25, p.~105452, Nov. 2022.

\bibitem{zhang_fourier_2022}
K.~Zhang, Y.~Zuo, H.~Zhao, X.~Ma, J.~Gu, J.~Wang, Y.~Yang, C.~Yao, and J.~Yao,
  ``Fourier {Neural} {Operator} for {Solving} {Subsurface} {Oil}/{Water}
  {Two}-{Phase} {Flow} {Partial} {Differential} {Equation},'' {\em SPE
  Journal}, vol.~27, pp.~1815--1830, June 2022.

\bibitem{yan_robust_2022}
B.~Yan, B.~Chen, D.~Robert~Harp, W.~Jia, and R.~J. Pawar, ``A robust deep
  learning workflow to predict multiphase flow behavior during geological {C}
  {O} 2 sequestration injection and {Post}-{Injection} periods,'' {\em Journal
  of Hydrology}, vol.~607, p.~127542, Apr. 2022.

\bibitem{wen_u-fnoenhanced_2022}
G.~Wen, Z.~Li, K.~Azizzadenesheli, A.~Anandkumar, and S.~M. Benson,
  ``U-{FNO}—{An} enhanced {Fourier} neural operator-based deep-learning model
  for multiphase flow,'' {\em Advances in Water Resources}, vol.~163,
  p.~104180, May 2022.

\bibitem{peng_attention-enhanced_2022}
W.~Peng, Z.~Yuan, and J.~Wang, ``Attention-enhanced neural network models for
  turbulence simulation,'' {\em Physics of Fluids}, vol.~34, p.~025111, Feb.
  2022.

\bibitem{you_learning_2022}
H.~You, Q.~Zhang, C.~J. Ross, C.-H. Lee, and Y.~Yu, ``Learning deep {Implicit}
  {Fourier} {Neural} {Operators} ({IFNOs}) with applications to heterogeneous
  material modeling,'' {\em Computer Methods in Applied Mechanics and
  Engineering}, vol.~398, p.~115296, Aug. 2022.

\bibitem{kuang_fast_2023}
T.~Kuang, J.~Liu, Z.~Yin, H.~Jing, Y.~Lan, Z.~Lan, and H.~Pan, ``Fast and
  {Robust} {Prediction} of {Multiphase} {Flow} in {Complex} {Fractured}
  {Reservoir} {Using} a {Fourier} {Neural} {Operator},'' {\em Energies},
  vol.~16, p.~3765, Apr. 2023.

\bibitem{costa_rocha_deep_2023}
P.~A. Costa~Rocha, S.~J. Johnston, V.~Oliveira~Santos, A.~A. Aliabadi, J.~V.~G.
  Thé, and B.~Gharabaghi, ``Deep {Neural} {Network} {Modeling} for {CFD}
  {Simulations}: {Benchmarking} the {Fourier} {Neural} {Operator} on the
  {Lid}-{Driven} {Cavity} {Case},'' {\em Applied Sciences}, vol.~13, p.~3165,
  Mar. 2023.

\bibitem{vaswani_attention_2017}
A.~Vaswani, N.~Shazeer, N.~Parmar, J.~Uszkoreit, L.~Jones, A.~N. Gomez,
  L.~Kaiser, and I.~Polosukhin, ``Attention is {All} you {Need},'' in {\em
  Advances in {Neural} {Information} {Processing} {Systems}}, vol.~30, Curran
  Associates, Inc., 2017.

\bibitem{cao_choose_2021}
S.~Cao, ``Choose a {Transformer}: {Fourier} or {Galerkin},'' Nov. 2021.
\newblock arXiv:2105.14995 [cs, math].

\bibitem{li_physics-informed_2023}
Z.~Li, H.~Zheng, N.~Kovachki, D.~Jin, H.~Chen, B.~Liu, K.~Azizzadenesheli, and
  A.~Anandkumar, ``Physics-{Informed} {Neural} {Operator} for {Learning}
  {Partial} {Differential} {Equations},'' Apr. 2023.
\newblock arXiv:2111.03794 [cs, math].

\bibitem{marcati_exponential_2023-1}
C.~Marcati, J.~A.~A. Opschoor, P.~C. Petersen, and C.~Schwab, ``Exponential
  {ReLU} {Neural} {Network} {Approximation} {Rates} for {Point} and {Edge}
  {Singularities},'' {\em Foundations of Computational Mathematics}, vol.~23,
  pp.~1043--1127, June 2023.

\bibitem{gonon_deep_2023}
L.~Gonon and C.~Schwab, ``Deep {ReLU} neural networks overcome the curse of
  dimensionality for partial integrodifferential equations,'' {\em Analysis and
  Applications}, vol.~21, pp.~1--47, Jan. 2023.

\bibitem{marcati_exponential_2023}
C.~Marcati and C.~Schwab, ``Exponential {Convergence} of {Deep} {Operator}
  {Networks} for {Elliptic} {Partial} {Differential} {Equations},'' {\em SIAM
  Journal on Numerical Analysis}, vol.~61, pp.~1513--1545, June 2023.

\bibitem{alvarez-aramberri_generation_2023}
J.~Ã.-A. Álvarez Aramberri, V.~D. Darrigrand, F.~C. Caro, and D.~P. Pardo,
  ``Generation of {Massive} {Databases} for {Deep} {Learning} {Inversion}: {A}
  {Goal}-{Oriented} hp-{Adaptive} {Strategy},'' {\em XI International
  Conference on Adaptive Modeling and Simulation (ADMOS 2023)},
  vol.~Applications of Goal-Oriented Error Estimation and Adaptivity, May 2023.

\bibitem{bolager_sampling_2023}
E.~L. Bolager, I.~Burak, C.~Datar, Q.~Sun, and F.~Dietrich, ``Sampling weights
  of deep neural networks,'' June 2023.
\newblock arXiv:2306.16830 [cs, math].

\bibitem{cohn_active_1994}
D.~Cohn, Z.~Ghahramani, and M.~Jordan, ``Active {Learning} with {Statistical}
  {Models},'' in {\em Advances in {Neural} {Information} {Processing}
  {Systems}}, vol.~7, MIT Press, 1994.

\bibitem{liu_knowledge_2021}
X.~Liu, C.~E. Athanasiou, N.~P. Padture, B.~W. Sheldon, and H.~Gao, ``Knowledge
  extraction and transfer in data-driven fracture mechanics,'' {\em Proceedings
  of the National Academy of Sciences}, vol.~118, p.~e2104765118, June 2021.

\bibitem{haasdonk_new_2023}
B.~Haasdonk, H.~Kleikamp, M.~Ohlberger, F.~Schindler, and T.~Wenzel, ``A {New}
  {Certified} {Hierarchical} and {Adaptive} {RB}-{ML}-{ROM} {Surrogate} {Model}
  for {Parametrized} {PDEs},'' {\em SIAM Journal on Scientific Computing},
  vol.~45, pp.~A1039--A1065, June 2023.

\bibitem{kalina_fetextrmann_2023}
K.~A. Kalina, L.~Linden, J.~Brummund, and M.~Kästner,
  ``{FE}\$\$\{\}{\textasciicircum}{\textbackslash}textrm\{{ANN}\}\$\$: an
  efficient data-driven multiscale approach based on physics-constrained neural
  networks and automated data mining,'' {\em Computational Mechanics}, vol.~71,
  pp.~827--851, May 2023.

\bibitem{pan_survey_2010}
S.~J. Pan and Q.~Yang, ``A {Survey} on {Transfer} {Learning},'' {\em IEEE
  Transactions on Knowledge and Data Engineering}, vol.~22, pp.~1345--1359,
  Oct. 2010.

\bibitem{yosinski_how_2014}
J.~Yosinski, J.~Clune, Y.~Bengio, and H.~Lipson, ``How transferable are
  features in deep neural networks?,'' in {\em Proceedings of the 27th
  {International} {Conference} on {Neural} {Information} {Processing} {Systems}
  - {Volume} 2}, {NIPS}'14, (Cambridge, MA, USA), pp.~3320--3328, MIT Press,
  Dec. 2014.

\bibitem{kollmannsberger_transfer_2023}
S.~Kollmannsberger, D.~Singh, and L.~Herrmann, ``Transfer {Learning} {Enhanced}
  {Full} {Waveform} {Inversion},'' Feb. 2023.
\newblock arXiv:2302.11259 [physics].

\bibitem{ballakur_empirical_2020}
A.~A. Ballakur and A.~Arya, ``Empirical {Evaluation} of {Gated} {Recurrent}
  {Neural} {Network} {Architectures} in {Aviation} {Delay} {Prediction},'' in
  {\em 2020 5th {International} {Conference} on {Computing}, {Communication}
  and {Security} ({ICCCS})}, pp.~1--7, Oct. 2020.

\bibitem{geneva_modeling_2020}
N.~Geneva and N.~Zabaras, ``Modeling the {Dynamics} of {PDE} {Systems} with
  {Physics}-{Constrained} {Deep} {Auto}-{Regressive} {Networks},'' {\em Journal
  of Computational Physics}, vol.~403, p.~109056, Feb. 2020.
\newblock arXiv: 1906.05747.

\bibitem{chang_compositional_2017}
M.~B. Chang, T.~Ullman, A.~Torralba, and J.~B. Tenenbaum, ``A {Compositional}
  {Object}-{Based} {Approach} to {Learning} {Physical} {Dynamics},'' Mar. 2017.

\bibitem{mrowca_flexible_2018}
D.~Mrowca, C.~Zhuang, E.~Wang, N.~Haber, L.~Fei-Fei, J.~B. Tenenbaum, and
  D.~L.~K. Yamins, ``Flexible {Neural} {Representation} for {Physics}
  {Prediction},'' Oct. 2018.

\bibitem{sanchez-gonzalez_graph_2018}
A.~Sanchez-Gonzalez, N.~Heess, J.~T. Springenberg, J.~Merel, M.~Riedmiller,
  R.~Hadsell, and P.~Battaglia, ``Graph networks as learnable physics engines
  for inference and control,'' June 2018.

\bibitem{li_propagation_2019}
Y.~Li, J.~Wu, J.-Y. Zhu, J.~B. Tenenbaum, A.~Torralba, and R.~Tedrake,
  ``Propagation {Networks} for {Model}-{Based} {Control} {Under} {Partial}
  {Observation},'' Apr. 2019.

\bibitem{lino_simulating_2021}
M.~Lino, C.~Cantwell, A.~A. Bharath, and S.~Fotiadis, ``Simulating {Continuum}
  {Mechanics} with {Multi}-{Scale} {Graph} {Neural} {Networks},'' June 2021.

\bibitem{alfarraj_petrophysical-property_2018}
M.~Alfarraj and G.~AlRegib, ``Petrophysical-property estimation from seismic
  data using recurrent neural networks,'' in {\em {SEG} {Technical} {Program}
  {Expanded} {Abstracts} 2018}, (Anaheim, California), pp.~2141--2146, Society
  of Exploration Geophysicists, Aug. 2018.

\bibitem{adler_deep_2019}
A.~Adler, M.~Araya-Polo, and T.~Poggio, ``Deep {Recurrent} {Architectures} for
  {Seismic} {Tomography},'' in {\em 81st {EAGE} {Conference} and {Exhibition}
  2019}, pp.~1--5, 2019.

\bibitem{fabien-ouellet_seismic_2020}
G.~Fabien-Ouellet and R.~Sarkar, ``Seismic velocity estimation: {A} deep
  recurrent neural-network approach,'' {\em GEOPHYSICS}, vol.~85, pp.~U21--U29,
  Jan. 2020.

\bibitem{vlachas_data-driven_2018}
P.~R. Vlachas, W.~Byeon, Z.~Y. Wan, T.~P. Sapsis, and P.~Koumoutsakos,
  ``Data-driven forecasting of high-dimensional chaotic systems with long
  short-term memory networks,'' {\em Proceedings of the Royal Society A:
  Mathematical, Physical and Engineering Sciences}, vol.~474, p.~20170844, May
  2018.

\bibitem{hou_machine_2019}
W.~Hou, D.~Darakananda, and J.~Eldredge, ``Machine {Learning} {Based}
  {Detection} of {Flow} {Disturbances} {Using} {Surface} {Pressure}
  {Measurements},'' in {\em {AIAA} {Scitech} 2019 {Forum}}, (San Diego,
  California), American Institute of Aeronautics and Astronautics, Jan. 2019.

\bibitem{heindel_virtual_2021}
L.~Heindel, P.~Hantschke, and M.~Kästner, ``A {Virtual} {Sensing} approach for
  approximating nonlinear dynamical systems using {LSTM} networks,'' {\em
  PAMM}, vol.~21, p.~e202100119, Dec. 2021.

\bibitem{heindel_data-driven_2022}
L.~Heindel, P.~Hantschke, and M.~Kästner, ``A data-driven approach for
  approximating non-linear dynamic systems using {LSTM} networks,'' {\em
  Procedia Structural Integrity}, vol.~38, pp.~159--167, Jan. 2022.

\bibitem{freitag_recurrent_2018}
S.~Freitag, B.~T. Cao, J.~Ninić, and G.~Meschke, ``Recurrent neural networks
  and proper orthogonal decomposition with interval data for real-time
  predictions of mechanised tunnelling processes,'' {\em Computers \&
  Structures}, vol.~207, pp.~258--273, Sept. 2018.

\bibitem{cao_artificial_2020}
B.~T. Cao, M.~Obel, S.~Freitag, P.~Mark, and G.~Meschke, ``Artificial neural
  network surrogate modelling for real-time predictions and control of building
  damage during mechanised tunnelling,'' {\em Advances in Engineering
  Software}, vol.~149, p.~102869, Nov. 2020.

\bibitem{cao_real-time_2022}
B.~T. Cao, M.~Obel, S.~Freitag, L.~Heußner, G.~Meschke, and P.~Mark,
  ``Real-{Time} {Risk} {Assessment} of {Tunneling}-{Induced} {Building}
  {Damage} {Considering} {Polymorphic} {Uncertainty},'' {\em ASCE-ASME Journal
  of Risk and Uncertainty in Engineering Systems, Part A: Civil Engineering},
  vol.~8, p.~04021069, Mar. 2022.

\bibitem{gruber_comparison_2022}
A.~Gruber, M.~Gunzburger, L.~Ju, and Z.~Wang, ``A comparison of neural network
  architectures for data-driven reduced-order modeling,'' {\em Computer Methods
  in Applied Mechanics and Engineering}, vol.~393, p.~114764, Apr. 2022.

\bibitem{gonzalez_deep_2018}
F.~J. Gonzalez and M.~Balajewicz, ``Deep convolutional recurrent autoencoders
  for learning low-dimensional feature dynamics of fluid systems,'' Aug. 2018.
\newblock arXiv:1808.01346 [physics].

\bibitem{holden_subspace_2019}
D.~Holden, B.~C. Duong, S.~Datta, and D.~Nowrouzezahrai, ``Subspace neural
  physics: fast data-driven interactive simulation,'' in {\em Proceedings of
  the 18th annual {ACM} {SIGGRAPH}/{Eurographics} {Symposium} on {Computer}
  {Animation}}, {SCA} '19, (New York, NY, USA), pp.~1--12, Association for
  Computing Machinery, July 2019.

\bibitem{fresca_deep_2020}
S.~Fresca, A.~Manzoni, L.~Dedè, and A.~Quarteroni, ``Deep learning-based
  reduced order models in cardiac electrophysiology,'' {\em PLOS ONE}, vol.~15,
  p.~e0239416, Oct. 2020.

\bibitem{fresca_comprehensive_2021}
S.~Fresca, L.~Dede', and A.~Manzoni, ``A {Comprehensive} {Deep}
  {Learning}-{Based} {Approach} to {Reduced} {Order} {Modeling} of {Nonlinear}
  {Time}-{Dependent} {Parametrized} {PDEs},'' {\em Journal of Scientific
  Computing}, vol.~87, p.~61, Apr. 2021.

\bibitem{fresca_pod-dl-rom_2022}
S.~Fresca and A.~Manzoni, ``{POD}-{DL}-{ROM}: {Enhancing} deep learning-based
  reduced order models for nonlinear parametrized {PDEs} by proper orthogonal
  decomposition,'' {\em Computer Methods in Applied Mechanics and Engineering},
  vol.~388, p.~114181, Jan. 2022.

\bibitem{ren_phycrnet_2022}
P.~Ren, C.~Rao, Y.~Liu, J.-X. Wang, and H.~Sun, ``{PhyCRNet}:
  {Physics}-informed convolutional-recurrent network for solving spatiotemporal
  {PDEs},'' {\em Computer Methods in Applied Mechanics and Engineering},
  vol.~389, p.~114399, Feb. 2022.

\bibitem{hu_accelerating_2022}
C.~Hu, S.~Martin, and R.~Dingreville, ``Accelerating phase-field predictions
  via recurrent neural networks learning the microstructure evolution in latent
  space,'' {\em Computer Methods in Applied Mechanics and Engineering},
  vol.~397, p.~115128, July 2022.

\bibitem{lee_model_2020}
K.~Lee and K.~T. Carlberg, ``Model reduction of dynamical systems on nonlinear
  manifolds using deep convolutional autoencoders,'' {\em Journal of
  Computational Physics}, vol.~404, p.~108973, Mar. 2020.

\bibitem{shen_high-order_2021}
S.~Shen, Y.~Yin, T.~Shao, H.~Wang, C.~Jiang, L.~Lan, and K.~Zhou, ``High-order
  {Differentiable} {Autoencoder} for {Nonlinear} {Model} {Reduction},'' Feb.
  2021.
\newblock arXiv:2102.11026 [cs].

\bibitem{schmid_dynamic_2010}
P.~J. Schmid, ``Dynamic mode decomposition of numerical and experimental
  data,'' {\em Journal of Fluid Mechanics}, vol.~656, pp.~5--28, Aug. 2010.

\bibitem{tu_dynamic_2013}
J.~H. Tu, C.~W. Rowley, D.~M. Luchtenburg, S.~L. Brunton, and J.~N. Kutz, ``On
  {Dynamic} {Mode} {Decomposition}: {Theory} and {Applications},'' Nov. 2013.
\newblock arXiv:1312.0041 [physics].

\bibitem{brunton_data_2017}
S.~L. Brunton and J.~N. Kutz, ``Data {Driven} {Science} \& {Engineering},''
  p.~572, 2017.

\bibitem{koopman_hamiltonian_1931}
B.~O. Koopman, ``Hamiltonian {Systems} and {Transformation} in {Hilbert}
  {Space},'' {\em Proceedings of the National Academy of Sciences}, vol.~17,
  pp.~315--318, May 1931.

\bibitem{williams_datadriven_2015}
M.~O. Williams, I.~G. Kevrekidis, and C.~W. Rowley, ``A {Data}–{Driven}
  {Approximation} of the {Koopman} {Operator}: {Extending} {Dynamic} {Mode}
  {Decomposition},'' {\em Journal of Nonlinear Science}, vol.~25,
  pp.~1307--1346, Dec. 2015.

\bibitem{li_extended_2017}
Q.~Li, F.~Dietrich, E.~M. Bollt, and I.~G. Kevrekidis, ``Extended dynamic mode
  decomposition with dictionary learning: {A} data-driven adaptive spectral
  decomposition of the {Koopman} operator,'' {\em Chaos: An Interdisciplinary
  Journal of Nonlinear Science}, vol.~27, p.~103111, Oct. 2017.

\bibitem{yeung_learning_2019}
E.~Yeung, S.~Kundu, and N.~Hodas, ``Learning {Deep} {Neural} {Network}
  {Representations} for {Koopman} {Operators} of {Nonlinear} {Dynamical}
  {Systems},'' in {\em 2019 {American} {Control} {Conference} ({ACC})},
  pp.~4832--4839, July 2019.
\newblock ISSN: 2378-5861.

\bibitem{takeishi_learning_2017}
N.~Takeishi, Y.~Kawahara, and T.~Yairi, ``Learning {Koopman} invariant
  subspaces for dynamic mode decomposition,'' in {\em Proceedings of the 31st
  {International} {Conference} on {Neural} {Information} {Processing}
  {Systems}}, {NIPS}'17, (Red Hook, NY, USA), pp.~1130--1140, Curran Associates
  Inc., Dec. 2017.

\bibitem{morton_deep_2018}
J.~Morton, F.~D. Witherden, A.~Jameson, and M.~J. Kochenderfer, ``Deep
  dynamical modeling and control of unsteady fluid flows,'' in {\em Proceedings
  of the 32nd {International} {Conference} on {Neural} {Information}
  {Processing} {Systems}}, {NIPS}'18, (Red Hook, NY, USA), pp.~9278--9288,
  Curran Associates Inc., Dec. 2018.

\bibitem{lusch_deep_2018}
B.~Lusch, J.~N. Kutz, and S.~L. Brunton, ``Deep learning for universal linear
  embeddings of nonlinear dynamics,'' {\em Nature Communications}, vol.~9,
  p.~4950, Nov. 2018.

\bibitem{otto_linearly_2019}
S.~E. Otto and C.~W. Rowley, ``Linearly {Recurrent} {Autoencoder} {Networks}
  for {Learning} {Dynamics},'' {\em SIAM Journal on Applied Dynamical Systems},
  vol.~18, pp.~558--593, Jan. 2019.

\bibitem{psichogios_hybrid_1992}
D.~C. Psichogios and L.~H. Ungar, ``A hybrid neural network-first principles
  approach to process modeling,'' {\em AIChE Journal}, vol.~38, pp.~1499--1511,
  Oct. 1992.

\bibitem{dissanayake_neural-network-based_1994}
M.~W. M.~G. Dissanayake and N.~Phan-Thien, ``Neural-network-based
  approximations for solving partial differential equations,'' {\em
  Communications in Numerical Methods in Engineering}, vol.~10, pp.~195--201,
  Mar. 1994.

\bibitem{lagaris_artificial_1998}
I.~Lagaris, A.~Likas, and D.~Fotiadis, ``Artificial neural networks for solving
  ordinary and partial differential equations,'' {\em IEEE Transactions on
  Neural Networks}, vol.~9, pp.~987--1000, Sept. 1998.

\bibitem{raissi_deep_2018}
M.~Raissi, ``Deep {Hidden} {Physics} {Models}: {Deep} {Learning} of {Nonlinear}
  {Partial} {Differential} {Equations},'' 2018.

\bibitem{karniadakis_physics-informed_2021}
G.~E. Karniadakis, I.~G. Kevrekidis, L.~Lu, P.~Perdikaris, S.~Wang, and
  L.~Yang, ``Physics-informed machine learning,'' {\em Nature Reviews Physics},
  vol.~3, pp.~422--440, June 2021.

\bibitem{cuomo_scientific_2022}
S.~Cuomo, V.~S. di~Cola, F.~Giampaolo, G.~Rozza, M.~Raissi, and F.~Piccialli,
  ``Scientific {Machine} {Learning} through {Physics}-{Informed} {Neural}
  {Networks}: {Where} we are and {What}'s next,'' {\em arXiv:2201.05624
  [physics]}, Jan. 2022.
\newblock arXiv: 2201.05624.

\bibitem{hao_physics-informed_2022}
Z.~Hao, S.~Liu, Y.~Zhang, C.~Ying, Y.~Feng, H.~Su, and J.~Zhu,
  ``Physics-{Informed} {Machine} {Learning}: {A} {Survey} on {Problems},
  {Methods} and {Applications},'' Nov. 2022.

\bibitem{haghighat_sciann_2021}
E.~Haghighat and R.~Juanes, ``{SciANN}: {A} {Keras}/{TensorFlow} wrapper for
  scientific computations and physics-informed deep learning using artificial
  neural networks,'' {\em Computer Methods in Applied Mechanics and
  Engineering}, vol.~373, p.~113552, Jan. 2021.

\bibitem{hennigh_nvidia_2021}
O.~Hennigh, S.~Narasimhan, M.~A. Nabian, A.~Subramaniam, K.~Tangsali, Z.~Fang,
  M.~Rietmann, W.~Byeon, and S.~Choudhry, ``{NVIDIA} {SimNet}™: {An}
  {AI}-{Accelerated} {Multi}-{Physics} {Simulation} {Framework},'' in {\em
  Computational {Science} – {ICCS} 2021} (M.~Paszynski, D.~Kranzlmüller,
  V.~V. Krzhizhanovskaya, J.~J. Dongarra, and P.~M. Sloot, eds.), Lecture
  {Notes} in {Computer} {Science}, (Cham), pp.~447--461, Springer International
  Publishing, 2021.

\bibitem{lu_deepxde_2021}
L.~Lu, X.~Meng, Z.~Mao, and G.~E. Karniadakis, ``{DeepXDE}: {A} {Deep}
  {Learning} {Library} for {Solving} {Differential} {Equations},'' {\em SIAM
  Review}, vol.~63, pp.~208--228, Jan. 2021.

\bibitem{baydin_automatic_2018}
A.~G. Baydin, B.~A. Pearlmutter, A.~A. Radul, and J.~M. Siskind, ``Automatic
  {Differentiation} in {Machine} {Learning}: a {Survey},'' {\em Journal of
  Machine Learning Research}, p.~43, 2018.

\bibitem{cai_deep_2020}
Z.~Cai, J.~Chen, M.~Liu, and X.~Liu, ``Deep least-squares methods: {An}
  unsupervised learning-based numerical method for solving elliptic {PDEs},''
  {\em Journal of Computational Physics}, vol.~420, p.~109707, Nov. 2020.

\bibitem{sirignano_dgm_2018}
J.~Sirignano and K.~Spiliopoulos, ``{DGM}: {A} deep learning algorithm for
  solving partial differential equations,'' {\em Journal of Computational
  Physics}, vol.~375, pp.~1339--1364, Dec. 2018.
\newblock arXiv: 1708.07469.

\bibitem{kharazmi_variational_2019}
E.~Kharazmi, Z.~Zhang, and G.~E. Karniadakis, ``Variational
  {Physics}-{Informed} {Neural} {Networks} {For} {Solving} {Partial}
  {Differential} {Equations},'' {\em arXiv:1912.00873 [physics, stat]}, Nov.
  2019.
\newblock arXiv: 1912.00873.

\bibitem{kharazmi_hp-vpinns_2021}
E.~Kharazmi, Z.~Zhang, and G.~E.~M. Karniadakis, ``hp-{VPINNs}: {Variational}
  physics-informed neural networks with domain decomposition,'' {\em Computer
  Methods in Applied Mechanics and Engineering}, vol.~374, p.~113547, Feb.
  2021.

\bibitem{morokoff_quasi-monte_1995}
W.~J. Morokoff and R.~E. Caflisch, ``Quasi-{Monte} {Carlo} {Integration},''
  {\em Journal of Computational Physics}, vol.~122, pp.~218--230, Dec. 1995.

\bibitem{pharr_14_2004}
``14 - {Monte} carlo integration {I}: {Basic} concepts,'' in {\em Physically
  {Based} {Rendering}} (M.~Pharr and G.~Humphreys, eds.), pp.~631--660,
  Burlington: Morgan Kaufmann, Jan. 2004.

\bibitem{novak_high_1996}
E.~Novak and K.~Ritter, ``High dimensional integration of smooth functions over
  cubes,'' {\em Numerische Mathematik}, vol.~75, pp.~79--97, Nov. 1996.

\bibitem{rivera_quadrature_2022}
J.~A. Rivera, J.~M. Taylor, A.~J. Omella, and D.~Pardo, ``On quadrature rules
  for solving {Partial} {Differential} {Equations} using {Neural} {Networks},''
  {\em Computer Methods in Applied Mechanics and Engineering}, vol.~393,
  p.~114710, Apr. 2022.

\bibitem{zang_weak_2020}
Y.~Zang, G.~Bao, X.~Ye, and H.~Zhou, ``Weak adversarial networks for
  high-dimensional partial differential equations,'' {\em Journal of
  Computational Physics}, vol.~411, p.~109409, June 2020.

\bibitem{nguyen-thanh_deep_2019}
V.~M. Nguyen-Thanh, X.~Zhuang, and T.~Rabczuk, ``A deep energy method for
  finite deformation hyperelasticity,'' {\em European Journal of Mechanics -
  A/Solids}, p.~103874, Oct. 2019.

\bibitem{e_deep_2018}
W.~E and B.~Yu, ``The {Deep} {Ritz} {Method}: {A} {Deep} {Learning}-{Based}
  {Numerical} {Algorithm} for {Solving} {Variational} {Problems},'' {\em
  Communications in Mathematics and Statistics}, vol.~6, pp.~1--12, Mar. 2018.

\bibitem{grossmann_can_2023}
T.~G. Grossmann, U.~J. Komorowska, J.~Latz, and C.-B. Schönlieb, ``Can
  {Physics}-{Informed} {Neural} {Networks} beat the {Finite} {Element}
  {Method}?,'' Feb. 2023.

\bibitem{kashefi_physics-informed_2022}
A.~Kashefi and T.~Mukerji, ``Physics-informed {PointNet}: {A} deep learning
  solver for steady-state incompressible flows and thermal fields on multiple
  sets of irregular geometries,'' {\em Journal of Computational Physics},
  vol.~468, p.~111510, Nov. 2022.

\bibitem{berg_unified_2018}
J.~Berg and K.~Nyström, ``A unified deep artificial neural network approach to
  partial differential equations in complex geometries,'' {\em Neurocomputing},
  vol.~317, pp.~28--41, Nov. 2018.
\newblock arXiv: 1711.06464.

\bibitem{henkes_physics_2022}
A.~Henkes, H.~Wessels, and R.~Mahnken, ``Physics informed neural networks for
  continuum micromechanics,'' {\em Computer Methods in Applied Mechanics and
  Engineering}, vol.~393, p.~114790, Apr. 2022.

\bibitem{lagaris_neural-network_2000}
I.~Lagaris, A.~Likas, and D.~Papageorgiou, ``Neural-network methods for
  boundary value problems with irregular boundaries,'' {\em IEEE Transactions
  on Neural Networks}, vol.~11, pp.~1041--1049, Sept. 2000.

\bibitem{ferrari_constrained_2008}
S.~Ferrari and M.~Jensenius, ``A {Constrained} {Optimization} {Approach} to
  {Preserving} {Prior} {Knowledge} {During} {Incremental} {Training},'' {\em
  IEEE Transactions on Neural Networks}, vol.~19, pp.~996--1009, June 2008.

\bibitem{rudd_constrained_2014}
K.~Rudd, G.~D. Muro, and S.~Ferrari, ``A {Constrained} {Backpropagation}
  {Approach} for the {Adaptive} {Solution} of {Partial} {Differential}
  {Equations},'' {\em IEEE Transactions on Neural Networks and Learning
  Systems}, vol.~25, pp.~571--584, Mar. 2014.

\bibitem{rudd_constrained_2015}
K.~Rudd and S.~Ferrari, ``A constrained integration ({CINT}) approach to
  solving partial differential equations using artificial neural networks,''
  {\em Neurocomputing}, vol.~155, pp.~277--285, May 2015.

\bibitem{wang_residual_2017}
F.~Wang, M.~Jiang, C.~Qian, S.~Yang, C.~Li, H.~Zhang, X.~Wang, and X.~Tang,
  ``Residual {Attention} {Network} for {Image} {Classification},'' in {\em 2017
  {IEEE} {Conference} on {Computer} {Vision} and {Pattern} {Recognition}
  ({CVPR})}, pp.~6450--6458, July 2017.
\newblock ISSN: 1063-6919.

\bibitem{zhang_occluded_2018}
S.~Zhang, J.~Yang, and B.~Schiele, ``Occluded {Pedestrian} {Detection}
  {Through} {Guided} {Attention} in {CNNs},'' in {\em 2018 {IEEE}/{CVF}
  {Conference} on {Computer} {Vision} and {Pattern} {Recognition}},
  pp.~6995--7003, June 2018.
\newblock ISSN: 2575-7075.

\bibitem{magiera_constraint-aware_2020}
J.~Magiera, D.~Ray, J.~S. Hesthaven, and C.~Rohde, ``Constraint-aware neural
  networks for {Riemann} problems,'' {\em Journal of Computational Physics},
  vol.~409, p.~109345, May 2020.

\bibitem{nandwani_primal_2019}
Y.~Nandwani, A.~Pathak, {Mausam}, and P.~Singla, ``A {Primal} {Dual}
  {Formulation} {For} {Deep} {Learning} {With} {Constraints},'' in {\em
  Advances in {Neural} {Information} {Processing} {Systems}}, vol.~32, Curran
  Associates, Inc., 2019.

\bibitem{mcclenny_self-adaptive_2022}
L.~McClenny and U.~Braga-Neto, ``Self-{Adaptive} {Physics}-{Informed} {Neural}
  {Networks} using a {Soft} {Attention} {Mechanism},'' Apr. 2022.
\newblock arXiv:2009.04544 [cs, stat].

\bibitem{lu_physics-informed_2021}
L.~Lu, R.~Pestourie, W.~Yao, Z.~Wang, F.~Verdugo, and S.~G. Johnson,
  ``Physics-{Informed} {Neural} {Networks} with {Hard} {Constraints} for
  {Inverse} {Design},'' {\em SIAM Journal on Scientific Computing}, vol.~43,
  pp.~B1105--B1132, Jan. 2021.

\bibitem{zeng_competitive_2022}
Q.~Zeng, Y.~Kothari, S.~H. Bryngelson, and F.~Schäfer, ``Competitive {Physics}
  {Informed} {Networks},'' Oct. 2022.
\newblock arXiv:2204.11144 [cs, math].

\bibitem{moser_modeling_2023}
P.~Moser, W.~Fenz, S.~Thumfart, I.~Ganitzer, and M.~Giretzlehner, ``Modeling of
  {3D} {Blood} {Flows} with {Physics}-{Informed} {Neural} {Networks}:
  {Comparison} of {Network} {Architectures},'' {\em Fluids}, vol.~8, p.~46,
  Jan. 2023.

\bibitem{zhu_physics-constrained_2019}
Y.~Zhu, N.~Zabaras, P.-S. Koutsourelakis, and P.~Perdikaris,
  ``Physics-{Constrained} {Deep} {Learning} for {High}-dimensional {Surrogate}
  {Modeling} and {Uncertainty} {Quantification} without {Labeled} {Data},''
  {\em Journal of Computational Physics}, vol.~394, pp.~56--81, Oct. 2019.
\newblock arXiv: 1901.06314.

\bibitem{han_flownet_2020}
J.~Han, J.~Tao, and C.~Wang, ``{FlowNet}: {A} {Deep} {Learning} {Framework} for
  {Clustering} and {Selection} of {Streamlines} and {Stream} {Surfaces},'' {\em
  IEEE Transactions on Visualization and Computer Graphics}, vol.~26,
  pp.~1732--1744, Apr. 2020.

\bibitem{bhatnagar_prediction_2019}
S.~Bhatnagar, Y.~Afshar, S.~Pan, K.~Duraisamy, and S.~Kaushik, ``Prediction of
  aerodynamic flow fields using convolutional neural networks,'' {\em
  Computational Mechanics}, vol.~64, pp.~525--545, Aug. 2019.

\bibitem{gao_phygeonet_2021}
H.~Gao, L.~Sun, and J.-X. Wang, ``{PhyGeoNet}: {Physics}-informed
  geometry-adaptive convolutional neural networks for solving parameterized
  steady-state {PDEs} on irregular domain,'' {\em Journal of Computational
  Physics}, vol.~428, p.~110079, Mar. 2021.

\bibitem{wandel_spline-pinn_2022}
N.~Wandel, M.~Weinmann, M.~Neidlin, and R.~Klein, ``Spline-{PINN}:
  {Approaching} {PDEs} without {Data} using {Fast}, {Physics}-{Informed}
  {Hermite}-{Spline} {CNNs},'' Mar. 2022.
\newblock arXiv:2109.07143 [physics].

\bibitem{gao_physics-informed_2022}
H.~Gao, M.~J. Zahr, and J.-X. Wang, ``Physics-informed graph neural {Galerkin}
  networks: {A} unified framework for solving {PDE}-governed forward and
  inverse problems,'' {\em Computer Methods in Applied Mechanics and
  Engineering}, vol.~390, p.~114502, Feb. 2022.

\bibitem{moller_physics-informed_2021}
M.~Möller, D.~Toshniwal, and F.~Van~Ruiten, ``Physics-{Informed} {Machine}
  {Learning} {Embedded} into {Isogeometric} {Analysis},'' {\em Mathematics: Key
  enabling technology for scientific machine learning}, 2021.

\bibitem{hughes_isogeometric_2005}
T.~J.~R. Hughes, J.~A. Cottrell, and Y.~Bazilevs, ``Isogeometric analysis:
  {CAD}, finite elements, {NURBS}, exact geometry and mesh refinement,'' {\em
  Computer Methods in Applied Mechanics and Engineering}, vol.~194,
  pp.~4135--4195, Oct. 2005.

\bibitem{meethal_finite_2022}
R.~E. Meethal, B.~Obst, M.~Khalil, A.~Ghantasala, A.~Kodakkal, K.-U.
  Bletzinger, and R.~Wüchner, ``Finite {Element} {Method}-enhanced {Neural}
  {Network} for {Forward} and {Inverse} {Problems},'' May 2022.
\newblock arXiv:2205.08321 [cs, math].

\bibitem{hughes_finite_2000}
T.~J.~R. Hughes, {\em The finite element method: linear static and dynamic
  finite element analysis}.
\newblock Mineola, NY: Dover Publications, 2000.

\bibitem{bathe_finite_2014}
K.-J. Bathe, {\em Finite element procedures}.
\newblock Englewood Cliffs, N.J: Prentice-Hall, 2nd ed~ed., 2014.
\newblock OCLC: ocn930843107.

\bibitem{berrone_variational_2022}
S.~Berrone, C.~Canuto, and M.~Pintore, ``Variational {Physics} {Informed}
  {Neural} {Networks}: the {Role} of {Quadratures} and {Test} {Functions},''
  {\em Journal of Scientific Computing}, vol.~92, p.~100, Aug. 2022.

\bibitem{badia_finite_2023}
S.~Badia, W.~Li, and A.~F. Martín, ``Finite element interpolated neural
  networks for solving forward and inverse problems,'' June 2023.
\newblock arXiv:2306.06304 [cs, math].

\bibitem{yazdani_systems_2020}
A.~Yazdani, L.~Lu, M.~Raissi, and G.~E. Karniadakis, ``Systems biology informed
  deep learning for inferring parameters and hidden dynamics,'' {\em PLOS
  Computational Biology}, vol.~16, p.~e1007575, Nov. 2020.

\bibitem{uriarte_finite_2022}
C.~Uriarte, D.~Pardo, and A.~J. Omella, ``A {Finite} {Element} based {Deep}
  {Learning} solver for parametric {PDEs},'' {\em Computer Methods in Applied
  Mechanics and Engineering}, vol.~391, p.~114562, Mar. 2022.

\bibitem{jagtap_conservative_2020}
A.~D. Jagtap, E.~Kharazmi, and G.~E. Karniadakis, ``Conservative
  physics-informed neural networks on discrete domains for conservation laws:
  {Applications} to forward and inverse problems,'' {\em Computer Methods in
  Applied Mechanics and Engineering}, vol.~365, p.~113028, June 2020.

\bibitem{shukla_parallel_2021}
K.~Shukla, A.~D. Jagtap, and G.~E. Karniadakis, ``Parallel physics-informed
  neural networks via domain decomposition,'' {\em Journal of Computational
  Physics}, vol.~447, p.~110683, Dec. 2021.

\bibitem{karniadakis_extended_2020}
A.~D. J. . G.~E. Karniadakis, ``Extended {Physics}-{Informed} {Neural}
  {Networks} ({XPINNs}): {A} {Generalized} {Space}-{Time} {Domain}
  {Decomposition} {Based} {Deep} {Learning} {Framework} for {Nonlinear}
  {Partial} {Differential} {Equations},'' {\em Communications in Computational
  Physics}, vol.~28, pp.~2002--2041, June 2020.

\bibitem{chen_transfer_2021}
X.~Chen, C.~Gong, Q.~Wan, L.~Deng, Y.~Wan, Y.~Liu, B.~Chen, and J.~Liu,
  ``Transfer learning for deep neural network-based partial differential
  equations solving,'' {\em Advances in Aerodynamics}, vol.~3, p.~36, Dec.
  2021.

\bibitem{goswami_transfer_2019}
S.~Goswami, C.~Anitescu, S.~Chakraborty, and T.~Rabczuk, ``Transfer learning
  enhanced physics informed neural network for phase-field modeling of
  fracture,'' {\em arXiv:1907.02531 [cs, stat]}, July 2019.
\newblock arXiv: 1907.02531.

\bibitem{he_deep_2023}
J.~He, C.~Chadha, S.~Kushwaha, S.~Koric, D.~Abueidda, and I.~Jasiuk, ``Deep
  energy method in topology optimization applications,'' {\em Acta Mechanica},
  vol.~234, pp.~1365--1379, Apr. 2023.

\bibitem{nabian_efficient_2021}
M.~A. Nabian, R.~J. Gladstone, and H.~Meidani, ``Efficient training of
  physics‐informed neural networks via importance sampling,'' {\em
  Computer-Aided Civil and Infrastructure Engineering}, vol.~36, pp.~962--977,
  Aug. 2021.

\bibitem{hanna_residual-based_2022}
J.~M. Hanna, J.~V. Aguado, S.~Comas-Cardona, R.~Askri, and D.~Borzacchiello,
  ``Residual-based adaptivity for two-phase flow simulation in porous media
  using {Physics}-informed {Neural} {Networks},'' {\em Computer Methods in
  Applied Mechanics and Engineering}, vol.~396, p.~115100, June 2022.

\bibitem{kollmannsberger_physics-informed_2021}
S.~Kollmannsberger, D.~D'Angella, M.~Jokeit, and L.~Herrmann,
  ``Physics-{Informed} {Neural} {Networks},'' in {\em Deep {Learning} in
  {Computational} {Mechanics}}, vol.~977, pp.~55--84, Cham: Springer
  International Publishing, 2021.
\newblock Series Title: Studies in Computational Intelligence.

\bibitem{anton_identification_2021}
D.~Anton and H.~Wessels, ``Identification of {Material} {Parameters} from
  {Full}-{Field} {Displacement} {Data} {Using} {Physics}-{Informed} {Neural}
  {Networks},'' 2021.

\bibitem{zong_improved_2023}
Y.~Zong, Q.~He, and A.~M. Tartakovsky, ``Improved training of physics-informed
  neural networks for parabolic differential equations with sharply perturbed
  initial conditions,'' {\em Computer Methods in Applied Mechanics and
  Engineering}, vol.~414, p.~116125, Sept. 2023.

\bibitem{yu_gradient-enhanced_2022}
J.~Yu, L.~Lu, X.~Meng, and G.~E. Karniadakis, ``Gradient-enhanced
  physics-informed neural networks for forward and inverse {PDE} problems,''
  {\em Computer Methods in Applied Mechanics and Engineering}, vol.~393,
  p.~114823, Apr. 2022.

\bibitem{taylor_deep_2023}
J.~M. Taylor, D.~Pardo, and I.~Muga, ``A {Deep} {Fourier} {Residual} method for
  solving {PDEs} using {Neural} {Networks},'' {\em Computer Methods in Applied
  Mechanics and Engineering}, vol.~405, p.~115850, Feb. 2023.

\bibitem{chiu_can-pinn_2022}
P.-H. Chiu, J.~C. Wong, C.~Ooi, M.~H. Dao, and Y.-S. Ong, ``{CAN}-{PINN}: {A}
  fast physics-informed neural network based on coupled-automatic–numerical
  differentiation method,'' {\em Computer Methods in Applied Mechanics and
  Engineering}, vol.~395, p.~114909, May 2022.

\bibitem{jagtap_adaptive_2020}
A.~D. Jagtap and G.~E. Karniadakis, ``Adaptive activation functions accelerate
  convergence in deep and physics-informed neural networks,'' {\em Journal of
  Computational Physics}, vol.~404, p.~109136, Mar. 2020.
\newblock arXiv: 1906.01170.

\bibitem{huang_extreme_2006}
G.-B. Huang, Q.-Y. Zhu, and C.-K. Siew, ``Extreme learning machine: {Theory}
  and applications,'' {\em Neurocomputing}, vol.~70, pp.~489--501, Dec. 2006.

\bibitem{huang_extreme_2011}
G.-B. Huang, D.~H. Wang, and Y.~Lan, ``Extreme learning machines: a survey,''
  {\em International Journal of Machine Learning and Cybernetics}, vol.~2,
  pp.~107--122, June 2011.

\bibitem{dong_local_2021}
S.~Dong and Z.~Li, ``Local extreme learning machines and domain decomposition
  for solving linear and nonlinear partial differential equations,'' {\em
  Computer Methods in Applied Mechanics and Engineering}, vol.~387, p.~114129,
  Dec. 2021.

\bibitem{dong_numerical_2022}
S.~Dong and J.~Yang, ``Numerical approximation of partial differential
  equations by a variable projection method with artificial neural networks,''
  {\em Computer Methods in Applied Mechanics and Engineering}, vol.~398,
  p.~115284, Aug. 2022.

\bibitem{haghighat_physics-informed_2021}
E.~Haghighat, M.~Raissi, A.~Moure, H.~Gomez, and R.~Juanes, ``A
  physics-informed deep learning framework for inversion and surrogate modeling
  in solid mechanics,'' {\em Computer Methods in Applied Mechanics and
  Engineering}, vol.~379, p.~113741, June 2021.

\bibitem{bai_introduction_2023}
J.~Bai, H.~Jeong, C.~P. Batuwatta-Gamage, S.~Xiao, Q.~Wang, C.~M. Rathnayaka,
  L.~Alzubaidi, G.-R. Liu, and Y.~Gu, ``An {Introduction} to {Programming}
  {Physics}-{Informed} {Neural} {Network}-{Based} {Computational} {Solid}
  {Mechanics},'' {\em International Journal of Computational Methods},
  p.~2350013, May 2023.

\bibitem{kissas_machine_2020}
G.~Kissas, Y.~Yang, E.~Hwuang, W.~R. Witschey, J.~A. Detre, and P.~Perdikaris,
  ``Machine learning in cardiovascular flows modeling: {Predicting} arterial
  blood pressure from non-invasive {4D} flow {MRI} data using physics-informed
  neural networks,'' {\em Computer Methods in Applied Mechanics and
  Engineering}, vol.~358, p.~112623, Jan. 2020.

\bibitem{raissi_hidden_2020}
M.~Raissi, A.~Yazdani, and G.~E. Karniadakis, ``Hidden fluid mechanics:
  {Learning} velocity and pressure fields from flow visualizations,'' {\em
  Science}, vol.~367, pp.~1026--1030, Feb. 2020.

\bibitem{sun_surrogate_2020}
L.~Sun, H.~Gao, S.~Pan, and J.-X. Wang, ``Surrogate modeling for fluid flows
  based on physics-constrained deep learning without simulation data,'' {\em
  Computer Methods in Applied Mechanics and Engineering}, vol.~361, p.~112732,
  Apr. 2020.

\bibitem{jin_nsfnets_2021}
X.~Jin, S.~Cai, H.~Li, and G.~E. Karniadakis, ``{NSFnets} ({Navier}-{Stokes}
  flow nets): {Physics}-informed neural networks for the incompressible
  {Navier}-{Stokes} equations,'' {\em Journal of Computational Physics},
  vol.~426, p.~109951, Feb. 2021.

\bibitem{cai_flow_2021}
S.~Cai, Z.~Wang, F.~Fuest, Y.~J. Jeon, C.~Gray, and G.~E. Karniadakis, ``Flow
  over an espresso cup: inferring 3-{D} velocity and pressure fields from
  tomographic background oriented {Schlieren} via physics-informed neural
  networks,'' {\em Journal of Fluid Mechanics}, vol.~915, p.~A102, May 2021.

\bibitem{fraces_physics_2021}
C.~G. Fraces and H.~Tchelepi, ``Physics {Informed} {Deep} {Learning} for {Flow}
  and {Transport} in {Porous} {Media},'' OnePetro, Oct. 2021.

\bibitem{zhang_simulation_2022}
W.~Zhang, D.~S. Li, T.~Bui-Thanh, and M.~S. Sacks, ``Simulation of the {3D}
  hyperelastic behavior of ventricular myocardium using a finite-element based
  neural-network approach,'' {\em Computer Methods in Applied Mechanics and
  Engineering}, vol.~394, p.~114871, May 2022.

\bibitem{wang_fluxnet_2023}
J.~C.~H. Wang and J.-P. Hickey, ``{FluxNet}: {A} physics-informed
  learning-based {Riemann} solver for transcritical flows with non-ideal
  thermodynamics,'' {\em Computer Methods in Applied Mechanics and
  Engineering}, vol.~411, p.~116070, June 2023.

\bibitem{amini_niaki_physics-informed_2021}
S.~Amini~Niaki, E.~Haghighat, T.~Campbell, A.~Poursartip, and R.~Vaziri,
  ``Physics-informed neural network for modelling the thermochemical curing
  process of composite-tool systems during manufacture,'' {\em Computer Methods
  in Applied Mechanics and Engineering}, vol.~384, p.~113959, Oct. 2021.

\bibitem{zhu_machine_2021}
Q.~Zhu, Z.~Liu, and J.~Yan, ``Machine learning for metal additive
  manufacturing: predicting temperature and melt pool fluid dynamics using
  physics-informed neural networks,'' {\em Computational Mechanics}, vol.~67,
  pp.~619--635, Feb. 2021.

\bibitem{markidis_old_2021}
S.~Markidis, ``The {Old} and the {New}: {Can} {Physics}-{Informed}
  {Deep}-{Learning} {Replace} {Traditional} {Linear} {Solvers}?,'' {\em
  Frontiers in Big Data}, vol.~4, 2021.

\bibitem{li_ref-nets_2022}
L.~Li, Y.~Li, Q.~Du, T.~Liu, and Y.~Xie, ``{ReF}-nets: {Physics}-informed
  neural network for {Reynolds} equation of gas bearing,'' {\em Computer
  Methods in Applied Mechanics and Engineering}, vol.~391, p.~114524, Mar.
  2022.

\bibitem{chen_physics-informed_2020}
Y.~Chen, L.~Lu, G.~E. Karniadakis, and L.~D. Negro, ``Physics-informed neural
  networks for inverse problems in nano-optics and metamaterials,'' {\em Optics
  Express}, vol.~28, p.~11618, Apr. 2020.

\bibitem{zhang_physics-informed_2020}
R.~Zhang, Y.~Liu, and H.~Sun, ``Physics-{Informed} {Multi}-{LSTM} {Networks}
  for {Metamodeling} of {Nonlinear} {Structures},'' {\em Computer Methods in
  Applied Mechanics and Engineering}, vol.~369, p.~113226, Sept. 2020.
\newblock arXiv: 2002.10253.

\bibitem{shukla_physics-informed_2020}
K.~Shukla, P.~C. Di~Leoni, J.~Blackshire, D.~Sparkman, and G.~E. Karniadakis,
  ``Physics-{Informed} {Neural} {Network} for {Ultrasound} {Nondestructive}
  {Quantification} of {Surface} {Breaking} {Cracks},'' {\em Journal of
  Nondestructive Evaluation}, vol.~39, p.~61, Aug. 2020.

\bibitem{anton_physics-informed_2022}
D.~Anton and H.~Wessels, ``Physics-{Informed} {Neural} {Networks} for
  {Material} {Model} {Calibration} from {Full}-{Field} {Displacement} {Data},''
  Dec. 2022.

\bibitem{herrmann_use_2023}
L.~Herrmann, T.~Bürchner, F.~Dietrich, and S.~Kollmannsberger, ``On the use of
  neural networks for full waveform inversion,'' {\em Computer Methods in
  Applied Mechanics and Engineering}, vol.~415, p.~116278, Oct. 2023.

\bibitem{rojas_parameter_2021}
C.~J.~G. Rojas, M.~L. Bitterncourt, and J.~L. Boldrini, ``Parameter
  identification for a damage model using a physics informed neural network,''
  June 2021.

\bibitem{li_physics_2021}
W.~Li and K.-M. Lee, ``Physics informed neural network for parameter
  identification and boundary force estimation of compliant and biomechanical
  systems,'' {\em International Journal of Intelligent Robotics and
  Applications}, vol.~5, pp.~313--325, Sept. 2021.

\bibitem{zhang_analyses_2022}
E.~Zhang, M.~Dao, G.~E. Karniadakis, and S.~Suresh, ``Analyses of internal
  structures and defects in materials using physics-informed neural networks,''
  {\em Science Advances}, vol.~8, p.~eabk0644, Feb. 2022.

\bibitem{depina_application_2022}
I.~Depina, S.~Jain, S.~Mar~Valsson, and H.~Gotovac, ``Application of
  physics-informed neural networks to inverse problems in unsaturated
  groundwater flow,'' {\em Georisk: Assessment and Management of Risk for
  Engineered Systems and Geohazards}, vol.~16, pp.~21--36, Jan. 2022.

\bibitem{xu_transfer_2023}
C.~Xu, B.~T. Cao, Y.~Yuan, and G.~Meschke, ``Transfer learning based
  physics-informed neural networks for solving inverse problems in engineering
  structures under different loading scenarios,'' {\em Computer Methods in
  Applied Mechanics and Engineering}, vol.~405, p.~115852, Feb. 2023.

\bibitem{sun_physics-informed_2023}
Y.~Sun, U.~Sengupta, and M.~Juniper, ``Physics-informed deep learning for
  simultaneous surrogate modeling and {PDE}-constrained optimization of an
  airfoil geometry,'' {\em Computer Methods in Applied Mechanics and
  Engineering}, vol.~411, p.~116042, June 2023.

\bibitem{rashtbehesht_physicsinformed_2022}
M.~Rasht‐Behesht, C.~Huber, K.~Shukla, and G.~E. Karniadakis,
  ``Physics‐{Informed} {Neural} {Networks} ({PINNs}) for {Wave} {Propagation}
  and {Full} {Waveform} {Inversions},'' {\em Journal of Geophysical Research:
  Solid Earth}, vol.~127, May 2022.

\bibitem{zehnder_ntopo_2021}
J.~Zehnder, Y.~Li, S.~Coros, and B.~Thomaszewski, ``{NTopo}: {Mesh}-free
  {Topology} {Optimization} using {Implicit} {Neural} {Representations},'' Nov.
  2021.

\bibitem{di_lorenzo_physics_2023}
D.~Di~Lorenzo, V.~Champaney, J.~Y. Marzin, C.~Farhat, and F.~Chinesta,
  ``Physics informed and data-based augmented learning in structural health
  diagnosis,'' {\em Computer Methods in Applied Mechanics and Engineering},
  vol.~414, p.~116186, Sept. 2023.

\bibitem{berg_data-driven_2019}
J.~Berg and K.~Nyström, ``Data-driven discovery of {PDEs} in complex
  datasets,'' {\em Journal of Computational Physics}, vol.~384, pp.~239--252,
  May 2019.

\bibitem{udrescu_ai_2020}
S.-M. Udrescu and M.~Tegmark, ``{AI} {Feynman}: {A} physics-inspired method for
  symbolic regression,'' {\em Science Advances}, vol.~6, p.~eaay2631, Apr.
  2020.

\bibitem{feynman_feynman_2011}
R.~P. Feynman, R.~B. Leighton, and M.~L. Sands, {\em The {Feynman} lectures on
  physics}.
\newblock New York: Basic Books, new millennium ed~ed., 2011.
\newblock OCLC: ocn671704374.

\bibitem{meng_ppinn_2020}
X.~Meng, Z.~Li, D.~Zhang, and G.~E. Karniadakis, ``{PPINN}: {Parareal}
  physics-informed neural network for time-dependent {PDEs},'' {\em Computer
  Methods in Applied Mechanics and Engineering}, vol.~370, p.~113250, Oct.
  2020.

\bibitem{mattey_novel_2022}
R.~Mattey and S.~Ghosh, ``A novel sequential method to train physics informed
  neural networks for {Allen} {Cahn} and {Cahn} {Hilliard} equations,'' {\em
  Computer Methods in Applied Mechanics and Engineering}, vol.~390, p.~114474,
  Feb. 2022.

\bibitem{iserles_first_2008}
A.~Iserles, {\em A {First} {Course} in the {Numerical} {Analysis} of
  {Differential} {Equations}}.
\newblock Cambridge University Press, Nov. 2008.
\newblock Google-Books-ID: 3acgAwAAQBAJ.

\bibitem{wessels_neural_2020}
H.~Wessels, C.~Weißenfels, and P.~Wriggers, ``The neural particle method –
  {An} updated {Lagrangian} physics informed neural network for computational
  fluid dynamics,'' {\em Computer Methods in Applied Mechanics and
  Engineering}, vol.~368, p.~113127, Aug. 2020.

\bibitem{bai_general_2022}
J.~Bai, Y.~Zhou, Y.~Ma, H.~Jeong, H.~Zhan, C.~Rathnayaka, E.~Sauret, and Y.~Gu,
  ``A general {Neural} {Particle} {Method} for hydrodynamics modeling,'' {\em
  Computer Methods in Applied Mechanics and Engineering}, vol.~393, p.~114740,
  Apr. 2022.

\bibitem{gonzalez-garcia_identification_1998}
R.~González-García, R.~Rico-Martínez, and I.~G. Kevrekidis, ``Identification
  of distributed parameter systems: {A} neural net based approach,'' {\em
  Computers \& Chemical Engineering}, vol.~22, pp.~S965--S968, Mar. 1998.

\bibitem{long_pde-net_2018}
Z.~Long, Y.~Lu, X.~Ma, and B.~Dong, ``{PDE}-{Net}: {Learning} {PDEs} from
  {Data},'' {\em arXiv:1710.09668 [cs, math, stat]}, Jan. 2018.
\newblock arXiv: 1710.09668.

\bibitem{long_pde-net_2019}
Z.~Long, Y.~Lu, and B.~Dong, ``{PDE}-{Net} 2.0: {Learning} {PDEs} from data
  with a numeric-symbolic hybrid deep network,'' {\em Journal of Computational
  Physics}, vol.~399, p.~108925, Dec. 2019.

\bibitem{hua_pointwise_2018}
B.-S. Hua, M.-K. Tran, and S.-K. Yeung, ``Pointwise {Convolutional} {Neural}
  {Networks},'' Mar. 2018.
\newblock arXiv:1712.05245 [cs].

\bibitem{brunton_discovering_2016}
S.~L. Brunton, J.~L. Proctor, and J.~N. Kutz, ``Discovering governing equations
  from data by sparse identification of nonlinear dynamical systems,'' {\em
  Proceedings of the National Academy of Sciences}, vol.~113, pp.~3932--3937,
  Apr. 2016.

\bibitem{rudy_data-driven_2017}
S.~H. Rudy, S.~L. Brunton, J.~L. Proctor, and J.~N. Kutz, ``Data-driven
  discovery of partial differential equations,'' {\em Science Advances},
  vol.~3, p.~e1602614, Apr. 2017.

\bibitem{schaeffer_learning_2017}
H.~Schaeffer, ``Learning partial differential equations via data discovery and
  sparse optimization,'' {\em Proceedings of the Royal Society A: Mathematical,
  Physical and Engineering Sciences}, vol.~473, p.~20160446, Jan. 2017.

\bibitem{champion_data-driven_2019}
K.~Champion, B.~Lusch, J.~N. Kutz, and S.~L. Brunton, ``Data-driven discovery
  of coordinates and governing equations,'' {\em Proceedings of the National
  Academy of Sciences}, vol.~116, pp.~22445--22451, Nov. 2019.

\bibitem{conti_reduced_2023}
P.~Conti, G.~Gobat, S.~Fresca, A.~Manzoni, and A.~Frangi, ``Reduced order
  modeling of parametrized systems through autoencoders and {SINDy} approach:
  continuation of periodic solutions,'' {\em Computer Methods in Applied
  Mechanics and Engineering}, vol.~411, p.~116072, June 2023.

\bibitem{kim_deep_2019}
B.~Kim, V.~C. Azevedo, N.~Thuerey, T.~Kim, M.~Gross, and B.~Solenthaler, ``Deep
  {Fluids}: {A} {Generative} {Network} for {Parameterized} {Fluid}
  {Simulations},'' {\em Computer Graphics Forum}, vol.~38, pp.~59--70, May
  2019.

\bibitem{ling_machine_2016}
J.~Ling, R.~Jones, and J.~Templeton, ``Machine learning strategies for systems
  with invariance properties,'' {\em Journal of Computational Physics},
  vol.~318, pp.~22--35, Aug. 2016.

\bibitem{ling_reynolds_2016}
J.~Ling, A.~Kurzawski, and J.~Templeton, ``Reynolds averaged turbulence
  modelling using deep neural networks with embedded invariance,'' {\em Journal
  of Fluid Mechanics}, vol.~807, pp.~155--166, Nov. 2016.

\bibitem{smith_isotropic_1965}
G.~F. Smith, ``On isotropic integrity bases,'' {\em Archive for Rational
  Mechanics and Analysis}, vol.~18, pp.~282--292, Jan. 1965.

\bibitem{lutter_deep_2019}
M.~Lutter, K.~Listmann, and J.~Peters, ``Deep {Lagrangian} {Networks} for
  end-to-end learning of energy-based control for under-actuated systems,'' in
  {\em 2019 {IEEE}/{RSJ} {International} {Conference} on {Intelligent} {Robots}
  and {Systems} ({IROS})}, pp.~7718--7725, Nov. 2019.
\newblock ISSN: 2153-0866.

\bibitem{lutter_deep_2019-1}
M.~Lutter, C.~Ritter, and J.~Peters, ``Deep {Lagrangian} {Networks}: {Using}
  {Physics} as {Model} {Prior} for {Deep} {Learning},'' July 2019.
\newblock arXiv:1907.04490 [cs, eess, stat].

\bibitem{cranmer_lagrangian_2020}
M.~Cranmer, S.~Greydanus, S.~Hoyer, P.~Battaglia, D.~Spergel, and S.~Ho,
  ``Lagrangian {Neural} {Networks},'' July 2020.
\newblock arXiv:2003.04630 [physics, stat].

\bibitem{greydanus_hamiltonian_2019}
S.~Greydanus, M.~Dzamba, and J.~Yosinski, ``Hamiltonian {Neural} {Networks},''
  Sept. 2019.
\newblock arXiv:1906.01563 [cs].

\bibitem{zhang_road_2016}
L.~Zhang, F.~Yang, Y.~Daniel~Zhang, and Y.~J. Zhu, ``Road crack detection using
  deep convolutional neural network,'' in {\em 2016 {IEEE} {International}
  {Conference} on {Image} {Processing} ({ICIP})}, pp.~3708--3712, Sept. 2016.

\bibitem{chen_nb-cnn_2018}
F.-C. Chen and M.~R. Jahanshahi, ``{NB}-{CNN}: {Deep} {Learning}-{Based}
  {Crack} {Detection} {Using} {Convolutional} {Neural} {Network} and {Naïve}
  {Bayes} {Data} {Fusion},'' {\em IEEE Transactions on Industrial Electronics},
  vol.~65, pp.~4392--4400, May 2018.

\bibitem{jaeger_infrared_2022}
B.~E. Jaeger, S.~Schmid, C.~U. Grosse, A.~Gögelein, and F.~Elischberger,
  ``Infrared {Thermal} {Imaging}-{Based} {Turbine} {Blade} {Crack}
  {Classification} {Using} {Deep} {Learning},'' {\em Journal of Nondestructive
  Evaluation}, vol.~41, p.~74, Oct. 2022.

\bibitem{korshunova_image-based_2020}
N.~Korshunova, J.~Jomo, G.~Lékó, D.~Reznik, P.~Balázs, and
  S.~Kollmannsberger, ``Image-based material characterization of complex
  microarchitectured additively manufactured structures,'' {\em Computers \&
  Mathematics with Applications}, vol.~80, pp.~2462--2480, Dec. 2020.

\bibitem{hall_barbosa_automation_1999}
C.~Hall~Barbosa, A.~Bruno, M.~Vellasco, M.~Pacheco, J.~Wikswo, and A.~Ewing,
  ``Automation of {SQUlD} nondestructive evaluation of steel plates by neural
  networks,'' {\em IEEE Transactions on Applied Superconductivity}, vol.~9,
  pp.~3475--3478, June 1999.

\bibitem{ovcharenko_deep_2019}
O.~Ovcharenko, V.~Kazei, M.~Kalita, D.~Peter, and T.~Alkhalifah, ``Deep
  learning for low-frequency extrapolation from multioffset seismic data,''
  {\em GEOPHYSICS}, vol.~84, pp.~R989--R1001, Nov. 2019.

\bibitem{sun_extrapolated_2020}
H.~Sun and L.~Demanet, ``Extrapolated full waveform inversion with deep
  learning,'' {\em GEOPHYSICS}, vol.~85, pp.~R275--R288, May 2020.

\bibitem{sun_deep_2022}
H.~Sun and L.~Demanet, ``Deep {Learning} for {Low}-{Frequency} {Extrapolation}
  of {Multicomponent} {Data} in {Elastic} {FWI},'' {\em IEEE Transactions on
  Geoscience and Remote Sensing}, vol.~60, pp.~1--11, 2022.

\bibitem{lewis_deep_2017}
W.~Lewis and D.~Vigh, ``Deep learning prior models from seismic images for
  full-waveform inversion,'' in {\em {SEG} {Technical} {Program} {Expanded}
  {Abstracts} 2017}, (Houston, Texas), pp.~1512--1517, Society of Exploration
  Geophysicists, Aug. 2017.

\bibitem{dyck_determining_1992}
D.~Dyck, D.~Lowther, and S.~McFee, ``Determining an approximate finite element
  mesh density using neural network techniques,'' {\em IEEE Transactions on
  Magnetics}, vol.~28, pp.~1767--1770, Mar. 1992.

\bibitem{chedid_automatic_1996}
R.~Chedid and N.~Najjar, ``Automatic finite-element mesh generation using
  artificial neural networks-{Part} {I}: {Prediction} of mesh density,'' {\em
  IEEE Transactions on Magnetics}, vol.~32, pp.~5173--5178, Sept. 1996.

\bibitem{triantafyllidis_automatic_2000}
D.~G. Triantafyllidis and D.~P. Labridis, ``An automatic mesh generator for
  handling small features in open boundary power transmission line problems
  using artificial neural networks,'' {\em Communications in Numerical Methods
  in Engineering}, vol.~16, pp.~177--190, Mar. 2000.

\bibitem{krzhizhanovskaya_meshingnet_2020}
Z.~Zhang, Y.~Wang, P.~K. Jimack, and H.~Wang, ``{MeshingNet}: {A} {New} {Mesh}
  {Generation} {Method} {Based} on {Deep} {Learning},'' in {\em Computational
  {Science} – {ICCS} 2020} (V.~V. Krzhizhanovskaya, G.~Závodszky, M.~H.
  Lees, J.~J. Dongarra, P.~M.~A. Sloot, S.~Brissos, and J.~Teixeira, eds.),
  vol.~12139, pp.~186--198, Cham: Springer International Publishing, 2020.
\newblock Series Title: Lecture Notes in Computer Science.

\bibitem{lock_meshing_2023}
C.~Lock, O.~Hassan, R.~Sevilla, and J.~Jones, ``Meshing using neural networks
  for improving the efficiency of computer modelling,'' {\em Engineering with
  Computers}, Apr. 2023.

\bibitem{fritzke_growing_1994}
B.~Fritzke, ``Growing cell structures—{A} self-organizing network for
  unsupervised and supervised learning,'' {\em Neural Networks}, vol.~7,
  pp.~1441--1460, Jan. 1994.

\bibitem{alfonzetti_automatic_1996}
S.~Alfonzetti, S.~Coco, S.~Cavalieri, and M.~Malgeri, ``Automatic mesh
  generation by the let-it-grow neural network,'' {\em IEEE Transactions on
  Magnetics}, vol.~32, pp.~1349--1352, May 1996.

\bibitem{triantafyllidis_finite-element_2002}
D.~Triantafyllidis and D.~Labridis, ``A finite-element mesh generator based on
  growing neural networks,'' {\em IEEE Transactions on Neural Networks},
  vol.~13, pp.~1482--1496, Nov. 2002.

\bibitem{lefik_artificial_2003}
M.~Lefik and B.~A. Schrefler, ``Artificial neural network as an incremental
  non-linear constitutive model for a finite element code,'' {\em Computer
  Methods in Applied Mechanics and Engineering}, vol.~192, pp.~3265--3283, July
  2003.

\bibitem{jang_machine_2021}
D.~P. Jang, P.~Fazily, and J.~W. Yoon, ``Machine learning-based constitutive
  model for {J2}- plasticity,'' {\em International Journal of Plasticity},
  vol.~138, p.~102919, Mar. 2021.

\bibitem{lin_application_2008}
Y.~C. Lin, J.~Zhang, and J.~Zhong, ``Application of neural networks to predict
  the elevated temperature flow behavior of a low alloy steel,'' {\em
  Computational Materials Science}, vol.~43, pp.~752--758, Oct. 2008.

\bibitem{li_artificial_2012}
H.-Y. Li, J.-D. Hu, D.-D. Wei, X.-F. Wang, and Y.-H. Li, ``Artificial neural
  network and constitutive equations to predict the hot deformation behavior of
  modified 2.{25Cr}–{1Mo} steel,'' {\em Materials \& Design}, vol.~42,
  pp.~192--197, Dec. 2012.

\bibitem{liu_mechanistically_2022}
D.~Liu, H.~Yang, K.~I. Elkhodary, S.~Tang, W.~K. Liu, and X.~Guo,
  ``Mechanistically informed data-driven modeling of cyclic plasticity via
  artificial neural networks,'' {\em Computer Methods in Applied Mechanics and
  Engineering}, vol.~393, p.~114766, Apr. 2022.

\bibitem{unger_neural_2009}
J.~F. Unger and C.~Könke, ``Neural networks as material models within a
  multiscale approach,'' {\em Computers \& Structures}, vol.~87,
  pp.~1177--1186, Oct. 2009.

\bibitem{hattori_contact_2015}
G.~Hattori and A.~L. Serpa, ``Contact stiffness estimation in {ANSYS} using
  simplified models and artificial neural networks,'' {\em Finite Elements in
  Analysis and Design}, vol.~97, pp.~43--53, May 2015.

\bibitem{oishi_new_1970}
A.~Oishi and S.~Yoshimura, ``A {New} {Local} {Contact} {Search} {Method}
  {Using} a {Multi}-{Layer} {Neural} {Network},'' {\em Computer Modeling in
  Engineering \& Sciences}, vol.~21, no.~2, pp.~93--104, 1970.

\bibitem{oishi_surface--surface_2020}
A.~Oishi and G.~Yagawa, ``A surface-to-surface contact search method enhanced
  by deep learning,'' {\em Computational Mechanics}, vol.~65, pp.~1125--1147,
  Apr. 2020.

\bibitem{singh_machine-learning-augmented_2017}
A.~P. Singh, S.~Medida, and K.~Duraisamy, ``Machine-{Learning}-{Augmented}
  {Predictive} {Modeling} of {Turbulent} {Separated} {Flows} over {Airfoils},''
  {\em AIAA Journal}, vol.~55, pp.~2215--2227, July 2017.

\bibitem{maulik_subgrid_2019}
R.~Maulik, O.~San, A.~Rasheed, and P.~Vedula, ``Subgrid modelling for
  two-dimensional turbulence using neural networks,'' {\em Journal of Fluid
  Mechanics}, vol.~858, pp.~122--144, Jan. 2019.

\bibitem{fabra_finite_2022}
A.~Fabra, J.~Baiges, and R.~Codina, ``Finite element approximation of wave
  problems with correcting terms based on training artificial neural networks
  with fine solutions,'' {\em Computer Methods in Applied Mechanics and
  Engineering}, vol.~399, p.~115280, Sept. 2022.

\bibitem{le_computational_2015}
B.~A. Le, J.~Yvonnet, and Q.~He, ``Computational homogenization of nonlinear
  elastic materials using neural networks,'' {\em International Journal for
  Numerical Methods in Engineering}, vol.~104, pp.~1061--1084, Dec. 2015.

\bibitem{lu_data-driven_2019}
X.~Lu, D.~G. Giovanis, J.~Yvonnet, V.~Papadopoulos, F.~Detrez, and J.~Bai, ``A
  data-driven computational homogenization method based on neural networks for
  the nonlinear anisotropic electrical response of graphene/polymer
  nanocomposites,'' {\em Computational Mechanics}, vol.~64, pp.~307--321, Aug.
  2019.

\bibitem{huang_learning_2020}
D.~Z. Huang, K.~Xu, C.~Farhat, and E.~Darve, ``Learning constitutive relations
  from indirect observations using deep neural networks,'' {\em Journal of
  Computational Physics}, vol.~416, p.~109491, Sept. 2020.

\bibitem{wang_multiscale_2018}
K.~Wang and W.~Sun, ``A multiscale multi-permeability poroplasticity model
  linked by recursive homogenizations and deep learning,'' {\em Computer
  Methods in Applied Mechanics and Engineering}, vol.~334, pp.~337--380, June
  2018.

\bibitem{li_multiscale_2020}
B.~Li and X.~Zhuang, ``Multiscale computation on feedforward neural network and
  recurrent neural network,'' {\em Frontiers of Structural and Civil
  Engineering}, vol.~14, pp.~1285--1298, Dec. 2020.

\bibitem{vlassis_geometric_2020}
N.~N. Vlassis, R.~Ma, and W.~Sun, ``Geometric deep learning for computational
  mechanics {Part} {I}: anisotropic hyperelasticity,'' {\em Computer Methods in
  Applied Mechanics and Engineering}, vol.~371, p.~113299, Nov. 2020.

\bibitem{frankenreiter_hybrid_2011}
I.~Frankenreiter, D.~Rosato, and C.~Miehe, ``Hybrid {Micro}-{Macro}-{Modeling}
  of {Evolving} {Anisotropies} and {Length} {Scales} in {Finite} {Plasticity}
  of {Polycrystals}: {Hybrid} {Micro}-{Macro}-{Modeling} of {Evolving}
  {Anisotropies} and {Length} {Scales} in {Finite} {Plasticity} of
  {Polycrystals},'' {\em PAMM}, vol.~11, pp.~515--518, Dec. 2011.

\bibitem{fish_practical_2013}
J.~Fish, {\em Practical multiscaling}.
\newblock Chichester, West Sussex, United Kingdom: John Wiley \& Sons Inc,
  2013.

\bibitem{linka_constitutive_2021}
K.~Linka, M.~Hillgärtner, K.~P. Abdolazizi, R.~C. Aydin, M.~Itskov, and C.~J.
  Cyron, ``Constitutive artificial neural networks: {A} fast and general
  approach to predictive data-driven constitutive modeling by deep learning,''
  {\em Journal of Computational Physics}, vol.~429, p.~110010, Mar. 2021.

\bibitem{mozaffar_deep_2019}
M.~Mozaffar, R.~Bostanabad, W.~Chen, K.~Ehmann, J.~Cao, and M.~A. Bessa, ``Deep
  learning predicts path-dependent plasticity,'' {\em Proceedings of the
  National Academy of Sciences}, vol.~116, pp.~26414--26420, Dec. 2019.

\bibitem{wu_recurrent_2022}
L.~Wu and L.~Noels, ``Recurrent {Neural} {Networks} ({RNNs}) with
  dimensionality reduction and break down in computational mechanics;
  application to multi-scale localization step,'' {\em Computer Methods in
  Applied Mechanics and Engineering}, vol.~390, p.~114476, Feb. 2022.

\bibitem{abueidda_deep_2021}
D.~W. Abueidda, S.~Koric, N.~A. Sobh, and H.~Sehitoglu, ``Deep learning for
  plasticity and thermo-viscoplasticity,'' {\em International Journal of
  Plasticity}, vol.~136, p.~102852, Jan. 2021.

\bibitem{hsu_using_2020}
Y.-C. Hsu, C.-H. Yu, and M.~J. Buehler, ``Using {Deep} {Learning} to {Predict}
  {Fracture} {Patterns} in {Crystalline} {Solids},'' {\em Matter}, vol.~3,
  pp.~197--211, July 2020.

\bibitem{lew_deep_2021}
A.~J. Lew, C.-H. Yu, Y.-C. Hsu, and M.~J. Buehler, ``Deep learning model to
  predict fracture mechanisms of graphene,'' {\em npj 2D Materials and
  Applications}, vol.~5, pp.~1--8, Apr. 2021.

\bibitem{liu_generic_2020}
M.~Liu, L.~Liang, and W.~Sun, ``A generic physics-informed neural network-based
  constitutive model for soft biological tissues,'' {\em Computer Methods in
  Applied Mechanics and Engineering}, vol.~372, p.~113402, Dec. 2020.

\bibitem{weber_constrained_2021}
P.~Weber, J.~Geiger, and W.~Wagner, ``Constrained neural network training and
  its application to hyperelastic material modeling,'' {\em Computational
  Mechanics}, vol.~68, pp.~1179--1204, Nov. 2021.

\bibitem{leng_predicting_2021}
Y.~Leng, V.~Tac, S.~Calve, and A.~B. Tepole, ``Predicting the {Mechanical}
  {Properties} of {Biopolymer} {Gels} {Using} {Neural} {Networks} {Trained} on
  {Discrete} {Fiber} {Network} {Data},'' {\em Computer Methods in Applied
  Mechanics and Engineering}, vol.~387, p.~114160, Dec. 2021.
\newblock arXiv:2101.11712 [cs, q-bio].

\bibitem{tac_data-driven_2022}
V.~Tac, F.~Sahli~Costabal, and A.~B. Tepole, ``Data-driven tissue mechanics
  with polyconvex neural ordinary differential equations,'' {\em Computer
  Methods in Applied Mechanics and Engineering}, vol.~398, p.~115248, Aug.
  2022.

\bibitem{linden_neural_2023}
L.~Linden, D.~K. Klein, K.~A. Kalina, J.~Brummund, O.~Weeger, and M.~Kästner,
  ``Neural networks meet hyperelasticity: {A} guide to enforcing physics,''
  Feb. 2023.
\newblock arXiv:2302.02403 [cs].

\bibitem{klein_finite_2022}
D.~K. Klein, R.~Ortigosa, J.~Martínez-Frutos, and O.~Weeger, ``Finite
  electro-elasticity with physics-augmented neural networks,'' {\em Computer
  Methods in Applied Mechanics and Engineering}, vol.~400, p.~115501, Oct.
  2022.

\bibitem{klein_polyconvex_2022}
D.~K. Klein, M.~Fernández, R.~J. Martin, P.~Neff, and O.~Weeger, ``Polyconvex
  anisotropic hyperelasticity with neural networks,'' {\em Journal of the
  Mechanics and Physics of Solids}, vol.~159, p.~104703, Feb. 2022.

\bibitem{asad_mechanics-informed_2023}
F.~As'ad and C.~Farhat, ``A {Mechanics}-{Informed} {Neural} {Network}
  {Framework} for {Data}-{Driven} {Nonlinear} {Viscoelasticity},'' in {\em
  {AIAA} {SCITECH} 2023 {Forum}}, (National Harbor, MD \& Online), American
  Institute of Aeronautics and Astronautics, Jan. 2023.

\bibitem{tac_data-driven_2023}
V.~Taç, M.~K. Rausch, F.~Sahli~Costabal, and A.~B. Tepole, ``Data-driven
  anisotropic finite viscoelasticity using neural ordinary differential
  equations,'' {\em Computer Methods in Applied Mechanics and Engineering},
  vol.~411, p.~116046, June 2023.

\bibitem{amos_input_2017}
B.~Amos, L.~Xu, and J.~Z. Kolter, ``Input {Convex} {Neural} {Networks},'' in
  {\em Proceedings of the 34th {International} {Conference} on {Machine}
  {Learning}}, pp.~146--155, PMLR, July 2017.

\bibitem{chen_neural_2019}
R.~T.~Q. Chen, Y.~Rubanova, J.~Bettencourt, and D.~Duvenaud, ``Neural
  {Ordinary} {Differential} {Equations},'' {\em arXiv:1806.07366 [cs, stat]},
  Dec. 2019.
\newblock arXiv: 1806.07366.

\bibitem{chen_polyconvex_2022}
P.~Chen and J.~Guilleminot, ``Polyconvex neural networks for hyperelastic
  constitutive models: {A} rectification approach,'' {\em Mechanics Research
  Communications}, vol.~125, p.~103993, Oct. 2022.

\bibitem{masi_thermodynamics-based_2021}
F.~Masi, I.~Stefanou, P.~Vannucci, and V.~Maffi-Berthier,
  ``Thermodynamics-based {Artificial} {Neural} {Networks} for constitutive
  modeling,'' {\em Journal of the Mechanics and Physics of Solids}, vol.~147,
  p.~104277, Feb. 2021.

\bibitem{masi_material_2021}
F.~Masi, I.~Stefanou, P.~Vannucci, and V.~Maffi-Berthier, ``Material {Modeling}
  via {Thermodynamics}-{Based} {Artificial} {Neural} {Networks},'' in {\em
  Geometric {Structures} of {Statistical} {Physics}, {Information} {Geometry},
  and {Learning}} (F.~Barbaresco and F.~Nielsen, eds.), Springer {Proceedings}
  in {Mathematics} \& {Statistics}, (Cham), pp.~308--329, Springer
  International Publishing, 2021.

\bibitem{masi_multiscale_2022}
F.~Masi and I.~Stefanou, ``Multiscale modeling of inelastic materials with
  {Thermodynamics}-based {Artificial} {Neural} {Networks} ({TANN}),'' {\em
  Computer Methods in Applied Mechanics and Engineering}, vol.~398, p.~115190,
  Aug. 2022.

\bibitem{ladeveze_updating_1994}
P.~Ladeveze, D.~Nedjar, and M.~Reynier, ``Updating of finite element models
  using vibration tests,'' {\em AIAA Journal}, vol.~32, pp.~1485--1491, July
  1994.

\bibitem{marchand_parameter_2019}
B.~Marchand, L.~Chamoin, and C.~Rey, ``Parameter identification and model
  updating in the context of nonlinear mechanical behaviors using a unified
  formulation of the modified {Constitutive} {Relation} {Error} concept,'' {\em
  Computer Methods in Applied Mechanics and Engineering}, vol.~345,
  pp.~1094--1113, Mar. 2019.

\bibitem{nguyen_mcre-based_2022}
H.~N. Nguyen, L.~Chamoin, and C.~Ha~Minh, ``{mCRE}-based parameter
  identification from full-field measurements: {Consistent} framework,
  integrated version, and extension to nonlinear material behaviors,'' {\em
  Computer Methods in Applied Mechanics and Engineering}, vol.~400, p.~115461,
  Oct. 2022.

\bibitem{benady_nn-mcre_2023}
A.~Benady, E.~Baranger, and L.~Chamoin, ``{NN}-{mCRE}: a modified
  {Constitutive} {Relation} {Error} framework for unsupervised learning of
  nonlinear state laws with physics-augmented {Neural} {Networks},'' 2023.

\bibitem{benady_modified_2023}
A.~B. Benady, L.~C. Chamoin, and E.~B. Baranger, ``A modified {Constitutive}
  {Relation} {Error} ({mCRE}) framework to learn nonlinear constitutive models
  from strain measurements with thermodynamics-consistent {Neural}
  {Networks},'' {\em XI International Conference on Adaptive Modeling and
  Simulation (ADMOS 2023)}, vol.~Advanced Techniques for Data Assimilation,
  Inverse Analysis, and Data-based Enrichment of Simulation Models, May 2023.

\bibitem{li_machine-learning_2019}
X.~Li, C.~C. Roth, and D.~Mohr, ``Machine-learning based temperature- and
  rate-dependent plasticity model: {Application} to analysis of fracture
  experiments on {DP} steel,'' {\em International Journal of Plasticity},
  vol.~118, pp.~320--344, July 2019.

\bibitem{thakolkaran_nn-euclid_2022}
P.~Thakolkaran, A.~Joshi, Y.~Zheng, M.~Flaschel, L.~De~Lorenzis, and S.~Kumar,
  ``{NN}-{EUCLID}: {Deep}-learning hyperelasticity without stress data,'' {\em
  Journal of the Mechanics and Physics of Solids}, vol.~169, p.~105076, Dec.
  2022.

\bibitem{li_predicting_2019}
X.~Li, Z.~Liu, S.~Cui, C.~Luo, C.~Li, and Z.~Zhuang, ``Predicting the effective
  mechanical property of heterogeneous materials by image based modeling and
  deep learning,'' {\em Computer Methods in Applied Mechanics and Engineering},
  vol.~347, pp.~735--753, Apr. 2019.

\bibitem{henkes_deep_2021}
A.~Henkes, I.~Caylak, and R.~Mahnken, ``A deep learning driven pseudospectral
  {PCE} based {FFT} homogenization algorithm for complex microstructures,''
  {\em Computer Methods in Applied Mechanics and Engineering}, vol.~385,
  p.~114070, Nov. 2021.

\bibitem{liu_estimation_2019}
M.~Liu, L.~Liang, and W.~Sun, ``Estimation of in vivo constitutive parameters
  of the aortic wall using a machine learning approach,'' {\em Computer Methods
  in Applied Mechanics and Engineering}, vol.~347, pp.~201--217, Apr. 2019.

\bibitem{lu_extraction_2020}
L.~Lu, M.~Dao, P.~Kumar, U.~Ramamurty, G.~E. Karniadakis, and S.~Suresh,
  ``Extraction of mechanical properties of materials through deep learning from
  instrumented indentation,'' {\em Proceedings of the National Academy of
  Sciences}, vol.~117, pp.~7052--7062, Mar. 2020.

\bibitem{meng_composite_2020}
X.~Meng and G.~E. Karniadakis, ``A composite neural network that learns from
  multi-fidelity data: {Application} to function approximation and inverse
  {PDE} problems,'' {\em Journal of Computational Physics}, vol.~401,
  p.~109020, Jan. 2020.

\bibitem{liu_machine_2020}
X.~Liu, C.~E. Athanasiou, N.~P. Padture, B.~W. Sheldon, and H.~Gao, ``A machine
  learning approach to fracture mechanics problems,'' {\em Acta Materialia},
  vol.~190, pp.~105--112, May 2020.

\bibitem{hambli_multiscale_2011}
R.~Hambli, H.~Katerchi, and C.-L. Benhamou, ``Multiscale methodology for bone
  remodelling simulation using coupled finite element and neural network
  computation,'' {\em Biomechanics and Modeling in Mechanobiology}, vol.~10,
  pp.~133--145, Feb. 2011.

\bibitem{flaschel_unsupervised_2021}
M.~Flaschel, S.~Kumar, and L.~De~Lorenzis, ``Unsupervised discovery of
  interpretable hyperelastic constitutive laws,'' {\em Computer Methods in
  Applied Mechanics and Engineering}, vol.~381, p.~113852, Aug. 2021.

\bibitem{tibshirani_regression_1996}
R.~Tibshirani, ``Regression {Shrinkage} and {Selection} via the {Lasso},'' {\em
  Journal of the Royal Statistical Society. Series B (Methodological)},
  vol.~58, no.~1, pp.~267--288, 1996.

\bibitem{flaschel_discovering_2022}
M.~Flaschel, S.~Kumar, and L.~De~Lorenzis, ``Discovering plasticity models
  without stress data,'' {\em npj Computational Materials}, vol.~8, p.~91, Apr.
  2022.
\newblock arXiv:2202.04916 [cs].

\bibitem{marino_automated_2023}
E.~Marino, M.~Flaschel, S.~Kumar, and L.~De~Lorenzis, ``Automated
  identification of linear viscoelastic constitutive laws with {EUCLID},'' {\em
  Mechanics of Materials}, vol.~181, p.~104643, June 2023.

\bibitem{flaschel_automated_2023}
M.~Flaschel, S.~Kumar, and L.~De~Lorenzis, ``Automated discovery of generalized
  standard material models with {EUCLID},'' {\em Computer Methods in Applied
  Mechanics and Engineering}, vol.~405, p.~115867, Feb. 2023.

\bibitem{joshi_bayesian-euclid_2022}
A.~Joshi, P.~Thakolkaran, Y.~Zheng, M.~Escande, M.~Flaschel, L.~De~Lorenzis,
  and S.~Kumar, ``Bayesian-{EUCLID}: {Discovering} hyperelastic material laws
  with uncertainties,'' {\em Computer Methods in Applied Mechanics and
  Engineering}, vol.~398, p.~115225, Aug. 2022.

\bibitem{linka_automated_2023}
K.~Linka, S.~R. St.~Pierre, and E.~Kuhl, ``Automated model discovery for human
  brain using {Constitutive} {Artificial} {Neural} {Networks},'' {\em Acta
  Biomaterialia}, vol.~160, pp.~134--151, Apr. 2023.

\bibitem{linka_new_2023}
K.~Linka and E.~Kuhl, ``A new family of {Constitutive} {Artificial} {Neural}
  {Networks} towards automated model discovery,'' {\em Computer Methods in
  Applied Mechanics and Engineering}, vol.~403, p.~115731, Jan. 2023.

\bibitem{oishi_computational_2017}
A.~Oishi and G.~Yagawa, ``Computational mechanics enhanced by deep learning,''
  {\em Computer Methods in Applied Mechanics and Engineering}, vol.~327,
  pp.~327--351, Dec. 2017.

\bibitem{jung_deep_2020}
J.~Jung, K.~Yoon, and P.-S. Lee, ``Deep learned finite elements,'' {\em
  Computer Methods in Applied Mechanics and Engineering}, vol.~372, p.~113401,
  Dec. 2020.

\bibitem{bar-sinai_learning_2019}
Y.~Bar-Sinai, S.~Hoyer, J.~Hickey, and M.~P. Brenner, ``Learning data-driven
  discretizations for partial differential equations,'' {\em Proceedings of the
  National Academy of Sciences}, vol.~116, pp.~15344--15349, July 2019.

\bibitem{pantidis_integrated_2023}
P.~Pantidis and M.~E. Mobasher, ``Integrated {Finite} {Element} {Neural}
  {Network} ({I}-{FENN}) for non-local continuum damage mechanics,'' {\em
  Computer Methods in Applied Mechanics and Engineering}, vol.~404, p.~115766,
  Feb. 2023.

\bibitem{arcones_neural_2022}
D.~A. Arcones, R.~E. Meethal, B.~Obst, and R.~Wüchner, ``Neural
  {Network}-{Based} {Surrogate} {Models} {Applied} to {Fluid}-{Structure}
  {Interaction} {Problems},'' {\em WCCM-APCOM 2022}, vol.~1700 Data Science,
  Machine Learning and Artificial Intelligence, July 2022.

\bibitem{han_artificial_2021}
C.~Han, P.~Zhang, D.~Bluestein, G.~Cong, and Y.~Deng, ``Artificial intelligence
  for accelerating time integrations in multiscale modeling,'' {\em Journal of
  Computational Physics}, vol.~427, p.~110053, Feb. 2021.

\bibitem{sluzalec_automatic_2023}
T.~Służalec, M.~Dobija, A.~Paszyńska, I.~Muga, M.~Łoś, and M.~Paszyński,
  ``Automatic stabilization of finite-element simulations using neural networks
  and hierarchical matrices,'' {\em Computer Methods in Applied Mechanics and
  Engineering}, vol.~411, p.~116073, June 2023.

\bibitem{casadei_geometric_2013}
F.~Casadei, J.~J. Rimoli, and M.~Ruzzene, ``A geometric multiscale finite
  element method for the dynamic analysis of heterogeneous solids,'' {\em
  Computer Methods in Applied Mechanics and Engineering}, vol.~263, pp.~56--70,
  Aug. 2013.

\bibitem{oztoprak_two-scale_2023}
O.~Oztoprak, A.~Paolini, P.~D'Acunto, E.~Rank, and S.~Kollmannsberger,
  ``Two-scale analysis and design of spaceframes with complex additive
  manufactured nodes,'' 2023.

\bibitem{koeppe_intelligent_2020}
A.~Koeppe, F.~Bamer, and B.~Markert, ``An intelligent nonlinear meta element
  for elastoplastic continua: deep learning using a new {Time}-distributed
  {Residual} {U}-{Net} architecture,'' {\em Computer Methods in Applied
  Mechanics and Engineering}, vol.~366, p.~113088, July 2020.

\bibitem{capuano_smart_2019}
G.~Capuano and J.~J. Rimoli, ``Smart finite elements: {A} novel machine
  learning application,'' {\em Computer Methods in Applied Mechanics and
  Engineering}, vol.~345, pp.~363--381, Mar. 2019.

\bibitem{yamaguchi_zooming_2021}
T.~Yamaguchi and H.~Okuda, ``Zooming method for {FEA} using a neural network,''
  {\em Computers \& Structures}, vol.~247, p.~106480, Apr. 2021.

\bibitem{yin_interfacing_2022}
M.~Yin, E.~Zhang, Y.~Yu, and G.~E. Karniadakis, ``Interfacing finite elements
  with deep neural operators for fast multiscale modeling of mechanics
  problems,'' {\em Computer Methods in Applied Mechanics and Engineering},
  vol.~402, p.~115027, Dec. 2022.

\bibitem{sigmund_usefulness_2011}
O.~Sigmund, ``On the usefulness of non-gradient approaches in topology
  optimization,'' {\em Structural and Multidisciplinary Optimization}, vol.~43,
  pp.~589--596, May 2011.

\bibitem{holl_learning_2020}
P.~Holl, V.~Koltun, and N.~Thuerey, ``Learning to {Control} {PDEs} with
  {Differentiable} {Physics},'' Jan. 2020.
\newblock arXiv:2001.07457 [physics, stat].

\bibitem{um_solver---loop_2020}
K.~Um, R.~Brand, Y.~R. Fei, P.~Holl, and N.~Thuerey, ``Solver-in-the-loop:
  learning from differentiable physics to interact with iterative
  {PDE}-solvers,'' in {\em Proceedings of the 34th {International} {Conference}
  on {Neural} {Information} {Processing} {Systems}}, {NIPS}'20, (Red Hook, NY,
  USA), pp.~6111--6122, Curran Associates Inc., Dec. 2020.

\bibitem{um_solver---loop_2021}
K.~Um, R.~Brand, {Yun}, {Fei}, P.~Holl, and N.~Thuerey, ``Solver-in-the-{Loop}:
  {Learning} from {Differentiable} {Physics} to {Interact} with {Iterative}
  {PDE}-{Solvers},'' Jan. 2021.
\newblock arXiv:2007.00016 [physics].

\bibitem{jensen_inversion_1999}
C.~Jensen, R.~Reed, R.~Marks, M.~El-Sharkawi, J.-B. Jung, R.~Miyamoto,
  G.~Anderson, and C.~Eggen, ``Inversion of feedforward neural networks:
  algorithms and applications,'' {\em Proceedings of the IEEE}, vol.~87,
  pp.~1536--1549, Sept. 1999.

\bibitem{yu_artificial_2019}
C.-H. Yu, Z.~Qin, and M.~J. Buehler, ``Artificial intelligence design algorithm
  for nanocomposites optimized for shear crack resistance,'' {\em Nano
  Futures}, vol.~3, p.~035001, Aug. 2019.

\bibitem{chen_generative_2020}
C.~Chen and G.~X. Gu, ``Generative {Deep} {Neural} {Networks} for {Inverse}
  {Materials} {Design} {Using} {Backpropagation} and {Active} {Learning},''
  {\em Advanced Science}, vol.~7, p.~1902607, Mar. 2020.

\bibitem{tanyu_deep_2022}
D.~N. Tanyu, J.~Ning, T.~Freudenberg, N.~Heilenkötter, A.~Rademacher, U.~Iben,
  and P.~Maass, ``Deep {Learning} {Methods} for {Partial} {Differential}
  {Equations} and {Related} {Parameter} {Identification} {Problems},'' Dec.
  2022.

\bibitem{zohdi_machine-learning_2023}
T.~I. Zohdi, ``A machine-learning digital-twin for rapid large-scale
  solar-thermal energy system design,'' {\em Computer Methods in Applied
  Mechanics and Engineering}, vol.~412, p.~115991, July 2023.

\bibitem{plessix_review_2006}
R.-E. Plessix, ``A review of the adjoint-state method for computing the
  gradient of a functional with geophysical applications,'' {\em Geophysical
  Journal International}, vol.~167, pp.~495--503, Nov. 2006.

\bibitem{givoli_tutorial_2021}
D.~Givoli, ``A tutorial on the adjoint method for inverse problems,'' {\em
  Computer Methods in Applied Mechanics and Engineering}, vol.~380, p.~113810,
  July 2021.

\bibitem{keshavarzzadeh_robust_2021}
V.~Keshavarzzadeh, R.~M. Kirby, and A.~Narayan, ``Robust topology optimization
  with low rank approximation using artificial neural networks,'' {\em
  Computational Mechanics}, vol.~68, pp.~1297--1323, Dec. 2021.

\bibitem{qian_accelerating_2021}
C.~Qian and W.~Ye, ``Accelerating gradient-based topology optimization design
  with dual-model artificial neural networks,'' {\em Structural and
  Multidisciplinary Optimization}, vol.~63, pp.~1687--1707, Apr. 2021.

\bibitem{chi_universal_2021}
H.~Chi, Y.~Zhang, T.~L.~E. Tang, L.~Mirabella, L.~Dalloro, L.~Song, and G.~H.
  Paulino, ``Universal machine learning for topology optimization,'' {\em
  Computer Methods in Applied Mechanics and Engineering}, vol.~375, p.~112739,
  Mar. 2021.

\bibitem{aulig_evolutionary_2013}
N.~Aulig and M.~Olhofer, ``Evolutionary generation of neural network update
  signals for the topology optimization of structures,'' in {\em Proceedings of
  the 15th annual conference companion on {Genetic} and evolutionary
  computation}, {GECCO} '13 {Companion}, (New York, NY, USA), pp.~213--214,
  Association for Computing Machinery, July 2013.

\bibitem{aulig_topology_2014}
N.~Aulig and M.~Olhofer, ``Topology {Optimization} by predicting sensitivities
  based on local state features,'' 2014.

\bibitem{aulig_neuro-evolutionary_2015}
N.~Aulig and M.~Olhofer, ``Neuro-evolutionary {Topology} {Optimization} with
  {Adaptive} {Improvement} {Threshold},'' in {\em Applications of
  {Evolutionary} {Computation}} (A.~M. Mora and G.~Squillero, eds.), Lecture
  {Notes} in {Computer} {Science}, (Cham), pp.~655--666, Springer International
  Publishing, 2015.

\bibitem{zhang_speeding_2021}
Y.~Zhang, H.~Chi, B.~Chen, T.~L.~E. Tang, L.~Mirabella, L.~Song, and G.~H.
  Paulino, ``Speeding up {Computational} {Morphogenesis} with {Online} {Neural}
  {Synthetic} {Gradients},'' Apr. 2021.

\bibitem{hunter_superadjoint_2023}
T.~H. Hunter, S.~H. Hulsoff, and A.~S. Sitaram, ``{SuperAdjoint}:
  {Super}-{Resolution} {Neural} {Networks} in {Adjoint}-based {Output} {Error}
  {Estimation},'' {\em XI International Conference on Adaptive Modeling and
  Simulation (ADMOS 2023)}, vol.~Recent Developments in Methods and
  Applications for Mesh Adaptation, May 2023.

\bibitem{fukami_machine-learning-based_2021}
K.~Fukami, K.~Fukagata, and K.~Taira, ``Machine-learning-based spatio-temporal
  super resolution reconstruction of turbulent flows,'' {\em Journal of Fluid
  Mechanics}, vol.~909, p.~A9, Feb. 2021.

\bibitem{senhora_machine_2022}
F.~V. Senhora, H.~Chi, Y.~Zhang, L.~Mirabella, T.~L.~E. Tang, and G.~H.
  Paulino, ``Machine learning for topology optimization: {Physics}-based
  learning through an independent training strategy,'' {\em Computer Methods in
  Applied Mechanics and Engineering}, vol.~398, p.~115116, Aug. 2022.

\bibitem{hsieh_learning_2019}
J.-T. Hsieh, S.~Zhao, S.~Eismann, L.~Mirabella, and S.~Ermon, ``Learning
  {Neural} {PDE} {Solvers} with {Convergence} {Guarantees},'' June 2019.
\newblock arXiv:1906.01200 [cs, stat].

\bibitem{ye_acceleration_2021}
H.-L. Ye, J.-C. Li, B.-S. Yuan, N.~Wei, and Y.-K. Sui, ``Acceleration {Design}
  for {Continuum} {Topology} {Optimization} by {Using} {Pix2pix} {Neural}
  {Network},'' {\em International Journal of Applied Mechanics}, vol.~13,
  p.~2150042, May 2021.

\bibitem{hoyer_neural_2019}
S.~Hoyer, J.~Sohl-Dickstein, and S.~Greydanus, ``Neural reparameterization
  improves structural optimization,'' Sept. 2019.

\bibitem{xu_neural_2019}
K.~Xu and E.~Darve, ``The {Neural} {Network} {Approach} to {Inverse} {Problems}
  in {Differential} {Equations},'' Jan. 2019.

\bibitem{berg_neural_2021}
J.~Berg and K.~Nyström, ``Neural networks as smooth priors for inverse
  problems for {PDEs},'' {\em Journal of Computational Mathematics and Data
  Science}, vol.~1, p.~100008, Sept. 2021.

\bibitem{chen_new_2021}
L.~Chen and M.-H.~H. Shen, ``A {New} {Topology} {Optimization} {Approach} by
  {Physics}-{Informed} {Deep} {Learning} {Process},'' {\em Advances in Science,
  Technology and Engineering Systems Journal}, vol.~6, pp.~233--240, July 2021.

\bibitem{halle_artificial_2021}
A.~Halle, L.~F. Campanile, and A.~Hasse, ``An {Artificial}
  {Intelligence}–{Assisted} {Design} {Method} for {Topology} {Optimization}
  without {Pre}-{Optimized} {Training} {Data},'' {\em Applied Sciences},
  vol.~11, p.~9041, Jan. 2021.

\bibitem{deng_topology_2020}
H.~Deng and A.~C. To, ``Topology optimization based on deep representation
  learning ({DRL}) for compliance and stress-constrained design,'' {\em
  Computational Mechanics}, vol.~66, pp.~449--469, Aug. 2020.

\bibitem{chandrasekhar_tounn_2021}
A.~Chandrasekhar and K.~Suresh, ``{TOuNN}: {Topology} {Optimization} using
  {Neural} {Networks},'' {\em Structural and Multidisciplinary Optimization},
  vol.~63, pp.~1135--1149, Mar. 2021.

\bibitem{chandrasekhar_length_2021}
A.~Chandrasekhar and K.~Suresh, ``Length {Scale} {Control} in {Topology}
  {Optimization} using {Fourier} {Enhanced} {Neural} {Networks},'' Sept. 2021.

\bibitem{chandrasekhar_multi-material_2021}
A.~Chandrasekhar and K.~Suresh, ``Multi-{Material} {Topology} {Optimization}
  {Using} {Neural} {Networks},'' {\em Computer-Aided Design}, vol.~136,
  p.~103017, July 2021.

\bibitem{park_deepsdf_2019}
J.~J. Park, P.~Florence, J.~Straub, R.~Newcombe, and S.~Lovegrove, ``{DeepSDF}:
  {Learning} {Continuous} {Signed} {Distance} {Functions} for {Shape}
  {Representation},'' in {\em 2019 {IEEE}/{CVF} {Conference} on {Computer}
  {Vision} and {Pattern} {Recognition} ({CVPR})}, pp.~165--174, June 2019.
\newblock ISSN: 2575-7075.

\bibitem{michalkiewicz_implicit_2019}
M.~Michalkiewicz, J.~K. Pontes, D.~Jack, M.~Baktashmotlagh, and A.~Eriksson,
  ``Implicit {Surface} {Representations} {As} {Layers} in {Neural}
  {Networks},'' in {\em 2019 {IEEE}/{CVF} {International} {Conference} on
  {Computer} {Vision} ({ICCV})}, pp.~4742--4751, Oct. 2019.
\newblock ISSN: 2380-7504.

\bibitem{gropp_implicit_2020}
A.~Gropp, L.~Yariv, N.~Haim, M.~Atzmon, and Y.~Lipman, ``Implicit geometric
  regularization for learning shapes,'' in {\em Proceedings of the 37th
  {International} {Conference} on {Machine} {Learning}}, vol.~119 of {\em
  {ICML}'20}, pp.~3789--3799, JMLR.org, July 2020.

\bibitem{sitzmann_implicit_2020}
V.~Sitzmann, J.~N.~P. Martel, A.~W. Bergman, D.~B. Lindell, and G.~Wetzstein,
  ``Implicit {Neural} {Representations} with {Periodic} {Activation}
  {Functions},'' {\em arXiv:2006.09661 [cs, eess]}, June 2020.
\newblock arXiv: 2006.09661.

\bibitem{huang_textbackslash_2021}
Z.~Huang, S.~Bai, and J.~Z. Kolter, ``({\textbackslash}textbackslash
  textrm{\textbackslash}lbrace {Implicit}{\textbackslash}rbrace
  ){\textasciicircum}2: {Implicit} {Layers} for {Implicit} {Representations},''
  in {\em Advances in {Neural} {Information} {Processing} {Systems}}, vol.~34,
  pp.~9639--9650, Curran Associates, Inc., 2021.

\bibitem{deng_parametric_2021}
H.~Deng and A.~C. To, ``A {Parametric} {Level} {Set} {Method} for {Topology}
  {Optimization} based on {Deep} {Neural} {Network} ({DNN}),'' Jan. 2021.

\bibitem{zhang_tonr_2021}
Z.~Zhang, Y.~Li, W.~Zhou, X.~Chen, W.~Yao, and Y.~Zhao, ``{TONR}: {An}
  exploration for a novel way combining neural network with topology
  optimization,'' {\em Computer Methods in Applied Mechanics and Engineering},
  vol.~386, p.~114083, Dec. 2021.

\bibitem{biswas_prestack_2019}
R.~Biswas, M.~K. Sen, V.~Das, and T.~Mukerji, ``Prestack and poststack
  inversion using a physics-guided convolutional neural network,'' {\em
  Interpretation}, vol.~7, pp.~SE161--SE174, Aug. 2019.

\bibitem{alfarraj_semi-supervised_2019}
M.~Alfarraj and G.~AlRegib, ``Semi-supervised learning for acoustic impedance
  inversion,'' in {\em {SEG} {Technical} {Program} {Expanded} {Abstracts}
  2019}, (San Antonio, Texas), pp.~2298--2302, Society of Exploration
  Geophysicists, Aug. 2019.

\bibitem{dong_learning_2014}
C.~Dong, C.~C. Loy, K.~He, and X.~Tang, ``Learning a {Deep} {Convolutional}
  {Network} for {Image} {Super}-{Resolution},'' in {\em Computer {Vision} –
  {ECCV} 2014} (D.~Fleet, T.~Pajdla, B.~Schiele, and T.~Tuytelaars, eds.),
  Lecture {Notes} in {Computer} {Science}, (Cham), pp.~184--199, Springer
  International Publishing, 2014.

\bibitem{dong_image_2015}
C.~Dong, C.~C. Loy, K.~He, and X.~Tang, ``Image {Super}-{Resolution} {Using}
  {Deep} {Convolutional} {Networks},'' July 2015.
\newblock arXiv:1501.00092 [cs].

\bibitem{fukami_super-resolution_2019}
K.~Fukami, K.~Fukagata, and K.~Taira, ``Super-resolution reconstruction of
  turbulent flows with machine learning,'' {\em Journal of Fluid Mechanics},
  vol.~870, pp.~106--120, July 2019.

\bibitem{napier_artificial_2020}
N.~Napier, S.-A. Sriraman, H.~T. Tran, and K.~A. James, ``An {Artificial}
  {Neural} {Network} {Approach} for {Generating} {High}-{Resolution} {Designs}
  {From} {Low}-{Resolution} {Input} in {Topology} {Optimization},'' {\em
  Journal of Mechanical Design}, vol.~142, p.~011402, Jan. 2020.

\bibitem{wang_deep_2021}
C.~Wang, S.~Yao, Z.~Wang, and J.~Hu, ``Deep super-resolution neural network for
  structural topology optimization,'' {\em Engineering Optimization}, vol.~53,
  pp.~2108--2121, Dec. 2021.

\bibitem{xue_efficient_2021}
L.~Xue, J.~Liu, G.~Wen, and H.~Wang, ``Efficient, high-resolution topology
  optimization method based on convolutional neural networks,'' {\em Frontiers
  of Mechanical Engineering}, vol.~16, pp.~80--96, Mar. 2021.

\bibitem{oishi_finite_2021}
A.~Oishi and G.~Yagawa, ``Finite {Elements} {Using} {Neural} {Networks} and a
  {Posteriori} {Error},'' {\em Archives of Computational Methods in
  Engineering}, vol.~28, pp.~3433--3456, Aug. 2021.

\bibitem{elingaard_-homogenization_2022}
M.~O. Elingaard, N.~Aage, J.~A. Bærentzen, and O.~Sigmund, ``De-homogenization
  using convolutional neural networks,'' {\em Computer Methods in Applied
  Mechanics and Engineering}, vol.~388, p.~114197, Jan. 2022.

\bibitem{wan_data-assisted_2018}
Z.~Y. Wan, P.~Vlachas, P.~Koumoutsakos, and T.~Sapsis, ``Data-assisted
  reduced-order modeling of extreme events in complex dynamical systems,'' {\em
  PLOS ONE}, vol.~13, p.~e0197704, May 2018.

\bibitem{sato_example-based_2018}
S.~Sato, Y.~Dobashi, T.~Kim, and T.~Nishita, ``Example-based turbulence style
  transfer,'' {\em ACM Transactions on Graphics}, vol.~37, pp.~84:1--84:9, July
  2018.

\bibitem{chu_data-driven_2017}
M.~Chu and N.~Thuerey, ``Data-driven synthesis of smoke flows with {CNN}-based
  feature descriptors,'' {\em ACM Transactions on Graphics}, vol.~36,
  pp.~69:1--69:14, July 2017.

\bibitem{yildiz_integrated_2003}
A.~Yildiz, N.~Öztürk, N.~Kaya, and F.~Öztürk, ``Integrated optimal topology
  design and shape optimization using neural networks,'' {\em Structural and
  Multidisciplinary Optimization}, vol.~25, pp.~251--260, Oct. 2003.

\bibitem{lin_artificial_2005}
C.-Y. Lin and S.-H. Lin, ``Artificial neural network based hole image
  interpretation techniques for integrated topology and shape optimization,''
  {\em Computer Methods in Applied Mechanics and Engineering}, vol.~194,
  pp.~3817--3837, Sept. 2005.

\bibitem{chen_output-based_2020}
G.~Chen and K.~Fidkowski, ``Output-{Based} {Error} {Estimation} and {Mesh}
  {Adaptation} {Using} {Convolutional} {Neural} {Networks}: {Application} to a
  {Scalar} {Advection}-{Diffusion} {Problem},'' in {\em {AIAA} {Scitech} 2020
  {Forum}}, (Orlando, FL), American Institute of Aeronautics and Astronautics,
  Jan. 2020.

\bibitem{takeuchi_neural_1994}
J.~Takeuchi and Y.~Kosugi, ``Neural network representation of finite element
  method,'' {\em Neural Networks}, vol.~7, pp.~389--395, Jan. 1994.

\bibitem{ramuhalli_finite-element_2005}
P.~Ramuhalli, L.~Udpa, and S.~Udpa, ``Finite-element neural networks for
  solving differential equations,'' {\em IEEE Transactions on Neural Networks},
  vol.~16, pp.~1381--1392, Nov. 2005.

\bibitem{sikora_artificial_1999}
R.~Sikora, J.~Sikora, E.~Cardelli, and T.~Chady, ``Artificial neural network
  application for material evaluation by electromagnetic methods,'' in {\em
  {IJCNN}'99. {International} {Joint} {Conference} on {Neural} {Networks}.
  {Proceedings} ({Cat}. {No}.{99CH36339})}, vol.~6, pp.~4027--4032 vol.6, July
  1999.

\bibitem{xu_application_1999}
G.~Xu, G.~Littlefair, R.~Penson, and R.~Callan, ``Application of {FE}-based
  neural networks to dynamic problems,'' in {\em {ICONIP}'99. {ANZIIS}'99 \&
  {ANNES}'99 \& {ACNN}'99. 6th {International} {Conference} on {Neural}
  {Information} {Processing}. {Proceedings} ({Cat}. {No}.{99EX378})}, vol.~3,
  pp.~1039--1044 vol.3, Nov. 1999.

\bibitem{guo_finite_1999}
F.~Guo, P.~Zhang, F.~Wang, X.~Ma, and G.~Qiu, ``Finite element analysis based
  {Hopfield} neural network model for solving nonlinear electromagnetic field
  problems,'' in {\em {IJCNN}'99. {International} {Joint} {Conference} on
  {Neural} {Networks}. {Proceedings} ({Cat}. {No}.{99CH36339})}, vol.~6,
  pp.~4399--4403 vol.6, July 1999.

\bibitem{lee_neural_1990}
H.~Lee and I.~S. Kang, ``Neural algorithm for solving differential equations,''
  {\em Journal of Computational Physics}, vol.~91, pp.~110--131, Nov. 1990.

\bibitem{kalkkuhl_fem-based_1999}
J.~Kalkkuhl, K.~Hunt, and H.~Fritz, ``{FEM}-based neural-network approach to
  nonlinear modeling with application to longitudinal vehicle dynamics
  control,'' {\em IEEE Transactions on Neural Networks}, vol.~10, pp.~885--897,
  July 1999.

\bibitem{xu_finite-element_2012}
C.~Xu, C.~Wang, F.~Ji, and X.~Yuan, ``Finite-{Element} {Neural}
  {Network}-{Based} {Solving} 3-{D} {Differential} {Equations} in {MFL},'' {\em
  IEEE Transactions on Magnetics}, vol.~48, pp.~4747--4756, Dec. 2012.

\bibitem{yang_efficient_2012}
Z.~Yang, M.~Ruess, S.~Kollmannsberger, A.~Düster, and E.~Rank, ``An efficient
  integration technique for the voxel-based finite cell method: {Efficient}
  {Integration} {Technique} {For} {Finite} {Cells},'' {\em International
  Journal for Numerical Methods in Engineering}, vol.~91, pp.~457--471, Aug.
  2012.

\bibitem{zhang_hierarchical_2021}
L.~Zhang, L.~Cheng, H.~Li, J.~Gao, C.~Yu, R.~Domel, Y.~Yang, S.~Tang, and W.~K.
  Liu, ``Hierarchical deep-learning neural networks: finite elements and
  beyond,'' {\em Computational Mechanics}, vol.~67, pp.~207--230, Jan. 2021.

\bibitem{saha_hierarchical_2021}
S.~Saha, Z.~Gan, L.~Cheng, J.~Gao, O.~L. Kafka, X.~Xie, H.~Li, M.~Tajdari,
  H.~A. Kim, and W.~K. Liu, ``Hierarchical {Deep} {Learning} {Neural} {Network}
  ({HiDeNN}): {An} artificial intelligence ({AI}) framework for computational
  science and engineering,'' {\em Computer Methods in Applied Mechanics and
  Engineering}, vol.~373, p.~113452, Jan. 2021.

\bibitem{zhang_hidenn-td_2022}
L.~Zhang, Y.~Lu, S.~Tang, and W.~K. Liu, ``{HiDeNN}-{TD}: {Reduced}-order
  hierarchical deep learning neural networks,'' {\em Computer Methods in
  Applied Mechanics and Engineering}, vol.~389, p.~114414, Feb. 2022.

\bibitem{liu_hidenn-fem_2023}
Y.~Liu, C.~Park, Y.~Lu, S.~Mojumder, W.~K. Liu, and D.~Qian, ``{HiDeNN}-{FEM}:
  a seamless machine learning approach to nonlinear finite element analysis,''
  {\em Computational Mechanics}, vol.~72, pp.~173--194, July 2023.

\bibitem{lu_convolution_2023}
Y.~Lu, H.~Li, L.~Zhang, C.~Park, S.~Mojumder, S.~Knapik, Z.~Sang, S.~Tang,
  D.~W. Apley, G.~J. Wagner, and W.~K. Liu, ``Convolution {Hierarchical}
  {Deep}-learning {Neural} {Networks} ({C}-{HiDeNN}): finite elements,
  isogeometric analysis, tensor decomposition, and beyond,'' {\em Computational
  Mechanics}, vol.~72, pp.~333--362, Aug. 2023.

\bibitem{park_convolution_2023}
C.~Park, Y.~Lu, S.~Saha, T.~Xue, J.~Guo, S.~Mojumder, D.~W. Apley, G.~J.
  Wagner, and W.~K. Liu, ``Convolution hierarchical deep-learning neural
  network ({C}-{HiDeNN}) with graphics processing unit ({GPU}) acceleration,''
  {\em Computational Mechanics}, vol.~72, pp.~383--409, Aug. 2023.

\bibitem{li_convolution_2023}
H.~Li, S.~Knapik, Y.~Li, C.~Park, J.~Guo, S.~Mojumder, Y.~Lu, W.~Chen, D.~W.
  Apley, and W.~K. Liu, ``Convolution {Hierarchical} {Deep}-{Learning} {Neural}
  {Network} {Tensor} {Decomposition} ({C}-{HiDeNN}-{TD}) for high-resolution
  topology optimization,'' {\em Computational Mechanics}, vol.~72,
  pp.~363--382, Aug. 2023.

\bibitem{yao_fea-net_2019}
H.~Yao, Y.~Ren, and Y.~Liu, ``{FEA}-{Net}: {A} {Deep} {Convolutional} {Neural}
  {Network} {With} {PhysicsPrior} {For} {Efficient} {Data} {Driven} {PDE}
  {Learning},'' in {\em {AIAA} {Scitech} 2019 {Forum}}, (San Diego,
  California), American Institute of Aeronautics and Astronautics, Jan. 2019.

\bibitem{yao_fea-net_2020}
H.~Yao, Y.~Gao, and Y.~Liu, ``{FEA}-{Net}: {A} physics-guided data-driven model
  for efficient mechanical response prediction,'' {\em Computer Methods in
  Applied Mechanics and Engineering}, vol.~363, p.~112892, May 2020.

\bibitem{mishra_nfdtd_2005}
R.~Mishra and P.~Hall, ``{NFDTD} concept,'' {\em IEEE Transactions on Neural
  Networks}, vol.~16, pp.~484--490, Mar. 2005.

\bibitem{richardson_seismic_2018}
A.~Richardson, ``Seismic {Full}-{Waveform} {Inversion} {Using} {Deep}
  {Learning} {Tools} and {Techniques},'' Jan. 2018.

\bibitem{sun_theory-guided_2020}
J.~Sun, Z.~Niu, K.~A. Innanen, J.~Li, and D.~O. Trad, ``A theory-guided
  deep-learning formulation and optimization of seismic waveform inversion,''
  {\em GEOPHYSICS}, vol.~85, pp.~R87--R99, Mar. 2020.

\bibitem{hughes_wave_2019}
T.~W. Hughes, I.~A.~D. Williamson, M.~Minkov, and S.~Fan, ``Wave physics as an
  analog recurrent neural network,'' {\em Science Advances}, vol.~5,
  p.~eaay6946, Dec. 2019.

\bibitem{liu_deep_2019}
Z.~Liu, C.~T. Wu, and M.~Koishi, ``A deep material network for multiscale
  topology learning and accelerated nonlinear modeling of heterogeneous
  materials,'' {\em Computer Methods in Applied Mechanics and Engineering},
  vol.~345, pp.~1138--1168, Mar. 2019.

\bibitem{liu_exploring_2019}
Z.~Liu and C.~T. Wu, ``Exploring the {3D} architectures of deep material
  network in data-driven multiscale mechanics,'' {\em Journal of the Mechanics
  and Physics of Solids}, vol.~127, pp.~20--46, June 2019.

\bibitem{haber_stable_2018}
E.~Haber and L.~Ruthotto, ``Stable {Architectures} for {Deep} {Neural}
  {Networks},'' {\em Inverse Problems}, vol.~34, p.~014004, Jan. 2018.
\newblock arXiv:1705.03341 [cs, math].

\bibitem{ruthotto_deep_2018}
L.~Ruthotto and E.~Haber, ``Deep {Neural} {Networks} {Motivated} by {Partial}
  {Differential} {Equations},'' Dec. 2018.
\newblock arXiv:1804.04272 [cs, math, stat].

\bibitem{lu_beyond_2020}
Y.~Lu, A.~Zhong, Q.~Li, and B.~Dong, ``Beyond {Finite} {Layer} {Neural}
  {Networks}: {Bridging} {Deep} {Architectures} and {Numerical} {Differential}
  {Equations},'' Mar. 2020.
\newblock arXiv:1710.10121 [cs, stat].

\bibitem{pontriagin_mathematical_1986}
L.~S. Pontriagin, L.~W. Neustadt, and L.~S. Pontriagin, {\em The mathematical
  theory of optimal processes}.
\newblock No.~v. 1 in Classics of {Soviet} mathematics, New York: Gordon and
  Breach Science Publishers, english ed~ed., 1986.

\bibitem{yu_physics-based_2018}
Y.~Yu, H.~Yao, and Y.~Liu, ``Physics-based {Learning} for {Aircraft} {Dynamics}
  {Simulation},'' {\em Annual Conference of the PHM Society}, vol.~10, Sept.
  2018.

\bibitem{ranade_discretizationnet_2021}
R.~Ranade, C.~Hill, and J.~Pathak, ``{DiscretizationNet}: {A} machine-learning
  based solver for {Navier}–{Stokes} equations using finite volume
  discretization,'' {\em Computer Methods in Applied Mechanics and
  Engineering}, vol.~378, p.~113722, May 2021.

\bibitem{foster_generative_2023}
D.~Foster, {\em Generative deep learning: teaching machines to paint, write,
  compose, and play}.
\newblock Sebastopol, CA: O'Reilly Media, Incorporated, second edition~ed.,
  2023.
\newblock OCLC: 1378390519.

\bibitem{rezende_stochastic_2014}
D.~J. Rezende, S.~Mohamed, and D.~Wierstra, ``Stochastic {Backpropagation} and
  {Approximate} {Inference} in {Deep} {Generative} {Models},'' May 2014.
\newblock arXiv:1401.4082 [cs, stat].

\bibitem{kingma_auto-encoding_2022}
D.~P. Kingma and M.~Welling, ``Auto-{Encoding} {Variational} {Bayes},'' Dec.
  2022.
\newblock arXiv:1312.6114 [cs, stat].

\bibitem{goodfellow_generative_2014}
I.~Goodfellow, J.~Pouget-Abadie, M.~Mirza, B.~Xu, D.~Warde-Farley, S.~Ozair,
  A.~Courville, and Y.~Bengio, ``Generative {Adversarial} {Nets},'' in {\em
  Advances in {Neural} {Information} {Processing} {Systems}}, vol.~27, Curran
  Associates, Inc., 2014.

\bibitem{nash_equilibrium_1950}
J.~F. Nash, ``Equilibrium points in \textit{n} -person games,'' {\em
  Proceedings of the National Academy of Sciences}, vol.~36, pp.~48--49, Jan.
  1950.

\bibitem{salimans_improved_2016}
T.~Salimans, I.~Goodfellow, W.~Zaremba, V.~Cheung, A.~Radford, X.~Chen, and
  X.~Chen, ``Improved {Techniques} for {Training} {GANs},'' in {\em Advances in
  {Neural} {Information} {Processing} {Systems}}, vol.~29, Curran Associates,
  Inc., 2016.

\bibitem{srivastava_veegan_2017}
A.~Srivastava, L.~Valkov, C.~Russell, M.~U. Gutmann, and C.~Sutton, ``{VEEGAN}:
  reducing mode collapse in {GANs} using implicit variational learning,'' in
  {\em Proceedings of the 31st {International} {Conference} on {Neural}
  {Information} {Processing} {Systems}}, {NIPS}'17, (Red Hook, NY, USA),
  pp.~3310--3320, Curran Associates Inc., Dec. 2017.

\bibitem{arjovsky_wasserstein_2017}
M.~Arjovsky, S.~Chintala, and L.~Bottou, ``Wasserstein {GAN},'' Dec. 2017.
\newblock arXiv:1701.07875 [cs, stat].

\bibitem{mirza_conditional_2014}
M.~Mirza and S.~Osindero, ``Conditional {Generative} {Adversarial} {Nets},''
  Nov. 2014.
\newblock arXiv:1411.1784 [cs, stat].

\bibitem{chen_infogan_2016}
X.~Chen, Y.~Duan, R.~Houthooft, J.~Schulman, I.~Sutskever, and P.~Abbeel,
  ``{InfoGAN}: interpretable representation learning by information maximizing
  generative adversarial nets,'' in {\em Proceedings of the 30th
  {International} {Conference} on {Neural} {Information} {Processing}
  {Systems}}, {NIPS}'16, (Red Hook, NY, USA), pp.~2180--2188, Curran Associates
  Inc., Dec. 2016.

\bibitem{bridle_unsupervised_1991}
J.~S. Bridle, A.~J.~R. Heading, and D.~J.~C. MacKay, ``Unsupervised
  classifiers, mutual information and 'phantom targets','' in {\em Proceedings
  of the 4th {International} {Conference} on {Neural} {Information}
  {Processing} {Systems}}, {NIPS}'91, (San Francisco, CA, USA), pp.~1096--1101,
  Morgan Kaufmann Publishers Inc., Dec. 1991.

\bibitem{larsen_autoencoding_2016}
A.~B.~L. Larsen, S.~K. Sønderby, H.~Larochelle, and O.~Winther, ``Autoencoding
  beyond pixels using a learned similarity metric,'' in {\em Proceedings of the
  33rd {International} {Conference} on {International} {Conference} on
  {Machine} {Learning} - {Volume} 48}, {ICML}'16, (New York, NY, USA),
  pp.~1558--1566, JMLR.org, June 2016.

\bibitem{sohl-dickstein_deep_2015}
J.~Sohl-Dickstein, E.~Weiss, N.~Maheswaranathan, and S.~Ganguli, ``Deep
  {Unsupervised} {Learning} using {Nonequilibrium} {Thermodynamics},'' in {\em
  Proceedings of the 32nd {International} {Conference} on {Machine}
  {Learning}}, pp.~2256--2265, PMLR, June 2015.

\bibitem{ho_denoising_2020}
J.~Ho, A.~Jain, and P.~Abbeel, ``Denoising {Diffusion} {Probabilistic}
  {Models},'' in {\em Advances in {Neural} {Information} {Processing}
  {Systems}}, vol.~33, pp.~6840--6851, Curran Associates, Inc., 2020.

\bibitem{nichol_improved_2021}
A.~Nichol and P.~Dhariwal, ``Improved {Denoising} {Diffusion} {Probabilistic}
  {Models},'' Feb. 2021.
\newblock arXiv:2102.09672 [cs, stat].

\bibitem{rezende_variational_2015}
D.~Rezende and S.~Mohamed, ``Variational {Inference} with {Normalizing}
  {Flows},'' in {\em Proceedings of the 32nd {International} {Conference} on
  {Machine} {Learning}}, pp.~1530--1538, PMLR, June 2015.

\bibitem{kobyzev_normalizing_2021}
I.~Kobyzev, S.~J.~D. Prince, and M.~A. Brubaker, ``Normalizing {Flows}: {An}
  {Introduction} and {Review} of {Current} {Methods},'' {\em IEEE Transactions
  on Pattern Analysis and Machine Intelligence}, vol.~43, pp.~3964--3979, Nov.
  2021.
\newblock arXiv:1908.09257 [cs, stat].

\bibitem{mosser_reconstruction_2017}
L.~Mosser, O.~Dubrule, and M.~J. Blunt, ``Reconstruction of three-dimensional
  porous media using generative adversarial neural networks,'' {\em Physical
  Review E}, vol.~96, p.~043309, Oct. 2017.

\bibitem{feng_reconstruction_2019}
J.~Feng, X.~He, Q.~Teng, C.~Ren, H.~Chen, and Y.~Li, ``Reconstruction of porous
  media from extremely limited information using conditional generative
  adversarial networks,'' {\em Physical Review E}, vol.~100, p.~033308, Sept.
  2019.

\bibitem{shams_coupled_2020}
R.~Shams, M.~Masihi, R.~B. Boozarjomehry, and M.~J. Blunt, ``Coupled generative
  adversarial and auto-encoder neural networks to reconstruct three-dimensional
  multi-scale porous media,'' {\em Journal of Petroleum Science and
  Engineering}, vol.~186, p.~106794, Mar. 2020.

\bibitem{xia_multi-scale_2022}
P.~Xia, H.~Bai, and T.~Zhang, ``Multi-scale reconstruction of porous media
  based on progressively growing generative adversarial networks,'' {\em
  Stochastic Environmental Research and Risk Assessment}, vol.~36,
  pp.~3685--3705, Nov. 2022.

\bibitem{henkes_three-dimensional_2022}
A.~Henkes and H.~Wessels, ``Three-dimensional microstructure generation using
  generative adversarial neural networks in the context of continuum
  micromechanics,'' {\em Computer Methods in Applied Mechanics and
  Engineering}, vol.~400, p.~115497, Oct. 2022.

\bibitem{rawat_novel_2019}
S.~Rawat and M.~H.~H. Shen, ``A novel topology design approach using an
  integrated deep learning network architecture,'' Jan. 2019.

\bibitem{yaji_data-driven_2022}
K.~Yaji, S.~Yamasaki, and K.~Fujita, ``Data-driven multifidelity topology
  design using a deep generative model: {Application} to forced convection heat
  transfer problems,'' {\em Computer Methods in Applied Mechanics and
  Engineering}, vol.~388, p.~114284, Jan. 2022.

\bibitem{lee_microstructure_2023}
K.-H. Lee and G.~J. Yun, ``Microstructure reconstruction using diffusion-based
  generative models,'' Jan. 2023.
\newblock arXiv:2211.10949 [cond-mat, physics:physics].

\bibitem{dureth_conditional_2023}
C.~Düreth, P.~Seibert, D.~Rücker, S.~Handford, M.~Kästner, and M.~Gude,
  ``Conditional diffusion-based microstructure reconstruction,'' {\em Materials
  Today Communications}, vol.~35, p.~105608, June 2023.

\bibitem{vlassis_denoising_2023}
N.~N. Vlassis and W.~Sun, ``Denoising diffusion algorithm for inverse design of
  microstructures with fine-tuned nonlinear material properties,'' {\em
  Computer Methods in Applied Mechanics and Engineering}, vol.~413, p.~116126,
  Aug. 2023.

\bibitem{feng_end--end_2020}
J.~Feng, Q.~Teng, B.~Li, X.~He, H.~Chen, and Y.~Li, ``An end-to-end
  three-dimensional reconstruction framework of porous media from a single
  two-dimensional image based on deep learning,'' {\em Computer Methods in
  Applied Mechanics and Engineering}, vol.~368, p.~113043, Aug. 2020.

\bibitem{kench_generating_2021}
S.~Kench and S.~J. Cooper, ``Generating three-dimensional structures from a
  two-dimensional slice with generative adversarial network-based
  dimensionality expansion,'' {\em Nature Machine Intelligence}, vol.~3,
  pp.~299--305, Apr. 2021.

\bibitem{li_cascaded_2022}
Y.~Li, P.~Jian, and G.~Han, ``Cascaded {Progressive} {Generative} {Adversarial}
  {Networks} for {Reconstructing} {Three}-{Dimensional} {Grayscale} {Core}
  {Images} {From} a {Single} {Two}-{Dimensional} {Image},'' {\em Frontiers in
  Physics}, vol.~10, 2022.

\bibitem{zhang_3d-pmrnn_2022}
F.~Zhang, X.~He, Q.~Teng, X.~Wu, and X.~Dong, ``{3D}-{PMRNN}: {Reconstructing}
  three-dimensional porous media from the two-dimensional image with recurrent
  neural network,'' {\em Journal of Petroleum Science and Engineering},
  vol.~208, p.~109652, Jan. 2022.

\bibitem{zheng_rockgpt_2022}
Q.~Zheng and D.~Zhang, ``{RockGPT}: reconstructing three-dimensional digital
  rocks from single two-dimensional slice with deep learning,'' {\em
  Computational Geosciences}, vol.~26, pp.~677--696, June 2022.

\bibitem{phan_size-invariant_2022}
J.~Phan, L.~Ruspini, G.~Kiss, and F.~Lindseth, ``Size-invariant {3D} generation
  from a single {2D} rock image,'' {\em Journal of Petroleum Science and
  Engineering}, vol.~215, p.~110648, Aug. 2022.

\bibitem{zhang_slice--voxel_2021}
F.~Zhang, Q.~Teng, H.~Chen, X.~He, and X.~Dong, ``Slice-to-voxel stochastic
  reconstructions on porous media with hybrid deep generative model,'' {\em
  Computational Materials Science}, vol.~186, p.~110018, Jan. 2021.

\bibitem{rawat_application_2019}
S.~Rawat and M.~H. Shen, ``Application of {Adversarial} {Networks} for {3D}
  {Structural} {Topology} {Optimization},'' pp.~2019--01--0829, Apr. 2019.

\bibitem{rawat_novel_2019-1}
S.~Rawat and M.-H.~H. Shen, ``A {Novel} {Topology} {Optimization} {Approach}
  using {Conditional} {Deep} {Learning},'' Jan. 2019.

\bibitem{shen_new_2019}
M.-H.~H. Shen and L.~Chen, ``A {New} {CGAN} {Technique} for {Constrained}
  {Topology} {Design} {Optimization},'' Apr. 2019.

\bibitem{wessels_computational_2022}
H.~Wessels, C.~Böhm, F.~Aldakheel, M.~Hüpgen, M.~Haist, L.~Lohaus, and
  P.~Wriggers, ``Computational {Homogenization} {Using} {Convolutional}
  {Neural} {Networks},'' in {\em Current {Trends} and {Open} {Problems} in
  {Computational} {Mechanics}} (F.~Aldakheel, B.~Hudobivnik, M.~Soleimani,
  H.~Wessels, C.~Weißenfels, and M.~Marino, eds.), pp.~569--579, Cham:
  Springer International Publishing, 2022.

\bibitem{mosser_stochastic_2020}
L.~Mosser, O.~Dubrule, and M.~J. Blunt, ``Stochastic {Seismic} {Waveform}
  {Inversion} {Using} {Generative} {Adversarial} {Networks} as a {Geological}
  {Prior},'' {\em Mathematical Geosciences}, vol.~52, pp.~53--79, Jan. 2020.

\bibitem{guo_indirect_2018}
T.~Guo, D.~J. Lohan, R.~Cang, M.~Y. Ren, and J.~T. Allison, ``An {Indirect}
  {Design} {Representation} for {Topology} {Optimization} {Using} {Variational}
  {Autoencoder} and {Style} {Transfer},'' in {\em 2018
  {AIAA}/{ASCE}/{AHS}/{ASC} {Structures}, {Structural} {Dynamics}, and
  {Materials} {Conference}}, {AIAA} {SciTech} {Forum}, American Institute of
  Aeronautics and Astronautics, Jan. 2018.

\bibitem{vulimiri_integrating_2021}
P.~S. Vulimiri, H.~Deng, F.~Dugast, X.~Zhang, and A.~C. To, ``Integrating
  {Geometric} {Data} into {Topology} {Optimization} via {Neural} {Style}
  {Transfer},'' {\em Materials}, vol.~14, p.~4551, Jan. 2021.

\bibitem{gatys_neural_2016}
L.~Gatys, A.~Ecker, and M.~Bethge, ``A {Neural} {Algorithm} of {Artistic}
  {Style},'' {\em Journal of Vision}, vol.~16, p.~326, Sept. 2016.

\bibitem{khan_shiphullgan_2023}
S.~Khan, K.~Goucher-Lambert, K.~Kostas, and P.~Kaklis, ``{ShipHullGAN}: {A}
  generic parametric modeller for ship hull design using deep convolutional
  generative model,'' {\em Computer Methods in Applied Mechanics and
  Engineering}, vol.~411, p.~116051, June 2023.

\bibitem{chen_inverse_2022}
Q.~Chen, J.~Wang, P.~Pope, W.~(Wayne)~Chen, and M.~Fuge, ``Inverse {Design} of
  {Two}-{Dimensional} {Airfoils} {Using} {Conditional} {Generative} {Models}
  and {Surrogate} {Log}-{Likelihoods},'' {\em Journal of Mechanical Design},
  vol.~144, p.~021712, Feb. 2022.

\bibitem{chen_beziergan_2021}
W.~Chen and M.~Fuge, ``B{\textbackslash}'{ezierGAN}: {Automatic} {Generation}
  of {Smooth} {Curves} from {Interpretable} {Low}-{Dimensional} {Parameters},''
  Jan. 2021.
\newblock arXiv:1808.08871 [cs, stat].

\bibitem{chen_mo-padgan_2021}
W.~Chen and F.~Ahmed, ``{MO}-{PaDGAN}: {Reparameterizing} {Engineering}
  {Designs} for augmented multi-objective optimization,'' {\em Applied Soft
  Computing}, vol.~113, p.~107909, Dec. 2021.

\bibitem{richardson_generative_2018}
A.~Richardson, ``Generative {Adversarial} {Networks} for {Model} {Order}
  {Reduction} in {Seismic} {Full}-{Waveform} {Inversion},'' June 2018.
\newblock arXiv:1806.00828 [physics].

\bibitem{zhang_da-vegan_2023}
Y.~Zhang, P.~Seibert, A.~Otto, A.~Raßloff, M.~Ambati, and M.~Kästner,
  ``{DA}-{VEGAN}: {Differentiably} {Augmenting} {VAE}-{GAN} for microstructure
  reconstruction from extremely small data sets,'' Feb. 2023.
\newblock arXiv:2303.03403 [cs].

\bibitem{chen_padgan_2021}
W.~Chen and F.~Ahmed, ``{PaDGAN}: {Learning} to {Generate} {High}-{Quality}
  {Novel} {Designs},'' {\em Journal of Mechanical Design}, vol.~143, p.~031703,
  Mar. 2021.

\bibitem{kulesza_determinantal_2012}
A.~Kulesza and B.~Taskar, ``Determinantal point processes for machine
  learning,'' {\em Foundations and Trends® in Machine Learning}, vol.~5,
  no.~2-3, pp.~123--286, 2012.
\newblock arXiv:1207.6083 [cs, stat].

\bibitem{bates_formulation_2003}
S.~J. Bates, J.~Sienz, and D.~S. Langley, ``Formulation of the
  {Audze}–{Eglais} {Uniform} {Latin} {Hypercube} design of experiments,''
  {\em Advances in Engineering Software}, vol.~34, pp.~493--506, Aug. 2003.

\bibitem{heyrani_nobari_creativegan_2021}
A.~Heyrani~Nobari, M.~F. Rashad, and F.~Ahmed, ``{CreativeGAN}: {Editing}
  {Generative} {Adversarial} {Networks} for {Creative} {Design} {Synthesis},''
  in {\em Volume {3A}: 47th {Design} {Automation} {Conference} ({DAC})},
  (Virtual, Online), p.~V03AT03A002, American Society of Mechanical Engineers,
  Aug. 2021.

\bibitem{bau_rewriting_2020}
D.~Bau, S.~Liu, T.~Wang, J.-Y. Zhu, and A.~Torralba, ``Rewriting a {Deep}
  {Generative} {Model},'' July 2020.
\newblock arXiv:2007.15646 [cs].

\bibitem{elgammal_can_2017}
A.~Elgammal, B.~Liu, M.~Elhoseiny, and M.~Mazzone, ``{CAN}: {Creative}
  {Adversarial} {Networks}, {Generating} "{Art}" by {Learning} {About} {Styles}
  and {Deviating} from {Style} {Norms},'' June 2017.
\newblock arXiv:1706.07068 [cs].

\bibitem{oh_deep_2019}
S.~Oh, Y.~Jung, S.~Kim, I.~Lee, and N.~Kang, ``Deep {Generative} {Design}:
  {Integration} of {Topology} {Optimization} and {Generative} {Models},'' {\em
  Journal of Mechanical Design}, vol.~141, p.~111405, Nov. 2019.

\bibitem{greminger_generative_2020}
M.~Greminger, ``Generative {Adversarial} {Networks} {With} {Synthetic}
  {Training} {Data} for {Enforcing} {Manufacturing} {Constraints} on {Topology}
  {Optimization},'' in {\em Volume {11A}: 46th {Design} {Automation}
  {Conference} ({DAC})}, (Virtual, Online), p.~V11AT11A005, American Society of
  Mechanical Engineers, Aug. 2020.

\bibitem{yoo_integrating_2021}
S.~Yoo, S.~Lee, S.~Kim, K.~H. Hwang, J.~H. Park, and N.~Kang, ``Integrating
  deep learning into {CAD}/{CAE} system: generative design and evaluation of
  {3D} conceptual wheel,'' {\em Structural and Multidisciplinary Optimization},
  vol.~64, pp.~2725--2747, Oct. 2021.

\bibitem{zhang_machine-learning_2023}
W.~Zhang, Y.~Wang, Z.~Du, C.~Liu, S.-K. Youn, and X.~Guo, ``Machine-learning
  assisted topology optimization for architectural design with artistic
  flavor,'' {\em Computer Methods in Applied Mechanics and Engineering},
  vol.~413, p.~116041, Aug. 2023.

\bibitem{bendsoe_topology_2003}
M.~P. Bendsøe and O.~Sigmund, {\em Topology optimization: theory, methods, and
  applications}.
\newblock Berlin ; New York: Springer, 2003.

\bibitem{yang_fwigan_2023}
F.~Yang and J.~Ma, ``{FWIGAN}: {Full}‐{Waveform} {Inversion} via a
  {Physics}‐{Informed} {Generative} {Adversarial} {Network},'' {\em Journal
  of Geophysical Research: Solid Earth}, vol.~128, p.~e2022JB025493, Apr. 2023.

\bibitem{radhakrishnan_creative_2018}
S.~Radhakrishnan, V.~Bharadwaj, V.~Manjunath, and R.~Srinath, ``Creative
  {Intelligence} – {Automating} {Car} {Design} {Studio} with {Generative}
  {Adversarial} {Networks} ({GAN}),'' in {\em Machine {Learning} and
  {Knowledge} {Extraction}} (A.~Holzinger, P.~Kieseberg, A.~M. Tjoa, and
  E.~Weippl, eds.), Lecture {Notes} in {Computer} {Science}, (Cham),
  pp.~160--175, Springer International Publishing, 2018.

\bibitem{chen_synthesizing_2019}
W.~Chen and M.~Fuge, ``Synthesizing {Designs} {With} {Interpart} {Dependencies}
  {Using} {Hierarchical} {Generative} {Adversarial} {Networks},'' {\em Journal
  of Mechanical Design}, vol.~141, p.~111403, Nov. 2019.

\bibitem{nie_topologygan_2020}
Z.~Nie, T.~Lin, H.~Jiang, and L.~B. Kara, ``{TopologyGAN}: {Topology}
  {Optimization} {Using} {Generative} {Adversarial} {Networks} {Based} on
  {Physical} {Fields} {Over} the {Initial} {Domain},'' Mar. 2020.

\bibitem{hertlein_generative_2021}
N.~Hertlein, P.~R. Buskohl, A.~Gillman, K.~Vemaganti, and S.~Anand,
  ``Generative adversarial network for early-stage design flexibility in
  topology optimization for additive manufacturing,'' {\em Journal of
  Manufacturing Systems}, vol.~59, pp.~675--685, Apr. 2021.

\bibitem{heyrani_nobari_range-gan_2021}
A.~Heyrani~Nobari, W.~W. Chen, and F.~Ahmed, ``{RANGE}-{GAN}: {Design}
  {Synthesis} {Under} {Constraints} {Using} {Conditional} {Generative}
  {Adversarial} {Networks},'' {\em Journal of Mechanical Design}, pp.~1--16,
  Sept. 2021.

\bibitem{wang_ih-gan_2022}
J.~Wang, W.~W. Chen, D.~Da, M.~Fuge, and R.~Rai, ``{IH}-{GAN}: {A} conditional
  generative model for implicit surface-based inverse design of cellular
  structures,'' {\em Computer Methods in Applied Mechanics and Engineering},
  vol.~396, p.~115060, June 2022.

\bibitem{duque_automated_2019}
L.~Duque, G.~Gutiérrez, C.~Arias, A.~Rüger, and H.~Jaramillo, ``Automated
  {Velocity} {Estimation} by {Deep} {Learning} {Based} {Seismic}-to-{Velocity}
  {Mapping},'' vol.~2019, pp.~1--5, European Association of Geoscientists \&
  Engineers, June 2019.

\bibitem{wang_seismic_2022}
Y.-Q. Wang, Q.~Wang, W.-K. Lu, Q.~Ge, and X.-F. Yan, ``Seismic impedance
  inversion based on cycle-consistent generative adversarial network,'' {\em
  Petroleum Science}, vol.~19, pp.~147--161, Feb. 2022.

\bibitem{zhu_unpaired_2020}
J.-Y. Zhu, T.~Park, P.~Isola, and A.~A. Efros, ``Unpaired {Image}-to-{Image}
  {Translation} using {Cycle}-{Consistent} {Adversarial} {Networks},'' Aug.
  2020.
\newblock arXiv:1703.10593 [cs].

\bibitem{li_non-iterative_2019}
B.~Li, C.~Huang, X.~Li, S.~Zheng, and J.~Hong, ``Non-iterative structural
  topology optimization using deep learning,'' {\em Computer-Aided Design},
  vol.~115, pp.~172--180, Oct. 2019.

\bibitem{xie_tempogan_2018}
Y.~Xie, E.~Franz, M.~Chu, and N.~Thuerey, ``{tempoGAN}: a temporally coherent,
  volumetric {GAN} for super-resolution fluid flow,'' {\em ACM Transactions on
  Graphics}, vol.~37, pp.~95:1--95:15, July 2018.

\bibitem{pang_deep_2022}
G.~Pang, C.~Shen, L.~Cao, and A.~V.~D. Hengel, ``Deep {Learning} for {Anomaly}
  {Detection}: {A} {Review},'' {\em ACM Computing Surveys}, vol.~54, pp.~1--38,
  Mar. 2022.

\bibitem{hawkins_outlier_2002}
S.~Hawkins, H.~He, G.~Williams, and R.~Baxter, ``Outlier {Detection} {Using}
  {Replicator} {Neural} {Networks},'' in {\em Data {Warehousing} and
  {Knowledge} {Discovery}} (Y.~Kambayashi, W.~Winiwarter, and M.~Arikawa,
  eds.), Lecture {Notes} in {Computer} {Science}, (Berlin, Heidelberg),
  pp.~170--180, Springer, 2002.

\bibitem{schlegl_unsupervised_2017}
T.~Schlegl, P.~Seeböck, S.~M. Waldstein, U.~Schmidt-Erfurth, and G.~Langs,
  ``Unsupervised {Anomaly} {Detection} with {Generative} {Adversarial}
  {Networks} to {Guide} {Marker} {Discovery},'' in {\em Information
  {Processing} in {Medical} {Imaging}} (M.~Niethammer, M.~Styner, S.~Aylward,
  H.~Zhu, I.~Oguz, P.-T. Yap, and D.~Shen, eds.), Lecture {Notes} in {Computer}
  {Science}, (Cham), pp.~146--157, Springer International Publishing, 2017.

\bibitem{zenati_efficient_2019}
H.~Zenati, C.~S. Foo, B.~Lecouat, G.~Manek, and V.~R. Chandrasekhar,
  ``Efficient {GAN}-{Based} {Anomaly} {Detection},'' May 2019.
\newblock arXiv:1802.06222 [cs, stat].

\bibitem{schlegl_f-anogan_2019}
T.~Schlegl, P.~Seeböck, S.~M. Waldstein, G.~Langs, and U.~Schmidt-Erfurth,
  ``f-{AnoGAN}: {Fast} unsupervised anomaly detection with generative
  adversarial networks,'' {\em Medical Image Analysis}, vol.~54, pp.~30--44,
  May 2019.

\bibitem{henkes_gan_2023}
A.~Henkes, L.~Herrmann, H.~Wessels, and S.~Kollmannsberger, ``{GAN} enables
  {Microstructure} {Monitoring} for {Additive} {Manufacturing} of {Complex}
  {Structures},'' {\em in preparation}, 2023.

\bibitem{mnih_human-level_2015}
V.~Mnih, K.~Kavukcuoglu, D.~Silver, A.~A. Rusu, J.~Veness, M.~G. Bellemare,
  A.~Graves, M.~Riedmiller, A.~K. Fidjeland, G.~Ostrovski, S.~Petersen,
  C.~Beattie, A.~Sadik, I.~Antonoglou, H.~King, D.~Kumaran, D.~Wierstra,
  S.~Legg, and D.~Hassabis, ``Human-level control through deep reinforcement
  learning,'' {\em Nature}, vol.~518, pp.~529--533, Feb. 2015.

\bibitem{silver_mastering_2017}
D.~Silver, J.~Schrittwieser, K.~Simonyan, I.~Antonoglou, A.~Huang, A.~Guez,
  T.~Hubert, L.~Baker, M.~Lai, A.~Bolton, Y.~Chen, T.~Lillicrap, F.~Hui,
  L.~Sifre, G.~van~den Driessche, T.~Graepel, and D.~Hassabis, ``Mastering the
  game of {Go} without human knowledge,'' {\em Nature}, vol.~550, pp.~354--359,
  Oct. 2017.

\bibitem{vinyals_grandmaster_2019}
O.~Vinyals, I.~Babuschkin, W.~M. Czarnecki, M.~Mathieu, A.~Dudzik, J.~Chung,
  D.~H. Choi, R.~Powell, T.~Ewalds, P.~Georgiev, J.~Oh, D.~Horgan, M.~Kroiss,
  I.~Danihelka, A.~Huang, L.~Sifre, T.~Cai, J.~P. Agapiou, M.~Jaderberg, A.~S.
  Vezhnevets, R.~Leblond, T.~Pohlen, V.~Dalibard, D.~Budden, Y.~Sulsky,
  J.~Molloy, T.~L. Paine, C.~Gulcehre, Z.~Wang, T.~Pfaff, Y.~Wu, R.~Ring,
  D.~Yogatama, D.~Wünsch, K.~McKinney, O.~Smith, T.~Schaul, T.~Lillicrap,
  K.~Kavukcuoglu, D.~Hassabis, C.~Apps, and D.~Silver, ``Grandmaster level in
  {StarCraft} {II} using multi-agent reinforcement learning,'' {\em Nature},
  vol.~575, pp.~350--354, Nov. 2019.

\bibitem{kober_reinforcement_2013}
J.~Kober, J.~A. Bagnell, and J.~Peters, ``Reinforcement learning in robotics:
  {A} survey,'' {\em The International Journal of Robotics Research}, vol.~32,
  pp.~1238--1274, Sept. 2013.

\bibitem{kim_autonomous_2003}
H.~Kim, M.~Jordan, S.~Sastry, and A.~Ng, ``Autonomous {Helicopter} {Flight} via
  {Reinforcement} {Learning},'' in {\em Advances in {Neural} {Information}
  {Processing} {Systems}}, vol.~16, MIT Press, 2003.

\bibitem{abbeel_application_2006}
P.~Abbeel, A.~Coates, M.~Quigley, and A.~Ng, ``An {Application} of
  {Reinforcement} {Learning} to {Aerobatic} {Helicopter} {Flight},'' in {\em
  Advances in {Neural} {Information} {Processing} {Systems}}, vol.~19, MIT
  Press, 2006.

\bibitem{abbeel_autonomous_2010}
P.~Abbeel, A.~Coates, and A.~Y. Ng, ``Autonomous {Helicopter} {Aerobatics}
  through {Apprenticeship} {Learning},'' {\em The International Journal of
  Robotics Research}, vol.~29, pp.~1608--1639, Nov. 2010.

\bibitem{sutton_reinforcement_2018}
R.~S. Sutton and A.~G. Barto, {\em Reinforcement learning: an introduction}.
\newblock Adaptive computation and machine learning series, Cambridge,
  Massachusetts: The MIT Press, second edition~ed., 2018.

\bibitem{novati_synchronised_2017}
G.~Novati, S.~Verma, D.~Alexeev, D.~Rossinelli, W.~M. van Rees, and
  P.~Koumoutsakos, ``Synchronised {Swimming} of {Two} {Fish},'' {\em
  Bioinspiration \& Biomimetics}, vol.~12, p.~036001, Mar. 2017.
\newblock arXiv:1610.04248 [physics].

\bibitem{verma_efficient_2018}
S.~Verma, G.~Novati, and P.~Koumoutsakos, ``Efficient collective swimming by
  harnessing vortices through deep reinforcement learning,'' {\em Proceedings
  of the National Academy of Sciences}, vol.~115, pp.~5849--5854, June 2018.

\bibitem{ma_fluid_2018}
P.~Ma, Y.~Tian, Z.~Pan, B.~Ren, and D.~Manocha, ``Fluid directed rigid body
  control using deep reinforcement learning,'' {\em ACM Transactions on
  Graphics}, vol.~37, pp.~96:1--96:11, July 2018.

\bibitem{rabault_artificial_2019}
J.~Rabault, M.~Kuchta, A.~Jensen, U.~Réglade, and N.~Cerardi, ``Artificial
  neural networks trained through deep reinforcement learning discover control
  strategies for active flow control,'' {\em Journal of Fluid Mechanics},
  vol.~865, pp.~281--302, Apr. 2019.

\bibitem{fan_reinforcement_2020}
D.~Fan, L.~Yang, Z.~Wang, M.~S. Triantafyllou, and G.~E. Karniadakis,
  ``Reinforcement learning for bluff body active flow control in experiments
  and simulations,'' {\em Proceedings of the National Academy of Sciences},
  vol.~117, pp.~26091--26098, Oct. 2020.

\bibitem{xu_learning_2019}
J.~Xu, T.~Du, M.~Foshey, B.~Li, B.~Zhu, A.~Schulz, and W.~Matusik, ``Learning
  to fly: computational controller design for hybrid {UAVs} with reinforcement
  learning,'' {\em ACM Transactions on Graphics}, vol.~38, pp.~42:1--42:12,
  July 2019.

\bibitem{lee_flow_2018}
X.~Y. Lee, A.~Balu, D.~Stoecklein, B.~Ganapathysubramanian, and S.~Sarkar,
  ``Flow {Shape} {Design} for {Microfluidic} {Devices} {Using} {Deep}
  {Reinforcement} {Learning},'' Nov. 2018.
\newblock arXiv:1811.12444 [cs, stat].

\bibitem{wang_meta-modeling_2019}
K.~Wang and W.~Sun, ``Meta-modeling game for deriving theory-consistent,
  microstructure-based traction–separation laws via deep reinforcement
  learning,'' {\em Computer Methods in Applied Mechanics and Engineering},
  vol.~346, pp.~216--241, Apr. 2019.

\bibitem{bendsoe_optimal_1989}
M.~P. Bendsøe, ``Optimal shape design as a material distribution problem,''
  {\em Structural optimization}, vol.~1, pp.~193--202, Dec. 1989.

\bibitem{bendsoe_topology_2004}
M.~P. Bendsøe and O.~Sigmund, {\em Topology {Optimization}}.
\newblock Berlin, Heidelberg: Springer, 2004.

\bibitem{hayashi_reinforcement_2020}
K.~Hayashi and M.~Ohsaki, ``Reinforcement {Learning} and {Graph} {Embedding}
  for {Binary} {Truss} {Topology} {Optimization} {Under} {Stress} and
  {Displacement} {Constraints},'' {\em Frontiers in Built Environment}, vol.~6,
  2020.

\bibitem{zhu_machine-specified_2021}
S.~Zhu, M.~Ohsaki, K.~Hayashi, and X.~Guo, ``Machine-specified ground
  structures for topology optimization of binary trusses using graph embedding
  policy network,'' {\em Advances in Engineering Software}, vol.~159,
  p.~103032, Sept. 2021.

\bibitem{sun_generative_2020}
H.~Sun and L.~Ma, ``Generative {Design} by {Using} {Exploration} {Approaches}
  of {Reinforcement} {Learning} in {Density}-{Based} {Structural} {Topology}
  {Optimization},'' {\em Designs}, vol.~4, p.~10, June 2020.

\bibitem{jang_generative_2022}
S.~Jang, S.~Yoo, and N.~Kang, ``Generative {Design} by {Reinforcement}
  {Learning}: {Enhancing} the {Diversity} of {Topology} {Optimization}
  {Designs},'' {\em Computer-Aided Design}, vol.~146, p.~103225, May 2022.

\bibitem{han_solving_2018}
J.~Han, A.~Jentzen, and W.~E, ``Solving high-dimensional partial differential
  equations using deep learning,'' {\em Proceedings of the National Academy of
  Sciences}, vol.~115, pp.~8505--8510, Aug. 2018.

\bibitem{e_deep_2017}
W.~E and B.~Yu, ``The {Deep} {Ritz} method: {A} deep learning-based numerical
  algorithm for solving variational problems,'' {\em arXiv:1710.00211 [cs,
  stat]}, Sept. 2017.
\newblock arXiv: 1710.00211.

\bibitem{yang_reinforcement_2023}
J.~Yang, T.~Dzanic, B.~Petersen, J.~Kudo, K.~Mittal, V.~Tomov, J.-S. Camier,
  T.~Zhao, H.~Zha, T.~Kolev, R.~Anderson, and D.~Faissol, ``Reinforcement
  {Learning} for {Adaptive} {Mesh} {Refinement},'' in {\em Proceedings of {The}
  26th {International} {Conference} on {Artificial} {Intelligence} and
  {Statistics}}, pp.~5997--6014, PMLR, Apr. 2023.

\bibitem{rabault_accelerating_2019}
J.~Rabault and A.~Kuhnle, ``Accelerating {Deep} {Reinforcement} {Learning}
  strategies of {Flow} {Control} through a multi-environment approach,'' {\em
  Physics of Fluids}, vol.~31, p.~094105, Sept. 2019.
\newblock arXiv:1906.10382 [physics].

\bibitem{novati_automating_2020}
G.~Novati, H.~L. de~Laroussilhe, and P.~Koumoutsakos, ``Automating {Turbulence}
  {Modeling} by {Multi}-{Agent} {Reinforcement} {Learning},'' {\em
  arXiv:2005.09023 [physics]}, Oct. 2020.
\newblock arXiv: 2005.09023.

\bibitem{liu_physics-informed_2021}
X.-Y. Liu and J.-X. Wang, ``Physics-informed {Dyna}-style model-based deep
  reinforcement learning for dynamic control,'' {\em Proceedings of the Royal
  Society A: Mathematical, Physical and Engineering Sciences}, vol.~477,
  p.~20210618, Nov. 2021.

\bibitem{shi_physics-informed_2023}
H.~Shi, Y.~Zhou, K.~Wu, S.~Chen, B.~Ran, and Q.~Nie, ``Physics-informed deep
  reinforcement learning-based integrated two-dimensional car-following control
  strategy for connected automated vehicles,'' {\em Knowledge-Based Systems},
  vol.~269, p.~110485, June 2023.

\bibitem{ramesh_physics-informed_2023}
A.~Ramesh and B.~Ravindran, ``Physics-{Informed} {Model}-{Based}
  {Reinforcement} {Learning},'' May 2023.
\newblock arXiv:2212.02179 [cs].

\bibitem{rodwell_physics-informed_2023}
C.~Rodwell and P.~Tallapragada, ``Physics-informed reinforcement learning for
  motion control of a fish-like swimming robot,'' {\em Scientific Reports},
  vol.~13, p.~10754, July 2023.

\bibitem{sutton_dyna_1991}
R.~S. Sutton, ``Dyna, an integrated architecture for learning, planning, and
  reacting,'' {\em ACM SIGART Bulletin}, vol.~2, pp.~160--163, July 1991.

\bibitem{janner_when_2019}
M.~Janner, J.~Fu, M.~Zhang, and S.~Levine, ``When to trust your model:
  model-based policy optimization,'' in {\em Proceedings of the 33rd
  {International} {Conference} on {Neural} {Information} {Processing}
  {Systems}}, no.~1122, pp.~12519--12530, Red Hook, NY, USA: Curran Associates
  Inc., Dec. 2019.

\bibitem{kaiser_model-based_2020}
L.~Kaiser, M.~Babaeizadeh, P.~Milos, B.~Osinski, R.~H. Campbell, K.~Czechowski,
  D.~Erhan, C.~Finn, P.~Kozakowski, S.~Levine, A.~Mohiuddin, R.~Sepassi,
  G.~Tucker, and H.~Michalewski, ``Model-{Based} {Reinforcement} {Learning} for
  {Atari},'' Feb. 2020.
\newblock arXiv:1903.00374 [cs, stat].

\bibitem{luo_algorithmic_2021}
Y.~Luo, H.~Xu, Y.~Li, Y.~Tian, T.~Darrell, and T.~Ma, ``Algorithmic {Framework}
  for {Model}-based {Deep} {Reinforcement} {Learning} with {Theoretical}
  {Guarantees},'' Feb. 2021.
\newblock arXiv:1807.03858 [cs, stat].

\bibitem{deisenroth_pilco_2011}
M.~P. Deisenroth and C.~E. Rasmussen, ``{PILCO}: a model-based and
  data-efficient approach to policy search,'' in {\em Proceedings of the 28th
  {International} {Conference} on {International} {Conference} on {Machine}
  {Learning}}, {ICML}'11, (Madison, WI, USA), pp.~465--472, Omnipress, June
  2011.

\bibitem{levine_learning_2014}
S.~Levine and P.~Abbeel, ``Learning {Neural} {Network} {Policies} with {Guided}
  {Policy} {Search} under {Unknown} {Dynamics},'' in {\em Advances in {Neural}
  {Information} {Processing} {Systems}}, vol.~27, Curran Associates, Inc.,
  2014.

\bibitem{heess_learning_2015}
N.~Heess, G.~Wayne, D.~Silver, T.~Lillicrap, T.~Erez, and Y.~Tassa, ``Learning
  {Continuous} {Control} {Policies} by {Stochastic} {Value} {Gradients},'' in
  {\em Advances in {Neural} {Information} {Processing} {Systems}}, vol.~28,
  Curran Associates, Inc., 2015.

\bibitem{clavera_model-augmented_2020}
I.~Clavera, V.~Fu, and P.~Abbeel, ``Model-{Augmented} {Actor}-{Critic}:
  {Backpropagating} through {Paths},'' May 2020.
\newblock arXiv:2005.08068 [cs, stat].

\bibitem{hafner_dream_2020}
D.~Hafner, T.~Lillicrap, J.~Ba, and M.~Norouzi, ``Dream to {Control}:
  {Learning} {Behaviors} by {Latent} {Imagination},'' Mar. 2020.
\newblock arXiv:1912.01603 [cs].

\bibitem{hafner_mastering_2022}
D.~Hafner, T.~Lillicrap, M.~Norouzi, and J.~Ba, ``Mastering {Atari} with
  {Discrete} {World} {Models},'' Feb. 2022.
\newblock arXiv:2010.02193 [cs, stat].

\bibitem{williams_simple_1992}
R.~J. Williams, ``Simple statistical gradient-following algorithms for
  connectionist reinforcement learning,'' {\em Machine Learning}, vol.~8,
  pp.~229--256, May 1992.

\bibitem{sutton_policy_1999}
R.~S. Sutton, D.~McAllester, S.~Singh, and Y.~Mansour, ``Policy {Gradient}
  {Methods} for {Reinforcement} {Learning} with {Function} {Approximation},''
  in {\em Advances in {Neural} {Information} {Processing} {Systems}}, vol.~12,
  MIT Press, 1999.

\bibitem{kakade_natural_2001}
S.~M. Kakade, ``A {Natural} {Policy} {Gradient},'' in {\em Advances in {Neural}
  {Information} {Processing} {Systems}}, vol.~14, MIT Press, 2001.

\bibitem{silver_deterministic_2014}
D.~Silver, G.~Lever, N.~Heess, T.~Degris, D.~Wierstra, and M.~Riedmiller,
  ``Deterministic {Policy} {Gradient} {Algorithms},'' in {\em Proceedings of
  the 31st {International} {Conference} on {Machine} {Learning}}, pp.~387--395,
  PMLR, Jan. 2014.

\bibitem{schulman_trust_2015}
J.~Schulman, S.~Levine, P.~Abbeel, M.~Jordan, and P.~Moritz, ``Trust {Region}
  {Policy} {Optimization},'' in {\em Proceedings of the 32nd {International}
  {Conference} on {Machine} {Learning}}, pp.~1889--1897, PMLR, June 2015.

\bibitem{watkins_q-learning_1992}
C.~J. C.~H. Watkins and P.~Dayan, ``Q-learning,'' {\em Machine Learning},
  vol.~8, pp.~279--292, May 1992.

\bibitem{hasselt_deep_2016}
H.~v. Hasselt, A.~Guez, and D.~Silver, ``Deep reinforcement learning with
  double {Q}-{Learning},'' in {\em Proceedings of the {Thirtieth} {AAAI}
  {Conference} on {Artificial} {Intelligence}}, {AAAI}'16, (Phoenix, Arizona),
  pp.~2094--2100, AAAI Press, Feb. 2016.

\bibitem{wang_dueling_2016}
Z.~Wang, T.~Schaul, M.~Hessel, H.~Van~Hasselt, M.~Lanctot, and N.~De~Freitas,
  ``Dueling network architectures for deep reinforcement learning,'' in {\em
  Proceedings of the 33rd {International} {Conference} on {International}
  {Conference} on {Machine} {Learning} - {Volume} 48}, {ICML}'16, (New York,
  NY, USA), pp.~1995--2003, JMLR.org, June 2016.

\bibitem{schulman_proximal_2017}
J.~Schulman, F.~Wolski, P.~Dhariwal, A.~Radford, and O.~Klimov, ``Proximal
  {Policy} {Optimization} {Algorithms},'' Aug. 2017.
\newblock arXiv:1707.06347 [cs].

\bibitem{bellman_markovian_1957}
R.~Bellman, ``A {Markovian} {Decision} {Process},'' {\em Journal of Mathematics
  and Mechanics}, vol.~6, no.~5, pp.~679--684, 1957.

\bibitem{dolcetta_approximate_1984}
I.~C. Dolcetta and H.~Ishii, ``Approximate solutions of the bellman equation of
  deterministic control theory,'' {\em Applied Mathematics \& Optimization},
  vol.~11, pp.~161--181, Feb. 1984.

\bibitem{sutton_learning_1988}
R.~S. Sutton, ``Learning to predict by the methods of temporal differences,''
  {\em Machine Learning}, vol.~3, pp.~9--44, Aug. 1988.

\bibitem{bradtke_linear_1996}
S.~J. Bradtke and A.~G. Barto, ``Linear {Least}-{Squares} algorithms for
  temporal difference learning,'' {\em Machine Learning}, vol.~22, pp.~33--57,
  Mar. 1996.

\end{thebibliography}

\end{document}